\documentclass{article}

\PassOptionsToPackage{numbers, compress}{natbib}
\usepackage[main, final]{neurips_2026}
\usepackage[utf8]{inputenc}
\usepackage[T1]{fontenc}
\usepackage{url}
\usepackage{booktabs}
\usepackage{amsfonts}
\usepackage{nicefrac}
\usepackage{microtype}
\usepackage{xcolor}
\usepackage{graphicx}
\usepackage{subcaption}
\usepackage{enumitem}
\usepackage{bbm}
\usepackage{mathrsfs}
\usepackage{amsmath}
\usepackage{amssymb}
\usepackage{mathtools}
\usepackage{amsthm}
\usepackage{aliascnt}
\usepackage[colorlinks,linkcolor={red!80!black},citecolor={blue},allbordercolors={1 1 1},urlcolor={blue!80!black},hypertexnames=false]{hyperref}
\usepackage{algorithm}
\usepackage{algpseudocode}
\usepackage{float}
\usepackage{wrapfig2}
\usepackage[normalem]{ulem}
\usepackage[capitalize,noabbrev]{cleveref}

\newlist{assumplist}{enumerate}{1}
\setlist[assumplist]{label=\textbf{Assumption \Alph*.},wide=0pt,ref=\Alph*,topsep=0pt,parsep=0pt,itemsep=0pt}

\newlist{assumplistm}{enumerate}{1}
\setlist[assumplistm]{label=\textbf{Assumption \Alph*.},wide=0pt,ref=\Alph*,topsep=0pt,parsep=0pt,itemsep=0pt}

\newlist{assumplista}{enumerate}{1}
\setlist[assumplista]{label=\textbf{Assumption \alph*.},wide=0pt,ref=\alph*,topsep=0pt,parsep=0pt,itemsep=0pt}

\newlist{assumpar}{enumerate}{1}
\setlist[assumpar]{label=\textbf{Assumption \arabic*.},wide=0pt,ref=\arabic*,topsep=0pt,parsep=0pt,itemsep=0pt}

\newlist{assumpNum}{enumerate}{1}
\setlist[assumpNum]{
    label=\textbf{Assumption \arabic*.},
    ref=\textbf{\arabic*},
    start=0,
    wide=0pt,
    topsep=5pt,
    itemsep=3pt,
    before=\crefalias{assumpNumi}{assumpNum}
}

\crefname{assumplisti}{Assumption}{Assumptions}
\Crefname{assumplisti}{Assumption}{Assumptions}
\crefname{assumplistmi}{Assumption}{Assumptions}
\Crefname{assumplistmi}{Assumption}{Assumptions}
\crefname{assumplistai}{Assumption}{Assumptions}
\Crefname{assumplistai}{Assumption}{Assumptions}
\crefname{assumpari}{Assumption}{Assumptions}
\crefname{assumpNum}{Assumption}{Assumptions}

\crefformat{assumplisti}{Assumption~(\textbf{#2#1#3})}
\Crefformat{assumplisti}{Assumption~(\textbf{#2#1#3})}

\crefformat{assumpari}{Assumption~(\textbf{#2#1#3})}
\Crefformat{assumpari}{Assumption~(\textbf{#2#1#3})}

\crefrangeformat{assumplisti}{Assumptions~(\textbf{#3#1#4}) to~(\textbf{#5#2#6})}
\Crefrangeformat{assumplisti}{Assumptions~(\textbf{#3#1#4}) to~(\textbf{#5#2#6})}

\crefrangeformat{assumpari}{Assumptions~(\textbf{#3#1#4}) to~(\textbf{#5#2#6})}
\Crefrangeformat{assumpari}{Assumptions~(\textbf{#3#1#4}) to~(\textbf{#5#2#6})}

\crefmultiformat{assumplisti}{Assumptions~(\textbf{#2#1#3})}{ and~(\textbf{#2#1#3})}{, (\textbf{#2#1#3})}{ and~(\textbf{#2#1#3})}
\Crefmultiformat{assumplisti}{Assumptions~(\textbf{#2#1#3})}{ and~(\textbf{#2#1#3})}{, (\textbf{#2#1#3})}{ and~(\textbf{#2#1#3})}

\crefmultiformat{assumpari}{Assumptions~(\textbf{#2#1#3})}{ and~(\textbf{#2#1#3})}{, (\textbf{#2#1#3})}{ and~(\textbf{#2#1#3})}
\Crefmultiformat{assumpari}{Assumptions~(\textbf{#2#1#3})}{ and~(\textbf{#2#1#3})}{, (\textbf{#2#1#3})}{ and~(\textbf{#2#1#3})}

\crefrangelabelformat{assumplisti}{(\textbf{#3#1#4}) to~(\textbf{#5#2#6})}

\theoremstyle{plain}
\newtheorem{theorem}{Theorem}[section]
\crefname{theorem}{Theorem}{Theorems}
\Crefname{theorem}{Theorem}{Theorems}
\newaliascnt{proposition}{theorem}
\newtheorem{proposition}[proposition]{Proposition}
\aliascntresetthe{proposition}
\crefname{proposition}{Proposition}{Propositions}
\Crefname{proposition}{Proposition}{Propositions}
\newaliascnt{lemma}{theorem}
\newtheorem{lemma}[lemma]{Lemma}
\aliascntresetthe{lemma}
\crefname{lemma}{Lemma}{Lemmas}
\Crefname{lemma}{Lemma}{Lemmas}
\newaliascnt{corollary}{theorem}
\newtheorem{corollary}[corollary]{Corollary}
\aliascntresetthe{corollary}
\crefname{corollary}{Corollary}{Corollaries}
\Crefname{corollary}{Corollary}{Corollaries}
\theoremstyle{definition}

\newaliascnt{definition}{theorem}
\newtheorem{definition}[definition]{Definition}
\aliascntresetthe{definition}
\crefname{definition}{Definition}{Definitions}
\Crefname{definition}{Definition}{Definitions}

\newaliascnt{assumption}{theorem}

\aliascntresetthe{assumption}
\crefname{assumption}{Assumption}{Assumptions}
\Crefname{assumption}{Assumption}{Assumptions}

\theoremstyle{remark}

\newaliascnt{remark}{theorem}
\newtheorem{remark}[remark]{Remark}
\aliascntresetthe{remark}
\crefname{remark}{Remark}{Remarks}
\Crefname{remark}{Remark}{Remarks}

\DeclareMathOperator*{\argmin}{\arg\!\min}

\newcommand{\R}{\mathbb{R}}
\newcommand{\E}{\mathbb{E}}

\newcommand{\X}{\mathcal{X}}

\newcommand{\LL}{L^2(\mathcal{X})}
\newcommand{\norm}[1]{\left\|#1\right\|}
\newcommand{\abs}[1]{\left|#1\right|}

\newcommand{\seq}[1]{\left(#1\right)}

\newcommand{\nin}{n_{in}}
\newcommand{\nout}{n_{out}}
\newcommand{\ftt}{f_{t}^{\mathrm{NN}}}

\newcommand{\fttntk}{f_{t}^{\mathrm{NTK}}}
\newcommand{\fttop}{f_{t_{\mathrm{op}}}^{\mathrm{NN}}}
\newcommand{\fttopntk}{f_{t_{\mathrm{op}}}^{\mathrm{NTK}}}

\newcommand{\ftttop}{\widetilde{f}_{t_{\mathrm{op}}}^{\mathrm{NN}}}

\newcommand{\kntk}{K^\mathrm{NTK}}
\newcommand{\hntk}{\mathcal{H}^\mathrm{NTK}}

\newcommand{\hntktw}{h_{t,\omega}^{\mathrm{NTK}}}
\newcommand{\hntktwop}{h_{t_{\mathrm{op}_h},\omega}^{\mathrm{NTK}}}
\newcommand{\Gram}{\mathbf{K}}
\newcommand{\Yom}{Y_\omega}

\title{On Generalization in Bilevel Least Squares with Overparameterized Neural Networks}

\author{
  Fares El~Khoury\thanks{Correspondence to: \href{mailto:fares.el-khoury@inria.fr}{\texttt{fares.el-khoury@inria.fr}}.}
  \\
  Universit\'e Grenoble Alpes, Inria, \\CNRS, Grenoble INP, LJK, \\38000 Grenoble, France
  \And
  Edouard Pauwels\\
  Toulouse School of Economics, \\Universit\'e Toulouse Capitole,\\ 31080 Toulouse, France\\
  \And
  Samuel Vaiter\\
  CNRS \& Universit\'e C\^ote d'Azur,\\ Laboratoire J. A. Dieudonn\'e, \\06108 Nice, France
  \And
  Michael Arbel\\
  Universit\'e Grenoble Alpes, Inria, \\CNRS, Grenoble INP, LJK, \\38000 Grenoble, France
}

\begin{document}

\addtocontents{toc}{\protect\setcounter{tocdepth}{0}}

\maketitle

\begin{abstract}
  Bilevel optimization addresses machine learning problems in which the outer objective is a \emph{value function} implicitly defined through the solution of an inner problem. While its algorithmic aspects are well developed, its generalization properties remain much less understood. In this paper, we study the generalization of inexact gradient methods when both objectives are least-squares regression problems and when the estimator of the value function's gradient, or \emph{hypergradient}, is obtained by implicit differentiation in function space. This approach allows us to derive stationarity bounds for these methods in terms of the uniform generalization errors of the two regression functions needed to estimate the hypergradient. When these regression functions are defined as overparameterized two-layer ReLU networks trained by gradient flow with \emph{early stopping}, we obtain explicit finite-sample bounds on these generalization errors in the Neural Tangent Kernel (NTK) regime. Combining these with our general inexact gradient analysis shows the effect of sample size on the overall convergence of the bilevel method. Numerical experiments on synthetic data illustrate our results. Beyond the NTK regime, further experiments reveal an \emph{initialization-dependent implicit bias} induced by the bilevel structure, which can lead to \emph{feature learning} at the inner level and overall performance improvements.
\end{abstract}

\section{Introduction}
Bilevel optimization is a class of \emph{hierarchical} optimization problems in which the solution to an \emph{outer} problem depends on the solution of an \emph{inner} problem \citep{Colson2007}. These methods naturally arise in numerous machine learning applications, including hyperparameter optimization \citep{Maclaurin2015,Feurer2019Hyperparameter,franceschi2017forward}, meta-learning \citep{franceschi2018bilevel,finn2017model}, neural architecture search \citep{Liu2018}, and reinforcement learning \citep{hong2023two,nikishin2022control}. The main challenge in bilevel problems is that the outer objective is defined \emph{implicitly} through the solution of the inner problem, making both gradient computation and \emph{generalization} analysis nontrivial \citep{pedregosa2016hyperparameter}.

Bilevel optimization poses two intertwined challenges: numerically, one must approximate not only the inner-level solution but also its sensitivity to outer variables in order to compute hypergradients \citep{domke2012generic,baydin2018automatic,ghadimi2018approximation}; and statistically, one must understand how the outer solution generalizes as the sample size grows \citep{elkhoury2025learning,ji2021bilevel,Arbel2022Amortized}. A large body of work in the well-specified setting with a unique inner level solution addresses the numerical difficulty. Crucially, they typically require the inner objective to be strongly convex, which ensures a unique, smoothly varying inner solution and is therefore amenable to stability and convergence analyses \citep{gould2016differentiating,ablin2020super,blondel2022efficient}. Modern machine learning applications, however, usually employ neural networks as inner models \citep{bao2020bi,xiao2025first,Liu2018}; the resulting inner problems are nonconvex and often overparameterized, so multiple inner solutions can exist and their dependence on the outer variables becomes ambiguous \citep{vicol2022implicit,AllenZhu2019Convergence}. This dependence can lead to instability, both because inner solutions may be non-unique and because small perturbations in the lower-level problem can induce large changes in the hypergradient \citep{liu2021towards, arbel2022non, bolte2026geometric}. As a consequence, strong generalization guarantees in the nonconvex setting are still largely limited, and the behavior of the resulting solutions on unseen data is still poorly understood. On the statistical side, most generalization results adopt a fixed finite-dimensional parametric viewpoint \citep{dagreou2022framework,hu2023contextual}. This perspective becomes restrictive for overparameterized networks and nonparametric methods, where the effective parameter space (or feature map) can grow with the sample size and the parameters obtained at different sample sizes are not directly comparable. Bridging this gap therefore requires analytical tools that can handle \emph{(i)} nonconvex inner landscapes with possibly multiple solutions arising from overparameterization, and \emph{(ii)} nonparametric function spaces, such as a Reproducing Kernel Hilbert Space (RKHS), in order to obtain meaningful stability and generalization guarantees for bilevel learning in modern settings.

To address these challenges, the \emph{functional bilevel optimization} framework, introduced in \citep{petrulionyte2024functional}, provides a principled framework that shifts focus from parameter-centric approaches to a function-space perspective. By formulating the inner problem as an optimization over an infinite-dimensional \emph{Hilbert} space of functions, such as an RKHS \citep{elkhoury2025learning}, this approach naturally accommodates multiple inner solutions, which frequently arise in overparameterized neural networks. A crucial advantage of this functional perspective is that it only requires strong convexity in the function space, rather than with respect to the parameter vector, the latter being rarely satisfied by modern neural networks \citep{liu2022loss}. This ensures the \emph{uniqueness} of the inner solution for any outer variable, providing a well-defined dependency of the inner problem on the outer parameters and making stability and generalization analyses tractable in practice. In a related line of work, \citet{elkhoury2025learning} provided generalization guarantees for functional bilevel optimization where the considered space is an RKHS. However, these results remain restrictive in the deep learning setting, as they do not accommodate highly expressive models such as overparameterized networks, leaving the question of generalization in bilevel optimization in such settings largely open.

In the present paper, we address this gap by studying \emph{Overparameterized Bilevel Optimization} (OBO) problems in which the lower-level model is \emph{overparameterized}, \emph{i.e.}, it contains more parameters than training samples and thus may have multiple global minimizers for the inner problem \citep{vicol2022implicit}. Formally, we consider problems of the form:
\begin{equation}\tag{OBO}\label{eq:obo}
    \min_{\omega\in\Omega}\mathcal{F}(\omega)\coloneqq L_{out}(\omega,h^\star_\omega)\quad\text{such that}\quad h^\star_\omega=\argmin_{h\in\mathcal{H}}L_{in}(\omega,h),
\end{equation}
where $\Omega\subset\mathbb{R}^d$ is nonempty, compact, and convex, $\mathcal{H}$ is a Hilbert space of functions on $\mathcal{X}\subset\mathbb{R}^p$ taking values in $\mathcal{Y}\subset\mathbb{R}$, and $L_{in},L_{out}:\Omega\times\mathcal{H}\to\mathbb{R}$ denote the inner and outer objective functions, respectively. Throughout this work, both $L_{in}$ and $L_{out}$ take the form of expected squared losses, a choice motivated by its practical relevance in machine learning \citep{franceschi2018bilevel,pedregosa2016hyperparameter} and its ubiquity as a theoretical framework \citep{caponnetto2007optimal,arora2019exact}.

\textbf{Contributions.} We study generalization in \emph{bilevel least squares}, \emph{i.e.}, bilevel optimization problems where both levels are least-squares regressions. This setting is advantageous, as it expresses the hypergradient $\nabla\mathcal{F}$ in terms of two regression functions, the inner solution and the outer regression function. The latter does not depend on the outer variable and can thus be estimated once and for all, which simplifies both the analysis and the algorithmic complexity. Our contributions are as follows.

\begin{itemize}[leftmargin=*,itemsep=2pt,topsep=2pt,parsep=0pt]
\item \label{item:algo}\textbf{(i) Convergence under inexact hypergradients.} We bound the near-stationarity measure of a projected mini-batch stochastic gradient descent (SGD) method driven by a plug-in hypergradient estimator, in terms of the two estimators' generalization errors (\cref{thm:eta-scaled}).

\item\label{item:ntk}\textbf{(ii) Generalization rates in the NTK regime.} When these two estimators are two-layer ReLU networks with bias in both layers trained by gradient flow with early stopping, we derive finite-sample generalization bounds in the neural tangent kernel (NTK) regime (\cref{prop:gen_bounds_main}). These bounds hold in expectation, and uniformly over the outer variable for the inner estimator. To obtain them, we truncate the trained networks, \emph{i.e.}, set them to zero, on a rare event outside of which they provably remain in the NTK regime.

\item \textbf{(iii) End-to-end sample complexity.} Plugging the rates of \textbf{(ii)} into the bound of \textbf{(i)} yields the sample complexity of the bilevel method in the NTK regime (\cref{cor:end_to_end}): the statistical error is governed by the slower of the two regression rates, so the inner and outer sample sizes should be balanced to prevent either level from becoming a bottleneck.

\item \textbf{(iv) Implicit bias beyond the NTK regime.} Synthetic experiments suggest an initialization-dependent implicit bias of the bilevel structure, which promotes inner-level feature learning and improves performance over the NTK regime.
\end{itemize}

\section{Overparameterized bilevel optimization with squared losses}\label{sec:obo}
In this section, we introduce the problem under consideration, derive the expression of its hypergradient, and present the gradient-based algorithm we analyze along with its convergence guarantee.

\subsection{Problem formulation}\label{subsec:prel}
In many machine learning problems with a bilevel structure, the inner and outer objectives $L_{in}$ and $L_{out}$ are typically \emph{expectations of pointwise losses} over some data distributions. In this work, we instantiate both as \emph{squared losses}, a choice that arises widely in regression \citep{hastie2009elements} and causal inference \citep{xu2021learning,kim2025optimality}, and that allows us to develop our theoretical analysis in a simplified setting. Specifically, we consider:
\begin{equation}\label{eq:squared_losses}
    L_{in}(\omega, h)=\frac{1}{2}\mathbb{E}_{(x,z)\sim\mathbb{P}_{XZ}}\left[\left|h(x)-g_\omega(z)\right|^2\right],\quad L_{out}(h)=\frac{1}{2}\mathbb{E}_{(x,y)\sim\mathbb{P}_{XY}}\left[\left|h(x)-y\right|^2\right],
\end{equation}
where $g_\omega:\mathcal{Z}\subset\mathbb{R}^r\to\mathbb{R}$ is a model parameterized by $\omega$. Note that we omit the explicit dependence of $L_{out}$ on $\omega$, as it occurs only implicitly through $h^\star_\omega$. The distributions $\mathbb{P}_{XZ}$ and $\mathbb{P}_{XY}$ are the marginals of a joint distribution $\mathbb{P}$ over triplets $(x,y,z)$ in $\mathcal{X}\times\mathcal{Y}\times\mathcal{Z}$, sharing the same marginal $\mathbb{P}_X$ over elements $x\in \mathcal{X}$. A natural space of functions $\mathcal{H}$ in this context is $\LL$, the space of square-integrable functions with respect to $\mathbb{P}_X$, defined on $\mathcal{X}$ and taking values in $\mathcal{Y}$. Provided that $z\mapsto g_{\omega}(z)$ is square integrable, this problem is guaranteed to admit a \emph{unique} solution to the inner problem, which takes the simple form of a conditional expectation, \emph{i.e.}, $h^\star_\omega(x)=\mathbb{E}_\mathbb{P}[g_{\omega}(z)\mid x]$, for any $\omega\in\Omega$. In practice, such conditional expectation is only estimated, typically using a neural network, and thus requires careful analysis of the generalization error, which is a central challenge in this study. Typical applications of this problem in machine learning include instrumental variable regression and value function fitting in reinforcement learning, as further discussed in \cref{app:examples}. In what follows, $\|\cdot\|$ denotes the Euclidean norm, $\|\cdot\|_{\LL}$ denotes the $L^2$ norm with respect to $\mathbb{P}_X$, and $\langle\cdot,\cdot\rangle_{\LL}$ denotes the corresponding inner product. For notational simplicity, we will write $\mathbb{P}$ without subscript in the remainder of the paper whenever the meaning can be inferred from context.

Throughout the paper, we assume that the model $g_\omega$ is uniformly smooth with bounded values and gradients, that the target space is bounded, that the input space is compact, and that $\mathbb{P}_X$ admits a density $\mathsf{p}_\mathbb{P}$ with respect to the Lebesgue measure on $\mathcal{X}$ that is bounded above and bounded away from zero on some open ball. These assumptions are formalized below.
\begin{assumplistm}
\item\label{ass:gome} (Parameterization properties of $g_\omega$) There exists an open convex set $U\supset\Omega$ such that, for all $z\in\mathcal{Z}$, the map $(\omega,z)\mapsto g_\omega(z)$ is differentiable on $U$ with respect to $\omega$ and its partial derivative $\partial_\omega g_\omega(z)\in\mathbb{R}^d$ is $L$-Lipschitz on $U$. Moreover, there exist constants $G,D>0$ such that:
\begin{equation*}
    \sup_{\omega\in U,\,z\in\mathcal{Z}}\left\|\partial_\omega g_\omega(z)\right\|\leq G\quad\text{and}\quad\sup_{\omega\in U,\,z\in\mathcal{Z}}\left|g_\omega(z)\right|\leq D.
\end{equation*}
\item\label{ass:bound_la} (Boundedness of $\mathcal{Y}$) The target space $\mathcal{Y}\subset\mathbb{R}$ is bounded, \emph{i.e.}, there exists a constant $B>0$ such that $|y|\leq B$ for any $y\in\mathcal{Y}$.
\item\label{ass:bound_in} (Compactness of $\mathcal{X}$) The input space $\mathcal{X}\subset\mathbb{R}^p$ is compact with a constant $A>0$ such that $\|x\|\leq A$ for any $x\in\mathcal{X}$.
\item\label{ass:bounded_density_P} (Bounded density) There exists a constant $M_\mathbb{P}>0$ such that $\mathsf{p}_\mathbb{P}\leq M_\mathbb{P}$. Moreover, $\mathsf{p}_\mathbb{P}$ is bounded below by a positive constant on some open ball contained in $\mathcal{X}$.
\end{assumplistm}
These are relatively mild conditions introduced mainly to simplify the analysis as further discussed in \cref{subapp:ass_obo}. Additional assumptions will be introduced in later sections whenever needed.

\subsection{Optimization algorithm for solving (\ref{eq:obo})}
We aim to solve \eqref{eq:obo} with a gradient-based algorithm, driven by the \emph{hypergradient} $\nabla\mathcal{F}$. The following result establishes the differentiability of $\mathcal{F}$ and provides an explicit expression for $\nabla\mathcal{F}$, by applying the functional implicit differentiation theorem \citep[Proposition~2.2]{petrulionyte2024functional}.

\begin{proposition}[Hypergradient of $\mathcal{F}$]\label{prop:hyper}
    Under \Cref{ass:bound_la,ass:gome}, $\mathcal{F}$ is differentiable on $U$, and its hypergradient is given, for any $\omega\in U$, by:
    \begin{equation}\label{eq:closed_hyper}
        \nabla\mathcal{F}(\omega)=-\mathbb{E}_\mathbb{P}\left[\partial_\omega g_\omega(z)a^\star_\omega(x)\right]\in\mathbb{R}^d,
    \end{equation}
    where $a^\star_\omega=-h^\star_\omega+b^\star\in\LL$ is the adjoint function, $h^\star_\omega:x\mapsto\mathbb{E}_\mathbb{P}[g_\omega(z)\mid x]$ is the inner solution, and $b^\star:x\mapsto\mathbb{E}_\mathbb{P}[y\mid x]$ is the regression function, defined as the minimizer of $L_{out}$ over $\LL$. Moreover, $\mathcal{F}$ is $\mathcal{L}$-smooth on $U$, in particular on $\Omega$, with $\mathcal{L}\coloneqq G^2+L(D+B)$ where the constants $G, L, D$ and $B$ are defined in \cref{ass:bound_la,ass:gome}.
\end{proposition}
 
While the expression in \cref{eq:closed_hyper} is exact, it requires full knowledge of $\mathbb{P}$, $h_{\omega}^{\star}$ and $b^{\star}$, which are unavailable in practice, making it necessary to estimate $\nabla\mathcal{F}(\omega)$ from data. Its structure is particularly appealing: since $\partial_\omega g_\omega(z)$ is known, the statistical difficulty reduces entirely to approximating the adjoint function $a^\star_\omega$, which itself decomposes into two regression functions $b^\star$ and $h^\star_\omega$. This structure is key to our analysis, as it allows us to approximate $h^\star_\omega$ and $b^\star$ independently and combine their estimates to obtain an approximation of the hypergradient using \emph{finite-sample} estimates.

\textbf{Empirical losses. }Estimating $h^\star_\omega$ and $b^\star$ requires a \emph{finite-sample approximation} of \eqref{eq:obo}. To this end, we assume access to two datasets $\mathcal{D}_{in}=\left\{(x_i,z_i)\right\}_{i=1}^{\nin}$ and $\mathcal{D}_{out}=\left\{(\tilde{x}_j,\tilde{y}_j)\right\}_{j=1}^{\nout}$ consisting of \emph{i.i.d.} samples drawn from the distributions $\mathbb{P}_{XZ}$ and $\mathbb{P}_{XY}$, respectively. This is a standard setting in machine learning, since it leads to \emph{tractable} approximations of \eqref{eq:obo}. We therefore consider the corresponding empirical problem, obtained by replacing the population losses $L_{in}$ and $L_{out}$ with their empirical counterparts:
\begin{equation}\label{eq:lin_lout_main}
    \widehat{L}_{in}(\omega,h)=\frac{1}{2\nin}\sum_{i=1}^{\nin}\left(h(x_i)-g_\omega(z_i)\right)^2,\quad\widehat{L}_{out}(h)=\frac{1}{2\nout}\sum_{j=1}^{\nout}\left(h(\tilde{x}_j)-\tilde{y}_j\right)^2.
\end{equation}
These serve as training objectives for the estimators $\hat{h}_\omega$ and $\hat{b}$ of $h^\star_\omega$ and $b^\star$: the former is trained by minimizing $\widehat{L}_{in}(\omega,\cdot)$ for a given $\omega$, while the latter is estimated by minimizing $\widehat{L}_{out}$.

\textbf{Hypergradient estimator. }With $\hat{h}_\omega$ and $\hat{b}$ at hand, it remains to approximate the expectation in \cref{eq:closed_hyper}. Accordingly, we assume access, at each use of the estimator, to a \emph{fresh} mini-batch $\mathcal{B}_g=\{(\underline{x}_l,\underline{z}_l)\}_{l=1}^{n_g}$ of \emph{i.i.d.} samples drawn from $\mathbb{P}_{XZ}$, independent of $\mathcal{D}_{in}$, $\mathcal{D}_{out}$ and of the previously drawn batches. We then employ a \emph{plug-in} estimator for $\nabla\mathcal{F}(\omega)$ obtained by replacing the population expectation in \cref{eq:closed_hyper} with the empirical average over $\mathcal{B}_g$:
\begin{equation}\label{eq:gradient_est}
    \widehat{\nabla\mathcal{F}}(\omega)=-\frac{1}{n_g}\sum_{l=1}^{n_g}\partial_\omega g_\omega(\underline{z}_l)(-\hat{h}_\omega(\underline{x}_l)+\hat{b}(\underline{x}_l)),
\end{equation}
where we approximate $a^\star_\omega$ by its finite-sample estimate of the form $-\hat{h}_\omega+\hat{b}$ since $a^\star_\omega=-h^\star_\omega+b^\star$.

\noindent
\setlength{\intextsep}{5pt}
\begin{wrapfigure}[12]{r}{0.465\textwidth}
\vspace{-\baselineskip}
\begin{minipage}{0.465\textwidth}
\begin{algorithm}[H]
\caption{Projected mini-batch biased SGD}\label{alg:pmbsgd}
\begin{algorithmic}[1]
\Require $\omega_0\in\Omega$, $\eta$, $K$, $n_g$, $\mathcal{D}_{out}$, $\mathcal{D}_{in}$
\State Estimate $\hat{b}$ on $\mathcal{D}_{out}$
\For{$k=0,\ldots,K-1$}
    \State Estimate $\hat{h}_{\omega_k}$ on $\mathcal{D}_{in}$
    \State Draw a fresh $\mathcal{B}^k_g$ of size $n_g$
    \State Form $\widehat{\nabla\mathcal{F}}(\omega_k)$ via \cref{eq:gradient_est}
    \State $\omega_{k+1}\gets\Pi_\Omega\!\left(\omega_k-\eta\widehat{\nabla\mathcal{F}}(\omega_k)\right)$
\EndFor
\State \Return $\omega_{k^\star}$, $k^\star$ uniform in $\{0,\ldots,K-1\}$
\end{algorithmic}
\end{algorithm}
\end{minipage}
\end{wrapfigure}
\textbf{Algorithm. }We use the plug-in estimator $\widehat{\nabla\mathcal{F}}$ defined in \cref{eq:gradient_est} to drive a gradient-based bilevel algorithm that approximately solves \eqref{eq:obo}. Starting from $\omega_0\in\Omega$, we first compute $\hat{b}$ from $\mathcal{D}_{out}$ once and for all, since it does not depend on the outer variable. At each outer iteration $k=0,\ldots,K-1$, we then re-estimate $\hat{h}_{\omega_k}$ from $\mathcal{D}_{in}$ at the current outer variable, draw a \emph{fresh} mini-batch $\mathcal{B}^k_g$ of \emph{fixed} size $n_g$, independent of the previous ones, and update $\omega_{k+1}=\Pi_\Omega(\omega_k-\eta\widehat{\nabla\mathcal{F}}(\omega_k))$, where $\eta>0$ is the stepsize and $\Pi_\Omega$ is the Euclidean projection onto the feasible set $\Omega$. Only the mini-batches are redrawn along the iterations: $\mathcal{D}_{in}$ and $\mathcal{D}_{out}$ are sampled \emph{once} and reused at every $\omega_k$, so that $\hat{h}_\omega$ is required to be accurate \emph{uniformly} over $\Omega$. Since $\hat{h}_{\omega}$ and $\hat{b}$ are finite-sample approximations of $h^\star_{\omega}$ and $b^\star$, the resulting $\widehat{\nabla\mathcal{F}}$ is a \emph{biased} estimator of $\nabla\mathcal{F}$. As the iterates are constrained to $\Omega$, we refer to this scheme as \emph{projected mini-batch biased stochastic gradient descent (SGD)}.

\begin{remark}[Full-batch variant]
Instead of drawing a fresh mini-batch at each outer iteration, one can draw a single dataset once and use all of it at every iteration. Since the iterates then depend on this dataset, the analysis requires the hypergradient estimate to be accurate simultaneously for all $\omega\in\Omega$, and not only at the current iterate. Such a uniform control can be obtained through a covering argument over $\Omega$, but it requires a dataset substantially larger than a single mini-batch, and thus full-batch gradient evaluations, which are costly. As both approaches require a comparable number of outer iterations, and each iteration processes its own batch, the full-batch variant is computationally \emph{more expensive} than \cref{alg:pmbsgd}.
\end{remark}

\begin{remark}[Computational benefit of the squared loss setting]\label{rmk:squared_loss}
In general functional bilevel problems, the adjoint $a^\star_\omega$ is defined as the minimizer of a quadratic problem depending on $\omega$. Hence, both $a^\star_\omega$ and $h^\star_\omega$ must be re-estimated at each outer iteration \citep{petrulionyte2024functional,elkhoury2025learning}. Our setting simplifies this step considerably. Indeed, in the estimator $-\hat{h}_\omega+\hat{b}$ of $a^\star_\omega$, only $\hat{h}_\omega$ depends on $\omega$, while $\hat{b}$ can be estimated \emph{once} and kept fixed throughout the optimization. This reduces the computational cost of the algorithm: each outer iteration involves a single re-estimation of $\hat{h}_{\omega_k}$, $n_g$ evaluations of the summands in \cref{eq:gradient_est} and one projection onto $\Omega$, so that $K$ outer iterations require $K+1$ functional estimations instead of $2K$ in the general case. Among these operations, the dominant one is the re-estimation of $\hat{h}_{\omega_k}$, the projection being inexpensive for simple sets such as Euclidean balls or boxes.
\end{remark}

\subsection{General convergence result for the projected mini-batch biased SGD}
Since $\mathcal{F}$ is nonconvex and the iterates are constrained to $\Omega$, we quantify the convergence of the algorithm through the \emph{gradient mapping} \citep[Definition~10.5]{beck2017first}, defined for any $\lambda>0$ by:
\begin{equation*}
    \mathcal{G}_{\lambda}(\omega)\coloneqq\lambda^{-1}\left(\omega-\Pi_\Omega\left(\omega-\lambda\nabla\mathcal{F}(\omega)\right)\right),
\end{equation*}
and adopt $\|\mathcal{G}_\lambda(\omega)\|$ as our near-stationarity measure for \eqref{eq:obo}. This quantity measures the displacement produced by a single \emph{exact} projected gradient step at stepsize $\lambda$, and vanishes precisely at the stationary points of \eqref{eq:obo}. We now state our main convergence result for \cref{alg:pmbsgd} when using general estimators $\hat{h}_{\omega}$ and $\hat{b}$.
\begin{theorem}[Convergence guarantee of projected mini-batch biased SGD]\label{thm:eta-scaled}
Assume \ref{ass:gome} and \ref{ass:bound_la}, and fix $\omega_0\in\Omega$. Let  $\Delta_\mathcal{F}\coloneqq\mathcal{F}(\omega_0)-\mathcal{F}^\star$ with $\mathcal{F}^\star\coloneqq\min_{\omega\in\Omega}\mathcal{F}(\omega)$ and $\{\omega_k\}_{k\geq0}$ be generated by $\omega_{k+1}=\Pi_\Omega(\omega_k-\eta\widehat{\nabla\mathcal{F}}(\omega_k))$ with stepsize $0<\eta\leq\frac{1}{2\mathcal{L}}$, where $\mathcal{L}$ is the smoothness constant of $\mathcal{F}$ provided in \cref{prop:hyper}. For any $K\geq1$, the iterates $\{\omega_k\}_{k\geq0}$ satisfy:
\begin{equation*}
    \frac{1}{K}\sum_{k=0}^{K-1}\mathbb{E}\left[\left\|\mathcal{G}_{1/\mathcal{L}}(\omega_k)\right\|^2\right]\leq c\left[\frac{\Delta_\mathcal{F}}{\eta K}+4G^2(e_h+e_b)+\frac{4\mathcal{L}G^2\eta}{n_g}\left((D+B)^2+2(e_h+e_b)\right)\right],
\end{equation*}
where $c=18+12\sqrt{2}$, $e_h=\mathbb{E}\left[\sup_{\omega\in\Omega}\|\hat{h}_\omega-h^\star_\omega\|_{\LL}^2\right]$, $e_b=\mathbb{E}\left[\|\hat{b}-b^\star\|_{\LL}^2\right]$, and $B$, $D$, $G$ are the constants of \Cref{ass:bound_la,ass:gome}. The expectation is over all sources of randomness, namely the samples $\mathcal{D}_{in}$ and $\mathcal{D}_{out}$ used to construct $\hat{h}_\omega$ and $\hat{b}$, and the mini-batches $\{\mathcal{B}_g^k\}_{k\geq0}$.
\end{theorem}
The proof of \cref{thm:eta-scaled} is given in \cref{sub:sgd_biased}. It proceeds by controlling the bias and the variance of the hypergradient estimator in terms of $e_h$ and $e_b$ (\cref{lem:bias_cont,lem:var_control}), combining these bounds with a one-step descent inequality along the projected update (\cref{lem:onestep-envelope}), and telescoping over the iterations. The descent inequality is established for the \emph{Moreau envelope} of $\mathcal{F}$, whose gradient is equivalent to $\mathcal{G}_{1/\mathcal{L}}$ up to absolute constants in our case \citep[Section~2.2]{davis2019stochastic}. Note that the same bound holds for $\mathbb{E}[\|\mathcal{G}_{1/\mathcal{L}}(\omega_{k^\star})\|^2]$, where $k^\star$ is drawn uniformly at random in $\{0,\ldots,K-1\}$, so that the guarantee applies to an iterate \cref{alg:pmbsgd} can return.

\textbf{Error decomposition. }The bound decomposes into three terms of distinct origins. The first is the usual optimization error, and vanishes as the number of iterations $K$ grows. The second is a \emph{bias} term: it quantifies the two-stage error incurred by replacing the adjoint $a^\star_\omega$ with $-\hat{h}_\omega+\hat{b}$, and is unaffected by $K$, $\eta$ and $n_g$. The third is a \emph{variance} term, reflecting the sampling noise of the empirical average over the mini-batch. For a \emph{fixed} batch size $n_g$, it is proportional to $\eta$ and can therefore be reduced by decreasing the stepsize. The stepsize thus trades off the first and third terms, while leaving the bias unchanged. Choosing $\eta$ achieving the best trade-off makes the algorithmic part of the error decay at the rate $\sqrt{\mathcal{L}\Delta_\mathcal{F}/(Kn_g)}$ (\cref{cor:balanced}).

\textbf{Towards explicit rates for $e_h$ and $e_b$. }\Cref{thm:eta-scaled} is stated in terms of the generalization errors $e_h$ and $e_b$ alone, and is therefore agnostic to how $\hat{h}_\omega$ and $\hat{b}$ are obtained. These errors are \emph{statistical} in nature and decrease only as the two regression estimates improve, through larger sample sizes $\nin$ and $\nout$ or hypothesis spaces better suited to $h^\star_\omega$ and $b^\star$. We make them explicit for overparameterized two-layer ReLU neural networks trained by gradient flow, whose dynamics we describe in \cref{eq:flow_dyn} and whose generalization bounds we derive in the NTK regime in \cref{prop:gen_bounds_main}.

\section{Explicit statistical rate for \texorpdfstring{\cref{alg:pmbsgd}}{Algorithm~\ref{alg:pmbsgd}} in the NTK regime}\label{sec:exp_gen_ntk}
In this section, we build estimators for both targets as truncated neural networks trained by gradient flow, derive their finite-sample generalization rates in the NTK regime, and deduce the sample complexity of \cref{alg:pmbsgd}.

\subsection{Neural network estimators: training dynamics and truncation}
\label{sec:estimators}
\textbf{Architecture.} 
We construct estimators  $\hat{h}_\omega$ and $\hat{b}$ of the target functions $h^\star_\omega$ and $b^\star$ using  \emph{overparameterized} neural networks with a common architecture:  two-layer ReLU networks of the form:
\begin{equation}\label{eq:nn_main}
    f_\theta(x)=\frac{1}{\sqrt{m}}\sum_{i=1}^{m} v_i\,\sigma(\boldsymbol{u}_i^\top x+c_i)+a,
\end{equation}
where $\sigma(z)=\max(0,z)$ is the ReLU activation function, $m$ is the network's (even) width, and $\theta=(\boldsymbol{v}, \boldsymbol{u}, \boldsymbol{c}, a)$ collects the parameters. The first-layer weights are $\boldsymbol{u}\in\mathbb{R}^{m\times p}$ with corresponding biases $\boldsymbol{c}\in\mathbb{R}^m$, the second-layer weights are $\boldsymbol{v}\in\mathbb{R}^m$, and $a\in\mathbb{R}$ is the output bias. The width $m$ and the parameters $\theta$ above are generic: we write $m_h$ and $\theta^h$ for the width and parameters of the network approximating $h^\star_\omega$, and $m_b$ and $\theta^b$ for those of the network approximating $b^\star$. The scaling $1/\sqrt{m}$ allows reaching the NTK regime \citep{jacot2018neural} as $m$ grows and under the initialization discussed next.

\textbf{Initialization.} We set the initial output bias at $a=0$, while the rest of the parameters of both networks are drawn, independently of each other and of the data, from a standard Gaussian distribution using  a \emph{mirrored initialization} \citep{lai2023generalization,li2024eigenvalue}, to ensure the output is exactly zero initially. Specifically, for $1\leq i\leq m/2$, we draw $\boldsymbol{u}_i$, $c_i$ and $v_i$ independently from a standard Gaussian and set $\boldsymbol{u}_{i+m/2}=\boldsymbol{u}_i$, $c_{i+m/2}=c_i$ and   $v_{i+m/2}=-v_i$, so that the contributions of neurons $i$ and $i+m/2$ cancel. 
For the network approximating $h^\star_\omega$, the parameters are drawn once and shared by all $\omega\in\Omega$, so that their dependence on $\omega$ arises only through training. 

\textbf{Gradient flow training. }While neural networks are commonly trained with discrete-time gradient-based algorithms, we consider here their continuous-time limit, \emph{gradient flow}, for theoretical tractability \citep{elkabetz2021continuous}. Starting from the mirrored initialization, and for a given $\omega\in \Omega$, the parameters of both networks evolve in time according to the following ordinary differential equations:
\begin{equation}\label{eq:flow_dyn}
    \frac{\mathrm{d}}{\mathrm{d}t}\theta^h_{t,\omega}=-\partial_\theta \widehat{\mathcal{L}}_{in}(\omega, \theta^h_{t}),\qquad\frac{\mathrm{d}}{\mathrm{d}t}\theta^b_t=-\partial_\theta\widehat{\mathcal{L}}_{out}(\theta^b_t),
\end{equation}
where we defined the training objectives in terms of the network parameters: $\widehat{\mathcal{L}}_{in}(\omega,\theta^{h})\coloneqq\widehat{L}_{in}(\omega, f_{\theta^h})$ and $\widehat{\mathcal{L}}_{out}(\theta^{b})\coloneqq\widehat{L}_{out}(f_{\theta^b})$, while $\partial_\theta$ denotes the gradient with respect to the network parameters. We next introduce the NTK, which is needed later to define our final estimators. As shown in \citep{jacot2018neural}, the evolution of the predictor functions $f_{\theta_t^h}$ and $f_{\theta_t^b}$ is governed by the \emph{finite-width} NTKs, defined for any $x,x'\in\mathcal{X}$ by:
\begin{equation*}
    K^{m_h}_{t}(x,x')=\big(\partial_\theta f_{\theta_t^h}(x)\big)^\top\partial_\theta f_{\theta_t^h}(x'),\qquad K^{m_b}_{t}(x,x')=\big(\partial_\theta f_{\theta_t^b}(x)\big)^\top\partial_\theta f_{\theta_t^b}(x').
\end{equation*} 
As the widths $m_h$ and $m_b$ increase, the network's scaling and initialization choices ensure that both kernels $K_{t}^{m_h}$ and $K_t^{m_b}$ converge to a common deterministic, time-invariant \emph{infinite-width} limit \citep{jacot2018neural,AllenZhu2019Convergence,lee2019wide}, denoted here by $K^{\mathrm{NTK}}$, whose explicit expression is given in \cref{eq:explicit_ntk} of \cref{app:nna_iwdtr}. Consequently, the predictor's dynamics converge to those of a kernel regression trained by gradient flow, which we refer to as \emph{NTK regression}.

\textbf{Early stopping.} Since we consider an overparameterized setting in which the network's width can greatly exceed the number of training samples, the predictors $f_{\theta_t^h}$ and $f_{\theta_t^b}$ may interpolate the training data as $t\to\infty$. Although interpolation can still yield good generalization performance \citep{hastie2022surprises,belkin2019reconciling}, precisely characterizing generalization in this regime is beyond the scope of this work. Instead, we adopt an \emph{early stopping} strategy, halting the gradient flows at carefully chosen stopping times $t_{\mathrm{op}_h}$ and $t_{\mathrm{op}_b}$, which allows us to leverage techniques similar to those developed in \citep{li2024eigenvalue}. Early stopping acts as an \emph{implicit regularizer} \citep{raskutti2014early,yao2007early}, with the stopping time playing the role of a regularization parameter that controls the bias-variance trade-off and thereby improves generalization.

\textbf{Truncated estimators.} We could, in principle, use the early-stopped predictors as our final estimators. However, obtaining an explicit finite-sample convergence guarantee from \cref{thm:eta-scaled} requires bounds on their expected generalization errors. Existing results, such as \citep[Proposition~13]{li2024eigenvalue}, are not directly applicable: they provide high-probability bounds, holding with probability at least $1-\delta$, under width requirements that depend on the training samples and on the confidence level $\delta$, preventing the usual integration over $\delta$ to obtain an expectation bound for a fixed-width network. Instead, one would need to \emph{(i)} address the dependence of the required width on the training data and \emph{(ii)} control rare events on which the early-stopped predictors fall outside the high-probability guarantee and may potentially take arbitrarily large values. The former is addressed through a finer high-probability bound in \cref{prop:gen_bound}, which makes the required width a function of the sample size and $\delta$ only, while the latter is addressed by instead considering truncated estimators: 
\begin{align}\label{def:in_cond_trunc_main}
\hat{h}_{\omega}=f_{\theta_{t_{\mathrm{op}_h}}}\mathbbm{1}_{\mathcal{E}^h},\qquad\hat{b}=f_{\theta_{t_{\mathrm{op}_b}} }\mathbbm{1}_{\mathcal{E}^b},
\end{align}
where $\mathcal{E}^h$ and $\mathcal{E}^b$ depend only on the training data and initialization. These events are constructed to occur with high probability and ensure that, whenever they occur, the early-stopped predictors satisfy the required probabilistic bounds. The truncated estimators therefore coincide with the early-stopped networks on truncation events, inheriting their high-probability guarantees, and are set to zero on the low-probability complements, where no precise control over the early-stopped networks is available.

\textbf{Truncation events.} 
We construct each event $\mathcal{E}^{h}$ and $\mathcal{E}^{b}$ as the intersection of four explicit high-probability events depending on the data and initialization used to estimate $h^{\star}_{\omega}$ and $b^{\star}$. Since these events have the same structure for both estimators, we drop the superscript $h/b$ for simplicity and state them using the notation of \cref{eq:nn_main}, for a width-$m$ network trained on inputs $\{x_i\}_{i=1}^n$. The first three events depend only on the random initialization $\theta_0$, whereas the fourth depends only on the training inputs. For conciseness, we provide only a high-level description here and refer the reader to \cref{par:events} for the detailed definitions and discussion.
\begin{itemize}[nosep, leftmargin=*]
    \item\textbf{Small initial weights event $(\mathcal{E}_{\theta_0})$:} the event when initial weights satisfy: $ \Vert\theta_0\Vert_{\infty} \leq \sqrt{3\ln m}$.
    \item \textbf{Initial pre-activation margin event $(\mathcal{R})$:} the event when, uniformly over the input domain, only a small fraction $\mathcal{O}(m^\kappa)$ of initial pre-activations are within a prescribed margin of $0$, for $\kappa <1$.
    \item \textbf{Initial NTK concentration event $(\mathcal{C})$:} the event when, uniformly over a grid on the input domain, the finite-width NTK at initialization is within order $\sqrt{\ln m/m}$ of its infinite-width limit.
    \item \textbf{NTK non-degeneracy event $(\mathcal{E}_{\lambda_0})$:} the event when the Gram matrix of the infinite-width NTK on training data has its smallest eigenvalue $\lambda_0$ bounded below by a prescribed quantity $\lambda_\star(n,\delta)>0$.
\end{itemize}
Under $(\mathcal{E}_{\theta_0})$, it is possible to show that the gradients remain bounded along training, which helps keep the parameters in a controlled neighborhood of $\theta_0$. Under $(\mathcal{R})$, it can be shown that the signs of their pre-activations cannot change unless the parameters move far. This property is essential in proving that wide neural networks trained in the NTK regime remain close to their initialization \citep{jacot2018neural,AllenZhu2019Convergence}. $(\mathcal{C})$ is expected for large $m$, since the finite-width NTK is an average over the $m$ neurons of terms that are random at initialization and whose mean is $K^\mathrm{NTK}$. Finally, $(\mathcal{E}_{\lambda_0})$ guarantees that the gradient flow of the NTK regression converges on the training data.

\subsection{Finite-sample generalization bounds}
To state our generalization rates, we need a \emph{source condition}, \emph{i.e.}, a notion of regularity for the target functions $h^\star_\omega$ and $b^\star$. A natural way to quantify this regularity is through how well these targets can be approximated by elements of the RKHS $\mathcal{H}^\mathrm{NTK}$ associated with the limiting kernel $K^\mathrm{NTK}$. Indeed, in the infinite-width limit, the  predictors used to approximate the targets are elements of $\mathcal{H}^\mathrm{NTK}$. Rather than requiring these targets to also belong to $\mathcal{H}^\mathrm{NTK}$, we will instead assume they lie in an \emph{interpolation space} between $\LL$ and $\mathcal{H}^\mathrm{NTK}$, as this allows capturing intermediate regularity properties. To formally define these spaces, we introduce the integral operator $\mathcal{T}:\LL\to\LL$ associated with $K^\mathrm{NTK}$, given for every $f\in\LL$ and $x\in\mathcal{X}$ by:
\begin{equation*}
    (\mathcal{T}f)(x)=\mathbb{E}_{X\sim\mathbb{P}_X}\left[K^\mathrm{NTK}(x,X)f(X)\right].
\end{equation*}
Since $K^\mathrm{NTK}$ is a Mercer kernel, $\mathcal{T}$ is compact, self-adjoint, and positive semi-definite. Hence, by the spectral theorem \citep[Theorem~A.5.13]{steinwart2008support}, $\mathcal{T}$ admits a spectral decomposition with non-increasing eigenvalues $(\lambda_i)_{i\geq1}$ and corresponding orthonormal eigenfunctions $(e_i)_{i\geq1}$ forming a basis of $\LL$. We then define the interpolation space $[\mathcal{H}^\mathrm{NTK}]^s$ with parameter $s\geq 0$ as the subspace of functions $f \in\LL$ such that $\|f\|_{[\mathcal{H}^\mathrm{NTK}]^s}<\infty$, where $\|\cdot\|_{[\mathcal{H}^\mathrm{NTK}]^s}$ is the interpolation norm defined as the sum of the squared projections onto each $e_i$, weighted by $\lambda_i^{-s}$, \emph{i.e.}, $\|f\|^2_{[\mathcal{H}^\mathrm{NTK}]^s}=\sum_{i\geq 1}\lambda_i^{-s} \langle f,e_i\rangle^2_{\LL}$\footnote{In our setting, the norm $\|f\|_{[\mathcal{H}^\mathrm{NTK}]^s}$ is equivalent to $\big(\sum_{i\geq 1}i^{s(p+1)/p}\langle f,e_i\rangle_{\LL}^2\big)^{1/2}$, since the eigenvalues decay at a rate exactly equal to $\frac{p+1}{p}$ (\cref{lem:edr_set}).}. We now give the source conditions needed to establish generalization bounds.
\begin{assumplistm}[resume]
\item\label{ass:bound_tar} (Source condition for $h^\star_\omega$) There exist $s_h,R_h>0$ such that $\|h^\star_\omega\|_{[\mathcal{H}^\mathrm{NTK}]^{s_h}}\leq R_h$ for every $\omega\in\Omega$, hence in particular at every iterate of \cref{alg:pmbsgd}.
\item\label{ass:adj_source} (Source condition for $b^\star$) There exist $s_b,R_b>0$ such that $\|b^\star\|_{[\mathcal{H}^\mathrm{NTK}]^{s_b}}\leq R_b$.
\end{assumplistm}
\cref{ass:bound_tar,ass:adj_source} are standard source conditions in kernel regression \citep{lin2020optimal,caponnetto2007optimal,fischer2020sobolev}, with $s_h$ and $s_b$ controlling the regularity of $h^\star_\omega$ and $b^\star$ relative to $\mathcal{H}^\mathrm{NTK}$: larger values correspond to smoother, easier to learn targets. In addition, \cref{ass:bound_tar} requires this regularity bound to hold uniformly over $\omega$, ensuring that the difficulty of estimating $h^\star_\omega$ remains controlled throughout the optimization. When $\mathcal{X}$ is a bounded Lipschitz domain, such as a ball or a cube, we show in \cref{app:sob} that $[\hntk]^s$ coincides with the Sobolev space $H^{\frac{s(p+1)}{2}}(\mathcal{X})$ for $0\leq s\leq 1$. Thus, when $s_h,s_b\leq1$, \cref{ass:bound_tar,ass:adj_source} can be expressed as Sobolev regularity conditions, uniformly over $\omega\in\Omega$ for $h^\star_\omega$.

We can now derive generalization bounds for the estimators $\hat{h}_{\omega}$ and $\hat{b}$ from \cref{sec:estimators}.
\begin{proposition}[Expected generalization bounds]\label{prop:gen_bounds_main}
    Assume $p\geq 2$ and let $\Omega\subset\mathbb{R}^d$ be compact. Let \Cref{ass:bound_in,ass:bound_la,ass:adj_source,ass:bound_tar,ass:gome,ass:bounded_density_P} hold with $s_h,s_b\in(\frac{1}{p+1},2]$, and choose the stopping times $t_{\mathrm{op}_h}=n_{in}^{\gamma_h}$ and $t_{\mathrm{op}_b}=n_{out}^{\gamma_b}$, with $\gamma_h=\frac{p+1}{s_h(p+1)+p}$ and $\gamma_b=\frac{p+1}{s_b(p+1)+p}$. There exist two polynomials $\mathrm{poly}_h$ and $\mathrm{poly}_b$ such that, when $n_{in}$ and $n_{out}$ are sufficiently large and the widths satisfy $m_h\geq\mathrm{poly}_h(n_{in},\ln n_{in})$ and $m_b\geq\mathrm{poly}_b(n_{out},\ln n_{out})$, the truncated predictors of \cref{def:in_cond_trunc_main} satisfy:
    \begin{equation*}
    \mathbb{E}\left[\sup_{\omega\in\Omega}\left\|\hat{h}_{\omega}-h^\star_\omega\right\|_{\LL}^2\right]\lesssim d^4(\ln n_{in})^4 n_{in}^{-s_h\gamma_h},\qquad \mathbb{E}\left[\left\|\hat{b}-b^\star\right\|_{\LL}^2\right]\lesssim(\ln n_{out})^4 n_{out}^{-s_b\gamma_b},
    \end{equation*}
    where $\lesssim$ hides constants independent of $n_{in}$, $n_{out}$ and $d$, and the expectations are taken over the training samples and the random initialization.
\end{proposition}
\Cref{prop:gen_bounds_main} shows that the truncated predictors achieve the minimax-optimal rates of the corresponding regression problems, up to logarithmic factors \citep{blanchard2018optimal,JMLR:v25:23-0383}. We defer the detailed proof, including the explicit dependence of the constants and polynomial coefficients on the problem parameters, to \cref{subsec:exp_gen,subsec:gen_reg_exp}. The proof first shows that, with high probability, sufficiently wide networks remain uniformly close to their NTK regression counterparts, and then relies on truncation to turn these high-probability guarantees into bounds in expectation. Compared with existing results in the NTK literature \citep{li2024eigenvalue,lai2023generalization,chen2024impacts}, the main challenge is to obtain a bound that holds uniformly over $\omega\in\Omega$. This requires deriving bounds with constants independent of $\omega$ and combining a covering argument over $\Omega$ with the Lipschitz continuity of the NTK regression error in $\omega$ (\cref{lem:lipschitz-omega}).

The above bounds are precisely the quantities $e_h$ and $e_b$ arising in \cref{thm:eta-scaled}. Thus combining both the convergence result in \cref{thm:eta-scaled} with the generalization bounds in \cref{prop:gen_bounds_main} yields the following explicit convergence result for the iterates produced by \cref{alg:pmbsgd} when using the truncated estimators in \cref{def:in_cond_trunc_main}. 
\begin{corollary}[Sample complexity of \cref{alg:pmbsgd}]\label{cor:end_to_end}
Let   \Cref{ass:bound_in,ass:bound_la,ass:adj_source,ass:bound_tar,ass:gome,ass:bounded_density_P} hold and consider the iterates $\omega_k$ obtained from   \cref{alg:pmbsgd} using stepsize $0<\eta\leq\frac{1}{2\mathcal{L}}$,   gradient estimator in \cref{eq:gradient_est} and truncated predictors in \cref{def:in_cond_trunc_main} with width parameters, sample sizes and stopping times as in \cref{prop:gen_bounds_main}. Then, for any $K\geq 1$, the iterates $\{\omega_k\}_{k\geq0}$ satisfy:
\begin{equation*}
    \frac{1}{K}\sum_{k=0}^{K-1}\mathbb{E}\Big[\big\|\mathcal{G}_{1/\mathcal{L}}(\omega_k)\big\|^2\Big]\lesssim \frac{\Delta_\mathcal{F}}{\eta K}+ \frac{\mathcal{L}G^2\eta(D+B)^2}{n_g}+ G^2\Big(d^4(\ln n_{in})^4\nin^{-s_h\gamma_h} + (\ln n_{out})^4\nout^{-s_b\gamma_b}\Big),
\end{equation*}
where $\lesssim$ hides constants which are independent of $K$, $\eta$, $n_g$, $\nin$, $\nout$ and $d$.
\end{corollary}
In \cref{cor:end_to_end}, the statistical error is dominated by the slower of the two rates $\nin^{-s_h\gamma_h}$ and $\nout^{-s_b\gamma_b}$. This suggests balancing the two sample sizes so that $\nin^{-s_h\gamma_h}\approx \nout^{-s_b\gamma_b}$ up to constant and logarithmic factors, ensuring that neither level becomes a bottleneck. Moreover, since the exponents $s_h\gamma_h$ and $s_b\gamma_b$ increase with $s_h$ and $s_b$, respectively, the level whose target has the larger smoothness parameter needs fewer samples to reach the same accuracy.

\section{Related work}\label{sec:rel_work}
\citet{elkhoury2025learning} and \citet{chen2025towards} study generalization in bilevel nonparametric settings: the former uses \emph{full-batch gradient descent} to optimize the inner level over an RKHS, while the latter relies on \emph{mean-field Langevin dynamics} for two-stage least squares. Both works assume a \emph{fixed} regularization parameter to control the bias-variance trade-off, leaving open the question of how the optimal choice of the regularization, as a function of the sample sizes, would affect sample complexity and whether it could lead to faster rates. In contrast, in our setting, this trade-off is achieved by the early stopping time, which we select as a function of the sample sizes, yielding explicit finite-sample bounds that quantify the sample complexity of the problem. Relatedly, \citet{patel2025solving} provide convergence guarantees for training overparameterized neural networks in \emph{min-max (zero-sum) games} in the NTK regime, which can be viewed as a particular bilevel structure. Our work differs in that we study bilevel learning and generalization, and quantify how stochastic, approximate NTK training affects hypergradient accuracy and excess risk rather than game-theoretic optimization dynamics. Also, \citet{kim2025optimality} analyze the deep feature instrumental variable regression, which shares structural similarities with our bilevel setup. They establish minimax optimal convergence rates when the target function belongs to a Besov space. However, their analysis focuses on statistical aspects, whereas we also account for optimization. Finally, \citet{elkhoury2026parametric} study the same problem as ours and \emph{debias} the hypergradient to attain \emph{parametric} rates, whereas we derive generalization bounds for the plug-in estimator with neural networks in the NTK regime.

\section{Numerical experiments}\label{sec:num_exp}

\subsection{Empirical validation of the theory in the NTK regime on synthetic data}
\begin{figure}[t]
  \centering
  \includegraphics[width=0.86\textwidth]{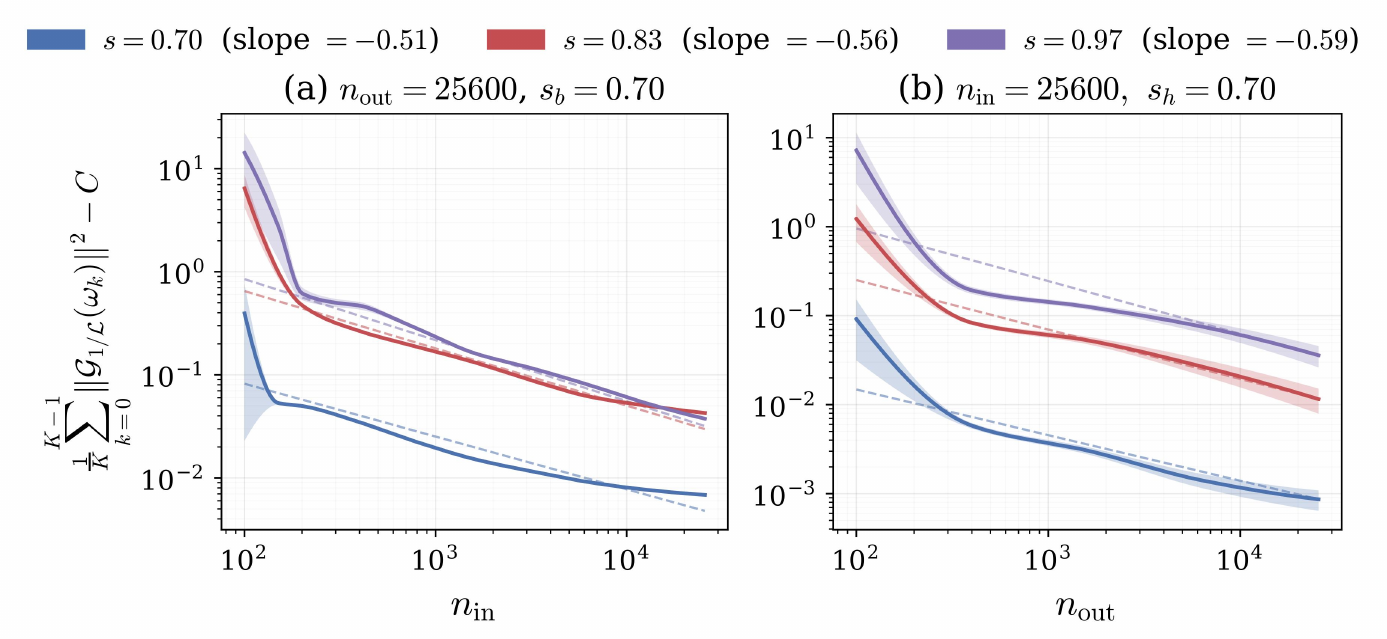}
  \caption{Illustration of \cref{alg:pmbsgd} on synthetic data with targets of known smoothness. Dashed lines show the theoretical slopes. The curves are averaged over 5 runs and displayed on a log scale. Solid lines represent the mean across runs, and shaded regions indicate the 95\% confidence intervals.}
  \label{fig:validate-rates}
\end{figure}

We empirically validate the convergence bound of \cref{cor:end_to_end} on targets whose smoothness is known in closed form. We consider inputs $x\in[0,1]^2$, outer parameters $\omega\in\mathbb{R}^4$, and cusp targets $h^\star_\omega(x)=\psi(x)^\top\omega$ and $b^\star(x)=\|x-x_0\|^{q_b}$, where $\psi(x)\coloneqq[\|x-c_1\|^{q_h},\ldots,\|x-c_4\|^{q_h}]^\top$, the centers $c_1,\ldots,c_4$ form the grid $\{\frac{1}{4},\frac{3}{4}\}^2$, and $x_0=(\frac{1}{2},\frac{1}{2})$. By \citep[Section~2.3.1,~Lemma~1]{RunstSickel+1996}, $h^\star_\omega$ and $b^\star$ satisfy the source conditions uniformly over $\omega$ for every $s_h<\bar s_h\coloneqq(2q_h+2)/3$ and $s_b<\bar s_b\coloneqq(2q_b+2)/3$, with $q_h,q_b\in\{0.05,0.25,0.45\}$, giving $\bar s_h,\bar s_b\in\{0.70,0.83,0.97\}$ and decay rates of $0.51$, $0.56$, and $0.59$, respectively. Both targets are approximated by networks of the form in \cref{eq:nn_main} trained by full-batch gradient descent. We follow \cref{alg:pmbsgd} with hypergradients evaluated on fresh batches of size $200$. The sample sizes $\nin$ and $\nout$ vary independently from $100$ to $25{,}600$. Further details about the setup are given in \cref{sec:validate-ntk}. \cref{fig:validate-rates} plots the near-stationarity measure averaged over the outer iterations against $\nin$ or $\nout$, minus a constant $C$ collecting the terms of the bound in \cref{cor:end_to_end} that are \emph{independent} of the varied sample size. In panel~(a), $\nin$ varies for $s_h\in\{0.70,0.83,0.97\}$, with $\nout=25{,}600$ and $s_b=0.70$; panel~(b) swaps the inner and outer roles. The curves follow the slopes predicted by \cref{cor:end_to_end}, thus validating our theory.

\subsection{Insights into feature learning beyond the NTK regime}

\begin{figure}[t]
    \centering
    \begin{subfigure}[t]{0.484\textwidth}
        \centering
        \includegraphics[width=0.9\textwidth]{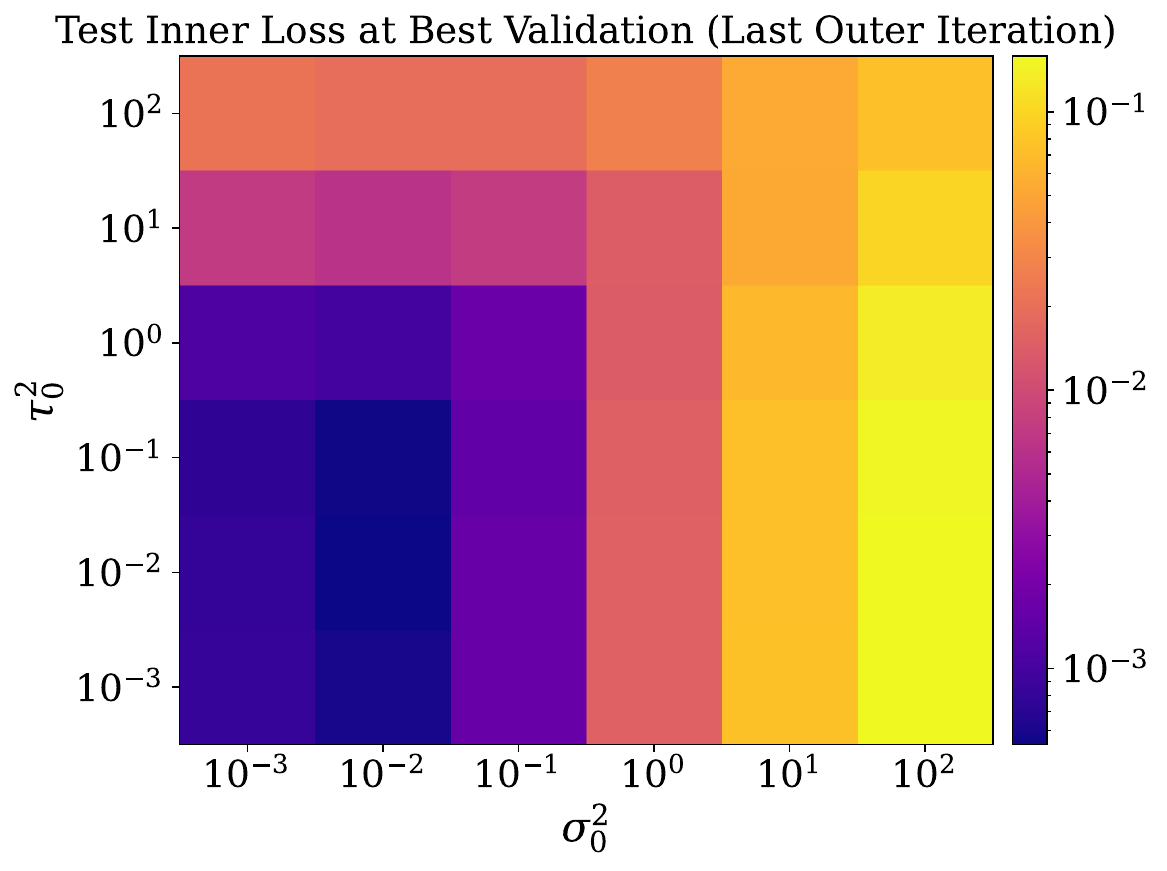}
    \end{subfigure}
    \hfill
    \begin{subfigure}[t]{0.484\textwidth}
        \centering
        \includegraphics[width=0.9\textwidth]{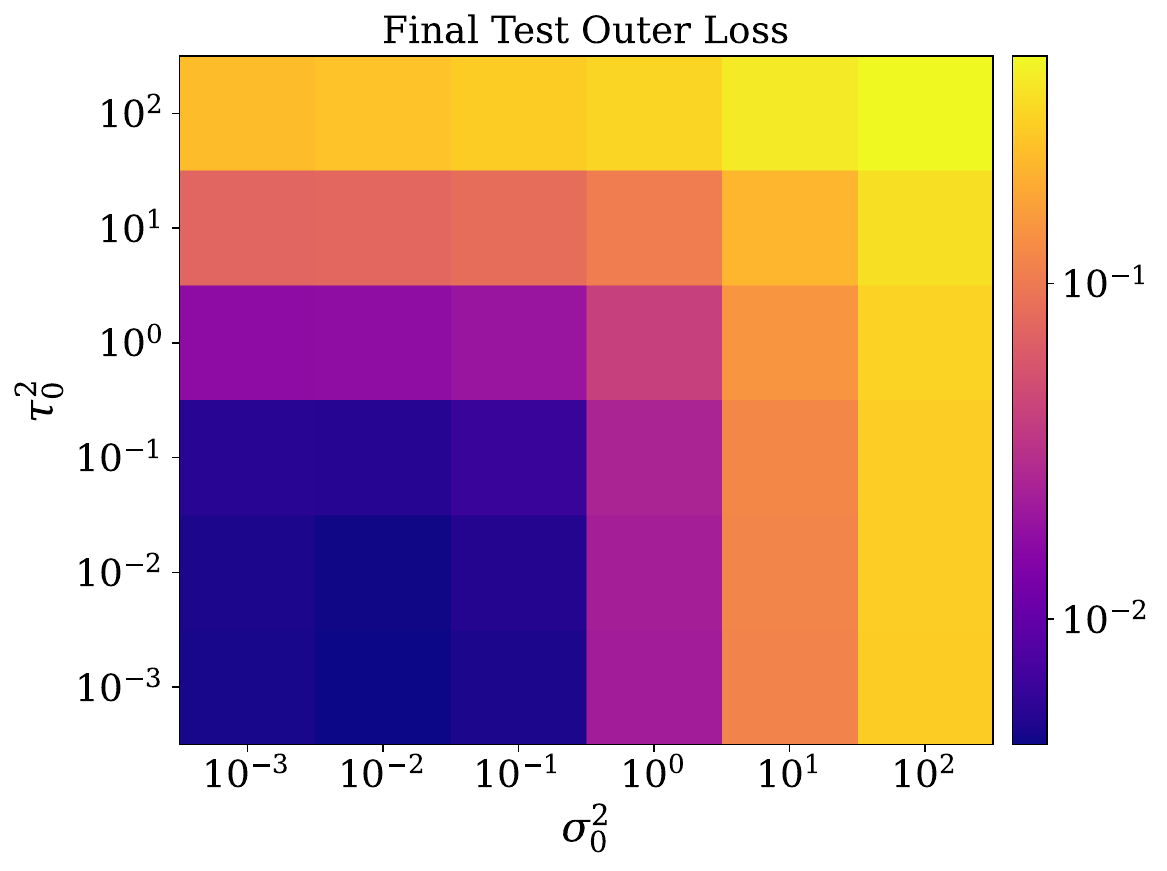}
    \end{subfigure}
    \caption{Test inner (left) and outer (right) losses at the last outer iteration as a function of $(\sigma_0^2, \tau_0^2)$: both losses are lowest when both initializations are small (feature learning) and degrade as either grows large (NTK). The heatmaps are averaged over 5 runs and displayed on a log scale.}
    \label{fig:heatmaps}
\end{figure}

Our analysis relies on the NTK regime, which assumes fixed features and thus precludes \emph{feature learning}. For a single network, the initialization scale governs the transition between the NTK and feature learning regimes \citep{woodworth2020kernel}. In the bilevel setting, the inner and outer initializations may \emph{jointly} determine the regime. We investigate this insight, which lies beyond the scope of our theory, on a synthetic problem in which $h^\star_\omega$ and $b^\star$ are estimated by two-layer networks of the form in \cref{eq:nn_main} normalized by $m$ instead of $\sqrt{m}$ to allow feature learning \citep{chizat2019lazy}, with first-layer initialization variances $\sigma_0^2$ and $\tau_0^2$, respectively. \cref{fig:heatmaps} shows that both the test inner and outer losses are lowest when both initializations are small (feature learning) and degrade as either grows (NTK), indicating a \emph{joint effect} of the two initializations on the effective regime. The full setup and additional results, showing that this ranking holds along the outer iterations and across sample sizes, are given in \cref{sec:beyond-ntk}.

\section{Conclusion and perspectives}\label{sec:con}
\textbf{Summary. }In this paper, we provided a general convergence guarantee for inexact bilevel gradient methods when both objectives are least-squares regressions. The resulting bound depends on the estimators of $h^\star_\omega$ and $b^\star$ only through their uniform generalization errors. In the NTK regime, we derived finite-sample bounds on these errors when both are estimated by overparameterized two-layer ReLU networks trained by gradient flow with early stopping. Plugging these rates into the general bound yields an explicit convergence guarantee for \cref{alg:pmbsgd}. Our experimental results suggest that, beyond the NTK regime, the bilevel formulation introduces an implicit bias that promotes feature learning and leads to improved performance.

\textbf{Limitations and future work.} Our theoretical analysis does not address minimax optimality, and establishing minimax rates for~\eqref{eq:obo} remains an open problem. While the transition between the NTK and feature learning regimes is well studied for single networks \citep{woodworth2020kernel, chizat2019lazy, yang2021tensor}, its theoretical characterization in the bilevel setting is still unexplored, which could perhaps be approached through the lens of \emph{grokking}~\citep{power2022grokking}. Indeed, this phenomenon has been interpreted as a transition between the NTK and feature learning regimes \citep{lyu2023dichotomy,kumar2023grokking,mohamadi2024you}, potentially offering an intermediate trade-off. Further directions include extending our analysis to deeper architectures, relaxing the polynomial width requirement typical of the NTK regime, and considering general losses, for which the adjoint function no longer reduces to two independently estimable regression functions.

\begin{ack}
FEK would like to thank Elie Louis for helpful discussions related to this paper. This work was supported by the ANR project BONSAI (grant ANR-23-CE23-0012-01). EP acknowledges funding from the AI Interdisciplinary Institute ANITI through the French Investments for the Future – PIA3 program under grant agreement ANR-19-PI3A-0004; the Air Force Office of Scientific Research, Air Force Materiel Command, USAF under grant number FA8655-22-1-7012; TSE-P; the ANR project Chess (grant ANR-17-EURE-0010); the ANR project Regulia; and the Institut Universitaire de France (IUF). EP and SV acknowledge support from the ANR project MAD. SV also thanks the PEPR PDE-AI program (ANR-23-PEIA-0004) and the 3IA Côte d’Azur Chair BOGL.
\end{ack}

\bibliography{references}
\bibliographystyle{abbrvnat}

\medskip


\newpage
\appendix

\appendix
\crefalias{section}{appendix}
\crefalias{subsection}{appendix}
\crefalias{subsubsection}{appendix}

\crefname{appendix}{Appendix}{Appendices}
\Crefname{appendix}{Appendix}{Appendices}

\textbf{\LARGE Appendices}
\vspace{0.2cm}

\textbf{Roadmap. }\cref{app:examples} presents concrete instances of \eqref{eq:obo}, namely instrumental variable regression and value function fitting in reinforcement learning. \cref{subapp:ass_obo} discusses \cref{ass:bound_in,ass:bound_la,ass:gome,ass:bounded_density_P}. \cref{sub:sgd_biased} establishes convergence guarantees for projected mini-batch biased SGD applied to the empirical approximation of \eqref{eq:obo}. Within it, \cref{app:hypergradient} derives the hypergradient and the smoothness of the value function, \cref{sec:stat_prop_grad_est} analyzes the statistical properties of the hypergradient estimator, and \cref{sec:moreau-alternative} proves the main convergence theorem. These guarantees in the NTK regime rely on the generalization bounds for \eqref{eq:obo} derived in \cref{app:genres}. Those bounds build on \cref{sec:ntk_single}, which establishes non-asymptotic generalization bounds for neural network gradient flow in the NTK regime and provides the theoretical foundation used to approximate both the inner solution~$h^\star_\omega$ and the regression function~$b^\star$. Differentiability of the objectives and properties of the optimal solutions are provided in \cref{sec:diff_ob_op_sol}. \cref{app:sob} shows that the RKHS of the infinite-width NTK and its interpolation spaces are Sobolev spaces. Background on integral and covariance operators as well as interpolation spaces is given in \cref{sec:int_space}. Auxiliary technical lemmas used throughout the proofs are collected in \cref{sec:aux_lem}. Finally, further experimental details and additional numerical results are provided in \cref{sec:exp_det}.

\textbf{Notations. }$\langle\cdot,\cdot\rangle$ and $\|\cdot\|$ denote the Euclidean inner product and its induced norm, $\langle\cdot,\cdot\rangle_{\mathcal{H}}$ and $\|\cdot\|_{\mathcal{H}}$ denote the inner product and norm of a Hilbert space $\mathcal{H}$, $\|\cdot\|_\mathrm{op}$ denotes the operator norm, $\|\cdot\|_\mathrm{F}$ denotes the Frobenius norm, and $\|\cdot\|_\mathrm{HS}$ denotes the Hilbert-Schmidt norm. $\odot$ denotes the Hadamard product. For any $v_1,\ldots,v_p\in\mathbb{R}$, $\mathbf{diag}(v_1,\ldots,v_p)\in\mathbb{R}^{p\times p}$ denotes the diagonal matrix of size $p\times p$, where the diagonal entries are $v_1,\ldots,v_p$ and all the off-diagonal entries are 0. $\mathbbm{1}_p$ denotes a vector of size $p$ where all entries are 1. $\mathbbm{1}_{p\times p}$ denotes the identity matrix of size $p$. For $x\in\mathbb{R}$, $\lfloor x\rfloor$ denotes the largest integer less than or equal to $x$, and $\lceil x\rceil$ denotes the smallest integer greater than or equal to $x$. For any event $\mathcal{E}$, $\mathbbm{1}_\mathcal{E}$ denotes the indicator function of $\mathcal{E}$. Given a compact set $\mathcal{K}$, $\mathrm{diam}(\mathcal{K})$ denotes its diameter. $\mathrm{Id}_{\mathcal{H}}$ denotes the identity operator on a Hilbert space $\mathcal{H}$. For a set $\mathcal{A}$, $\mathcal{A}^{\complement}$ denotes its complement. $\mathcal{H}_1\cong\mathcal{H}_2$ means $\mathcal{H}_1=\mathcal{H}_2$ as sets with equivalent norms. $a\asymp b$ means that there exist constants $c,C>0$ such that $cb\leq a\leq Cb$.
\addtocontents{toc}{\protect\setcounter{tocdepth}{2}}
{
    \hypersetup{linkcolor=black}
    \tableofcontents
}

\section{Concrete Examples of (\ref{eq:obo})}\label{app:examples}

\textbf{Instrumental variable regression.} It is an approach that handles endogeneity in statistical modeling, relying on instruments to identify causal relationships \citep{newey2003instrumental}. The aim is to learn a function $z\mapsto g_\omega(z)$, parameterized by a vector $\omega$, satisfying $y=g_\omega(z)+\epsilon$, where $y\in\mathcal{Y}$ denotes the observed outcome, $z\in\mathcal{Z}$ the treatment, and $\epsilon$ the error term. The central difficulty lies in the endogeneity of $z$, namely its correlation with $\epsilon$, which renders direct regression inconsistent. Exogeneity no longer holds, since $z$ and $\epsilon$ are no longer independent, and the estimates of $g_\omega$ are distorted as a result. A remedy is to bring in an instrumental variable $x\in\mathcal{X}$, correlated with $z$ but not with $\epsilon$, which retrieves the link between $y$ and $z$ while remaining immune to the distortion carried by $\epsilon$, classically through two-stage least squares regression \citep{angrist1995two}. Following \citep{xu2021learning}, this formulation translates directly into a bilevel problem whose inner and outer objectives are those given in \cref{eq:squared_losses}, where $h$ is taken in $L^2(\mathcal{X})$.

\textbf{Value function fitting in reinforcement learning.} A second instance arises in reinforcement learning. Let $x=(s,a)$ be a state--action pair and $z=(s,a,r,s')$ the transition it generates, where $r$ is the reward received and $s'$ is the successor state. Let $g_\omega(z)=r+\gamma V_\omega(s')$ be the one-step Bellman target associated with a parametric value function $V_\omega$, where $\gamma\in(0,1)$ is the discount factor. This target is random given $x$, since it depends on which reward and successor state are drawn. The inner problem averages it over that randomness, returning the Bellman backup $h^\star_\omega(s,a)=\mathbb{E}_{r,s'}[r+\gamma V_\omega(s')\mid s,a]$, which fitted value iteration estimates by regressing the target on $(s,a)$ \citep{munos2008finite}. The outer problem then fits this backup to the discounted return $y$ observed along a trajectory started at $(s,a)$, whose conditional expectation given $(s,a)$ is the true action-value function. Minimizing $\mathcal{F}$ therefore selects the value function whose averaged one-step target $h^\star_\omega(s,a)$ is closest, in $L^2(\mathcal{X})$, to the observed return $y$.
\section{Discussion of \texorpdfstring{\protect\cref{ass:bound_in,ass:bound_la,ass:gome,ass:bounded_density_P}}{Assumptions}}\label{subapp:ass_obo}

\cref{ass:bound_in} is mild and standard in practice, as inputs are naturally bounded in most applications, \emph{e.g.}, images and physical measurements, or can be made so through preprocessing procedures such as normalization or clipping. \cref{ass:bound_la} is common in supervised learning, as the output naturally takes values in a bounded range in most regression tasks, \emph{e.g.}, probabilities in $[0,1]$, normalized scores, or physical measurements on a known scale, and can always be enforced by output normalization otherwise. \cref{ass:gome} holds in practice for a broad class of models, including linear models with bounded features, \emph{e.g.}, random Fourier features \citep{rahimi2007random}, when the parameter space is compact. It also applies to feedforward neural networks or transformers with smooth activations, \emph{e.g.}, SiLU \citep{ramachandran2017searching} or GeLU \citep{hendrycks2016gelu}, provided that the parameter space and the input domain are compact. \cref{ass:bounded_density_P} is a standard regularity condition on the input data distribution. Together with \cref{ass:bound_in}, it is satisfied, for instance, by the uniform distribution on any compact set with nonempty interior, by Gaussian distributions truncated to a ball, and by finite mixtures of such distributions. It excludes, however, distributions without a Lebesgue density, such as discrete inputs (\emph{e.g.}, binary or one-hot features) and distributions concentrated on lower-dimensional sets (\emph{e.g.}, the uniform distribution on the unit sphere $\mathbb{S}^{p-1}$, or compositional data constrained to the simplex), as well as densities with singularities (\emph{e.g.}, the arcsine distribution $\mathrm{Beta}(\tfrac12,\tfrac12)$ on $[0,1]$, whose density diverges at both endpoints).
\section{Convergence of Projected Mini-Batch Biased SGD}\label{sub:sgd_biased}

\subsection{Hypergradient and smoothness property of the value function}\label{app:hypergradient}

\begin{proof}[Proof of \cref{prop:hyper}]
    The differentiability of $\mathcal{F}$ as well as the expression of its hypergradient $\nabla\mathcal{F}$ are a direct consequence of \citep[Proposition~2.2]{petrulionyte2024functional} together with \cref{prop:diff_lout,prop:diff_lin}. Next, we prove the smoothness property of $\mathcal{F}$. Let $\omega_1,\omega_2\in U$. We have:
    \begin{align}
        &\left\|\nabla\mathcal{F}(\omega_1)-\nabla\mathcal{F}(\omega_2)\right\|\nonumber\\
        &=\left\|\mathbb{E}_\mathbb{P}\left[\partial_\omega g_{\omega_1}(z)a^\star_{\omega_1}(x)-\partial_\omega g_{\omega_2}(z)a^\star_{\omega_2}(x)\right]\right\|\nonumber\\
        &\leq\mathbb{E}_\mathbb{P}\left[\left\|\partial_\omega g_{\omega_1}(z)a^\star_{\omega_1}(x)-\partial_\omega g_{\omega_2}(z)a^\star_{\omega_2}(x)\right\|\right]\nonumber\\
        &=\mathbb{E}_\mathbb{P}\left[\left\|\partial_\omega g_{\omega_1}(z)(a^\star_{\omega_1}(x)-a^\star_{\omega_2}(x))+(\partial_\omega g_{\omega_1}(z)-\partial_\omega g_{\omega_2}(z))a^\star_{\omega_2}(x)\right\|\right]\nonumber\\
        &\leq\mathbb{E}_\mathbb{P}\left[\left\|\partial_\omega g_{\omega_1}(z)\right\|\left|a^\star_{\omega_1}(x)-a^\star_{\omega_2}(x)\right|+\left\|\partial_\omega g_{\omega_1}(z)-\partial_\omega g_{\omega_2}(z)\right\|\left|a^\star_{\omega_2}(x)\right|\right]\nonumber\\
        &=\mathbb{E}_\mathbb{P}\left[\left\|\partial_\omega g_{\omega_1}(z)\right\|\left|h^\star_{\omega_1}(x)-h^\star_{\omega_2}(x)\right|+\left\|\partial_\omega g_{\omega_1}(z)-\partial_\omega g_{\omega_2}(z)\right\|\left|a^\star_{\omega_2}(x)\right|\right]\label{eq:sm_F},
    \end{align}
    where, in the third line, we use Jensen's inequality, and in the last line, we use $a^\star_{\omega_1}-a^\star_{\omega_2}=-(h^\star_{\omega_1}-h^\star_{\omega_2})$, the function $b^\star$ being independent of $\omega$. By \Cref{ass:gome}, $\sup_{\omega\in U,z\in\mathcal{Z}}\|\partial_\omega g_\omega(z)\|\leq G$ and $\omega\mapsto g_\omega(z)$ is $L$-smooth for any $z\in\mathcal{Z}$, that is, $\|\partial_\omega g_{\omega_1}(z)-\partial_\omega g_{\omega_2}(z)\|\leq L\|\omega_1-\omega_2\|$. Furthermore, by \cref{lem:bound_targets}, $\|a^\star_{\omega_2}\|_\infty\leq\sup_{\omega\in U}\|a^\star_\omega\|_\infty\leq D+B$. Combining these bounds in \cref{eq:sm_F}, we obtain:
    \begin{align*}
        \left\|\nabla\mathcal{F}(\omega_1)-\nabla\mathcal{F}(\omega_2)\right\|
        &\leq G\,\mathbb{E}_\mathbb{P}\left[\left|h^\star_{\omega_1}(x)-h^\star_{\omega_2}(x)\right|\right]+L(D+B)\left\|\omega_1-\omega_2\right\|\\
        &=G\left\|h^\star_{\omega_1}-h^\star_{\omega_2}\right\|_{L^1(\mathcal{X})}+L(D+B)\left\|\omega_1-\omega_2\right\|\\
        &\leq G\left\|h^\star_{\omega_1}-h^\star_{\omega_2}\right\|_{\LL}+L(D+B)\left\|\omega_1-\omega_2\right\|\\
        &\leq\underbrace{\left(G^2+L(D+B)\right)}_{\mathcal{L}}\left\|\omega_1-\omega_2\right\|,
    \end{align*}
    where, in the third line, we use $\|\cdot\|_{L^1(\mathcal{X})}\leq\|\cdot\|_{\LL}$, valid since $\mathbb{P}_X$ is a probability measure, and where $L^1(\mathcal{X})$ denotes the space of integrable functions on $\mathcal{X}$ with respect to $\mathbb{P}_X$, endowed with the norm $\|f\|_{L^1(\mathcal{X})}=\mathbb{E}_{x\sim\mathbb{P}_X}[|f(x)|]$. In the last line, we use the fact that $\omega\mapsto h^\star_\omega$ is $G$-Lipschitz continuous, as established in \cref{prop:lip_con_hstar}. This concludes the proof.
\end{proof}

\subsection{Statistical properties of the hypergradient estimator}\label{sec:stat_prop_grad_est}

\begin{lemma}[Conditional mean of the hypergradient estimator]\label{lem:oracle_gradient}
    Fix $\omega\in\Omega$. Let $\hat{h}_\omega$ and $\hat{b}$ be any estimators of the inner solution $h^\star_\omega$ and of the regression function $b^\star$, built from $\mathcal{D}_{in}$ and $\mathcal{D}_{out}$, and set $\hat{a}_\omega=-\hat{h}_\omega+\hat{b}$. Then:
    \begin{equation*}
        \widetilde{\nabla\mathcal{F}}(\omega)\coloneqq\mathbb{E}\left[\widehat{\nabla\mathcal{F}}(\omega)\;\Big|\;\hat{a}_\omega\right]=-\mathbb{E}_\mathbb{P}\left[\partial_\omega g_\omega(z)\hat{a}_\omega(x)\right],
    \end{equation*}
    where $\mathbb{E}_\mathbb{P}$ integrates over $(x,z)\sim\mathbb{P}_{XZ}$ only, $\hat{a}_\omega$ being held fixed.
\end{lemma}
\begin{proof}
    Let $\omega\in\Omega$. By \cref{eq:gradient_est}, we have:
    \begin{equation*}
        \widehat{\nabla\mathcal{F}}(\omega)=-\frac{1}{n_g}\sum_{l=1}^{n_g}\partial_\omega g_\omega(\underline{z}_l)\hat{a}_\omega(\underline{x}_l),
    \end{equation*}
    where $\mathcal{B}_g=\{(\underline{x}_l,\underline{z}_l)\}_{l=1}^{n_g}$ is the mini-batch used at $\omega$. Since $\hat{a}_\omega$ is built from $\mathcal{D}_{in}$ and $\mathcal{D}_{out}$, which are independent of $\mathcal{B}_g$, the pairs $(\underline{x}_l,\underline{z}_l)$ remain \emph{i.i.d.} with law $\mathbb{P}_{XZ}$ conditionally on $\hat{a}_\omega$. Hence, each summand has the same conditional expectation:
    \begin{equation*}
        \mathbb{E}\left[\partial_\omega g_\omega(\underline{z}_l)\hat{a}_\omega(\underline{x}_l)\;\Big|\;\hat{a}_\omega\right]=\mathbb{E}_\mathbb{P}\left[\partial_\omega g_\omega(z)\hat{a}_\omega(x)\right],\qquad l=1,\dots,n_g,
    \end{equation*}
    and averaging over $l$ yields the announced identity by linearity of the expectation.
\end{proof}

\begin{lemma}[Bias control]\label{lem:bias_cont}
Let \Cref{ass:gome} hold. Let $\{\hat{h}_\omega\}_{\omega\in\Omega}$ and $\hat{b}$ be any estimators of $\{h^\star_\omega\}_{\omega\in\Omega}$ and $b^\star$ built from $\mathcal{D}_{in}$ and $\mathcal{D}_{out}$, possibly using additional randomness independent of the mini-batches, and set $\hat{a}_\omega=-\hat{h}_\omega+\hat{b}$ and $a^\star_\omega=-h^\star_\omega+b^\star$ for every $\omega\in\Omega$. Then:
\begin{equation*}
    \mathbb{E}\left[\sup_{\omega\in\Omega}\left\|\widetilde{\nabla\mathcal{F}}(\omega)-\nabla\mathcal{F}(\omega)\right\|^2\right]\leq G^2\,\mathbb{E}\left[\sup_{\omega\in\Omega}\left\|\hat{a}_\omega-a^\star_\omega\right\|_{\LL}^2\right]\leq2G^2\left(e_h+e_b\right),
\end{equation*}
where $\widetilde{\nabla\mathcal{F}}(\omega)=\mathbb{E}[\widehat{\nabla\mathcal{F}}(\omega)\mid\hat{a}_\omega]$, $e_h=\mathbb{E}[\sup_{\omega\in\Omega}\|\hat{h}_\omega-h^\star_\omega\|_{\LL}^2]$, $e_b=\mathbb{E}[\|\hat{b}-b^\star\|_{\LL}^2]$, and the expectation is taken over the randomness used to construct $\hat{h}_\omega$ and $\hat{b}$.
\end{lemma}

\begin{proof}
Fix $\omega\in\Omega$. By \cref{lem:oracle_gradient} and \cref{prop:hyper}, we have:
\begin{equation*}
    \widetilde{\nabla\mathcal{F}}(\omega)-\nabla\mathcal{F}(\omega)=-\mathbb{E}_\mathbb{P}\left[\partial_\omega g_\omega(z)\left(\hat{a}_\omega(x)-a^\star_\omega(x)\right)\right].
\end{equation*}
Since $\|\cdot\|^2$ is convex, Jensen's inequality gives:
\begin{equation*}
    \left\|\widetilde{\nabla\mathcal{F}}(\omega)-\nabla\mathcal{F}(\omega)\right\|^2
    \leq\mathbb{E}_\mathbb{P}\left[\left\|\partial_\omega g_\omega(z)\right\|^2\left(\hat{a}_\omega(x)-a^\star_\omega(x)\right)^2\right]
    \leq G^2\left\|\hat{a}_\omega-a^\star_\omega\right\|_{\LL}^2,
\end{equation*}
where the last inequality uses $\sup_{\omega\in\Omega,z\in\mathcal{Z}}\|\partial_\omega g_\omega(z)\|\leq G$ from \Cref{ass:gome}, together with the fact that $\LL$ is taken with respect to $\mathbb{P}_X$, so that $\mathbb{E}_\mathbb{P}[(\hat{a}_\omega(x)-a^\star_\omega(x))^2]=\|\hat{a}_\omega-a^\star_\omega\|^2_{\LL}$. This inequality holds for every $\omega\in\Omega$ and every realization of the randomness used to construct $\hat{h}_\omega$ and $\hat{b}$, so we may take the supremum over $\omega\in\Omega$ on both sides and then the expectation, which gives the first bound. For the second, the triangle inequality and $(u+v)^2\leq2u^2+2v^2$ yield, for every $\omega\in\Omega$:
\begin{align*}
    \left\|\hat{a}_\omega-a^\star_\omega\right\|_{\LL}^2\leq&\left(\left\|\hat{h}_\omega-h^\star_\omega\right\|_{\LL}+\left\|\hat{b}-b^\star\right\|_{\LL}\right)^2\\
    \leq&2\left(\left\|\hat{h}_\omega-h^\star_\omega\right\|_{\LL}^2+\left\|\hat{b}-b^\star\right\|_{\LL}^2\right).
\end{align*}
Taking the supremum over $\omega\in\Omega$ and then the expectation concludes the proof.
\end{proof}

\begin{lemma}[Variance control]\label{lem:var_control}
    Let \Cref{ass:bound_la,ass:gome} hold. Let $\{\hat{h}_\omega\}_{\omega\in\Omega}$ and $\hat{b}$ be any estimators of $\{h^\star_\omega\}_{\omega\in\Omega}$ and $b^\star$ built from $\mathcal{D}_{in}$ and $\mathcal{D}_{out}$, possibly using additional randomness independent of the mini-batches, and set $\hat{a}_\omega=-\hat{h}_\omega+\hat{b}$ and $a^\star_\omega=-h^\star_\omega+b^\star$ for every $\omega\in\Omega$. Then, for every $\omega\in\Omega$, we have:
    \begin{align*}
        \mathbb{E}\left[\left\|\widehat{\nabla\mathcal{F}}(\omega)-\widetilde{\nabla\mathcal{F}}(\omega)\right\|^2\right]&\leq\frac{2G^2}{n_g}\left(\mathbb{E}\left[\sup_{\omega'\in\Omega}\left\|\hat{a}_{\omega'}-a^\star_{\omega'}\right\|_{\LL}^2\right]+(D+B)^2\right)\\
        &\leq\frac{2G^2}{n_g}\left[2\left(e_h+e_b\right)+(D+B)^2\right],
    \end{align*}
    where $\widetilde{\nabla\mathcal{F}}(\omega)=\mathbb{E}[\widehat{\nabla\mathcal{F}}(\omega)\mid\hat{a}_\omega]$, $e_h=\mathbb{E}[\sup_{\omega\in\Omega}\|\hat{h}_\omega-h^\star_\omega\|_{\LL}^2]$, $e_b=\mathbb{E}[\|\hat{b}-b^\star\|_{\LL}^2]$, and the expectation is taken over all sources of randomness.
\end{lemma}

\begin{proof}
Fix $\omega\in\Omega$. Let $\mathcal{B}_g=\{(\underline{x}_l,\underline{z}_l)\}_{l=1}^{n_g}$ be the mini-batch used at $\omega$, and set:
\begin{equation*}
    U_l(\omega)=-\partial_\omega g_\omega(\underline{z}_l)\hat{a}_\omega(\underline{x}_l),\qquad l=1,\dots,n_g,
\end{equation*}
so that $\widehat{\nabla\mathcal{F}}(\omega)=\frac{1}{n_g}\sum_{l=1}^{n_g}U_l(\omega)$ and, by \cref{lem:oracle_gradient}, $\widetilde{\nabla\mathcal{F}}(\omega)=\mathbb{E}[U_l(\omega)\mid\hat{a}_\omega]$ for every $l$. As $\mathcal{B}_g$ is independent of $\hat{a}_\omega$, the variables $U_1(\omega),\dots,U_{n_g}(\omega)$ are \emph{i.i.d.} conditionally on $\hat{a}_\omega$, so that the conditional variance of their average scales as $1/n_g$:
\begin{equation*}
    \mathbb{E}\left[\left\|\widehat{\nabla\mathcal{F}}(\omega)-\widetilde{\nabla\mathcal{F}}(\omega)\right\|^2\;\middle|\;\hat{a}_\omega\right]\leq\frac{1}{n_g}\mathbb{E}\left[\left\|U_1(\omega)\right\|^2\;\middle|\;\hat{a}_\omega\right].
\end{equation*}
By \Cref{ass:gome}, $\|U_1(\omega)\|^2\leq G^2|\hat{a}_\omega(\underline{x}_1)|^2$, and since $\underline{x}_1\sim\mathbb{P}_X$ conditionally on $\hat{a}_\omega$:
\begin{align*}
    \mathbb{E}\left[\left\|U_1(\omega)\right\|^2\;\middle|\;\hat{a}_\omega\right]&\leq G^2\left\|\hat{a}_\omega\right\|_{\LL}^2\leq2G^2\left(\left\|\hat{a}_\omega-a^\star_\omega\right\|_{\LL}^2+\left\|a^\star_\omega\right\|_{\LL}^2\right)\\
    &\leq2G^2\left(\sup_{\omega'\in\Omega}\left\|\hat{a}_{\omega'}-a^\star_{\omega'}\right\|_{\LL}^2+(D+B)^2\right),
\end{align*}
where the second inequality follows from the elementary inequality $\|u+v\|^2\leq2(\|u\|^2+\|v\|^2)$, and the third bounds $\|\hat{a}_\omega-a^\star_\omega\|_{\LL}^2$ by its supremum over $\Omega$ and uses $\|a^\star_\omega\|_{\LL}\leq\|a^\star_\omega\|_\infty$, valid since $\mathbb{P}_X$ is a probability measure, together with the bound $\|a^\star_\omega\|_\infty\leq D+B$ established in \cref{lem:bound_targets}. Taking the total expectation gives the first bound. For the second, the triangle inequality and $(u+v)^2\leq2u^2+2v^2$ yield, for every $\omega'\in\Omega$:
\begin{equation*}
    \left\|\hat{a}_{\omega'}-a^\star_{\omega'}\right\|_{\LL}^2\leq2\left\|\hat{h}_{\omega'}-h^\star_{\omega'}\right\|_{\LL}^2+2\left\|\hat{b}-b^\star\right\|_{\LL}^2,
\end{equation*}
so that taking the supremum over $\omega'\in\Omega$ and then the expectation gives $\mathbb{E}[\sup_{\omega'\in\Omega}\|\hat{a}_{\omega'}-a^\star_{\omega'}\|_{\LL}^2]\leq2(e_h+e_b)$, since $\hat{b}$ does not depend on $\omega'$.
\end{proof}

\subsection{Main theorem: convergence of projected mini-batch biased SGD}\label{sec:moreau-alternative}

We write $\delta_\Omega$ for the \emph{indicator function} of $\Omega$, that is $\delta_\Omega(\omega)=0$ if $\omega\in\Omega$ and $+\infty$ otherwise --- a standard device that folds the constraint $\omega\in\Omega$ into the objective itself, so that minimizing $\mathcal{F}+\delta_\Omega$ over all of $\R^d$ is the same problem as minimizing $\mathcal{F}$ over $\Omega$ alone. We set $f\coloneqq\mathcal{F}+\delta_\Omega$. For $\omega$ with $f(\omega)<\infty$, $\partial f(\omega)$ denotes the (limiting) subdifferential of $f$ at $\omega$: a \emph{set} of generalized slopes, reducing to $\{\nabla f(\omega)\}$ where $f$ is differentiable, with $0\in\partial f(\omega^\star)$ the first-order stationarity condition for $\min_{\omega\in\Omega}\mathcal{F}(\omega)$. For a point $p$ and a set $S$, $\mathrm{dist}(p,S)\coloneqq\inf_{s\in S}\|p-s\|$ is the ordinary distance from $p$ to $S$; we use it only as $\mathrm{dist}(0,\partial f(\omega))$, a continuous relaxation of exact stationarity (zero exactly when $0\in\partial f(\omega)$, small when \emph{some} element of $\partial f(\omega)$ is nearly zero). A function $\phi$ is $\rho$-\emph{weakly convex} if $\phi+\frac{\rho}{2}\|\cdot\|^2$ is convex. We note that every $\mathcal{L}$-smooth function is $\mathcal{L}$-weakly convex.

\textbf{The Moreau envelope.} Since $\mathcal{F}$ is $\mathcal{L}$-smooth on $\Omega$ (\cref{prop:hyper}), $\mathcal{F}+\frac{\mathcal{L}}{2}\|\cdot\|^2$ is convex, so $\mathcal{F}$ is $\mathcal{L}$-weakly convex. Moreover, since $\Omega$ is convex, $\delta_\Omega$ is convex, so $f=\mathcal{F}+\delta_\Omega$ is $\mathcal{L}$-weakly convex as well, being a sum of the two. Fix $\lambda\in(0,\mathcal{L}^{-1})$ and define, for $\omega\in\Omega$, the Moreau envelope and the proximal map of $\mathcal{F}$ as:
\begin{equation}\label{eq:mor_prox}
    \mathcal{F}_\lambda(\omega)\coloneqq\min_{y\in\Omega}\left\{\mathcal{F}(y)+\frac{1}{2\lambda}\|y-\omega\|^2\right\},\qquad\mathrm{prox}_{\lambda\mathcal{F}}(\omega)\coloneqq\argmin_{y\in\Omega}\left\{\mathcal{F}(y)+\frac{1}{2\lambda}\|y-\omega\|^2\right\}.
\end{equation}
Since $\lambda<\mathcal{L}^{-1}$, the minimized objective is $(\lambda^{-1}-\mathcal{L})$-strongly convex on $\Omega$, so the minimum is attained at a unique point and $\mathrm{prox}_{\lambda\mathcal{F}}(\omega)$ is well defined. By the standard Moreau-envelope calculus for weakly convex functions \citep{moreau1965proximite,rockafellar1998variational}, $\mathcal{F}_\lambda$ is $C^1$ with:
\begin{equation}\label{eq:moreau}
    \nabla\mathcal{F}_\lambda(\omega)=\frac{1}{\lambda}\left(\omega-\mathrm{prox}_{\lambda\mathcal{F}}(\omega)\right),\qquad\left\|\mathrm{prox}_{\lambda\mathcal{F}}(\omega)-\omega\right\|=\lambda\left\|\nabla\mathcal{F}_\lambda(\omega)\right\|.
\end{equation}
Writing $u_k\coloneqq\mathrm{prox}_{\lambda\mathcal{F}}(\omega_k)$, we have $f(u_k)\leq f(\omega_k)$, since $\omega_k$ itself is a feasible, zero-penalty competitor in the minimization defining $\mathcal{F}_\lambda(\omega_k)$, and $\mathrm{dist}(0,\partial f(u_k))\leq\|\nabla\mathcal{F}_\lambda(\omega_k)\|$, since the first-order optimality condition for $u_k$ reads exactly $\nabla\mathcal{F}_\lambda(\omega_k)\in\partial f(u_k)$ by the subdifferential sum rule, valid here because the added quadratic term is smooth. A small $\|\nabla\mathcal{F}_\lambda(\omega_k)\|$ therefore certifies that $\omega_k$ is close to a point $u_k$ that is itself near-stationary for \eqref{eq:obo}, with no loss of objective value. This is the $C^1$ surrogate for $\mathrm{dist}(0,\partial f(\omega_k))$ itself, which is generally discontinuous in $\omega_k$ and hence unsuitable for a descent-type stochastic analysis.

\begin{lemma}[One-step envelope descent]\label{lem:onestep-envelope}
Let \Cref{ass:bound_la,ass:gome} hold and let $\lambda\in(0,\mathcal{L}^{-1})$, where $\mathcal{L}$ is the smoothness parameter of $\mathcal{F}$ provided in \cref{prop:hyper}. Let $\{\omega_k\}_{k\geq0}$ be generated by $\omega_{k+1}=\Pi_\Omega\left(\omega_k-\eta\widehat{\nabla\mathcal{F}}(\omega_k)\right)$ with $0<\eta\leq\frac{1}{2\mathcal{L}}$. Then, for every $k\geq0$:
\begin{align*}
    &\mathbb{E}\left[\mathcal{F}_\lambda(\omega_{k+1})\mid\hat{a}_{\omega_k}\right]\\\leq{}&
    \mathcal{F}_\lambda(\omega_k)-\frac{3\eta}{4}\left(1-\lambda\mathcal{L}\right)\left\|\nabla\mathcal{F}_\lambda(\omega_k)\right\|^2+\eta\left(\frac{1}{1-\lambda\mathcal{L}}+\frac{2\eta}{\lambda}\right)\left\|\widetilde{\nabla\mathcal{F}}(\omega_k)-\nabla\mathcal{F}(\omega_k)\right\|^2\\
    &+\frac{2\eta^2}{\lambda}\,\mathbb{E}\left[\left\|\widehat{\nabla\mathcal{F}}(\omega_k)-\widetilde{\nabla\mathcal{F}}(\omega_k)\right\|^2\;\middle|\;\hat{a}_{\omega_k}\right]+\frac{\eta}{\lambda}\Big(\mathcal{F}(\omega_k)-\mathbb{E}\left[\mathcal{F}(\omega_{k+1})\mid\hat{a}_{\omega_k}\right]\Big).
\end{align*}
\end{lemma}

\begin{proof}
Let $k\geq0$. Write $u_k\coloneqq\mathrm{prox}_{\lambda\mathcal{F}}(\omega_k)\in\Omega$ for the proximal point of $\mathcal{F}$ at $\omega_k$, that is, the unique minimizer in \cref{eq:mor_prox}, so that $\|u_k-\omega_k\|=\lambda\|\nabla\mathcal{F}_\lambda(\omega_k)\|$ (\cref{eq:moreau}). Write also $\mathbb{E}_k[\cdot]\coloneqq\mathbb{E}[\cdot\mid\hat{a}_{\omega_k}]$\footnote{Precisely, $\hat{a}_{\omega_k}$ is shorthand for the pair $(\omega_k,\hat{a}_{\omega_k})$: Step 1 treats $u_k-\omega_k$, $\mathcal{F}_\lambda(\omega_k)$, $\mathcal{F}(\omega_k)$ and $b_k$ as known given $\mathbb{E}_k$, which requires $\omega_k$ itself to be fixed, not just the function $\hat{a}_{\omega_k}$. This causes no loss of generality: the fresh mini-batch $\mathcal{B}^k_g$ is independent of the \emph{entire} history up to step $k$ (not merely of $\hat{a}_{\omega_k}$), so every conditional expectation of a quantity built from $(\omega_k,\hat{a}_{\omega_k},\mathcal{B}^k_g)$ takes the same value whether one conditions on $(\omega_k,\hat{a}_{\omega_k})$ or on the full history.}, with $g_k=\widehat{\nabla\mathcal{F}}(\omega_k)$, $v_k\coloneqq\mathbb{E}_k[\widehat{\nabla\mathcal{F}}(\omega_k)]=\widetilde{\nabla\mathcal{F}}(\omega_k)$, $b_k\coloneqq v_k-\nabla\mathcal{F}(\omega_k)$, $\xi_k\coloneqq g_k-v_k$, and $\Delta_k\coloneqq\omega_{k+1}-\omega_k$.

\textbf{Step 1: A one-step bound on $\mathbb{E}_k[\mathcal{F}_\lambda]$. }By construction, it holds that $\omega_{k+1}=\Pi_\Omega(\omega_k-\eta g_k)=\argmin_{y\in\Omega}h(y)$ with $h(y)\coloneqq\langle g_k,y-\omega_k\rangle+\frac{1}{2\eta}\|y-\omega_k\|^2$, which is $\frac1\eta$-strongly convex. Since $h$ is quadratic, for any $y\in\Omega$, one has the exact expansion $h(y)=h(\omega_{k+1})+\langle\nabla h(\omega_{k+1}),y-\omega_{k+1}\rangle+\frac{1}{2\eta}\|y-\omega_{k+1}\|^2$. In addition, the first-order optimality condition for the constrained minimizer $\omega_{k+1}$ gives $\langle\nabla h(\omega_{k+1}),y-\omega_{k+1}\rangle\geq0$ for all $y\in\Omega$ \citep[Proposition~1.3]{bubeck2015convex}. Hence, taking $y=u_k\in\Omega$, we obtain:
\begin{equation*}
    h(u_k)\ \geq\ h(\omega_{k+1})+\frac{1}{2\eta}\|u_k-\omega_{k+1}\|^2.
\end{equation*}
Expanding both sides via the definition of $h$ and multiplying through by $2\eta>0$ yields:
\begin{equation}\label{eq:s1}
    \|u_k-\omega_{k+1}\|^2\leq\|u_k-\omega_k\|^2-\|\Delta_k\|^2+2\eta\langle g_k,u_k-\omega_{k+1}\rangle.
\end{equation}
This transfers to $\mathcal{F}_\lambda$ as follows. Since $u_k\in\Omega$ is feasible for the minimization defining $\mathcal{F}_\lambda(\omega_{k+1})$, but not necessarily its minimizer, we get:
\begin{equation*}
    \mathcal{F}_\lambda(\omega_{k+1})\leq\mathcal{F}(u_k)+\frac{1}{2\lambda}\|u_k-\omega_{k+1}\|^2,
\end{equation*}
and by definition $\mathcal{F}(u_k)+\frac{1}{2\lambda}\|u_k-\omega_k\|^2=\mathcal{F}_\lambda(\omega_k)$ exactly. Multiplying \cref{eq:s1} by $\frac{1}{2\lambda}$ and substituting:
\begin{equation}\label{eq:s2}
    \mathcal{F}_\lambda(\omega_{k+1})\leq\mathcal{F}_\lambda(\omega_k)-\frac{1}{2\lambda}\|\Delta_k\|^2+\frac{\eta}{\lambda}\langle g_k,u_k-\omega_{k+1}\rangle.
\end{equation}
Since $g_k=\nabla\mathcal{F}(\omega_k)+b_k+\xi_k$ and $u_k-\omega_{k+1}=(u_k-\omega_k)-\Delta_k$, we have:
\begin{equation*}
    \langle g_k,u_k-\omega_{k+1}\rangle=\langle g_k,u_k-\omega_k\rangle-\langle\nabla\mathcal{F}(\omega_k),\Delta_k\rangle-\langle b_k,\Delta_k\rangle-\langle\xi_k,\Delta_k\rangle.
\end{equation*}
By $\mathcal{L}$-smoothness of $\mathcal{F}$ (\cref{prop:hyper}), it holds that $-\langle\nabla\mathcal{F}(\omega_k),\Delta_k\rangle\leq\mathcal{F}(\omega_k)-\mathcal{F}(\omega_{k+1})+\frac{\mathcal{L}}{2}\|\Delta_k\|^2$ \citep[Lemma~1.2.3]{nesterov2013introductory}. By Young's inequality with parameter $\frac{1}{4\eta}$, $-\langle b_k,\Delta_k\rangle\leq2\eta\|b_k\|^2+\frac{1}{8\eta}\|\Delta_k\|^2$ and likewise $-\langle\xi_k,\Delta_k\rangle\leq2\eta\|\xi_k\|^2+\frac{1}{8\eta}\|\Delta_k\|^2$. Multiplying these three bounds by $\frac{\eta}{\lambda}$ and inserting into \cref{eq:s2}, the coefficient of $\|\Delta_k\|^2$ becomes $\frac{1}{\lambda}\big(-\frac12+\frac{\eta\mathcal{L}}{2}+\frac18+\frac18\big)=\frac{1}{\lambda}\big(\frac{\eta\mathcal{L}}{2}-\frac14\big)$, which is $\leq0$ for $\eta\leq\frac{1}{2\mathcal{L}}$, independently of $\lambda$. Dropping this nonpositive term, we get:
\begin{equation}\label{eq:s3}
    \mathcal{F}_\lambda(\omega_{k+1})\leq\mathcal{F}_\lambda(\omega_k)+\frac{\eta}{\lambda}\langle g_k,u_k-\omega_k\rangle+\frac{\eta}{\lambda}\big(\mathcal{F}(\omega_k)-\mathcal{F}(\omega_{k+1})\big)+\frac{2\eta^2}{\lambda}\big(\|b_k\|^2+\|\xi_k\|^2\big).
\end{equation}
The quantities $\mathcal{F}_\lambda(\omega_k)$, $\mathcal{F}(\omega_k)$ and $b_k$ are $\hat{a}_{\omega_k}$-measurable, and $\mathbb{E}_k[\xi_k]=0$ since $\mathcal{B}^k_g$ is fresh, so that $\mathbb{E}_k[\langle g_k,u_k-\omega_k\rangle]=\langle v_k,u_k-\omega_k\rangle$. Taking $\mathbb{E}_k$ in \cref{eq:s3} gives:
\begin{equation}\label{eq:s4}
    \mathbb{E}_k[\mathcal{F}_\lambda(\omega_{k+1})]\leq\mathcal{F}_\lambda(\omega_k)+\frac{\eta}{\lambda}\langle v_k,u_k-\omega_k\rangle+\frac{\eta}{\lambda}\big(\mathcal{F}(\omega_k)-\mathbb{E}_k[\mathcal{F}(\omega_{k+1})]\big)+\frac{2\eta^2}{\lambda}\big(\|b_k\|^2+\mathbb{E}_k[\|\xi_k\|^2]\big).
\end{equation}

\textbf{Step 2: Control on the cross term $\langle v_k,u_k-\omega_k\rangle$. }Write $v_k=\nabla\mathcal{F}(\omega_k)+b_k$, so:
\begin{equation}\label{eq:s5}
    \langle v_k,u_k-\omega_k\rangle=\langle\nabla\mathcal{F}(\omega_k),u_k-\omega_k\rangle+\langle b_k,u_k-\omega_k\rangle.
\end{equation}
Since $\mathcal{F}$ is $\mathcal{L}$-smooth (\cref{prop:hyper}), \citep[Lemma~1.2.3]{nesterov2013introductory} gives:
\begin{equation*}
    \mathcal{F}(u_k)\geq\mathcal{F}(\omega_k)+\langle\nabla\mathcal{F}(\omega_k),u_k-\omega_k\rangle-\frac{\mathcal{L}}{2}\|u_k-\omega_k\|^2,
\end{equation*}
so
\begin{equation}\label{eq:s6}
    \langle\nabla\mathcal{F}(\omega_k),u_k-\omega_k\rangle\leq\mathcal{F}(u_k)-\mathcal{F}(\omega_k)+\frac{\mathcal{L}}{2}\|u_k-\omega_k\|^2.
\end{equation}
We have by Young's inequality with parameter $\frac{1}{2}\left(\lambda^{-1}-\mathcal{L}\right)>0$:
\begin{equation}\label{eq:s7}
    \langle b_k,u_k-\omega_k\rangle\leq\frac{1}{\lambda^{-1}-\mathcal{L}}\|b_k\|^2+\frac{\lambda^{-1}-\mathcal{L}}{4}\|u_k-\omega_k\|^2.
\end{equation}
The map $\psi_k(y)\coloneqq\mathcal{F}(y)+\frac{1}{2\lambda}\|y-\omega_k\|^2$ is $(\lambda^{-1}-\mathcal{L})$-strongly convex on $\Omega$ (sum of the convex function $\mathcal{F}+\frac{\mathcal{L}}{2}\|\cdot\|^2$, up to an affine term, and a quadratic of curvature $\lambda^{-1}-\mathcal{L}$, which is positive since $\lambda<\mathcal{L}^{-1}$), with unique minimizer $u_k$. Comparing to the feasible point $y=\omega_k$, strong convexity gives:
\begin{equation*}
    \psi_k(\omega_k)\geq\psi_k(u_k)+\frac{\lambda^{-1}-\mathcal{L}}{2}\|\omega_k-u_k\|^2.
\end{equation*}
Since $\psi_k(\omega_k)=\mathcal{F}(\omega_k)$ and $\psi_k(u_k)=\mathcal{F}(u_k)+\frac{1}{2\lambda}\|u_k-\omega_k\|^2$, this rearranges to:
\begin{equation}\label{eq:s8}
    \mathcal{F}(u_k)-\mathcal{F}(\omega_k)\leq-\left(\lambda^{-1}-\frac{\mathcal{L}}{2}\right)\|u_k-\omega_k\|^2.
\end{equation}
Adding \cref{eq:s6} and \cref{eq:s7} reconstructs the right-hand side of \cref{eq:s5}. Substituting \cref{eq:s8} for $\mathcal{F}(u_k)-\mathcal{F}(\omega_k)$ and combining the three $\|u_k-\omega_k\|^2$ coefficients, $-\left(\lambda^{-1}-\frac{\mathcal{L}}{2}\right)+\frac{\mathcal{L}}{2}+\frac{\lambda^{-1}-\mathcal{L}}{4}=-\frac34\left(\lambda^{-1}-\mathcal{L}\right)$:
\begin{equation}\label{eq:s9}
    \langle v_k,u_k-\omega_k\rangle\leq-\frac34\left(\lambda^{-1}-\mathcal{L}\right)\|u_k-\omega_k\|^2+\frac{1}{\lambda^{-1}-\mathcal{L}}\|b_k\|^2.
\end{equation}

\textbf{Step 3: Assembling the descent inequality. }Plugging \cref{eq:s9} into \cref{eq:s4} and using $\|u_k-\omega_k\|^2=\lambda^2\|\nabla\mathcal{F}_\lambda(\omega_k)\|^2$, the descent term becomes:
\begin{equation*}
    -\frac{\eta}{\lambda}\cdot\frac34\left(\lambda^{-1}-\mathcal{L}\right)\cdot\lambda^2\|\nabla\mathcal{F}_\lambda(\omega_k)\|^2=-\frac{3\eta}{4}\left(1-\lambda\mathcal{L}\right)\|\nabla\mathcal{F}_\lambda(\omega_k)\|^2,
\end{equation*}
while the coefficient of $\|b_k\|^2$ becomes $\frac{\eta}{\lambda}\cdot\frac{1}{\lambda^{-1}-\mathcal{L}}+\frac{2\eta^2}{\lambda}=\eta\left(\frac{1}{1-\lambda\mathcal{L}}+\frac{2\eta}{\lambda}\right)$. Hence:
\begin{align*}
    \mathbb{E}_k[\mathcal{F}_\lambda(\omega_{k+1})]\leq{}&\mathcal{F}_\lambda(\omega_k)-\frac{3\eta}{4}\left(1-\lambda\mathcal{L}\right)\|\nabla\mathcal{F}_\lambda(\omega_k)\|^2+\eta\left(\frac{1}{1-\lambda\mathcal{L}}+\frac{2\eta}{\lambda}\right)\|b_k\|^2\\
    &+\frac{2\eta^2}{\lambda}\,\mathbb{E}_k[\|\xi_k\|^2]+\frac{\eta}{\lambda}\big(\mathcal{F}(\omega_k)-\mathbb{E}_k[\mathcal{F}(\omega_{k+1})]\big).
\end{align*}
Recalling $b_k=\widetilde{\nabla\mathcal{F}}(\omega_k)-\nabla\mathcal{F}(\omega_k)$ and $\xi_k=\widehat{\nabla\mathcal{F}}(\omega_k)-\widetilde{\nabla\mathcal{F}}(\omega_k)$ gives exactly the claimed inequality.
\end{proof}

\begin{theorem}[Convergence guarantee of projected mini-batch biased SGD]\label{thm:eta-scaled_lam}
Fix $\omega_0\in\Omega$. Let $\mathcal{F}^\star\coloneqq\min_{\omega\in\Omega}\mathcal{F}(\omega)$ and $\Delta_\mathcal{F}\coloneqq\mathcal{F}(\omega_0)-\mathcal{F}^\star$. Let $\{\omega_k\}_{k\geq0}$ be generated by $\omega_{k+1}=\Pi_\Omega(\omega_k-\eta\widehat{\nabla\mathcal{F}}(\omega_k))$ with stepsize $0<\eta\leq\frac{1}{2\mathcal{L}}$, where $\mathcal{L}$ is the smoothness parameter of $\mathcal{F}$ provided in \cref{prop:hyper}. Under \Cref{ass:bound_la,ass:gome}, for any $K\geq1$ and any $\lambda\in(0,\mathcal{L}^{-1})$, the iterates $\{\omega_k\}_{k\geq0}$ satisfy:
\begin{align*}
    \frac{1}{K}\sum_{k=0}^{K-1}\mathbb{E}\left[\left\|\nabla\mathcal{F}_\lambda(\omega_k)\right\|^2\right]\leq{}&\frac{4}{3\left(1-\lambda\mathcal{L}\right)}\Bigg[\left(\frac{1}{\eta}+\frac{1}{\lambda}\right)\frac{\Delta_\mathcal{F}}{K}+2G^2\left(\frac{1}{1-\lambda\mathcal{L}}+\frac{2\eta}{\lambda}\right)\left(e_h+e_b\right)\\
    &\qquad\qquad\quad+\frac{4G^2\eta}{\lambda n_g}\left((D+B)^2+2\left(e_h+e_b\right)\right)\Bigg],
\end{align*}
where $e_h=\mathbb{E}[\sup_{\omega\in\Omega}\|\hat{h}_\omega-h^\star_\omega\|_{\LL}^2]$, $e_b=\mathbb{E}[\|\hat{b}-b^\star\|_{\LL}^2]$, and $B$, $D$, $G$ are the constants of \Cref{ass:bound_la,ass:gome}. The expectation is over all sources of randomness, namely the samples $\mathcal{D}_{in}$ and $\mathcal{D}_{out}$ used to construct $\hat{h}_\omega$ and $\hat{b}$, and the mini-batches $\{\mathcal{B}_g^k\}_{k\geq0}$; the expectations defining $e_h$ and $e_b$ are over $\mathcal{D}_{in}$ and $\mathcal{D}_{out}$ only.
\end{theorem}

\begin{proof}
Let $k\geq0$, and recall that $b_k\coloneqq\widetilde{\nabla\mathcal{F}}(\omega_k)-\nabla\mathcal{F}(\omega_k)$ is the bias of the plug-in hypergradient estimator $\widehat{\nabla\mathcal{F}}(\omega_k)$ conditional on $\hat{a}_{\omega_k}$, and $\xi_k\coloneqq\widehat{\nabla\mathcal{F}}(\omega_k)-\widetilde{\nabla\mathcal{F}}(\omega_k)$ is its conditionally zero-mean noise.

Taking full expectation in \cref{lem:onestep-envelope} and summing over $k=0,\ldots,K-1$, both the $\mathcal{F}_\lambda$- and the $\mathcal{F}$-sums telescope, and $\omega_0$ is deterministic, so:
\begin{align*}
    &\frac{3\eta}{4}\left(1-\lambda\mathcal{L}\right)\sum_{k=0}^{K-1}\mathbb{E}\left[\left\|\nabla\mathcal{F}_\lambda(\omega_k)\right\|^2\right]\\
    \leq{}&\mathcal{F}_\lambda(\omega_0)-\mathbb{E}\left[\mathcal{F}_\lambda(\omega_K)\right]+\frac{\eta}{\lambda}\big(\mathcal{F}(\omega_0)-\mathbb{E}\left[\mathcal{F}(\omega_K)\right]\big)+\eta\left(\frac{1}{1-\lambda\mathcal{L}}+\frac{2\eta}{\lambda}\right)\sum_{k=0}^{K-1}\mathbb{E}\left[\|b_k\|^2\right]\\
    &+\frac{2\eta^2}{\lambda}\sum_{k=0}^{K-1}\mathbb{E}\left[\|\xi_k\|^2\right].
\end{align*}
Since $\mathcal{F}_\lambda(\omega)\leq\mathcal{F}(\omega)$ and $\mathcal{F}_\lambda(\omega),\mathcal{F}(\omega)\geq\mathcal{F}^\star$ for every $\omega\in\Omega$, the two differences on the right-hand side are bounded by $\left(1+\frac{\eta}{\lambda}\right)\Delta_\mathcal{F}$. Dividing by $\frac{3\eta}{4}(1-\lambda\mathcal{L})K$, which is positive because $\lambda<\mathcal{L}^{-1}$, we get:
\begin{align}\label{eq:t3}
    &\frac{1}{K}\sum_{k=0}^{K-1}\mathbb{E}\left[\left\|\nabla\mathcal{F}_\lambda(\omega_k)\right\|^2\right]\nonumber\\
    \leq{}&\frac{4}{3\left(1-\lambda\mathcal{L}\right)}\left(\frac{1}{\eta}+\frac{1}{\lambda}\right)\frac{\Delta_\mathcal{F}}{K}+\frac{4}{3\left(1-\lambda\mathcal{L}\right)}\left(\frac{1}{1-\lambda\mathcal{L}}+\frac{2\eta}{\lambda}\right)\frac{1}{K}\sum_{k=0}^{K-1}\mathbb{E}\left[\|b_k\|^2\right]\notag\\
    &+\frac{8\eta}{3\lambda\left(1-\lambda\mathcal{L}\right)}\frac{1}{K}\sum_{k=0}^{K-1}\mathbb{E}\left[\|\xi_k\|^2\right].
\end{align}
The iterate $\omega_k$ depends on $\mathcal{D}_{in}$ and $\mathcal{D}_{out}$, so we bound both error terms uniformly over $\Omega$. By \cref{lem:bias_cont,lem:var_control}, for every $k$, we have:
\begin{equation*}
    \mathbb{E}\left[\|b_k\|^2\right]\leq2G^2\left(e_h+e_b\right)\quad\text{and}\quad\mathbb{E}\left[\|\xi_k\|^2\right]\leq\frac{2G^2}{n_g}\left[(D+B)^2+2\left(e_h+e_b\right)\right].
\end{equation*}
Substituting these bounds into \cref{eq:t3} gives exactly the claimed bound.
\end{proof}

\begin{proof}[Proof of \cref{thm:eta-scaled}]
We specialize \cref{thm:eta-scaled_lam} to $\lambda=\frac{1}{2\mathcal{L}}$, which is admissible since $\frac{1}{2\mathcal{L}}<\mathcal{L}^{-1}$. With this choice, $\lambda\mathcal{L}=\frac12$, so the prefactor equals $\frac{4}{3(1-\lambda\mathcal{L})}=\frac{8}{3}$ and $\frac{1}{\lambda}=2\mathcal{L}$. Using $\eta\leq\frac{1}{2\mathcal{L}}$, that is $2\mathcal{L}\leq\frac{1}{\eta}$ and $\eta\mathcal{L}\leq\frac12$, the three brackets are bounded by:
\begin{equation*}
    \frac{1}{\eta}+\frac{1}{\lambda}=\frac{1}{\eta}+2\mathcal{L}\leq\frac{2}{\eta},
    \quad
    2G^2\left(\frac{1}{1-\lambda\mathcal{L}}+\frac{2\eta}{\lambda}\right)=2G^2\left(2+4\eta\mathcal{L}\right)\leq8G^2,
    \quad
    \frac{4G^2\eta}{\lambda n_g}=\frac{8\mathcal{L}G^2\eta}{n_g},
\end{equation*}
so that \cref{thm:eta-scaled_lam} gives:
\begin{equation}\label{eq:spec_half}
    \frac{1}{K}\sum_{k=0}^{K-1}\mathbb{E}\left[\left\|\nabla\mathcal{F}_{1/2\mathcal{L}}(\omega_k)\right\|^2\right]\leq\frac{16\Delta_\mathcal{F}}{3\eta K}+\frac{64G^2\left(e_h+e_b\right)}{3}+\frac{64\mathcal{L}G^2\eta}{3n_g}\left((D+B)^2+2\left(e_h+e_b\right)\right).
\end{equation}
It remains to pass from the Moreau envelope to the gradient mapping. Applying \citep[Section~2.2]{davis2019stochastic}, we obtain, for every $\omega\in\Omega$:
\begin{equation*}
    \left\|\mathcal{G}_{1/\mathcal{L}}(\omega)\right\|\leq\frac{3}{2}\left(1+\frac{1}{\sqrt{2}}\right)\left\|\nabla\mathcal{F}_{1/2\mathcal{L}}(\omega)\right\|,
    \qquad\text{hence}\qquad
    \left\|\mathcal{G}_{1/\mathcal{L}}(\omega)\right\|^2\leq c\left\|\nabla\mathcal{F}_{1/2\mathcal{L}}(\omega)\right\|^2,
\end{equation*}
with $c=\frac{9}{4}\left(\frac{3}{2}+\sqrt{2}\right)$. Taking expectations, averaging over $k=0,\ldots,K-1$ and combining with \cref{eq:spec_half} gives the claimed bound.
\end{proof}

\begin{corollary}[Balanced stepsize]\label{cor:balanced}
Let the assumptions of \Cref{thm:eta-scaled} hold, assume $\Delta_\mathcal{F}>0$, and set $\sigma^2\coloneqq(D+B)^2+2(e_h+e_b)$. For any $K\geq1$, the choice
\begin{equation}\label{eq:eta_balanced}
    \eta=\min\left\{\frac{1}{2\mathcal{L}},\ \eta_\star\right\},\qquad \eta_\star\coloneqq\frac{1}{2G\sigma}\sqrt{\frac{\Delta_\mathcal{F}\,n_g}{\mathcal{L}K}},
\end{equation}
is admissible and yields:
\begin{equation}\label{eq:rate_balanced}
    \frac{1}{K}\sum_{k=0}^{K-1}\mathbb{E}\left[\|\mathcal{G}_{1/\mathcal{L}}(\omega_k)\|^2\right]\leq 2c\left[\frac{\mathcal{L}\Delta_\mathcal{F}}{K}+2G\sigma\sqrt{\frac{\mathcal{L}\Delta_\mathcal{F}}{K n_g}}+2G^2(e_h+e_b)\right],
\end{equation}
where $c=18+12\sqrt{2}$ is the constant of \Cref{thm:eta-scaled}.
\end{corollary}

\begin{proof}
Write the bound of \Cref{thm:eta-scaled} as $c\left[c_{\mathrm{opt}}/\eta+c_{\mathrm{var}}\eta+c_{\mathrm{bias}}\right]$, valid for every $0<\eta\leq\frac{1}{2\mathcal{L}}$, where:
\begin{equation*}
    c_{\mathrm{opt}}=\frac{\Delta_\mathcal{F}}{K},\qquad c_{\mathrm{var}}=\frac{4\mathcal{L}G^2\sigma^2}{n_g},\qquad c_{\mathrm{bias}}=4G^2(e_h+e_b),\qquad c=18+12\sqrt{2}.
\end{equation*}
Since the constant $c$ does not depend on $\eta$, minimizing this bound amounts to minimizing $\eta\mapsto c_{\mathrm{opt}}/\eta+c_{\mathrm{var}}\eta$, which is convex on $(0,+\infty)$ and attains its minimum at $\sqrt{c_{\mathrm{opt}}/c_{\mathrm{var}}}$, and
\begin{equation*}
    \sqrt{\frac{c_{\mathrm{opt}}}{c_{\mathrm{var}}}}=\sqrt{\frac{\Delta_\mathcal{F}/K}{4\mathcal{L}G^2\sigma^2/n_g}}=\frac{1}{2G\sigma}\sqrt{\frac{\Delta_\mathcal{F}n_g}{\mathcal{L}K}}=\eta_\star,
\end{equation*}
so that \cref{eq:eta_balanced} reads $\eta=\min\{\frac{1}{2\mathcal{L}},\eta_\star\}$ and is admissible in \Cref{thm:eta-scaled}. Since $\frac{1}{\eta}=\max\{2\mathcal{L},\eta_\star^{-1}\}\leq2\mathcal{L}+\eta_\star^{-1}$ and $\eta\leq\eta_\star$, we obtain:
\begin{equation*}
    \frac{c_{\mathrm{opt}}}{\eta}+c_{\mathrm{var}}\eta\ \leq\ 2\mathcal{L}c_{\mathrm{opt}}+\frac{c_{\mathrm{opt}}}{\eta_\star}+c_{\mathrm{var}}\eta_\star\ =\ 2\mathcal{L}c_{\mathrm{opt}}+2\sqrt{c_{\mathrm{opt}}c_{\mathrm{var}}},
\end{equation*}
where we used $c_{\mathrm{opt}}/\eta_\star=c_{\mathrm{opt}}\sqrt{c_{\mathrm{var}}/c_{\mathrm{opt}}}=\sqrt{c_{\mathrm{opt}}c_{\mathrm{var}}}$ and $c_{\mathrm{var}}\eta_\star=c_{\mathrm{var}}\sqrt{c_{\mathrm{opt}}/c_{\mathrm{var}}}=\sqrt{c_{\mathrm{opt}}c_{\mathrm{var}}}$. Finally:
\begin{equation*}
    2\mathcal{L}c_{\mathrm{opt}}=\frac{2\mathcal{L}\Delta_\mathcal{F}}{K},\qquad 2\sqrt{c_{\mathrm{opt}}c_{\mathrm{var}}}=2\sqrt{\frac{\Delta_\mathcal{F}}{K}\cdot\frac{4\mathcal{L}G^2\sigma^2}{n_g}}=4\,G\sigma\sqrt{\frac{\mathcal{L}\Delta_\mathcal{F}}{Kn_g}},
\end{equation*}
which, together with $c_{\mathrm{bias}}$ and after multiplying by $c$, gives \cref{eq:rate_balanced}.
\end{proof}

\begin{remark}[Dominant regime]\label{rmk:balanced}
The second branch of \cref{eq:eta_balanced} is active as soon as $K\geq\mathcal{L}\Delta_\mathcal{F}n_g/(G^2\sigma^2)$. In that regime, the ratio of the first term of \cref{eq:rate_balanced} to the second equals $\frac{1}{2G\sigma}\sqrt{\mathcal{L}\Delta_\mathcal{F}n_g/K}$, which is at most $\frac{1}{2}$, so that the algorithmic error is of order $G\sigma\sqrt{\mathcal{L}\Delta_\mathcal{F}/(K n_g)}$. Only the total number $K n_g$ of hypergradient samples then matters: a given accuracy is reached either with many small batches or with fewer larger ones, as long as $K$ remains above this threshold.
\end{remark}
\section{Generalization Bounds for (\ref{eq:obo}) in the NTK Regime}\label{app:genres}

\cref{app:genres} proves the bounds on $e_h$ and $e_b$ stated in \cref{prop:gen_bounds_main}. The bound on $e_b$ follows directly from the single-regression result of \cref{prop:exp_gen_bound}. The bound on $e_h$ is more delicate because of the supremum over $\omega$, and proceeds in four steps: (i) we split the error of $\hat h_\omega$ into the gap between the finite-width network and its infinite-width NTK counterpart, which is controlled uniformly in $\omega$ since the truncation event does not depend on $\omega$, and the error of the NTK predictor itself; (ii) we show that the NTK predictor is linear in the target vector, so that its error is Lipschitz continuous in $\omega$; (iii) we control the NTK error on a finite net of $\Omega$ by a union bound and extend it to all of $\Omega$ by Lipschitz continuity; (iv) we choose the confidence level $\delta$ to balance the resulting terms.

Throughout \cref{app:genres}, we fix a nonempty, compact and convex parameter set $\Omega\subset\R^d$ with diameter $\mathrm{diam}(\Omega)$. We write $\Gram=[\kntk(x_i,x_j)]_{i,j=1}^{\nin}\in\R^{\nin\times\nin}$ for the inner infinite-width NTK Gram matrix on the samples of $\mathcal{D}_{\mathrm{in}}$, and $\lambda_0$ for its smallest eigenvalue. For $\omega\in\Omega$, let $\Yom=[g_\omega(z_1),\dots,g_\omega(z_{\nin})]^\top\in\R^{\nin}$ denote the inner target vector. The event $\mathcal{E}^h_{\lambda_0}=\{\lambda_0\geq\lambda_\star\}$ is the NTK non-degeneracy event of \cref{sec:estimators,subsec:exp_gen}, whose threshold $\lambda_\star=\lambda_\star(\nin,\delta)>0$ depends only on $\nin$, $\delta$, $p$, $A$ and $M_\mathbb{P}$. We denote by $\hntktw$ the inner NTK gradient flow predictor of $h^\star_\omega$ whose dynamics are described as follows for any $x\in\mathcal{X}$:

\begin{equation}\label{eq:grad_flow_time_var_inf_ntk}
    \frac{\mathrm{d}}{\mathrm{d}t}h_{t,\omega}^\mathrm{NTK}(x)=-\frac{1}{\nin}\sum_{i=1}^{\nin}K^\mathrm{NTK}(x,x_i)(h_{t,\omega}^\mathrm{NTK}(x_i)-g_\omega(z_i)),\qquad h_{0,\omega}^\mathrm{NTK}=0.
\end{equation}

For brevity, we state the generalization bounds without writing the exact lower bounds on the inner and outer sample sizes $\nin$ and $\nout$ required for the results to hold. We refer the reader to \cref{prop:exp_gen_bound} for a detailed statement including these conditions.

\subsection{Lipschitz continuity of the inner NTK approximation error in the outer variable}

\begin{lemma}[Linear representation of the inner NTK predictor]\label{lem:phi-linear}
On the event $\mathcal{E}^h_{\lambda_0}$, the inner NTK gradient flow predictor started from the mirrored initialization admits, for every $\omega\in\Omega$ and every $t\ge0$, the following representation:
\begin{equation*}
    \hntktw=\Phi_t\Yom,\qquad\Phi_t\coloneqq\frac{1}{\nin}\,\kntk(\cdot, \boldsymbol{X})^\top\seq{\mathbbm{1}_{\nin\times\nin}-\exp\left(-\frac{t}{\nin}\Gram\right)}\seq{\frac{1}{\nin}\Gram}^{-1},
\end{equation*}
where $\kntk(\cdot,\boldsymbol{X})=[\kntk(\cdot,x_1),\dots,\kntk(\cdot,x_{\nin})]^\top$. The operator $\Phi_t$ depends only on $\Gram$, the inputs $x_1,\ldots,x_{\nin}$, and $t$; all dependence on $\omega$ enters through $\Yom$.
\end{lemma}

\begin{proof}
We write $\mathbf{A}\coloneqq\frac1\nin\Gram$ and let $H(t)\coloneqq[\hntktw(x_1),\dots,\hntktw(x_{\nin})]^\top$. Evaluating the kernel gradient flow in \cref{eq:grad_flow_time_var_inf_ntk} at the training points gives the linear inhomogeneous ODE $\dot H(t)=-\mathbf{A}(H(t)-\Yom)$ with $H(0)=0$ due to the mirrored initialization. Its unique solution is $H(t)=(\mathbbm{1}_{\nin\times\nin}-\exp(-t\mathbf{A}))\Yom$, so the training residual is $H(t)-\Yom=-\exp(-t\mathbf{A})\Yom$. Substituting into the flow at an arbitrary $x\in\X$, we get:
\begin{equation*}
    \frac{\mathrm{d}}{\mathrm{d}t}\hntktw(x)=-\frac1\nin\kntk(x,\boldsymbol{X})^\top(H(t)-\Yom)=\frac1\nin\kntk(x,\boldsymbol{X})^\top\exp(-t\mathbf{A})\Yom.
\end{equation*}
The right-hand side does not involve $\hntktw(x)$. Integrating from $0$ to $t$ with $h^\mathrm{NTK}_{0,\omega}=0$ and using $\int_0^t \exp(-s\mathbf{A})\mathrm{d}s=\mathbf{A}^{-1}(\mathbbm{1}_{\nin\times\nin}-\exp(-t\mathbf{A}))$ (valid since $\mathbf{A}$ is invertible on $\mathcal{E}^h_{\lambda_0}$) yields $\hntktw(x)=\frac1\nin\kntk(x,\boldsymbol{X})^\top\mathbf{A}^{-1}(\mathbbm{1}_{\nin\times\nin}-\exp(-t\mathbf{A}))\Yom$, which concludes the proof.
\end{proof}

\begin{lemma}[Operator-norm bound on $\Phi_t$]\label{lem:phi-norm}
Let \cref{ass:bound_in} hold. On the event $\mathcal{E}^h_{\lambda_0}$, for every $t\ge0$, we have:
\begin{equation*}
\norm{\Phi_t}_{\ell^2 \to \LL}\le\frac{\kappa^{\mathrm{NTK}}\sqrt{\nin}}{\lambda_0},\text{ with }\kappa^{\mathrm{NTK}}\coloneqq(p+2)(A^2+1)+1,
\end{equation*}
where $\norm{\Phi_t}_{\ell^2\to\LL}\coloneqq\sup_{Y\neq0}\norm{\Phi_tY}_{\LL}/\norm{Y}$ denotes the operator norm of $\Phi_t$ as a map from $(\R^{\nin},\norm{\cdot})$ to $\LL$.
\end{lemma}

\begin{proof}
We factor $\Phi_t=\mathcal{E}M_t$, where $M_t\coloneqq(\mathbbm{1}_{\nin\times\nin}-\exp(-\frac t\nin\Gram))(\frac1\nin\Gram)^{-1}\in\mathbb{R}^{\nin\times\nin}$ and $\mathcal{E}:\R^{\nin}\to\LL$ is the kernel-evaluation operator defined by:
\begin{equation*}
    (\mathcal{E}c)(x) = \frac1\nin \sum_{i=1}^{\nin}\kntk(x,x_i)c_i,\quad\text{for }c\in\mathbb{R}^{\nin}\text{ and }x\in\mathcal{X}.
\end{equation*}

\textbf{Control of $\norm{M_t}_{\mathrm{op}}$.} $M_t=g_t(\Gram)$ is a function of $\Gram$ with scalar filter $g_t(\lambda)=\frac{\nin}{\lambda}(1-\exp(-\frac t\nin\lambda))$. Since $\Gram$ is symmetric positive definite on $\mathcal{E}^h_{\lambda_0}$, $\norm{M_t}_{\mathrm{op}}=\max_{1\leq i\leq\nin} g_t(\lambda_i)$ where $\{\lambda_i\}_{i=1}^{\nin}$ are the eigenvalues of $\Gram$. Using $0\le1-\exp(-a)\le1$ provided that $a\geq 0$, we obtain:
\begin{equation*}
    g_t(\lambda)=\frac{\nin}{\lambda}\seq{1-\exp\left(-\frac t\nin\lambda\right)}\le\frac{\nin}{\lambda},\quad\text{hence }\norm{M_t}_{\mathrm{op}}\le\frac{\nin}{\min_{1\leq i\leq\nin} \lambda_i}=\frac{\nin}{\lambda_0},
\end{equation*}
a bound independent of $t$.

\textbf{Control of $\norm{\mathcal{E}}_{\ell^2\to\LL}$.} Fix $c\in\R^{\nin}$ and $x\in\X$. By the Cauchy-Schwarz inequality and $\abs{\kntk(x,x_i)}\le\kappa^{\mathrm{NTK}}$ (\cref{lem:bounded_ntk_kernel}), we get:
\begin{equation*}
    \abs{\frac1\nin\sum_{i=1}^{\nin} \kntk(x,x_i)c_i}\le\frac1\nin\norm{\kntk(x,\boldsymbol{X})}\norm{c}\le\frac{\kappa^{\mathrm{NTK}}}{\sqrt{\nin}}\norm{c},
\end{equation*}
where $\kntk(x,\boldsymbol{X})=[\kntk(x,x_1),\dots,\kntk(x,x_{\nin})]^\top$. Since the marginal on $\X$ is a probability measure, we have:
\begin{equation*}
    \norm{\mathcal{E}c}_{\LL}^2\le\seq{\frac{\kappa^{\mathrm{NTK}}}{\sqrt{\nin}}}^2 \norm{c}^2,\text{ which implies that }\norm{\mathcal{E}}_{\ell^2\to\LL}\le\frac{\kappa^{\mathrm{NTK}}}{\sqrt{\nin}}.
\end{equation*}
By submultiplicativity of the operator norm, we get:
\begin{equation*}
\norm{\Phi_t}_{\ell^2\to\LL}\le\norm{\mathcal{E}}_{\ell^2\to\LL}\norm{M_t}_{\mathrm{op}}\le\frac{\kappa^{\mathrm{NTK}}}{\sqrt{\nin}}\cdot\frac{\nin}{\lambda_0}=\frac{\kappa^{\mathrm{NTK}}\sqrt{\nin}}{\lambda_0}.\qedhere
\end{equation*}
\end{proof}

\begin{lemma}[Lipschitz continuity in $\omega$ of the inner NTK error]\label{lem:lipschitz-omega}
Let \Cref{ass:bound_in,ass:bound_la,ass:gome} hold. On the event $\mathcal{E}^h_{\lambda_0}$, for any $t\geq0$, the map $\Psi:\omega\mapsto\hntktw-h^\star_\omega$ is $L_\Psi$-Lipschitz continuous from $\Omega$ to $\LL$ with:
\begin{equation*}
L_\Psi\coloneqq G\seq{1+\frac{\kappa^{\mathrm{NTK}}\,\nin}{\lambda_0}},\text{ with }\kappa^{\mathrm{NTK}}\coloneqq(p+2)(A^2+1)+1.
\end{equation*}
\end{lemma}

\begin{proof}
Let $t\geq0$ and $\omega,\omega'\in\Omega$. By the triangle inequality, we have:
\begin{equation*}
    \norm{\Psi(\omega)-\Psi(\omega')}_{\LL}\le\norm{\hntktw-h^{\mathrm{NTK}}_{t,\omega'}}_{\LL}+\norm{h^\star_\omega-h^\star_{\omega'}}_{\LL}.
\end{equation*}
For the target term, \cref{prop:lip_con_hstar} gives $\norm{h^\star_\omega-h^\star_{\omega'}}_{\LL}\le G\norm{\omega-\omega'}$. For the estimator term, \cref{lem:phi-linear} and the linearity of $\Phi_t$ give $\hntktw-h^{\mathrm{NTK}}_{t,\omega'}=\Phi_t(\Yom-Y_{\omega'})$, hence by \cref{lem:phi-norm}:
\begin{equation*}
    \norm{\hntktw-h^{\mathrm{NTK}}_{t,\omega'}}_{\LL}\le\norm{\Phi_t}_{\ell^2\to\LL}\norm{\Yom-Y_{\omega'}}\le\frac{\kappa^{\mathrm{NTK}}\sqrt{\nin}}{\lambda_0}\norm{\Yom-Y_{\omega'}}.
\end{equation*}
Since the open set $U\supset\Omega$ of \cref{ass:gome} is convex, the segment $\omega_s\coloneqq \omega'+s(\omega-\omega')$, $s\in[0,1]$, lies in $U$. Hence, applying the Cauchy-Schwarz inequality and \cref{ass:gome}, we get for each $1\leq i\leq \nin$:
\begin{equation*}
    \left|g_\omega(z_i)-g_{\omega'}(z_i)\right|=\left|\int_0^1\left\langle\partial_\omega g_{\omega_s}(z_i),\omega-\omega'\right\rangle\mathrm{d}s\right|\leq G\left\|\omega-\omega'\right\|,
\end{equation*}
and therefore $\norm{\Yom-Y_{\omega'}}\le\sqrt{\nin}G\norm{\omega-\omega'}$. Combining both, we get:
\begin{equation*}
    \norm{\hntktw-h^{\mathrm{NTK}}_{t,\omega'}}_{\LL}\le\frac{\kappa^{\mathrm{NTK}}\nin}{\lambda_0}G\norm{\omega-\omega'}.    
\end{equation*}
Adding the two contributions concludes the proof.
\end{proof}

\subsection{Generalization bounds for the inner solution, regression and adjoint functions}\label{subsec:gen_reg_exp}

\begin{proposition}[Uniform-in-$\omega$ expected error in the inner solution]\label{prop:err_h}
Consider $\nin$ i.i.d. samples $\{(x_i,z_i)\}_{i=1}^{\nin}\sim\mathbb{P}_{XZ}$ and assume $p\geq 2$. Fix a compact $\Omega\subset\R^d$ and let \Cref{ass:bound_in,ass:gome,ass:bound_tar,ass:bounded_density_P} hold. If we choose a stopping time $t_{\mathrm{op}_h}=\nin^{\frac{p+1}{s_h(p+1)+p}}$ with $\frac{1}{p+1}<s_h\leq 2$, then, considering the early-stopped truncated neural network predictor $\hat{h}_\omega$ for the inner solution $h^\star_\omega$, there exist a positive constant $C_1>0$ and a polynomial $\mathrm{poly}_h$ such that when $\nin$ is sufficiently large and the width satisfies $m_h\geq\mathrm{poly}_h(\nin,\ln\nin)$, the following holds:
\begin{equation*}
    \mathbb{E}\left[\sup_{\omega\in\Omega}\norm{\hat{h}_\omega-h^\star_\omega}_{\LL}^2\right]\leq C_1d^4(\ln\nin)^4\nin^{-\frac{s_h(p+1)}{s_h(p+1)+p}},
\end{equation*}
where the constant $C_1>0$ is independent of $\nin$ and $t_{\mathrm{op}_h}$, but depends on $s_h$, $p$, $A$, $D$, $G$, $M_\mathbb{P}$, $\mathrm{diam}(\Omega)$, and $R_h$, and the coefficients of $\mathrm{poly}_h$ depend on $s_h$, $A$, $D$, $G$, $M_\mathbb{P}$, $\mathrm{diam}(\Omega)$, $p$, and $R_h$. The expectation here is with respect to random samples and random initialization.
\end{proposition}

\begin{proof}
Throughout, we write $r_h\coloneqq\frac12 s_h\gamma_h$ with $\gamma_h=\frac{p+1}{s_h(p+1)+p}$. By the triangle inequality and the identity $(a+b)^2\le 2(a^2+b^2)$, we have:
\begin{align}\label{eq:u-split}
    \E\left[\sup_{\omega\in\Omega}\norm{\hat{h}_\omega-h^\star_\omega}_{\LL}^2\right]\le&2\Bigg(\E\left[\sup_{\omega\in\Omega}\norm{\hat{h}_\omega-\hntktwop}_{\LL}^2\right]\nonumber\\
    &\qquad+\E\left[\sup_{\omega\in\Omega}\norm{\hntktwop-h^\star_\omega}_{\LL}^2\right]\Bigg).
\end{align}

\textbf{Upper-bound on $\mathbb{E}\left[\sup_{\omega\in\Omega}\left\|\hat{h}_\omega-\hntktwop\right\|_{\LL}^2\right]$.} The bound on this term is obtained by following the first part of the proof of \cref{prop:exp_gen_bound}, the only difference being the presence of the supremum over $\Omega$. The supremum is harmless here for two reasons. First, the favorable event $\mathcal{E}^h$ is independent of $\omega$: the truncated predictor $\hat{h}_\omega$ therefore vanishes on $(\mathcal{E}^h)^\complement$ for all $\omega\in\Omega$ at once, so the decomposition over $\mathcal{E}^h$ and $(\mathcal{E}^h)^\complement$ and the ensuing control of $\mathrm{Pr}((\mathcal{E}^h)^\complement)$ are unchanged. Second, the outer variable enters the argument only through the uniform target bound $\abs{g_\omega(z_i)}\le D$ (\cref{ass:gome}), so all constants remain $\omega$-independent and each estimate holds uniformly over $\Omega$. We thus obtain, exactly as in \cref{prop:exp_gen_bound}:
\begin{equation}\label{eq:u-termI-final}
    \mathbb{E}\left[\sup_{\omega\in\Omega}\norm{\hat{h}_\omega-\hntktwop}_{\LL}^2\right]\le c_1^2m_h^{-\frac{1}{12}}\ln m_h+4(\kappa^{\mathrm{NTK}})^2D^2\nin^{2\gamma_h}\delta,
\end{equation}
where $\delta\in(0,1)$, $\kappa^{\mathrm{NTK}}=(p+2)(A^2+1)+1$, and $c_1>0$ is a constant that is independent of $\delta,\nin$, and $t_{\mathrm{op}_h}$, but depends on $A$. We note that \cref{eq:u-termI-final} is valid whenever $m_h\geq\mathrm{poly}(\nin,1/\delta,\ln\nin,\ln(1/\delta))$, where the coefficients of $\mathrm{poly}$ depend on $A$, $D$, $M_\mathbb{P}$, and $p$.

\textbf{Upper-bound on $\mathbb{E}\left[\sup_{\omega\in\Omega}\left\|\hntktwop-h^\star_\omega\right\|_{\LL}^2\right]$.} Let $\lambda_\star\coloneqq\lambda_\star(\nin,\delta)>0$ denote the threshold in the definition of $\mathcal{E}_{\lambda_0}^h$, so that $\mathcal{E}_{\lambda_0}^h=\{\lambda_0\geq\lambda_\star\}$. It depends only on $\nin,\delta,p,A$ and $M_{\mathbb{P}}$, and not on $\mathcal{D}_{in}$. Define the deterministic constants:
\begin{equation*}
L_\Psi^\star\coloneqq G\seq{1+\frac{\kappa^{\mathrm{NTK}}\,\nin}{\lambda_\star}}\qquad\text{and}\qquad\varepsilon\coloneqq(L_\Psi^\star)^{-1}\nin^{-r_h}.
\end{equation*}
On $\mathcal{E}_{\lambda_0}^h$, the Lipschitz constant $L_\Psi$ of \cref{lem:lipschitz-omega} satisfies $L_\Psi\leq L_\Psi^\star$, since $\lambda_0\geq\lambda_\star$ there. Let $\mathcal{N}=\{\omega_1,\dots,\omega_N\}\subset\Omega$ be a maximal $\varepsilon$-separated subset of $\Omega$, \emph{i.e.}, $\|\omega_j-\omega_k\|\geq\varepsilon$ for all $j\neq k$ and no point of $\Omega$ can be added to $\mathcal{N}$ without violating this property. By maximality, every $\omega\in\Omega$ satisfies $\|\omega-\omega_k\|<\varepsilon$ for some $1\leq k\leq N$, so $\mathcal{N}$ is an $\varepsilon$-net of $\Omega$ whose points lie in $\Omega$. In particular, \cref{lem:lipschitz-omega} applies to every pair $(\omega,\omega_k)$. Note that since $\varepsilon$ is deterministic, $\mathcal{N}$ and $N$ are fixed before the data are drawn, and are therefore independent of $\mathcal{D}_{in}$. Let also $\mathcal{E}_{\mathrm{net}}\coloneqq\bigcap_{k=1}^{N}\mathcal{E}^{\mathrm{NTK}}_{\omega_k}$, where, for each $1\leq k\leq N$, $\mathcal{E}^{\mathrm{NTK}}_{\omega_k}$ is the event of \cref{lem:adap_gen_ntk} at
$\omega_k$ with deviation parameter $\delta/N$, namely:
\begin{equation*}
    \mathcal{E}^{\mathrm{NTK}}_{\omega_k}\coloneqq\left\{\left\|h^{\mathrm{NTK}}_{t_{\mathrm{op}_h},\omega_k}-h^\star_{\omega_k}\right\|_{\LL}\leq c_2\left(\ln\frac{6N}{\delta}\right)^2 \nin^{-r_h}\right\},
\end{equation*}
where $c_2>0$ is a constant that is independent of $\delta$, $\nin, t_{\mathrm{op}_h}$, and $\omega_k$, but depends on $s_h$, $p$, $A$, $D$, $M_\mathbb{P}$, and $R_h$. Since $\omega_k$ is deterministic, \cref{lem:adap_gen_ntk} applies to the regression problem with target $h^\star_{\omega_k}$ and gives $\mathrm{Pr}\big((\mathcal{E}^{\mathrm{NTK}}_{\omega_k})^\complement\big)\le\delta/N$. Therefore, by a union bound, $\mathrm{Pr}(\mathcal{E}_{\mathrm{net}}^\complement)\le\delta$. On $\mathcal{E}_{\mathrm{net}}\cap \mathcal{E}^h_{\lambda_0}$, for every $\omega\in\Omega$ and its nearest net point $\omega_k$, \emph{i.e.}, $\|\omega-\omega_k\|\leq\varepsilon$, applying \cref{lem:lipschitz-omega} together with $L_\Psi\leq L_\Psi^\star$ gives:
\begin{equation*}
    \norm{\hntktwop-h^\star_\omega}_{\LL}\le\norm{h^{\mathrm{NTK}}_{t_{\mathrm{op}_h},\omega_k}-h^\star_{\omega_k}}_{\LL}+L_\Psi^\star\varepsilon\leq c_2\left(\ln\frac{6N}{\delta}\right)^2\nin^{-r_h}+\nin^{-r_h},
\end{equation*}
by the choice $L_\Psi^\star\varepsilon=\nin^{-r_h}$. Setting $\mathscr{F}\coloneqq\mathcal{E}_{\mathrm{net}}\cap \mathcal{E}^h_{\lambda_0}$, so that $\mathrm{Pr}(\mathscr{F}^\complement)\leq2\delta$, decomposing over $\mathscr{F}$ and $\mathscr{F}^\complement$, and using the identity $(a+b)^2\le2(a^2+b^2)$, we obtain:
\begin{align}\label{eq:u-termII}
&\mathbb{E}\left[\sup_{\omega\in\Omega}\norm{\hntktwop-h^\star_\omega}_{\LL}^2\right]\nonumber\\
=&\mathbb{E}\left[\sup_{\omega\in\Omega}\norm{\hntktwop-h^\star_\omega}_{\LL}^2\mathbbm{1}_\mathscr{F}\right]+\mathbb{E}\left[\sup_{\omega\in\Omega}\norm{\hntktwop-h^\star_\omega}_{\LL}^2\mathbbm{1}_{\mathscr{F}^\complement}\right]\nonumber\\
\leq& 2c_2^2\left(\ln\frac{6N}{\delta}\right)^4\nin^{-2r_h}+2\nin^{-2r_h}+4\left((\kappa^{\mathrm{NTK}})^2D^2\nin^{2\gamma_h}+D^2\right)\delta,    
\end{align}
where we apply \cref{lem:bound_ftntk,lem:bound_targets} on the event $\mathscr{F}^\complement$ to bound $\left\|\hntktwop\right\|_{\LL}^2$ by $(\kappa^{\mathrm{NTK}})^2D^2\nin^{2\gamma_h}$ and $\left\|h^\star_\omega\right\|_{\LL}^2$ by $D^2$ for any $\omega\in\Omega$.

\textbf{Expected generalization bound. }Plugging \cref{eq:u-termI-final,eq:u-termII} into \cref{eq:u-split} and recalling that $\nin^{-2r_h}=\nin^{-s_h\gamma_h}$, we obtain:
\begin{equation}\label{eq:assembled}
\E\left[\sup_{\omega\in\Omega}\norm{\hat{h}_\omega-h^\star_\omega}_{\LL}^2\right]\le 2c_1^2m_h^{-\frac{1}{12}}\ln m_h+4c_2^2\left(\ln\frac{6N}{\delta}\right)^4\nin^{-s_h\gamma_h}+4\nin^{-s_h\gamma_h}+ a_1\nin^{2\gamma_h}\delta,
\end{equation}
where $a_1\coloneqq 16(\kappa^{\mathrm{NTK}})^2D^2+8D^2$ collects the three $\delta$-terms, using $\nin^{2\gamma_h}\ge1$. For each fixed $\delta\in(0,1)$, the net above is deterministic, so \cref{eq:assembled} holds for every $\delta\in(0,1)$. We now control the $\delta$-dependent terms. The balls $B(\omega_k,\varepsilon/2)$, $1\leq k\leq N$, are pairwise disjoint and are all contained in $B(\omega_1,\mathrm{diam}(\Omega)+\varepsilon/2)$, since $\|\omega_k-\omega_1\|\leq\mathrm{diam}(\Omega)$. Comparing volumes, and using $\varepsilon\leq\mathrm{diam}(\Omega)$ for $\nin$ large, we obtain by \citep[Proposition~4.2.10]{Vershynin_2026}:
\begin{equation*}
    N\le\left(1+\frac{2\mathrm{diam}(\Omega)}{\varepsilon}\right)^{d}\le\left(\frac{3\mathrm{diam}(\Omega)}{\varepsilon}\right)^{d},\text{ hence }\ln N\le d\Big[\ln\big(3\mathrm{diam}(\Omega)\big)+\ln L_\Psi^\star+r_h\ln \nin\Big].
\end{equation*}
By \cref{cor:up_bound_lam_in}, the deterministic threshold $\lambda_\star$ satisfies:
\begin{equation*}
    \lambda_\star^{-1}\leq c_3\nin^\frac{8}{p}\delta^{-\frac{4}{p}}\left(\ln\nin+\ln\frac{1}{\delta}\right)^3,
\end{equation*}
where $c_3>0$ depends only on $p,A$ and $M_\mathbb{P}$. Since $L_\Psi^\star=G(1+\kappa^{\mathrm{NTK}}\nin/\lambda_\star)\le 2G\kappa^{\mathrm{NTK}}\nin\,\lambda_\star^{-1}$ for $\nin$ large, taking logarithms, and using the bound on $\lambda_\star^{-1}$ above and the inequality $\ln y\leq y$ for $y>0$ to bound the last term, we get:
\begin{align*}
    \ln L_\Psi^\star\le&\ln\left(2G\kappa^{\mathrm{NTK}}\right)+\ln\nin+\ln\lambda_\star^{-1}\\
    \le&\ln\left(2G\kappa^{\mathrm{NTK}}c_3\right)+\left(1+\frac8p\right)\ln\nin+\frac4p\ln\frac1\delta+3\ln\left(\ln\nin+\ln\frac1\delta\right)\\
    \leq&C_\Psi\left(\ln\nin+\ln\frac{1}{\delta}\right),    
\end{align*}
with $C_\Psi\coloneqq\left|\ln(2G\kappa^
\mathrm{NTK}c_3)\right|+4+\frac{8}{p}>0$, where the last inequality uses $\ln\nin\geq1$ for $\nin\geq3$. Using also $r_h\le1$ and $\ln\nin\geq1$, we obtain:
\begin{equation*}
    \ln N\le cd\left(\ln\nin+\ln\frac1\delta\right),\qquad\text{with}\qquad c\coloneqq\left|\ln(3\mathrm{diam}(\Omega))\right|+C_\Psi+1.
\end{equation*}
This implies that $\ln\frac{6N}{\delta}=\ln6+\ln N+\ln\frac1\delta\le c'd\big(\ln\nin+\ln\frac1\delta\big)$ with $c'\coloneqq c+2$, using $\ln 6\leq2\ln\nin$ for $\nin\geq3$ and $d\geq1$, where $c'>0$ is a constant depending on $\mathrm{diam}(\Omega),p,A,M_\mathbb{P}$, and $G$. Hence, using the identity $(u+v)^4\le 8(u^4+v^4)$ and $\ln\frac{1}{\delta}\leq\ln\frac{6}{\delta}$, we obtain:
\begin{equation*}
    \left(\ln\frac{6N}{\delta}\right)^4\leq 8c'^4 d^4\left[(\ln\nin)^4+\left(\ln\frac1\delta\right)^4\right]\leq 8c'^4 d^4\left[(\ln\nin)^4+\left(\ln\frac6\delta\right)^4\right].
\end{equation*}
Substituting into \cref{eq:assembled}, the $\delta$-dependent part of the bound is at most $\varphi(\delta)$, where:
\begin{equation*}
\varphi(\delta)\coloneqq a_2d^4\left(\ln\frac6\delta\right)^4\nin^{-s_h\gamma_h}+a_1\nin^{2\gamma_h}\delta,\qquad a_2\coloneqq 32c_2^2c'^4,
\end{equation*}
and the remaining, $\delta$-free part is $2c_1^2m_h^{-\frac{1}{12}}\ln m_h+\big(a_2d^4(\ln \nin)^4+4\big)\nin^{-s_h\gamma_h}$. Minimizing $\varphi$ over $\delta \in (0,1)$ exactly as in the proof of \cref{prop:exp_gen_bound} gives:
\begin{equation}\label{eq:deltastarh}
    \delta^\star=\frac{108a_2 d^4}{a_1}n_{in}^{-(s_h+2)\gamma_h}W_0^3(z_h),\qquad z_h\coloneqq\left(\frac{a_1}{18a_2 d^4}\right)^{\frac{1}{3}}n_{in}^{\frac{1}{3}(s_h+2)\gamma_h},
\end{equation}
and $\varphi(\delta^\star)=27a_2 d^4W_0^3(z_h)\big(3W_0(z_h)+4\big) n_{in}^{-s_h \gamma_h}$, where $W_0$ is the principal branch of the Lambert $W$ function. Since $d\geq1$, we have $z_h \leq (a_1/(18 a_2))^{1/3} n_{in}^{(s_h+2)\gamma_h/3}$, so \cref{lem:b_w0} bounds $W_0^3(z_h)(3W_0(z_h)+4)$ by a constant multiple of $(\ln n_{in})^4$ that does not involve $d$. This yields the rate $d^4 (\ln n_{in})^4 n_{in}^{-s_h\gamma_h}$. For the width, note that taking $\nin$ large enough so that $\nin^{(s_h+2)\gamma_h} \geq 18 e^3 a_2 d^4 / a_1$ gives $z_h \geq e$, hence $W_0(z_h) \geq 1$ and
\begin{equation*}
    \delta^\star\geq\frac{108\, a_2 d^4}{a_1} n_{in}^{-(s_h+2)\gamma_h}\geq\frac{108\, a_2}{a_1}n_{in}^{-(s_h+2)\gamma_h},
\end{equation*}
so that, using $(s_h+2)\gamma_h \leq 4$ and $\ln n_{in} \geq 1$, we obtain:
\begin{equation*}
    \frac{1}{\delta^\star}\leq \frac{a_1}{108a_2}n_{in}^{4},\qquad\ln\frac{1}{\delta^\star}\leq\left( \left|\ln\frac{a_1}{108a_2} \right|+4\right)\ln n_{in}.
\end{equation*}
Both bounds are free of $d$, so the condition $m_h \geq \mathrm{poly}_h(n_{in}, 1/\delta^\star, \ln n_{in}, \ln(1/\delta^\star))$ is implied by $m_h \geq \mathrm{poly}_h(n_{in}, \ln n_{in})$. As a consequence, we obtain:
\begin{equation*}
    \mathbb{E}\left[\sup_{\omega\in\Omega}\norm{\hat{h}_\omega-h^\star_\omega}_{\LL}^2\right]\leq C_1d^4(\ln\nin)^4\nin^{-s_h\gamma_h},
\end{equation*}
provided that $m_h\geq\mathrm{poly}_h(\nin,\ln\nin)$, where $C_1>0$ is a constant independent of $\nin$ and $t_{\mathrm{op}_h}$, but depends on $s_h$, $p$, $A$, $D$, $G$, $M_\mathbb{P}$, $\mathrm{diam}(\Omega)$, and $R_h$, and the coefficients of $\mathrm{poly}_h$ depend on $s_h$, $A$, $D$, $G$, $M_\mathbb{P}$, $\mathrm{diam}(\Omega)$, $p$, and $R_h$. This concludes the proof.
\end{proof}

\begin{proposition}[Expected error in the regression function]\label{prop:err_cond_exp}
    Consider $\nout$ i.i.d. samples $\{(\tilde{x}_j,\tilde{y}_j)\}_{j=1}^{\nout}\sim\mathbb{P}_{XY}$ and assume $p\geq 2$. Let \Cref{ass:bound_in,ass:bound_la,ass:adj_source,ass:bounded_density_P} hold. If we choose a stopping time $t_{\mathrm{op}_b}=\nout^{\frac{p+1}{s_b(p+1)+p}}$ with $\frac{1}{p+1}<s_b\leq 2$, then, considering the early-stopped truncated neural network predictor $\hat{b}$ for the regression function $b^\star$, there exist a positive constant $C_2>0$ and a polynomial $\mathrm{poly}_b$ such that when $\nout$ is sufficiently large and the width satisfies $m_b\geq\mathrm{poly}_b(\nout,\ln\nout)$, the following holds:
    \begin{equation*}
        \mathbb{E}\left[\left\|\hat{b}-b^\star\right\|_{\LL}^2\right]\leq C_2\left(\ln\nout\right)^4 \nout^{-\frac{s_b(p+1)}{s_b(p+1)+p}},
    \end{equation*}
    where $C_2>0$ is a constant that is independent of $\nout$ and $t_{\mathrm{op}_b}$, but depends on $s_b$, $p$, $A$, $B$, $M_\mathbb{P}$, and $R_b$, and the coefficients of $\mathrm{poly}_b$ depend on $s_b$, $A$, $B$, $M_\mathbb{P}$, $R_b$, and $p$. The expectation here is with respect to random samples and random initialization.
\end{proposition}

\begin{proof}
    This is a direct consequence of \Cref{prop:exp_gen_bound}, where the neural network approximating $b^\star$ of the form in \cref{eq:nn_main} is trained using the empirical loss $\widehat{L}_{out}$, defined for any $h\in\LL$ by:
    \begin{equation*}
        \widehat{L}_{out}(h)=\frac{1}{2\nout}\sum_{j=1}^{\nout}\left(h(\tilde{x}_j)-\tilde{y}_j\right)^2.\qedhere
    \end{equation*}
\end{proof}

\begin{proposition}[Uniform-in-$\omega$ expected error in the adjoint function]\label{prop:expected_adjoint_error}
    Consider $\nin$ i.i.d. samples $\{(x_i,z_i)\}_{i=1}^{\nin}\sim\mathbb{P}_{XZ}$, $\nout$ i.i.d. samples $\{(\tilde{x}_j,\tilde{y}_j)\}_{j=1}^{\nout}\sim\mathbb{P}_{XY}$, and assume $p\geq 2$. Let \Cref{ass:bound_in,ass:bound_la,ass:adj_source,ass:bound_tar,ass:gome,ass:bounded_density_P} hold. If we choose stopping times $t_{\mathrm{op}_h}=\nin^{\frac{p+1}{s_h(p+1)+p}}$ and $t_{\mathrm{op}_b}=\nout^{\frac{p+1}{s_b(p+1)+p}}$ with $\frac{1}{p+1}<s_h\leq 2$ and $\frac{1}{p+1}< s_b\leq 2$, then, for every $\omega\in\Omega$, considering the early-stopped truncated neural network predictors $\hat{h}_\omega$ and $\hat{b}$ for the inner solution $h^\star_\omega$ and the regression function $b^\star$, respectively, there exist two positive constants $C_1,C_2>0$ and two polynomials $\mathrm{poly}_h$ and $\mathrm{poly}_b$ such that when $\nin$ and $\nout$ are sufficiently large and the widths satisfy $m_h\geq\mathrm{poly}_h(\nin,\ln\nin)$ and $m_b\geq\mathrm{poly}_b(\nout,\ln\nout)$, the following holds:
\begin{equation*}
    \mathbb{E}\left[\sup_{\omega\in\Omega}\left\|\hat{a}_\omega-a^\star_\omega\right\|_{\LL}^2\right]\leq 2C_1d^4\left(\ln\nin\right)^4 \nin^{-\frac{s_h(p+1)}{s_h(p+1)+p}}+2C_2\left(\ln\nout\right)^4 \nout^{-\frac{s_b(p+1)}{s_b(p+1)+p}},
\end{equation*}
where $\hat{a}_\omega=-\hat{h}_\omega+\hat{b}$ and $a^\star_\omega=-h^\star_\omega+b^\star$. $C_1,C_2>0$ and $\mathrm{poly}_h,\mathrm{poly}_b$ are the same as in \cref{prop:err_h,prop:err_cond_exp}, respectively. The expectation here is with respect to random samples and random initialization.
\end{proposition}

\begin{proof}
    Let $\omega\in\Omega$. Using the triangle inequality and the identity $(a+b)^2\leq 2(a^2+b^2)$, we have:
\begin{align*}
    \left\|\hat{a}_\omega-a^\star_\omega\right\|_{\LL}^2=&\left\|\left(-\hat{h}_\omega+\hat{b}\right)-\left(-h^\star_\omega+b^\star\right)\right\|_{\LL}^2\\
    \leq&2\left(\left\|\hat{h}_\omega-h^\star_\omega\right\|_{\LL}^2+\left\|\hat{b}-b^\star\right\|_{\LL}^2\right).
\end{align*}
Taking the supremum over $\omega\in\Omega$ followed by the expectation over the random samples and random initialization on both sides, we obtain:
\begin{equation}\label{eq:adjoint_gene}
    \mathbb{E}\left[\sup_{\omega\in\Omega}\left\|\hat{a}_\omega-a^\star_\omega\right\|_{\LL}^2\right]\leq2\left(\mathbb{E}\left[\sup_{\omega\in\Omega}\left\|\hat{h}_\omega-h^\star_\omega\right\|_{\LL}^2\right]+\mathbb{E}\left[\left\|\hat{b}-b^\star\right\|_{\LL}^2\right]\right).
\end{equation}
By \Cref{prop:err_h}, we know that there exist a positive constant $C_1>0$ and a polynomial $\mathrm{poly}_h$ such that when $\nin$ is sufficiently large and $m_h\geq\mathrm{poly}_h(\nin,\ln \nin)$, the following holds:
\begin{equation*}
    \mathbb{E}\left[\sup_{\omega\in\Omega}\left\|\hat{h}_\omega-h^\star_\omega\right\|_{\LL}^2\right]\leq C_1d^4\left(\ln\nin\right)^4 \nin^{-\frac{s_h(p+1)}{s_h(p+1)+p}},
\end{equation*}
where the exact dependence of $C_1$ and the coefficients of $\mathrm{poly}_h$ are specified in \Cref{prop:err_h}. Additionally, by \Cref{prop:err_cond_exp}, we know that there exist a positive constant $C_2>0$ and a polynomial $\mathrm{poly}_b$ such that when $\nout$ is sufficiently large and $m_b\geq\mathrm{poly}_b(\nout,\ln\nout)$, the following holds:
\begin{equation*}
    \mathbb{E}\left[\left\|\hat{b}-b^\star\right\|_{\LL}^2\right]\leq C_2\left(\ln\nout\right)^4 \nout^{-\frac{s_b(p+1)}{s_b(p+1)+p}},
\end{equation*}
where the exact dependence of $C_2>0$ and the coefficients of $\mathrm{poly}_b$ are provided in \Cref{prop:err_cond_exp}. After combining both upper-bounds in \Cref{eq:adjoint_gene}, we get:
\begin{equation*}
    \mathbb{E}\left[\sup_{\omega\in\Omega}\left\|\hat{a}_\omega-a^\star_\omega\right\|_{\LL}^2\right]\leq 2C_1d^4\left(\ln\nin\right)^4 \nin^{-\frac{s_h(p+1)}{s_h(p+1)+p}}+2C_2\left(\ln\nout\right)^4 \nout^{-\frac{s_b(p+1)}{s_b(p+1)+p}}.\qedhere
\end{equation*}
\end{proof}
\section{Generalization Analysis of Neural Network Gradient Flow in the NTK Regime}\label{sec:ntk_single}
\subsection{Problem setting}
Let $p$ be a fixed positive integer, $\mathcal{X}\subset\mathbb{R}^p$, $\mathcal{Y}\subset\mathbb{R}$, and $\rho$ be a \emph{fixed} but \emph{unknown} probability distribution supported on $\mathcal{X}\times\mathcal{Y}$. We denote by $\mu$ the marginal distribution of $\rho$ on $\mathcal{X}$, which we assume is \emph{absolutely continuous} with respect to the Lebesgue measure on $\mathbb{R}^p$, \emph{i.e.}, it admits a \emph{density function} $f_\mu$. Let $L^2(\mathcal{X},\mu)$ and $L^\infty(\mathcal{X})$ denote respectively the spaces of \emph{square-integrable} with respect to $\mu$ and \emph{essentially bounded} functions defined on the input space $\mathcal{X}$ and taking values in $\mathcal{Y}$, endowed with the following norms:
\begin{equation*}
    \left\|f\right\|_{L^2(\mathcal{X},\mu)}=\left(\mathbb{E}_{x\sim\mu}\left[\left|f(x)\right|^2\right]\right)^\frac{1}{2},\quad \left\|f\right\|_\infty=\operatorname*{ess\,sup}\limits_{x\in\mathcal{X}}\left|f(x)\right|.
\end{equation*}
For conciseness, we write $\LL$ for $L^2(\mathcal{X},\mu)$. Suppose we are given a training set of $n$ \emph{i.i.d.} samples $\{(x_i,y_i)\}_{i=1}^n\subset\mathcal{X}\times\mathcal{Y}$ drawn from the distribution $\rho$. Using those $n$ samples, our goal is to \emph{learn a neural network} that approximates the regression function $f^\star:\mathcal{X}\to\mathcal{Y}$ in $\LL$, defined for any $x\in\mathcal{X}$ by $f^\star(x)=\mathbb{E}_{y\sim\rho}[y\mid x]$, which minimizes the following \emph{population loss} over $\LL$:
\begin{equation*}
    L(f)=\frac{1}{2}\mathbb{E}_{(x,y)\sim\rho}\left[\left|f(x)-y\right|^2\right].
\end{equation*}
Throughout \cref{sec:ntk_single}, we denote by $\boldsymbol{X}=[x_1,\ldots,x_n]^\top\in\mathbb{R}^{n\times p}$ the design matrix, by $\boldsymbol{Y}=[y_1,\ldots,y_n]^\top\in\mathbb{R}^{n}$ the target vector, and by $\mathbb{S}^{p}=\{x\in\mathbb{R}^{p+1}\mid\|x\|=1\}$ the unit sphere in $\mathbb{R}^{p+1}$.

\subsection{Neural network architecture and infinite-width training dynamics}\label{app:nna_iwdtr}
\textbf{Neural network architecture.} We consider a \emph{two-layer} ReLU neural network predictor $f^\mathrm{NN}_\theta$ for $f^\star$, defined for any $x\in\mathcal{X}$ by:
\begin{equation}\label{eq:shallow}
    f^\mathrm{NN}_\theta(x)=\frac{1}{\sqrt{m}}\sum_{i=1}^{m} v_i\sigma(\tilde{\boldsymbol{u}}_i^\top \tilde{x}) + a,\quad\text{with parameters }\theta=(\boldsymbol{v}, \boldsymbol{u}, \boldsymbol{b}, a)\in\Theta,
\end{equation}
where the augmented input is $\tilde{x}=[x^\top,1]^\top\in\mathbb{R}^{p+1}$ and the augmented inner weights $\tilde{\boldsymbol{u}}_i \in \mathbb{R}^{p+1}$ incorporate a neuron-wise bias $b_i$ via:
\begin{equation*}
    \tilde{\boldsymbol{u}}_i = \begin{bmatrix}\boldsymbol{u}_i\\b_i\end{bmatrix}\in\mathbb{R}^{p+1},
\end{equation*}
where $\boldsymbol{u}=[\boldsymbol{u}_1,\ldots,\boldsymbol{u}_m]^\top\in\mathbb{R}^{m\times p}$ are the inner layer weights, $\boldsymbol{b}=[b_1,\ldots,b_m]^\top\in\mathbb{R}^m$ are the inner layer biases, $\boldsymbol{v}=[v_1,\ldots,v_m]^\top\in\mathbb{R}^{m}$ are the outer layer weights, $a\in\mathbb{R}$ is the outer bias, $\Theta=\mathbb{R}^{m}\times\mathbb{R}^{m\times p}\times\mathbb{R}^{m}\times\mathbb{R}$ is the parameter space, and $\sigma$ is the ReLU activation function, defined as $\sigma(x)=\max(0,x)$. The derivative of $\sigma$ is $\sigma'(x)=1$ for $x>0$ and $\sigma'(x)=0$ for $x<0$. Since $\sigma$ is not differentiable at $0$, we adopt the convention $\sigma'(0)=0$; hence we write $\sigma'(x)=\mathbbm{1}_{x>0}$.

\textbf{Initialization.} We use a \emph{mirrored initialization} to ensure that the network output is exactly zero at initialization. Specifically, the $m$ neurons are arranged in $m/2$ pairs (assuming $m$ is even). For each pair $(i, i+m/2)$ with $1\leq i\leq m/2$, the inner weights and biases are initialized symmetrically:
\begin{equation*}
    \boldsymbol{u}_i = \boldsymbol{u}_{i+m/2} \sim \mathcal{N}(0,\mathbbm{1}_{p\times p}), \qquad b_i = b_{i+m/2} \sim \mathcal{N}(0, 1),
\end{equation*}
and the outer weights are initialized with opposite signs:
\begin{equation*}
    v_i = -v_{i+m/2} \sim \mathcal{N}(0,1).
\end{equation*}
The outer bias is initialized as $a=0$. Moreover, $v_i,\boldsymbol{u}_i$, and $b_i$ are drawn \emph{independently} from one another. Under this initialization, for each pair $(i, i+m/2)$ and any input $x \in \mathcal{X}$:
\begin{equation*}
    v_i\,\sigma(\tilde{\boldsymbol{u}}_i^\top\tilde{x}) + v_{i+m/2}\,\sigma(\tilde{\boldsymbol{u}}_{i+m/2}^\top\tilde{x}) = v_i\,\sigma(\tilde{\boldsymbol{u}}_i^\top\tilde{x}) - v_i\,\sigma(\tilde{\boldsymbol{u}}_i^\top\tilde{x}) = 0.
\end{equation*}
Summing over all pairs, the network output at initialization satisfies $f^\mathrm{NN}_\theta(x) = 0$ for all $x\in\mathcal{X}$.

\textbf{Training dynamics.} Given a training set of $n$ \emph{i.i.d.} samples $\{(x_i,y_i)\}_{i=1}^n\subset\mathcal{X}\times\mathcal{Y}$ drawn from the distribution $\rho$, we consider the following \emph{empirical loss}:
\begin{equation*}
    \widehat{L}(f^\mathrm{NN}_\theta)\coloneqq \widehat{\mathscr{L}}(\theta)=\frac{1}{2n}\sum_{i=1}^n\left(f^\mathrm{NN}_\theta(x_i)-y_i\right)^2,
\end{equation*}
and we train the network by \emph{gradient flow}, where the parameters are updated according to the following ordinary differential equation:
\begin{equation*}
    \frac{\text{d}}{\text{d}t}\theta_t=-\partial_\theta \widehat{\mathscr{L}}(\theta_t)=-\frac{1}{n}\sum_{i=1}^n\partial_\theta f_{\theta_t}^\mathrm{NN}(x_i)\left(f_{\theta_t}^\mathrm{NN}(x_i)-y_i\right),
\end{equation*}
where $\theta_t=(\boldsymbol{v}_t,\boldsymbol{u}_t, \boldsymbol{b}_t, a_t)$ and $\partial_\theta f_{\theta_t}^\mathrm{NN}(x_i)\in\mathbb{R}^{\mathrm{dim}(\Theta)}$ denotes the Jacobian. Using the chain rule, for any $x\in\mathcal{X}$, the network's output evolves as follows:
\begin{equation*}
    \frac{\text{d}}{\text{d}t}f_{\theta_t}^\mathrm{NN}(x)=\left(\partial_\theta f_{\theta_t}^\mathrm{NN}(x)\right)^\top\frac{\text{d}}{\text{d}t}\theta_t=-\frac{1}{n}\sum_{i=1}^n\left(\partial_\theta f_{\theta_t}^\mathrm{NN}(x)\right)^\top\partial_\theta f_{\theta_t}^\mathrm{NN}(x_i)\left(f_{\theta_t}^\mathrm{NN}(x_i)-y_i\right).
\end{equation*}
We define the \emph{finite-width Neural Tangent Kernel} (NTK) $K_\theta^m$ for any $x,x'\in\mathcal{X}$ by: 
\begin{equation}\label{eq:nnk}
    K_\theta^m(x,x')=\left(\partial_\theta f_{\theta}^\mathrm{NN}(x)\right)^\top\partial_\theta f_{\theta}^\mathrm{NN}(x')\in\mathbb{R}.
\end{equation}
Using this, the output dynamics becomes:
\begin{equation}\label{eq:grad_flow_ftnn}
    \frac{\text{d}}{\text{d}t}f_{\theta_t}^\mathrm{NN}(x)=-\frac{1}{n}\sum_{i=1}^n K_{\theta_t}^m(x,x_i)\left(f_{\theta_t}^\mathrm{NN}(x_i)-y_i\right).
\end{equation}
For the neural network defined above in \cref{eq:shallow}, $K_\theta^m$ admits an explicit expression. Indeed, computing the gradients with respect to the parameters for any $1\leq i\leq m$, we get:
\begin{equation*}
    \partial_{v_i} f^\mathrm{NN}_\theta(x) = \frac{1}{\sqrt{m}} \sigma(\tilde{\boldsymbol{u}}_i^\top \tilde{x}), 
\quad
\partial_{\tilde{\boldsymbol{u}}_i} f^\mathrm{NN}_\theta(x) = \frac{1}{\sqrt{m}} v_i \, \sigma'(\tilde{\boldsymbol{u}}_i^\top \tilde{x})\, \tilde{x},
\quad
\partial_{a} f^\mathrm{NN}_\theta(x) = 1.
\end{equation*}
Using the definition of $K_\theta^m$ in \cref{eq:nnk}, we obtain:
\begin{align}\label{eq:shallow_ntkf}
    K_\theta^m(x,x')=&\sum_{i=1}^m\left\langle \partial_{v_i} f_{\theta}^\mathrm{NN}(x), \partial_{v_i} f_{\theta}^\mathrm{NN}(x')\right\rangle+\sum_{i=1}^m \left\langle \partial_{\tilde{\boldsymbol{u}}_i} f_{\theta}^\mathrm{NN}(x), \partial_{\tilde{\boldsymbol{u}}_i} f_{\theta}^\mathrm{NN}(x') \right\rangle+1\nonumber\\
    =&\frac{1}{m} \sum_{i=1}^m \sigma(\tilde{\boldsymbol{u}}_i^\top \tilde{x})\sigma(\tilde{\boldsymbol{u}}_i^\top \tilde{x}')+\frac{1}{m} \sum_{i=1}^m v_i^2 \, \sigma'(\tilde{\boldsymbol{u}}_i^\top \tilde{x})\sigma'(\tilde{\boldsymbol{u}}_i^\top \tilde{x}') \, \tilde{x}^\top \tilde{x}'+1.
\end{align}

\textbf{Infinite-width limit.} When the network's width $m$ goes to $+\infty$, $K_{\theta_t}^m$ converges to a \emph{time-invariant deterministic} kernel $K^\mathrm{NTK}$, referred to as the \emph{infinite-width} NTK \citep{AllenZhu2019Convergence,jacot2018neural,lee2019wide}. Let $f_{\theta_t}^\mathrm{NTK}$ denote the function that describes the limiting training dynamics of $f_{\theta_t}^\mathrm{NN}$ when $m\to+\infty$. Then, in this regime, the dynamics of $f_{\theta_t}^\mathrm{NTK}$ are described by:
\begin{equation}\label{eq:grad_flow_ftntk}
    \frac{\text{d}}{\text{d}t}f_{\theta_t}^\mathrm{NTK}(x)=-\frac{1}{n}\sum_{i=1}^n K^\mathrm{NTK}(x,x_i)\left(f_{\theta_t}^\mathrm{NTK}(x_i)-y_i\right).
\end{equation}
Similarly to the neural network initialization, we also have that $f_{\theta_0}^\mathrm{NTK}=0$ due to the mirrored initialization. For the neural network defined above in \cref{eq:shallow}, $K^\mathrm{NTK}$ also admits an explicit expression. Indeed, as $m\to+\infty$, by the law of large numbers and the Gaussian initialization of the parameters, the kernel $K_\theta^m$, whose closed-form expression is provided in \cref{eq:shallow_ntkf}, converges to a deterministic kernel $K^\mathrm{NTK}$ given for any $x,x'\in\mathcal{X}$ by: 
\begin{align}\label{eq:shallow_ntk}
    K^\mathrm{NTK}(x,x')=& \mathbb{E}_{\tilde{\boldsymbol{u}} \sim \mathcal{N}(0,\mathbbm{1}_{(p+1)\times (p+1)})} \left[ \sigma(\tilde{\boldsymbol{u}}^\top \tilde{x})\sigma(\tilde{\boldsymbol{u}}^\top \tilde{x}') \right]\nonumber\\
    &+ \mathbb{E}_{v\sim\mathcal{N}(0,1),\tilde{\boldsymbol{u}} \sim \mathcal{N}(0,\mathbbm{1}_{(p+1)\times (p+1)})} \left[ v^2\sigma'(\tilde{\boldsymbol{u}}^\top \tilde{x})\sigma'(\tilde{\boldsymbol{u}}^\top \tilde{x}') \right] \tilde{x}^\top \tilde{x}'+1\nonumber\\
    =& \mathbb{E}_{\tilde{\boldsymbol{u}} \sim \mathcal{N}(0,\mathbbm{1}_{(p+1)\times (p+1)})} \left[ \sigma(\tilde{\boldsymbol{u}}^\top \tilde{x})\sigma(\tilde{\boldsymbol{u}}^\top \tilde{x}') \right]\nonumber\\
    &+ \mathbb{E}_{\tilde{\boldsymbol{u}} \sim \mathcal{N}(0,\mathbbm{1}_{(p+1)\times (p+1)})} \left[\sigma'(\tilde{\boldsymbol{u}}^\top \tilde{x})\sigma'(\tilde{\boldsymbol{u}}^\top \tilde{x}') \right] \tilde{x}^\top \tilde{x}'+1,
\end{align}
where in the second equality, we use the independence between $v$ and $\tilde{\boldsymbol{u}}$, as well as $\mathbb{E}_{v\sim\mathcal{N}(0,1)}[v^2]=1$. The NTK matrix $\boldsymbol{K}^\mathrm{NTK}=[K^\mathrm{NTK}(x_i,x_j)]_{i,j=1}^n$ can thus be expressed as: 
\begin{align}\label{eq:ntk_inf_matrix}
    \boldsymbol{K}^\mathrm{NTK}=&\mathbb{E}_{\tilde{\boldsymbol{u}}\sim\mathcal{N}(0,\mathbbm{1}_{(p+1)\times (p+1)})}\left[\sigma(\tilde{\boldsymbol{X}}\tilde{\boldsymbol{u}})\sigma(\tilde{\boldsymbol{u}}^\top\tilde{\boldsymbol{X}}^\top)\right]\nonumber\\
    &+\mathbb{E}_{\tilde{\boldsymbol{u}}\sim\mathcal{N}(0,\mathbbm{1}_{(p+1)\times (p+1)})}\left[\sigma'(\tilde{\boldsymbol{X}}\tilde{\boldsymbol{u}})\sigma'(\tilde{\boldsymbol{u}}^\top\tilde{\boldsymbol{X}}^\top)\right]\odot\tilde{\boldsymbol{X}}\tilde{\boldsymbol{X}}^\top+\mathbbm{1}_n\mathbbm{1}_n^\top,
\end{align}
where $\tilde{\boldsymbol{X}}=[\tilde{x}_1,\ldots,\tilde{x}_n]^\top\in\mathbb{R}^{n\times (p+1)}$ is the design matrix of the dataset $\{\tilde{x}_i\}_{i=1}^n$ and $\odot$ is the Hadamard product. In \cref{eq:ntk_inf_matrix}, $\sigma$ and $\sigma'$ are applied entrywise to vectors or matrices. Moreover, we have the following explicit formula for the infinite-width NTK \citep{cho2009kernel}:
\begin{equation}\label{eq:explicit_ntk}
    K^\mathrm{NTK}(x,x')=\frac{\|\tilde{x}\|\|\tilde{x}'\|}{2\pi}(\sin\varphi+(\pi-\varphi)\cos\varphi)+\frac{\pi-\varphi}{2\pi}\cdot\tilde{x}^\top\tilde{x}'+1,
\end{equation}
where $\varphi=\arccos\left(\frac{\tilde{x}^\top\tilde{x}'}{\|\tilde{x}\|\|\tilde{x}'\|}\right)$ is the angle between the augmented inputs $\tilde{x}=[x^\top,1]^\top$ and $\tilde{x}'=[x'^\top,1]^\top$. In the following, for notational simplicity, we denote $f_{\theta_t}^\mathrm{NN}$, $f_{\theta_t}^\mathrm{NTK}$, and $K_{\theta_t}^m$ by $\ftt$, $\fttntk$, and $K_t^m$, respectively. In the remainder of \cref{sec:ntk_single}, we denote by $\hntk$ the RKHS associated with the infinite-width kernel $\kntk$, by $[\hntk]^s$ its interpolation space, and by $(\lambda_i)_{i\geq 1}\subset(0,+\infty)$ the non-increasing sequence of non-zero eigenvalues of the integral operator $\mathcal{T}$ (equivalently, of the covariance operator $\Sigma$), with associated orthonormal family $(e_i)_{i\geq 1}$ in $\LL$. We refer the reader to \cref{sec:int_space} for details on interpolation spaces and the spectral decomposition. We further denote by $\lambda_0$ the smallest eigenvalue of the infinite-width NTK matrix $\boldsymbol{K}^\mathrm{NTK}=[K^\mathrm{NTK}(x_i,x_j)]_{i,j=1}^n\in\mathbb{R}^{n\times n}$, which controls the convergence of the NTK gradient flow. 
\subsection{Assumptions}\label{sec:ass_cons}
We make the following assumptions.
\begin{assumplista}
\item\label{ass:bound_in_res} (Compactness of $\mathcal{X}$) The input space $\mathcal{X}\subset\mathbb{R}^p$ is compact with a constant $A>0$ such that $\|x\|\leq A$ for any $x\in\mathcal{X}$.
\item\label{ass:bound_la_res} (Boundedness of $\mathcal{Y}$) The target space $\mathcal{Y}\subset\mathbb{R}$ is bounded, \emph{i.e.}, there exists a constant $B>0$ such that $|y|\leq B$ for any $y\in\mathcal{Y}$.
\item\label{ass:source_cond} (Source condition for $f^\star$) There exist $s,R>0$ such that $\left\|f^\star\right\|_{[\hntk]^s}\leq R$. 
\item\label{ass:bounded_density} (Bounded density) There exists a positive constant $M_\mu>0$ such that $f_\mu\leq M_\mu$. Moreover, there exist an open ball $\Omega_\mu\subset\mathcal{X}$ and a constant $c_\mu>0$ such that $f_\mu\geq c_\mu$ on $\Omega_\mu$.
\end{assumplista}

We now derive some consequences of these assumptions that will be useful later.

\begin{lemma}[Boundedness of $f^\star$]\label{ass:tar_b}
    Under \cref{ass:bound_la_res}, $f^\star\in L^\infty(\mathcal{X})$ and $\|f^\star\|_\infty\leq B$.
\end{lemma}
\begin{proof}
    For any $x\in\mathcal{X}$, we have $|f^\star(x)|=|\mathbb{E}_{y\sim\rho}[y\mid x]|\leq B$ by Jensen's inequality and \cref{ass:bound_la_res}. Taking the supremum over $x\in\mathcal{X}$ allows us to conclude.
\end{proof}

\begin{lemma}[Boundedness of $\kntk$]\label{lem:bounded_ntk_kernel}
    Under \cref{ass:bound_in_res}, $K^\mathrm{NTK}$ is bounded by $\kappa^\mathrm{NTK}\coloneqq(p+2)(A^2+1)+1$ on $\mathcal{X}\times\mathcal{X}$.
\end{lemma}

\begin{proof}
    We start by recalling the expression of $K^\mathrm{NTK}$ given in \cref{eq:shallow_ntk} for any $x,x'\in\mathcal{X}$ by:
    \begin{align*}
        K^\mathrm{NTK}(x,x')=& \mathbb{E}_{\tilde{\boldsymbol{u}} \sim \mathcal{N}(0,\mathbbm{1}_{(p+1)\times (p+1)})} \left[ \sigma(\tilde{\boldsymbol{u}}^\top \tilde{x})\sigma(\tilde{\boldsymbol{u}}^\top \tilde{x}') \right]\\
        &+ \mathbb{E}_{\tilde{\boldsymbol{u}} \sim \mathcal{N}(0,\mathbbm{1}_{(p+1)\times (p+1)})} \left[\sigma'(\tilde{\boldsymbol{u}}^\top \tilde{x})\sigma'(\tilde{\boldsymbol{u}}^\top \tilde{x}') \right] \tilde{x}^\top \tilde{x}'+1,
    \end{align*}
    By the triangle inequality, we have:
    \begin{align}\label{eq:b_ntk}
        \left|K^\mathrm{NTK}(x,x')\right|\leq& \left|\mathbb{E}_{\tilde{\boldsymbol{u}} \sim \mathcal{N}(0,\mathbbm{1}_{(p+1)\times (p+1)})} \left[ \sigma(\tilde{\boldsymbol{u}}^\top \tilde{x})\sigma(\tilde{\boldsymbol{u}}^\top \tilde{x}') \right]\right|\nonumber\\
        &+ \left|\mathbb{E}_{\tilde{\boldsymbol{u}} \sim \mathcal{N}(0,\mathbbm{1}_{(p+1)\times (p+1)})} \left[\sigma'(\tilde{\boldsymbol{u}}^\top \tilde{x})\sigma'(\tilde{\boldsymbol{u}}^\top \tilde{x}') \right] \tilde{x}^\top \tilde{x}'\right|+1,
    \end{align}
    Using the triangle inequality, the identities $\abs{\sigma(z)}\leq\abs{z}$ and $\sigma'(z)\leq 1$, Cauchy-Schwarz inequality, and \cref{ass:bound_in_res}, we get:
    \begin{align*}
        \left|\mathbb{E}_{\tilde{\boldsymbol{u}} \sim \mathcal{N}(0,\mathbbm{1}_{(p+1)\times (p+1)})} \left[ \sigma(\tilde{\boldsymbol{u}}^\top \tilde{x})\sigma(\tilde{\boldsymbol{u}}^\top \tilde{x}') \right]\right|\leq&\mathbb{E}_{\tilde{\boldsymbol{u}} \sim \mathcal{N}(0,\mathbbm{1}_{(p+1)\times (p+1)})} \left[ \left|\tilde{\boldsymbol{u}}^\top \tilde{x}\right|\left|\tilde{\boldsymbol{u}}^\top \tilde{x}'\right| \right]\\
        \leq&\underbrace{\mathbb{E}_{\tilde{\boldsymbol{u}} \sim \mathcal{N}(0,\mathbbm{1}_{(p+1)\times (p+1)})} \left[\|\tilde{\boldsymbol{u}}\|^2\right]}_{=p+1}\|\tilde{x}\|\|\tilde{x}'\|\\
        \leq& (p+1)(A^2+1),\\
        \text{and}\quad\left|\mathbb{E}_{\tilde{\boldsymbol{u}} \sim \mathcal{N}(0,\mathbbm{1}_{(p+1)\times (p+1)})} \left[\sigma'(\tilde{\boldsymbol{u}}^\top \tilde{x})\sigma'(\tilde{\boldsymbol{u}}^\top \tilde{x}') \right] \tilde{x}^\top \tilde{x}'\right|\leq&\|\tilde{x}\|\|\tilde{x}'\|\leq A^2+1.
    \end{align*}
    Substituting both inequalities into \cref{eq:b_ntk} concludes the proof.
\end{proof}

\begin{remark}
    As pointed out in \cref{sec:int_space}, \cref{lem:bounded_ntk_kernel} implies that the RKHS $\hntk$ is included in $\LL$. This allows us to consider the closure of $\hntk$ in $\LL$, as we do in \cref{lem:fstar_closure}.
\end{remark}

\begin{lemma}\label{lem:fstar_closure}
    Under \cref{ass:bound_in_res,ass:source_cond}, $f^\star$ lies in the closure of $\hntk$ in $\LL$.
\end{lemma}

\begin{proof}
     It suffices to show that there exists a sequence in $\hntk$ that converges to $f^\star$ in $\LL$. Under \cref{ass:source_cond}, we know that $f^\star\in[\hntk]^s$ for some $s>0$. Therefore, $f^\star$ can be written as (see \cref{sec:int_space}):
\begin{equation*}
    f^\star=\sum_{i\geq 1}a_i\lambda_i^\frac{s}{2}e_i,\qquad\text{with }\sum_{i\geq 1}a_i^2<+\infty.
\end{equation*}
Let $(h_n)_{n\geq 1}$ be a sequence in $\hntk$ defined for any $n\geq 1$ by:
\begin{equation*}
    h_n=\sum_{i=1}^n a_i\lambda_i^\frac{s}{2}e_i=\sum_{i=1}^n \left(a_i\lambda_i^\frac{s-1}{2}\right)\lambda_i^\frac{1}{2}e_i.
\end{equation*}
Indeed, $h_n\in\hntk$ for any $n\geq 1$ since $h_n$ is a finite linear combination of the basis functions $\lambda_i^\frac{1}{2}e_i$. Moreover, we have:
\begin{equation*}
    \left\|h_n-f^\star\right\|_{\LL}^2=\left\|\sum_{i>n}a_i\lambda_i^\frac{s}{2}e_i\right\|_{\LL}^2=\sum_{i>n}a_i^2\lambda_i^s\leq\lambda_1^s\sum_{i>n}a_i^2,
\end{equation*}
where the second equality follows from Parseval's identity, and the inequality uses the fact that the eigenvalues $(\lambda_i)_{i\geq 1}$ are sorted in non-increasing order. Given that the series $(a_i^2)_{i\geq 1}$ is convergent, we know that its tail $\sum_{i>n}a_i^2\longrightarrow 0$ as $n\to\infty$. Hence, we obtain:
\begin{equation*}
    \left\|h_n-f^\star\right\|_{\LL}^2\xrightarrow[n\to\infty]{} 0,
\end{equation*}
which shows that the sequence $(h_n)_{n\geq 1}$ converges to $f^\star$ in $\LL$. This concludes the proof.
\end{proof}

\subsection{Lower-bound on the smallest eigenvalue of \texorpdfstring{$\boldsymbol{K}^\mathrm{NTK}$}{K\textasciicircum NTK}}
Recall that we are given a dataset $x_1,\ldots,x_n\in\mathcal{X}\subset\mathbb{R}^p$ sampled from a distribution $\mu$ and an augmented dataset $\tilde{x}_1,\ldots,\tilde{x}_n\in\mathbb{R}^{p+1}$ where $\tilde{x}_i=[x_i^\top,1]^\top$ for $1\leq i\leq n$, with design matrices $\boldsymbol{X}=[x_1,\ldots,x_n]^\top\in\mathbb{R}^{n\times p}$ and $\tilde{\boldsymbol{X}}=[\tilde{x}_1,\ldots,\tilde{x}_n]^\top\in\mathbb{R}^{n\times(p+1)}$, respectively. Recall also that $\sigma$ is the ReLU activation function, and $\sigma'$ is its derivative. For $\psi\in\{\sqrt{p+1}\sigma,\sigma'\}$, let $\boldsymbol{K}_\psi^\infty\in\mathbb{R}^{n\times n}$ be the matrix defined as:
\begin{equation*}
    \boldsymbol{K}_\psi^\infty=\mathbb{E}_{w\sim\mathcal{U}(\mathbb{S}^{p})}\left[\psi(\hat{\boldsymbol{X}} w)\psi(w^\top\hat{\boldsymbol{X}}^\top)\right],
\end{equation*}
where $\mathcal{U}(\mathbb{S}^{p})$ denotes the uniform distribution on the unit sphere $\mathbb{S}^{p}$, $\psi$ is applied entrywise to the vector $\hat{\boldsymbol{X}} w\in\mathbb{R}^n$ and its transpose, with $\hat{\boldsymbol{X}}=[\hat{x}_1,\ldots,\hat{x}_n]^\top\in\mathbb{R}^{n\times(p+1)}$ being the design matrix of the normalized counterparts of $\tilde{x}_1,\ldots,\tilde{x}_n$, defined by:
\begin{equation}\label{eq:hom_norm_points}
    \hat{x}_i=\frac{\tilde{x}_i}{\|\tilde{x}_i\|}=\frac{1}{\sqrt{\|x_i\|^2+1}}\begin{bmatrix}x_i\\1\end{bmatrix}\in\mathbb{S}^p\quad\text{for }1\leq i\leq n.
\end{equation}

\begin{definition}[$\epsilon$-separated dataset \citep{karhadkar2024bounds}]\label{def:eps_sep}
    A dataset $\{x_i\}_{i=1}^n$ is $\epsilon$-separated for $\epsilon>0$ if:
    \begin{equation*}
        \min_{i\neq k}\min\left(\left\|x_i-x_k\right\|,\left\|x_i+x_k\right\|\right)\geq\epsilon.
    \end{equation*}
\end{definition}

\begin{lemma}[Lemma~7 in \citep{karhadkar2024bounds}]\label{lem:min_eig_sep}
    Let $\hat{\boldsymbol{X}}\in\mathbb{R}^{n\times(p+1)}$ be the design matrix of the dataset $\hat{x}_1,\ldots,\hat{x}_n\in\mathbb{S}^{p}$ with $p\geq 2$, and $\psi\in\{\sqrt{p+1}\sigma,\sigma'\}$. Assume $0<\epsilon\leq 1$. Suppose that this dataset is $\epsilon$-separated. Then, for all $w\in\mathbb{R}^n$ with $\|w\|= 1$, we have:
    \begin{equation*}
        \left\langle\boldsymbol{K}_\psi^\infty w,w\right\rangle\geq\begin{cases}
            C(1+g(\epsilon))^{-3}\epsilon^2\quad\text{if }\psi=\sigma'\\
            C(1+g(\epsilon))^{-3}\epsilon^4\quad\text{if }\psi=\sqrt{p+1}\sigma,
        \end{cases}
    \end{equation*}
    where $C>0$ is an absolute positive constant and $g(\epsilon)=\frac{(p+1)\ln(1/\epsilon)}{\ln (p+1)}$.
\end{lemma}

\begin{lemma}[Deterministic lower-bound on the smallest eigenvalue of $\boldsymbol{K}^\mathrm{NTK}$]\label{lem:sm_eig_ti_ntk}
    Let $\tilde{\boldsymbol{X}}\in\mathbb{R}^{n\times(p+1)}$ be the design matrix of the augmented dataset $\tilde{x}_i=[x_i^\top,1]^\top$ and $\hat{\boldsymbol{X}}\in\mathbb{R}^{n\times(p+1)}$ be the design matrix of the normalized dataset $\hat{x}_i=\tilde{x}_i/\|\tilde{x}_i\|$, with $p\geq 2$. Assume $0<\epsilon\leq 1$. Suppose that the normalized dataset $\hat{x}_1,\ldots,\hat{x}_n\in\mathbb{S}^p$ is $\epsilon$-separated. Then, we have the following lower-bound on the smallest eigenvalue $\lambda_0$ of $\boldsymbol{K}^\mathrm{NTK}$:
    \begin{equation*}
        \lambda_0\geq C(1+g(\epsilon))^{-3}\epsilon^4,
    \end{equation*}
    where $C>0$ is an absolute positive constant and $g(\epsilon)=\frac{(p+1)\ln(1/\epsilon)}{\ln(p+1)}$.
\end{lemma}

\begin{proof}
    By \cref{eq:ntk_inf_matrix}, the infinite-width NTK can be written as:
    \begin{align*}
        \boldsymbol{K}^\mathrm{NTK}=&\mathbb{E}_{\tilde{\boldsymbol{u}}\sim\mathcal{N}(0,\mathbbm{1}_{(p+1)\times(p+1)})}\left[\sigma(\tilde{\boldsymbol{X}}\tilde{\boldsymbol{u}})\sigma(\tilde{\boldsymbol{u}}^\top\tilde{\boldsymbol{X}}^\top)\right]\\
        &+\mathbb{E}_{\tilde{\boldsymbol{u}}\sim\mathcal{N}(0,\mathbbm{1}_{(p+1)\times(p+1)})}\left[\sigma'(\tilde{\boldsymbol{X}}\tilde{\boldsymbol{u}})\sigma'(\tilde{\boldsymbol{u}}^\top\tilde{\boldsymbol{X}}^\top)\right]\odot\tilde{\boldsymbol{X}}\tilde{\boldsymbol{X}}^\top+\mathbbm{1}_n\mathbbm{1}_n^\top.
    \end{align*}
    Since $\tilde{\boldsymbol{u}}\sim\mathcal{N}(0,\mathbbm{1}_{(p+1)\times(p+1)})$, using the Gaussian polar decomposition, it holds that $\tilde{\boldsymbol{u}}=r\boldsymbol{w}$, where $r=\|\tilde{\boldsymbol{u}}\|$, $\boldsymbol{w}\sim\mathcal{U}(\mathbb{S}^{p})$, and $r$ and $\boldsymbol{w}$ are independent. We note also that $r^2=\|\tilde{\boldsymbol{u}}\|^2=\tilde{u}_1^2+\cdots+\tilde{u}_{p+1}^2$, where $\tilde{u}_1,\ldots,\tilde{u}_{p+1}$ are independent and sampled from the standard Gaussian distribution $\mathcal{N}(0,1)$. This implies that $r^2\sim\chi_{p+1}^2$ (the chi-squared distribution with $p+1$ degrees of freedom) with $\mathbb{E}[r^2]=p+1$. Writing $\tilde{x}_i=\|\tilde{x}_i\|\hat{x}_i$ and letting $\boldsymbol{D}=\mathbf{diag}(\|\tilde{x}_1\|,\ldots,\|\tilde{x}_n\|)$, so that $\tilde{\boldsymbol{X}}\boldsymbol{w}=\boldsymbol{D}\hat{\boldsymbol{X}}\boldsymbol{w}$, and using the facts that $\sigma$ is homogeneous of degree $1$ and $\sigma'$ is scale-invariant together with the independence of $r$ and $\boldsymbol{w}$, we obtain:
    \begin{align*}
        &\boldsymbol{K}^\mathrm{NTK}\\
        =&\mathbb{E}_{r,\boldsymbol{w}\sim\mathcal{U}(\mathbb{S}^{p})}\left[\sigma(\tilde{\boldsymbol{X}} r\boldsymbol{w})\sigma(r\boldsymbol{w}^\top\tilde{\boldsymbol{X}}^\top)\right]+\mathbb{E}_{r,\boldsymbol{w}\sim\mathcal{U}(\mathbb{S}^{p})}\left[\sigma'(\tilde{\boldsymbol{X}} r\boldsymbol{w})\sigma'(r\boldsymbol{w}^\top\tilde{\boldsymbol{X}}^\top)\right]\odot\tilde{\boldsymbol{X}}\tilde{\boldsymbol{X}}^\top+\mathbbm{1}_n\mathbbm{1}_n^\top\\
        =&\underbrace{\mathbb{E}_{r^2\sim\chi_{p+1}^2}[r^2]}_{=p+1}\mathbb{E}_{\boldsymbol{w}\sim\mathcal{U}(\mathbb{S}^{p})}\left[\sigma(\tilde{\boldsymbol{X}}\boldsymbol{w})\sigma(\boldsymbol{w}^\top\tilde{\boldsymbol{X}}^\top)\right]
        +\mathbb{E}_{\boldsymbol{w}\sim\mathcal{U}(\mathbb{S}^{p})}\left[\sigma'(\tilde{\boldsymbol{X}}\boldsymbol{w})\sigma'(\boldsymbol{w}^\top\tilde{\boldsymbol{X}}^\top)\right]\odot\tilde{\boldsymbol{X}}\tilde{\boldsymbol{X}}^\top\\
        &+\mathbbm{1}_n\mathbbm{1}_n^\top\\
        =&\,\boldsymbol{D}\boldsymbol{K}_{\sqrt{p+1}\sigma}^\infty\boldsymbol{D}+\boldsymbol{K}_{\sigma'}^\infty\odot\tilde{\boldsymbol{X}}\tilde{\boldsymbol{X}}^\top+\mathbbm{1}_n\mathbbm{1}_n^\top,
    \end{align*}
    where the last line uses $\sigma(\tilde{\boldsymbol{X}}\boldsymbol{w})=\boldsymbol{D}\,\sigma(\hat{\boldsymbol{X}}\boldsymbol{w})$ and $\sigma'(\tilde{\boldsymbol{X}}\boldsymbol{w})=\sigma'(\hat{\boldsymbol{X}}\boldsymbol{w})$. The matrices $\boldsymbol{D}\boldsymbol{K}_{\sqrt{p+1}\sigma}^\infty\boldsymbol{D}$, $\boldsymbol{K}_{\sigma'}^\infty\odot\tilde{\boldsymbol{X}}\tilde{\boldsymbol{X}}^\top$, and $\mathbbm{1}_n\mathbbm{1}_n^\top$ are all positive semi-definite (the second by the Schur product theorem). Hence, by Weyl's inequality, we obtain:
    \begin{equation}\label{eq:int_low_bound_tilde_lambda0}
        \lambda_0\geq\lambda_\mathrm{min}(\boldsymbol{D}\boldsymbol{K}_{\sqrt{p+1}\sigma}^\infty\boldsymbol{D})
        +\lambda_\mathrm{min}(\boldsymbol{K}_{\sigma'}^\infty\odot\tilde{\boldsymbol{X}}\tilde{\boldsymbol{X}}^\top)
        +\lambda_\mathrm{min}(\mathbbm{1}_n\mathbbm{1}_n^\top)
        \geq\lambda_\mathrm{min}(\boldsymbol{D}\boldsymbol{K}_{\sqrt{p+1}\sigma}^\infty\boldsymbol{D}),
    \end{equation}
    where we use the fact that $\lambda_\mathrm{min}(\boldsymbol{K}_{\sigma'}^\infty\odot\tilde{\boldsymbol{X}}\tilde{\boldsymbol{X}}^\top)\geq 0$ and $\lambda_\mathrm{min}(\mathbbm{1}_n\mathbbm{1}_n^\top)\geq 0$ given that $\boldsymbol{K}_{\sigma'}^\infty\odot\tilde{\boldsymbol{X}}\tilde{\boldsymbol{X}}^\top$ and $\mathbbm{1}_n\mathbbm{1}_n^\top$ are both positive semi-definite. By definition, for any unit vector $w\in\mathbb{R}^n$, we have:
    \begin{align*}
        \lambda_\mathrm{min}(\boldsymbol{D}\boldsymbol{K}_{\sqrt{p+1}\sigma}^\infty\boldsymbol{D})
        &=\inf_{\|w\|=1}\langle\boldsymbol{D}\boldsymbol{K}_{\sqrt{p+1}\sigma}^\infty\boldsymbol{D}w,w\rangle\\
        &=\inf_{\|w\|=1}\|\boldsymbol{D}w\|^2\left\langle\boldsymbol{K}_{\sqrt{p+1}\sigma}^\infty\frac{\boldsymbol{D}w}{\|\boldsymbol{D}w\|},\frac{\boldsymbol{D}w}{\|\boldsymbol{D}w\|}\right\rangle\\
        &\geq\lambda_\mathrm{min}(\boldsymbol{K}_{\sqrt{p+1}\sigma}^\infty)\cdot\lambda_\mathrm{min}(\boldsymbol{D}^2),
    \end{align*}
    since $\boldsymbol{D}w/\|\boldsymbol{D}w\|$ is a unit vector and $\|\boldsymbol{D}w\|^2\geq\lambda_\mathrm{min}(\boldsymbol{D}^2)\|w\|^2=\lambda_\mathrm{min}(\boldsymbol{D}^2)$. Since each $\tilde{x}_i$ has an appended $1$, we have $\lambda_\mathrm{min}(\boldsymbol{D}^2)=\min_i\|\tilde{x}_i\|^2=\min_i(\|x_i\|^2+1)\geq 1$. Moreover, by definition, we have:
    \begin{equation*}
        \lambda_\mathrm{min}(\boldsymbol{K}_{\sqrt{p+1}\sigma}^\infty)=\inf_{\|w\|=1}\left\langle\boldsymbol{K}_{\sqrt{p+1}\sigma}^\infty w,w\right\rangle.
    \end{equation*}
    It then follows from \cref{lem:min_eig_sep} that:
    \begin{equation*}
        \lambda_\mathrm{min}(\boldsymbol{D}\boldsymbol{K}_{\sqrt{p+1}\sigma}^\infty\boldsymbol{D})\geq C(1+g(\epsilon))^{-3}\epsilon^4,
    \end{equation*}
    where $C>0$ is an absolute positive constant and $g(\epsilon)=\frac{(p+1)\ln(1/\epsilon)}{\ln(p+1)}$. Substituting this lower-bound into \cref{eq:int_low_bound_tilde_lambda0} concludes the proof.
\end{proof}

\begin{lemma}[High-probability separation of homogeneous normalized points on $\mathbb{S}^p$]\label{lem:prob_sep}
    Let \cref{ass:bound_in_res,ass:bounded_density} hold. Define the homogeneous normalized points $\{\hat{x}_i\}_{i=1}^n$ as in \cref{eq:hom_norm_points}. Then, for any $\delta\in(0,1)$, with probability at least $1-\delta$ with respect to random samples, the normalized dataset $\{\hat{x}_i\}_{i=1}^n$ is $\epsilon$-separated in the sense of \cref{def:eps_sep} with
    \begin{equation*}
        \epsilon\leq\min\left(\frac{2}{\sqrt{A^2+1}},\frac{2}{\pi}\left(\frac{2\delta}{M_\mu(A^2+1)^{(p+1)/2}v_p n^2}\right)^{\frac{1}{p}}\right),
    \end{equation*}
    where $v_p=\mathrm{Vol}(B(0,1))$ is the volume of the unit ball in $\mathbb{R}^p$.
\end{lemma}

\begin{proof}
Let $\phi:\mathcal{X}\subset\mathbb{R}^p\rightarrow\mathbb{S}^p\subset\mathbb{R}^{p+1}$ be the normalization map defined for any $x\in\mathcal{X}$ by:
\begin{equation*}
    \phi(x)=\frac{1}{\sqrt{\|x\|^2+1}}\begin{bmatrix}x\\1\end{bmatrix},
\end{equation*}
so that $\hat{x}_i=\phi(x_i)$. Define the event that the dataset is not $\epsilon$-separated:
\begin{equation*}
    \mathcal{E}_\epsilon=\left\{\min_{i\neq k}\min\left(\|\hat{x}_i-\hat{x}_k\|,\|\hat{x}_i+\hat{x}_k\|\right)<\epsilon\right\},
\end{equation*}
\emph{i.e.}, there exists at least one pair $i\neq k$ for which the points are either too close together, or too close to antipodal. Hence, we obtain the following inclusion:
\begin{equation*}
    \mathcal{E}_\epsilon\subset\bigcup_{i<k}\left(\{\|\hat{x}_i-\hat{x}_k\|<\epsilon\}\cup\{\|\hat{x}_i+\hat{x}_k\|<\epsilon\}\right).
\end{equation*}
Note that the antipodal term $\|\hat{x}_i+\hat{x}_k\|<\epsilon$ need not be considered provided that $\epsilon$ is chosen small enough such that $\epsilon\leq 2/\sqrt{A^2+1}$. Indeed, for any $x\in\mathcal{X}$ with $\|x\|\leq A$ (\cref{ass:bound_in_res}), the last coordinate of $\phi(x)$ is $1/\sqrt{\|x\|^2+1}\geq 1/\sqrt{A^2+1}>0$, so all points $\hat{x}_i$ lie in the open upper hemisphere $\mathbb{S}^p_+\coloneqq\{z\in\mathbb{S}^p\mid z_{p+1}>0\}$, bounded away from the equator by the constant $1/\sqrt{A^2+1}$. Therefore $(\hat{x}_i+\hat{x}_k)_{p+1}\geq 2/\sqrt{A^2+1}$ for all $i\neq k$. Since the last coordinate of a vector is dominated by its Euclidean norm, we have $\|\hat{x}_i+\hat{x}_k\|\geq(\hat{x}_i+\hat{x}_k)_{p+1}\geq 2/\sqrt{A^2+1}$. Hence, provided that $\epsilon\leq 2/\sqrt{A^2+1}$, the event $\{\|\hat{x}_i+\hat{x}_k\|<\epsilon\}$ is empty for every pair $i\neq k$. This implies that the antipodal event cannot occur. Thus:
\begin{equation*}
    \mathcal{E}_\epsilon\subset\bigcup_{i<k}\left\{\|\hat{x}_i-\hat{x}_k\|<\epsilon\right\}.
\end{equation*}
Therefore, by the union bound, we get:
\begin{equation}\label{eq:u_b}
    \Pr(\mathcal{E}_\epsilon)\leq\sum_{i<k}\Pr\left(\|\hat{x}_i-\hat{x}_k\|<\epsilon\right).
\end{equation}
Let $\nu=\phi_\#\mu$ be the pushforward distribution of $\mu$ under the map $\phi$, so that $\hat{x}_i\sim\nu$ on $\mathbb{S}^p$. Since $\hat{x}_i$ and $\hat{x}_k$ are independent and identically distributed samples from $\nu$ on $\mathbb{S}^p$, conditioning on $\hat{x}_k=z$ and applying the tower property yield:
\begin{equation}\label{eq:exp_nu_ball}
    \Pr\left(\|\hat{x}_i-\hat{x}_k\|<\epsilon\right)=\mathbb{E}\left[\Pr\left(\|\hat{x}_i-z\|<\epsilon\mid\hat{x}_k=z\right)\right]=\mathbb{E}\left[\nu\left(B_{\mathbb{S}^p}(z,\epsilon)\right)\right],
\end{equation}
where $B_{\mathbb{S}^p}(z,\epsilon)=\{\hat{x}\in\mathbb{S}^p\mid\|\hat{x}-z\|<\epsilon\}$ is the Euclidean ball of radius $\epsilon$ centered at $z$ intersected with $\mathbb{S}^p$. We now derive the density of $\nu$ with respect to the surface measure $\sigma$ on $\mathbb{S}^p$. For any measurable set $\mathcal{M}\subset\mathbb{S}^p$, we have:
\begin{equation}\label{eq:nu_A}
    \nu(\mathcal{M})=\mu(\phi^{-1}(\mathcal{M}))=\int_{\phi^{-1}(\mathcal{M})}f_\mu(x)\,\mathrm{d}x=\int_{\mathbb{R}^p}g(x)J_\phi(x)\,\mathrm{d}x,
\end{equation}
where $g(x)=f_\mu(x)/J_\phi(x)\cdot\mathbbm{1}_{\phi(x)\in \mathcal{M}}$, $J_\phi(x)=\sqrt{\det\left(D\phi(x)^\top D\phi(x)\right)}$ with $D\phi(x)\in\mathbb{R}^{(p+1)\times p}$ being the Jacobian matrix. Since $\phi$ is $C^\infty$ on $\mathbb{R}^p$, it is Lipschitz on the compact domain $\mathcal{X}$. The area formula \citep[Theorem~3.9]{EvansGariepy2015} therefore gives:
\begin{equation*}
    \int_{\mathbb{R}^p}g(x)J_\phi(x)\,\mathrm{d}x=\int_{\mathbb{S}^p}\left[\sum_{x\in\phi^{-1}(z)}g(x)\right]\mathrm{d}\sigma(z).
\end{equation*}
Given that $\phi$ is injective on $\mathbb{R}^p$, the preimage $\phi^{-1}(z)$ contains exactly one point for each $z\in\mathbb{S}^p_+$ and is empty for $z\notin\mathbb{S}^p_+$. Using injectivity so that the sum has exactly one term for $z\in\mathbb{S}^p_+$, we obtain:
\begin{equation*}
    \int_{\mathbb{R}^p}g(x)J_\phi(x)\,\mathrm{d}x=\int_{\mathbb{S}^p}g(\phi^{-1}(z))\mathrm{d}\sigma(z)=\int_\mathcal{M}\frac{f_\mu(\phi^{-1}(z))}{J_\phi(\phi^{-1}(z))}\mathrm{d}\sigma(z).
\end{equation*}
Next, we find the expression of $J_\phi(x)$. Write $s=\|x\|^2+1$ for brevity. $D\phi(x)$ has entries, for $1\leq i\leq p$ and $1\leq j\leq p$, obtained by the product rule applied to $\phi_i(x)=x_is^{-1/2}$:
\begin{equation*}
    \frac{\partial\phi_i}{\partial x_j}(x)=\delta_{ij}s^{-1/2}+x_i\cdot\left(-x_js^{-3/2}\right)=\frac{\delta_{ij}s-x_ix_j}{s^{3/2}},
\end{equation*}
where $\delta_{ij}$ is the Kronecker delta, and for the last row, using $\phi_{p+1}(x)=s^{-1/2}$ and the chain rule:
\begin{equation*}
    \frac{\partial\phi_{p+1}}{\partial x_j}(x)=-\frac{x_j}{s^{3/2}}.
\end{equation*}
Therefore, the full Jacobian matrix is:
\begin{equation*}
    D\phi(x)=\frac{1}{s^{3/2}}\begin{bmatrix}s\mathbbm{1}_{p\times p}-xx^\top\\-x^\top\end{bmatrix}\in\mathbb{R}^{(p+1)\times p}.
\end{equation*}
We compute the Gram matrix $D\phi(x)^\top D\phi(x)\in\mathbb{R}^{p\times p}$. Expanding the block matrix product and using $(xx^\top)(xx^\top)=\|x\|^2xx^\top=(s-1)xx^\top$, we obtain:
\begin{align*}
    D\phi(x)^\top D\phi(x)=&\frac{1}{s^3}\left[(s\mathbbm{1}_{p\times p}-xx^\top)^2+xx^\top\right]=\frac{1}{s^3}\left[s^2\mathbbm{1}_{p\times p}-(s+1)xx^\top+xx^\top\right]\\
    =&\frac{1}{s}\left(\mathbbm{1}_{p\times p}-\frac{xx^\top}{s}\right).
\end{align*}
Applying the matrix determinant lemma $\det(\mathbbm{1}_{p\times p}+uv^\top)=1+v^\top u$ with $u=-x/\sqrt{s}$ and $v=x/\sqrt{s}$:
\begin{equation*}
    \det\left(\mathbbm{1}_{p\times p}-\frac{xx^\top}{s}\right)=1-\frac{\|x\|^2}{s}=1-\frac{s-1}{s}=\frac{1}{s}.
\end{equation*}
Therefore:
\begin{equation*}
    \det\left(D\phi(x)^\top D\phi(x)\right)=\frac{1}{s^p}\cdot\frac{1}{s}=\frac{1}{(\|x\|^2+1)^{p+1}},
\end{equation*}
and the Jacobian factor is:
\begin{equation*}
    J_\phi(x)=\sqrt{\det\left(D\phi(x)^\top D\phi(x)\right)}=\frac{1}{(\|x\|^2+1)^{(p+1)/2}}.
\end{equation*}
Consequently, we obtain:
\begin{equation*}
    \int_{\mathbb{R}^p}g(x)J_\phi(x)\,\mathrm{d}x=\int_\mathcal{M}f_\mu\!\left(\phi^{-1}(z)\right)\cdot\left(\|\phi^{-1}(z)\|^2+1\right)^{(p+1)/2}\mathrm{d}\sigma(z).
\end{equation*}
Substituting this into \cref{eq:nu_A}, we get:
\begin{equation*}
    \nu(\mathcal{M})=\int_\mathcal{M} f_\mu\!\left(\phi^{-1}(z)\right)\cdot\left(\|\phi^{-1}(z)\|^2+1\right)^{(p+1)/2}\mathrm{d}\sigma(z).
\end{equation*}
Since this holds for any measurable set $\mathcal{M}\subset\mathbb{S}^p$, by uniqueness of the Radon-Nikodym derivative, $\nu$ has the following density with respect to $\sigma$:
\begin{equation*}
    f_\nu(z)=f_\mu\!\left(\phi^{-1}(z)\right)\cdot\left(\|\phi^{-1}(z)\|^2+1\right)^{(p+1)/2}.
\end{equation*}
The inverse map is $\phi^{-1}(z)=z_{1:p}/z_{p+1}$, where $z_{1:p}=[z_1,\ldots,z_p]^\top$ denotes the first $p$ coordinates of $z$. Since $z\in\mathbb{S}^p$ we have $\|z_{1:p}\|^2+z_{p+1}^2=1$, therefore:
\begin{equation*}
    \|\phi^{-1}(z)\|^2+1=\frac{\|z_{1:p}\|^2}{z_{p+1}^2}+1=\frac{\|z_{1:p}\|^2+z_{p+1}^2}{z_{p+1}^2}=\frac{1}{z_{p+1}^2}=\|x\|^2+1.
\end{equation*}
Given that $\|x\|\leq A$ for all $x\in\mathcal{X}$ by \cref{ass:bound_in_res}, we have $\|\phi^{-1}(z)\|^2+1\leq A^2+1$. Moreover, we have by \cref{ass:bounded_density} that $\|f_\mu\|_\infty\leq M_\mu$. This yields:
\begin{equation*}
    \|f_\nu\|_\infty\leq\|f_\mu\|_\infty\cdot\sup_{x\in\mathcal{X}}\left(\|x\|^2+1\right)^{(p+1)/2}\leq M_\mu(A^2+1)^{(p+1)/2}.
\end{equation*}
From this bound and \cref{lem:sur_euc_geo_balls}, we bound the $\nu$-measure of $B_{\mathbb{S}^p}(z,\epsilon)$ as follows:
\begin{equation*}
    \nu\left(B_{\mathbb{S}^p}(z,\epsilon)\right)=\int_{B_{\mathbb{S}^p}(z,\epsilon)}f_\nu(\hat{x})\,\mathrm{d}\sigma(\hat{x})\leq\|f_\nu\|_\infty\cdot\sigma\left(B_{\mathbb{S}^p}(z,\epsilon)\right)\leq M_\mu(A^2+1)^{(p+1)/2}v_p\left(\frac{\pi}{2}\epsilon\right)^p,
\end{equation*}
where $v_p$ is the volume of the unit ball in $\mathbb{R}^p$. Since this bound is uniform in $z$, after substituting it into \cref{eq:exp_nu_ball}, we obtain:
\begin{equation*}
    \Pr\left(\|\hat{x}_i-\hat{x}_k\|<\epsilon\right)\leq M_\mu(A^2+1)^{(p+1)/2}v_p\left(\frac{\pi}{2}\epsilon\right)^p.
\end{equation*}
Putting this bound into the union bound in \cref{eq:u_b} and using $\binom{n}{2}\leq n^2/2$, we get:
\begin{align*}
    \Pr(\mathcal{E}_\epsilon)&\leq\sum_{i<k}M_\mu(A^2+1)^{(p+1)/2}v_p\left(\frac{\pi}{2}\epsilon\right)^p\\
    &=M_\mu(A^2+1)^{(p+1)/2}v_p\left(\frac{\pi}{2}\epsilon\right)^p\underbrace{\sum_{i<k}1}_{\text{number of pairs with }i<k}\\
    &=M_\mu(A^2+1)^{(p+1)/2}v_p\left(\frac{\pi}{2}\epsilon\right)^p\binom{n}{2}\\
    &\leq M_\mu(A^2+1)^{(p+1)/2}v_p\left(\frac{\pi}{2}\epsilon\right)^p\frac{n^2}{2}.
\end{align*}
Setting this quantity less than or equal to $\delta$ and solving for $\epsilon$:
\begin{equation*}
    M_\mu(A^2+1)^{(p+1)/2}v_p\left(\frac{\pi}{2}\epsilon\right)^p\frac{n^2}{2}\leq\delta\implies\epsilon\leq\frac{2}{\pi}\left(\frac{2\delta}{M_\mu(A^2+1)^{(p+1)/2}v_p n^2}\right)^{\frac{1}{p}}.
\end{equation*}
Therefore, for any $\epsilon\leq\frac{2}{\pi}\left(\frac{2\delta}{M_\mu(A^2+1)^{(p+1)/2}v_p n^2}\right)^{1/p}$, we have $\Pr(\mathcal{E}_\epsilon)\leq\delta$, \emph{i.e.}, the normalized dataset $\{\hat{x}_i\}_{i=1}^n$ is $\epsilon$-separated with probability at least $1-\delta$. Combining both upper-bounds on $\epsilon$, we obtain:
\begin{equation*}
    \epsilon\leq\min\left(\frac{2}{\sqrt{A^2+1}},\frac{2}{\pi}\left(\frac{2\delta}{M_\mu(A^2+1)^{(p+1)/2}v_p n^2}\right)^{\frac{1}{p}}\right),
\end{equation*}
which concludes the proof.
\end{proof}

\begin{lemma}[High-probability lower-bound on the smallest eigenvalue of $\boldsymbol{K}^\mathrm{NTK}$]\label{lem:proba_low_bound_min_eig}
    Fix $p\geq 2$ and let \cref{ass:bound_in_res,ass:bounded_density} hold. Define the homogeneous normalized points $\{\hat{x}_i\}_{i=1}^n$ as in \cref{eq:hom_norm_points}. Then, for any $\delta\in(0,1)$, with probability at least $1-\delta$ with respect to random samples, the smallest eigenvalue $\lambda_0$ of $\boldsymbol{K}^\mathrm{NTK}$ satisfies:
    \begin{equation*}
        \lambda_0\geq C (1+g(\epsilon))^{-3}\epsilon^4\quad\text{for every}\quad0<\epsilon\leq\min\left(1,\frac{2}{\sqrt{A^2+1}},\frac{2}{\pi}\left(\frac{2\delta}{M_\mu(A^2+1)^{(p+1)/2}v_p n^2}\right)^{\frac{1}{p}}\right),
    \end{equation*}
    where $C>0$ is an absolute positive constant, $g(\epsilon)=\frac{(p+1)\ln(1/\epsilon)}{\ln(p+1)}$, and $v_{p}=\mathrm{Vol}(B(0,1))$ is the volume of the unit ball in $\mathbb{R}^{p}$.
\end{lemma}

\begin{proof}
    By \cref{lem:prob_sep}, with probability at least $1-\delta$ with respect to random samples, the dataset $\{\hat{x}_i\}_{i=1}^n$ is $\bar\epsilon$-separated with:
    \begin{equation*}
        \bar\epsilon\leq\min\left(\frac{2}{\sqrt{A^2+1}},\frac{2}{\pi}\left(\frac{2\delta}{M_\mu(A^2+1)^{(p+1)/2}v_p n^2}\right)^{\frac{1}{p}}\right).
    \end{equation*}
    On this event, it is also $\epsilon$-separated for every $0<\epsilon\leq\bar\epsilon$, since the condition in \cref{def:eps_sep} becomes weaker as $\epsilon$ decreases. Consequently, by \cref{lem:sm_eig_ti_ntk}, we obtain on this event, for every $0<\epsilon\leq\min(1,\bar\epsilon)$, the following lower-bound on $\lambda_0$:
    \begin{equation*}
        \lambda_0\geq C (1+g(\epsilon))^{-3}\epsilon^4,
    \end{equation*}
    where $C>0$ is an absolute positive constant and $g(\epsilon)=\frac{(p+1)\ln(1/\epsilon)}{\ln(p+1)}$. Note that we require $\epsilon\leq1$ to ensure that $g(\epsilon)\geq 0$, and hence $1+g(\epsilon)\geq 1$, so that the lower-bound is well defined. This concludes the proof.
\end{proof}

\begin{corollary}[High-probability upper-bound on the inverse of the smallest eigenvalue of $\boldsymbol{K}^\mathrm{NTK}$]\label{cor:up_bound_lam_in}
    Fix $p\geq 2$ and let \cref{ass:bound_in_res,ass:bounded_density} hold. Define the homogeneous normalized points $\{\hat{x}_i\}_{i=1}^n$ as in \cref{eq:hom_norm_points}, and, for $n\geq 1$ and $\delta\in(0,1)$, the deterministic threshold $\lambda_\star(n,\delta)\coloneqq h(\min(\epsilon_A,\epsilon_0(n,\delta)))>0$, where $h(\epsilon)\coloneqq C(1+g(\epsilon))^{-3}\epsilon^4$, $\epsilon_A\coloneqq\min(1,2/\sqrt{A^2+1})$ and $\epsilon_0(n,\delta)\coloneqq\frac{2}{\pi}\left(\frac{2\delta}{M_\mu(A^2+1)^{(p+1)/2}v_p n^2}\right)^{\frac{1}{p}}$, with $C$, $g$ and $v_p$ as in \cref{lem:proba_low_bound_min_eig}. Then, there exists a constant $C'>0$, depending only on $A$, $M_\mu$, and $p$, such that, for any $n\geq 3$ and $\delta\in(0,1)$, with probability $1-\delta$ with respect to random samples, the smallest eigenvalue $\lambda_0$ of $\boldsymbol{K}^\mathrm{NTK}$ satisfies $\lambda_0\geq\lambda_\star(n,\delta)$, and hence:
    \begin{equation*}
        \lambda_0^{-1}\leq\lambda_\star(n,\delta)^{-1}\leq C'n^{8/p}\delta^{-4/p}\left(\ln n+\ln(1/\delta)\right)^3.
    \end{equation*}
\end{corollary}

\begin{proof}
    Since $\epsilon_A\leq1$, we have $g(\epsilon_A)\geq0$, hence $1+g(\epsilon_A)\geq 1$ and $\kappa\coloneqq h(\epsilon_A)>0$, a constant depending on $A$ and $p$. Similarly, $\lambda_\star(n,\delta)>0$, since $\min(\epsilon_A,\epsilon_0(n,\delta))\in(0,1]$, which is admissible in \cref{lem:proba_low_bound_min_eig}. Therefore, with probability at least $1-\delta$ with respect to random samples, $\lambda_0\geq\lambda_\star(n,\delta)$. It remains to bound $\lambda_\star(n,\delta)^{-1}$. To this end, we define:
    \begin{equation*}
        L\coloneqq\ln n+\ln(1/\delta),\qquad F\coloneqq n^{8/p}\,\delta^{-4/p}L^3,\qquad M\coloneqq M_\mu(A^2+1)^{(p+1)/2}v_p.
    \end{equation*}
    Since $n\geq3$, we have $\ln n\geq\ln 3>1$, and since $\delta\in(0,1)$, we have $\ln(1/\delta)>0$; hence $L\geq1$. Moreover, $n^{8/p}\geq1$ and $\delta^{-4/p}\geq1$, so that $F\geq1$. We distinguish two cases.
 
    \textbf{Case 1: $\epsilon_0(n,\delta)>\epsilon_A$. }Then $\lambda_\star(n,\delta)=h(\epsilon_A)=\kappa$, so that, using $F\geq1$, we get $\lambda_\star(n,\delta)^{-1}=\kappa^{-1}\leq\kappa^{-1}F$.
 
    \textbf{Case 2: $\epsilon_0(n,\delta)\leq\epsilon_A$. }Then $\lambda_\star(n,\delta)=h(\epsilon_0(n,\delta))$. Since $\epsilon_0(n,\delta)\leq 1$, the quantity
    \begin{equation*}
        \Lambda\coloneqq\ln\frac{1}{\epsilon_0(n,\delta)}=\ln\left(\frac{\pi}{2}\right)+\frac{1}{p}\ln\left(\frac{Mn^2}{2\delta}\right)
    \end{equation*}
    is non-negative, and $1+g(\epsilon_0(n,\delta))=1+\frac{(p+1)\Lambda}{\ln(p+1)}\geq1$. Substituting $\epsilon_0(n,\delta)$ gives:
    \begin{equation*}
        \lambda_\star(n,\delta)^{-1}=h(\epsilon_0(n,\delta))^{-1}=C^{-1}\left(\frac{\pi}{2}\right)^4\left(\frac{M}{2}\right)^{\frac{4}{p}}\left(1+\frac{(p+1)\Lambda}{\ln(p+1)}\right)^{3}n^{8/p}\,\delta^{-4/p}.
    \end{equation*}
    Expanding the logarithm, we have $\Lambda=\ln(\frac{\pi}{2})+\frac{1}{p}\ln\frac{M}{2}+\frac{2}{p}\ln n+\frac{1}{p}\ln\frac{1}{\delta}$. Since $p\geq2$ and $\ln n,\ln(1/\delta)\geq0$, this gives $\Lambda\leq a_0+L$, where $a_0\coloneqq\left|\ln(\frac{\pi}{2})+\frac{1}{p}\ln\frac{M}{2}\right|$. Using $L\geq1$, we have $a_0+L\leq(a_0+1)L$, hence:
    \begin{equation*}
        1+\frac{(p+1)\Lambda}{\ln(p+1)}\leq L+\frac{(p+1)(a_0+1)}{\ln(p+1)}L=bL,
    \end{equation*}
    where $b\coloneqq1+\frac{(p+1)(a_0+1)}{\ln(p+1)}$. Both sides being positive, cubing preserves the inequality, and we obtain:
    \begin{equation*}
        \lambda_\star(n,\delta)^{-1}\leq cF,\qquad\text{where}\qquad c\coloneqq C^{-1}\left(\frac{\pi}{2}\right)^4\left(\frac{M}{2}\right)^{\frac{4}{p}}b^3.
    \end{equation*}
    Combining both cases yields $\lambda_\star(n,\delta)^{-1}\leq\max(\kappa^{-1},c)F\leq(\kappa^{-1}+c)F$. Setting $C'\coloneqq\kappa^{-1}+c$ concludes the proof, since $C$ is an absolute constant, $\kappa$ depends on $A$ and $p$, and $M$, $a_0$ and $b$ depend on $A$, $M_\mu$ and $p$.
\end{proof}

\subsection{Eigenvalue decay rate of \texorpdfstring{$\hntk$}{NTK}}
We now derive the eigenvalue decay rate of $\hntk$ in our setting in the sense of \cref{def:edr}. The eigenvalues of $\hntk$ are understood as those of its associated integral/covariance operator (see \cref{sec:int_space} for more details). Throughout, $\lambda_i(k;E,\nu)$ denotes the $i$-th largest eigenvalue of the integral operator of a kernel $k$ on the space $L^2(E,\nu)$.

\begin{definition}[Eigenvalue decay rate]\label{def:edr}
    Let $K:\mathcal{X}\times\mathcal{X}\to\mathbb{R}$ be a continuous kernel associated to an RKHS $\mathcal{H}$, and $(\lambda_i)_{i\geq 1}$ be the non-zero eigenvalues of its associated integral operator in non-increasing order. We say that $\mathcal{H}$ has eigenvalue decay rate $\beta>1$ if there exist two constants $k_1,k_2>0$ such that $k_1 i^{-\beta} \leq \lambda_i \leq k_2 i^{-\beta}$, for any $i \geq 1$.
\end{definition}

\begin{lemma}[Eigenvalue decay rate of $\hntk$]\label{lem:edr_set}
Under \cref{ass:bound_in_res,ass:bounded_density}, the eigenvalue decay rate of the integral/covariance operator associated with $K^{\mathrm{NTK}}$ in \cref{eq:shallow_ntk} is $\frac{p+1}{p}$. 
\end{lemma}
 
\begin{proof}
Let $\phi(x) \coloneqq \tilde{x}/\|\tilde{x}\| \in \mathbb{S}^p$, $\rho(x) \coloneqq \|\tilde{x}\|/\sqrt{2}$ and $K_0(y,y') \coloneqq \kappa(\langle y, y' \rangle)$ with $y,y'\in\mathbb{S}^p$, where, for any $u\in[-1,1]$, $\kappa(u)=u\kappa_0(u)+\kappa_1(u)$ with $\kappa_0(u)\coloneqq \frac{1}{\pi}(\pi-\arccos(u))$ and $\kappa_1(u)\coloneqq\frac{1}{\pi}(u(\pi-\arccos(u))+\sqrt{1-u^2})$. Thus, we can write $K^{\mathrm{NTK}}=K_1+1$, where $K_1(x,x') \coloneqq \rho(x)K_0(\phi(x),\phi(x'))\rho(x')$. Let $\tilde{\nu} \coloneqq\phi_\#(\rho^2\mu)$ be the push-forward measure.
 
\textbf{Upper bound.} Since $\phi$ is injective, \citep[Propositions~1~and~3]{li2024eigenvalue} give:
\begin{equation*}
    \lambda_i(K_1;\mathcal{X},\mu)=\lambda_i(K_0\circ(\phi\times\phi);\mathcal{X},\rho^2\mu)=\lambda_i(K_0;\mathbb{S}^p,\tilde{\nu}).
\end{equation*}
By the change of variables computed in the proof of \cref{lem:prob_sep}, $\tilde{\nu}$ has the density $q(y) = \frac{1}{2}(\|x\|^2+1)^{(p+3)/2} f_\mu(x)$, with $x = \phi^{-1}(y)$, with respect to the surface measure $\sigma$ on $\mathbb{S}^p$, and $q=0$ outside $\phi(\mathcal{X})$. By \cref{ass:bound_in_res,ass:bounded_density}, $q\le C_q \coloneqq \frac{1}{2}(A^2+1)^{(p+3)/2}M_\mu$. Hence \citep[Lemma~2]{li2024eigenvalue} gives $\lambda_i(K_0;\mathbb{S}^p,\tilde\nu)\le C_q\lambda_i(K_0;\mathbb{S}^p,\sigma)$. By \citep[Lemma~B.1]{haas2023mind}, $\lambda_i(K_0;\mathbb{S}^p,\sigma) \le C i^{-(p+1)/p}$ for all $i \ge 1$, where $C>0$. Finally, the constant kernel $1$ has rank one, so Weyl's inequality gives $\lambda_{i+1}(K^{\mathrm{NTK}}; \mathcal{X}, \mu) \le \lambda_i(K_1;\mathcal{X},\mu)$ for all $i \ge 1$. Together with $\lambda_1(K^\mathrm{NTK}; \mathcal{X}, \mu) \le \int_\mathcal{X} K^\mathrm{NTK}(x,x)\mathrm{d}\mu(x) \le \kappa^\mathrm{NTK}$ (\cref{lem:bounded_ntk_kernel}), this yields $\lambda_i(K^\mathrm{NTK};\mathcal{X},\mu) \le k_2i^{-(p+1)/p}$ for all $i \ge 1$, with $k_2 \coloneqq \max\bigl(\kappa^\mathrm{NTK}, 2^{(p+1)/p}C_q C\bigr)$.
 
\textbf{Lower bound.} By \cref{ass:bounded_density}, $f_\mu \ge c_\mu$ on the open ball $\Omega_\mu \subset \mathcal{X}$. The integral operator of $K^\mathrm{NTK}$ on $L^2(\Omega_\mu, \mu)$ is the compression of the one on $L^2(\mathcal{X}, \mu)$ to functions vanishing outside $\Omega_\mu$, so the minimax principle \citep[Lemma~49]{li2024eigenvalue} gives $\lambda_i(K^\mathrm{NTK};\mathcal{X},\mu) \ge \lambda_i(K^\mathrm{NTK};\Omega_\mu,\mu)$. Since $\mathrm{d}\mu/\mathrm{d}x \ge c_\mu$ on $\Omega_\mu$, \citep[Lemma~2]{li2024eigenvalue} gives $\lambda_i(K^\mathrm{NTK};\Omega_\mu,\mu) \ge c_\mu\lambda_i(K^\mathrm{NTK};\Omega_\mu,\mathrm{d}x)$. The ball $\Omega_\mu$ is a bounded Lipschitz domain, so \cref{thm:sobolev_ntk} shows that the RKHS of $K^\mathrm{NTK}$ restricted to $\Omega_\mu$ equals the Sobolev space $H^{\frac{p+1}{2}}(\Omega_\mu)$, with equivalent norms. The eigenvalues $\lambda_i(K^\mathrm{NTK};\Omega_\mu,\mathrm{d}x)$ are the squares of the singular values of the embedding of this RKHS into $L^2(\Omega_\mu)$ \citep{steinwart2012mercer}, and equivalent norms change these singular values by at most constant factors. The singular values of the embedding $H^{\frac{p+1}{2}}(\Omega_\mu) \hookrightarrow L^2(\Omega_\mu)$ are of exact order $i^{-\frac{p+1}{2p}}$ \citep[\S~3.3.4]{edmunds1996function}. Hence $\lambda_i(K^\mathrm{NTK};\Omega_\mu,\mathrm{d}x) \ge c' i^{-(p+1)/p}$, and therefore $\lambda_i(K^\mathrm{NTK};\mathcal{X},\mu) \ge k_1 i^{-(p+1)/p}$ with $k_1 \coloneqq c_\mu c'$, where $c'$ depends on $p$, $A$ and the radius of $\Omega_\mu$.
\end{proof}

\subsection{Bounds on the training dynamics}

For $t\geq0$, we recall that $f_t^\mathrm{NN}$ and $f_t^\mathrm{NTK}$ are the solutions of the gradient flows provided in \cref{eq:grad_flow_ftnn,eq:grad_flow_ftntk}, respectively.
\begin{lemma}[Training residual bounds]\label{lem:bound_rntk}
    Consider $n$ i.i.d. samples $\{(x_i,y_i)\}_{i=1}^n\sim\rho$. For any $t\geq 0$, let:
    \begin{align*}
        r^\mathrm{NTK}(t)=&f_t^\mathrm{NTK}(\boldsymbol{X})-\boldsymbol{Y}=[f_t^\mathrm{NTK}(x_1)-y_1,\ldots,f_t^\mathrm{NTK}(x_n)-y_n]^\top\in\mathbb{R}^{n},\\
        r^\mathrm{NN}(t)=&f_t^\mathrm{NN}(\boldsymbol{X})-\boldsymbol{Y}=[f_t^\mathrm{NN}(x_1)-y_1,\ldots,f_t^\mathrm{NN}(x_n)-y_n]^\top\in\mathbb{R}^{n}.
    \end{align*}
    Under \cref{ass:bound_la_res}, it holds that:
    \begin{align*}
        \left\|r^\mathrm{NTK}(t)\right\|\leq&\exp\left(-\frac{\lambda_0}{n}t\right)\sqrt{n}B,\\
        \left\|r^\mathrm{NN}(t)-r^\mathrm{NTK}(t)\right\|\leq&\Delta t\exp\left(\left(\Delta -\frac{\lambda_0}{n}\right)t\right)\sqrt{n}B,
    \end{align*}
    where $\lambda_0$ is the smallest eigenvalue of the infinite-width NTK matrix $\boldsymbol{K}^\mathrm{NTK}$ and
    \begin{equation*}
        \Delta=\sup_{t\geq0}\sup_{\|x\|\leq A,\|x'\|\leq A}\left|K_t^m(x,x')-K^\mathrm{NTK}(x,x')\right|.
    \end{equation*}
\end{lemma}

\begin{remark}
    The quantity $\Delta$ is random, as it depends on both the random initialization of the network and the random training samples. \Cref{lem:nn_ntk} shows that, for sufficiently large width $m$, it is finite with high probability, and all subsequent results involving $\Delta$ are understood to hold on this high-probability event.
\end{remark}

\begin{proof}[Proof of \cref{lem:bound_rntk}]
Let $t\geq0$. We recall that the gradient flows of $\ftt$ and $\fttntk$ evaluated at the samples $\{(x_i,y_i)\}_{i=1}^n$ are:
\begin{align}
    \frac{\text{d}}{\text{d}t}r^\mathrm{NN}(t)=&-\frac{1}{n}\boldsymbol{K}^m_t r^\mathrm{NN}(t),\\
    \frac{\text{d}}{\text{d}t}r^\mathrm{NTK}(t)=&-\frac{1}{n}\boldsymbol{K}^\mathrm{NTK}r^\mathrm{NTK}(t),\label{eq:ode_ntk}
\end{align}
with $r^\mathrm{NN}(0)=r^\mathrm{NTK}(0)=-\boldsymbol{Y}$.

\textbf{Upper-bound on $\left\|r^\mathrm{NTK}(t)\right\|$. }A closed-form expression of the solution to the ordinary differential equation in \cref{eq:ode_ntk} is:
\begin{equation*}
    r^\mathrm{NTK}(t)=\exp\left(-\frac{1}{n}t\boldsymbol{K}^\mathrm{NTK}\right)r^\mathrm{NTK}(0)=-\exp\left(-\frac{1}{n}t\boldsymbol{K}^\mathrm{NTK}\right)\boldsymbol{Y},
\end{equation*}
where the second equality holds because $f_0^\mathrm{NTK}=0$. Thus, under \cref{ass:bound_la_res}, we get:
\begin{equation}\label{eq:op_norm_rntk}
    \left\|r^\mathrm{NTK}(t)\right\|\leq\left\|\exp\left(-\frac{1}{n}t\boldsymbol{K}^\mathrm{NTK}\right)\right\|_\mathrm{op}\left\|\boldsymbol{Y}\right\|=\exp\left(-\frac{\lambda_0}{n}t\right)\left\|\boldsymbol{Y}\right\|\leq\exp\left(-\frac{\lambda_0}{n}t\right)\sqrt{n}B,
\end{equation}
where we use \cref{lem:op_norm_exp_mat} to find the operator norm of the matrix.

\textbf{Upper-bound on $\left\|r^\mathrm{NN}(t)-r^\mathrm{NTK}(t)\right\|$. }Define the error $e(t)=r^\mathrm{NN}(t)-r^\mathrm{NTK}(t)$. Using the gradient flow expressions, we obtain:
    \begin{equation*}
        \frac{\text{d}}{\text{d}t}e(t)=\frac{\text{d}}{\text{d}t}r^\mathrm{NN}(t)-\frac{\text{d}}{\text{d}t}r^\mathrm{NTK}(t)=-\frac{1}{n}\boldsymbol{K}_t^m r^\mathrm{NN}(t)+\frac{1}{n}\boldsymbol{K}^\mathrm{NTK} r^\mathrm{NTK}(t).
    \end{equation*}
    Adding and subtracting $\frac{1}{n}\boldsymbol{K}^\mathrm{NTK} r^\mathrm{NN}(t)$ leads to:
    \begin{equation*}
        \frac{\text{d}}{\text{d}t}e(t)=-\frac{1}{n}\left(\boldsymbol{K}_t^m-\boldsymbol{K}^\mathrm{NTK}\right)r^\mathrm{NN}(t)-\frac{1}{n}\boldsymbol{K}^\mathrm{NTK}\underbrace{\left(r^\mathrm{NN}(t)-r^\mathrm{NTK}(t)\right)}_{e(t)}.
    \end{equation*}
    This is a linear ordinary differential equation with initialization $e(0)=0$. Applying Duhamel's principle, \emph{i.e.}, the variation of constants method, we get:
    \begin{equation*}
        e(t)=\int_0^t\exp\left(-\frac{1}{n}(t-s)\boldsymbol{K}^\mathrm{NTK}\right)\left(-\frac{1}{n}\left(\boldsymbol{K}_s^m-\boldsymbol{K}^\mathrm{NTK}\right)r^\mathrm{NN}(s)\right)\text{d}s.
    \end{equation*}
    Taking norms and applying the triangle inequality together with \cref{lem:op_norm_exp_mat} gives:
    \begin{align*}
        \left\|e(t)\right\|\leq&\frac{1}{n}\int_0^t\left\|\exp\left(-\frac{1}{n}(t-s)\boldsymbol{K}^\mathrm{NTK}\right)\right\|_\mathrm{op}\left\|\boldsymbol{K}_s^m-\boldsymbol{K}^\mathrm{NTK}\right\|_\mathrm{op}\left\|r^\mathrm{NN}(s)\right\|\text{d}s\\
        =&\frac{1}{n}\int_0^t\exp\left(-\frac{1}{n}\lambda_0 (t-s)\right)\left\|\boldsymbol{K}_s^m-\boldsymbol{K}^\mathrm{NTK}\right\|_\mathrm{op}\left\|r^\mathrm{NN}(s)\right\|\text{d}s.
    \end{align*}
    Noticing by the triangle inequality that $\left\|r^\mathrm{NN}(s)\right\|\leq\left\|e(s)\right\|+\left\|r^\mathrm{NTK}(s)\right\|$ yields:
    \begin{align*}
        \left\|e(t)\right\|\leq&\frac{1}{n}\int_0^t\exp\left(-\frac{1}{n}\lambda_0 (t-s)\right)\left\|\boldsymbol{K}_s^m-\boldsymbol{K}^\mathrm{NTK}\right\|_\mathrm{op}\left\|r^\mathrm{NTK}(s)\right\|\text{d}s\\
        &+\frac{1}{n}\int_0^t\exp\left(-\frac{1}{n}\lambda_0 (t-s)\right)\left\|\boldsymbol{K}_s^m-\boldsymbol{K}^\mathrm{NTK}\right\|_\mathrm{op}\left\|e(s)\right\|\text{d}s
    \end{align*}
    Multiplying both sides by $\exp\left(\frac{1}{n}\lambda_0 t\right)$ gives $f(t)\leq g(t)+\int_0^t h(s)f(s)\text{d}s$, where:
\begin{align*}
    f(t)=&\exp\left(\frac{1}{n}\lambda_0 t\right)\left\|e(t)\right\|,\\
    g(t)=&\frac{1}{n}\int_0^t\exp\left(\frac{1}{n}\lambda_0 s\right)\left\|\boldsymbol{K}_s^m-\boldsymbol{K}^\mathrm{NTK}\right\|_\mathrm{op}\left\|r^\mathrm{NTK}(s)\right\|\mathrm{d}s,\\
    h(t)=&\frac{1}{n}\left\|\boldsymbol{K}_t^m-\boldsymbol{K}^\mathrm{NTK}\right\|_\mathrm{op}.
\end{align*}
    We notice that $t\mapsto h(t)$ is non-negative and $t\mapsto g(t)$ is non-decreasing. Hence, it follows from Grönwall's inequality that:
    \begin{equation*}
        f(t)\leq g(t)\exp\left(\int_0^t h(s)\text{d}s\right).
    \end{equation*}
    Plugging in the definitions of $f(t)$, $g(t)$, and $h(t)$, then dividing by $\exp\left(\frac{1}{n}\lambda_0 t\right)$, we get:
    \begin{align*}
        \left\|e(t)\right\|\leq&\frac{1}{n}\int_0^t\exp\left(-\frac{1}{n}\lambda_0 (t-s)\right)\left\|\boldsymbol{K}_s^m-\boldsymbol{K}^\mathrm{NTK}\right\|_\mathrm{op}\left\|r^\mathrm{NTK}(s)\right\|\mathrm{d}s\\
        &\times\exp\left(\int_0^t\frac{1}{n}\left\|\boldsymbol{K}_s^m-\boldsymbol{K}^\mathrm{NTK}\right\|_\mathrm{op}\text{d}s\right).
    \end{align*}
    Let
    \begin{equation*}
        \Delta=\sup_{t\geq0}\sup_{\|x\|\leq A,\|x'\|\leq A}\left|K_t^m(x,x')-K^\mathrm{NTK}(x,x')\right|.
    \end{equation*}
    Using \cref{eq:op_norm_rntk} to bound $\left\|r^\mathrm{NTK}(s)\right\|$ and the fact that $\left\|\boldsymbol{K}_s^m-\boldsymbol{K}^\mathrm{NTK}\right\|_\mathrm{op}\leq n\Delta$, we obtain:
    \begin{equation*}
        \left\|e(t)\right\|\leq\Delta t\exp\left(\left(\Delta -\frac{\lambda_0}{n}\right)t\right)\sqrt{n}B.\qedhere
    \end{equation*}
\end{proof}

\begin{lemma}[Bound on $\|f_t^\mathrm{NTK}\|_{\LL}$]\label{lem:bound_ftntk}
    Let $t\geq0$. Under \cref{ass:bound_in_res,ass:bound_la_res}, it holds that:
    \begin{equation*}
        \left\|f_t^\mathrm{NTK}\right\|_{\LL}\leq\kappa^\mathrm{NTK}Bt,
    \end{equation*}
    where $\kappa^\mathrm{NTK}\coloneqq(p+2)(A^2+1)+1$.
\end{lemma}

\begin{proof}
    Let $x\in\mathcal{X}$ and $t\geq 0$. We recall that the gradient flow of $\fttntk$ is the following:
    \begin{equation}\label{eq:grad_flow_ntk}
    \frac{\text{d}}{\text{d}t}\fttntk(x)=-\frac{1}{n}\sum_{i=1}^n K^\mathrm{NTK}(x,x_i)r^\mathrm{NTK}_i(t),
    \end{equation}
where $r^\mathrm{NTK}_i(t)=\fttntk(x_i)-y_i$. Using the fact that $f_0^{\mathrm{NTK}}=0$ and the triangle inequality, we obtain:
\begin{equation}\label{eq:bound_fttntk}
    \left|\fttntk(x)\right|=\left|\fttntk(x)-f_0^\mathrm{NTK}(x)\right|=\left|\int_0^t\frac{\text{d}}{\text{d}s}f_s^\mathrm{NTK}(x)\text{d}s\right|\leq\int_0^t\left|\frac{\text{d}}{\text{d}s}f_s^\mathrm{NTK}(x)\right|\text{d}s.
\end{equation}
Next, we find an upper-bound on $\left|\frac{\text{d}}{\text{d}s}f_s^\mathrm{NTK}(x)\right|$. We have:
\begin{equation*}
    \left|\frac{\text{d}}{\text{d}s}f_s^\mathrm{NTK}(x)\right|=\frac{1}{n}\left|\sum_{i=1}^n K^\mathrm{NTK}(x,x_i)r^\mathrm{NTK}_i(s)\right|\leq\frac{\kappa^\mathrm{NTK}}{n}\sum_{i=1}^n\left|r^\mathrm{NTK}_i(s)\right|,
\end{equation*}
where we use \cref{lem:bounded_ntk_kernel} to bound $K^\mathrm{NTK}(x,x_i)$ by $\kappa^\mathrm{NTK}\coloneqq(p+2)(A^2+1)+1$ in the inequality. Using the notation $r^\mathrm{NTK}(s)=[r^\mathrm{NTK}_1(s),\ldots,r^\mathrm{NTK}_n(s)]^\top\in\mathbb{R}^n$, we obtain by the Cauchy-Schwarz inequality:
\begin{equation}\label{eq:norm_diff_ntk_s}
    \left|\frac{\text{d}}{\text{d}s}f_s^\mathrm{NTK}(x)\right|\leq\frac{\kappa^\mathrm{NTK}}{\sqrt{n}}\left\|r^\mathrm{NTK}(s)\right\|.
\end{equation}
By \cref{lem:bound_rntk}, we get:
\begin{equation*}
    \left\|r^\mathrm{NTK}(s)\right\|\leq\exp\left(-\frac{\lambda_0}{n}s\right)\sqrt{n}B,
\end{equation*}
where $\lambda_0$ is the smallest eigenvalue of the matrix $\boldsymbol{K}^\mathrm{NTK}$. Substituting this bound into \cref{eq:norm_diff_ntk_s}, we get:
\begin{equation*}
    \left|\frac{\text{d}}{\text{d}s}f_s^\mathrm{NTK}(x)\right|\leq \kappa^\mathrm{NTK}B\exp\left(-\frac{\lambda_0}{n}s\right).
\end{equation*}
Inserting this bound into \cref{eq:bound_fttntk} gives:
\begin{equation*}
    \left|\fttntk(x)\right|\leq\kappa^\mathrm{NTK}B\int_0^t\exp\left(-\frac{\lambda_0}{n}s\right)\text{d}s=\frac{\kappa^\mathrm{NTK}B n}{\lambda_0}\left(1-\exp\left(-\frac{\lambda_0}{n}t\right)\right)\leq\kappa^\mathrm{NTK}Bt,
\end{equation*}
where we use the identity $1-\exp(-a)\leq a$ for all $a\geq0$. Since $\mu$ is a probability measure, the claimed $\LL$ bound follows.
\end{proof}

\begin{lemma}[High-probability entrywise upper-bound on the parameters at initialization]\label{lem:high_prob_upp_bnd_ini}
    Let $\theta_0 = (\boldsymbol{v}_0, \boldsymbol{u}_0, \boldsymbol{b}_0, a_0)$ be the parameters at initialization, where $\boldsymbol{u}_{0,i} \sim \mathcal{N}(0, \mathbbm{1}_{p\times p})$, $b_{0,i} \sim \mathcal{N}(0,1)$, and $v_{0,i} \sim \mathcal{N}(0,1)$ are drawn independently for $i = 1, \ldots, m/2$, with mirrored copies $\boldsymbol{u}_{0,i+m/2} = \boldsymbol{u}_{0,i}$, $b_{0,i+m/2} = b_{0,i}$, $v_{0,i+m/2} = -v_{0,i}$, and $a_0=0$. Let $m\geq2$. Then, the following holds with probability at least $1-\frac{p+2}{\sqrt{2\pi}}m^{-1/2}$ with respect to random initialization:
    \begin{equation*}
        \max_{1\leq i\leq m}\max\left(\left|v_{0,i}\right|,\left|b_{0,i}\right|,\left\|\boldsymbol{u}_{0,i}\right\|_\infty\right)\leq \sqrt{3\ln m}.
    \end{equation*}
\end{lemma}

\begin{proof}
By the mirrored initialization, the parameters of neuron $i+m/2$ coincide with those of neuron $i$, up to the sign of the outer weight. It therefore suffices to control the $m(p+2)/2$ independent standard Gaussian variables $(u_{0,i})_j$, $b_{0,i}$, and $v_{0,i}$, for $1\leq i\leq m/2$ and $1\leq j\leq p$. For $Z\sim\mathcal{N}(0,1)$ and $t>0$, the classical Gaussian tail bound gives $\mathrm{Pr}(|Z|\geq t)\leq\frac{2}{\sqrt{2\pi}t}\exp(-t^2/2)$. Since $m\geq2$, we have $\sqrt{3\ln m}\geq\sqrt{3\ln 2}>1$, hence $\mathrm{Pr}(|Z|\geq\sqrt{3\ln m})\leq\frac{2}{\sqrt{2\pi}}\exp(-3\ln m/2)=\frac{2}{\sqrt{2\pi}}m^{-3/2}$. A union bound over the $m(p+2)/2$ variables shows that the bound fails with probability at most $\frac{m(p+2)}{2}\times \frac{2}{\sqrt{2\pi}}m^{-3/2}=\frac{p+2}{\sqrt{2\pi}}m^{-1/2}$, which concludes the proof.
\end{proof}

\subsection{Uniform NTK approximation bounds}
\cref{lem:nn_ntk_ini_k}, borrowed from \citep{li2024eigenvalue}, establishes that the empirical NTK of a sufficiently wide fully connected ReLU neural network with mirrored initialization converges uniformly to the infinite-width NTK, both over the entire training trajectory and over all inputs in a bounded domain. This is the key result that allows us to approximate the dynamics of the neural network by those of its NTK surrogate throughout training, which we use in \cref{lem:nn_ntk}.
\begin{proposition}[Kernel uniform convergence -- Proposition~45 in \citep{li2024eigenvalue}]\label{lem:nn_ntk_ini_k}
    Consider $n$ i.i.d. samples $\{(x_i,y_i)\}_{i=1}^n$ and their corresponding augmented inputs $\{\tilde{x}_i\}_{i=1}^n$, where $\tilde{x}_i=[x_i^\top,1]^\top$ for $1\leq i\leq n$. Let $\tilde{M}_X=\sum_{i=1}^n\|\tilde{x}_i\|$, $\mathcal{B}_r=\{x\in\mathcal{X}\mid 1\leq\|\tilde{x}\|\leq r\}$ for $r\geq 1$, and $\lambda_0$ be the smallest eigenvalue of the infinite-width NTK kernel matrix. There exist a constant $C>0$ and a polynomial $\mathrm{poly}$ such that, for any $\delta\in(0,1)$ and $k>0$, when the width satisfies $m\geq\mathrm{poly}(n,\tilde{M}_X,\lambda_0^{-1},|y_1|,\ldots,|y_n|,\ln(1/\delta),k)$ and $m\geq r^k$, with probability at least $1-\delta$ with respect to random samples and random initialization, we have:
    \begin{equation*}
        \sup_{t\geq0}\sup_{x,x'\in \mathcal{B}_r}\left|K_t^m(x,x')-K^\mathrm{NTK}(x,x')\right|\leq Cr^2m^{-\frac{1}{12}}\sqrt{\ln m},
    \end{equation*}
    where the constant $C>0$ is independent of $\delta$ and $n$.
\end{proposition}

\begin{lemma}[Kernel uniform convergence -- Application of \cref{lem:nn_ntk_ini_k} to our setting]\label{lem:nn_ntk}
    Consider $n\geq 3$ i.i.d. samples $\{(x_i,y_i)\}_{i=1}^n\sim\rho$ and fix $p\geq 2$. Under \Cref{ass:bound_in_res,ass:bound_la_res,ass:bounded_density}, there exist a constant $C>0$ and a polynomial $\mathrm{poly}$ such that, for any $\delta\in(0,1)$, when the width satisfies $m\geq\mathrm{poly}(n,1/\delta,\ln n,\ln(1/\delta))$, with probability at least $1-\delta$ with respect to random samples and random initialization, we have:
    \begin{equation*}
        \sup_{t\geq0}\sup_{\|x\|\leq A,\|x'\|\leq A}\left|K_t^m(x,x')-K^\mathrm{NTK}(x,x')\right|\leq C m^{-\frac{1}{12}}\sqrt{\ln m},
    \end{equation*}
    where the constant $C>0$ is independent of $\delta$ and $n$, but depends on $A$, and the coefficients of $\mathrm{poly}$ depend on $A$, $B$, $M_\mu$, and $p$.
\end{lemma}

\begin{proof}
    In this proof, we apply the statement of \cref{lem:nn_ntk_ini_k} to our setting. Under \Cref{ass:bound_in_res}, we know that $\|x\|\leq A$, for any $x\in\mathcal{X}$, which implies that $\tilde{M}_X=\sum_{i=1}^n\|\tilde{x}_i\|\leq n\sqrt{A^2+1}$. Moreover, under \Cref{ass:bound_la_res}, we know that $|y_i|\leq B$, for $1\leq i\leq n$. By taking $k=1$ and $r=\sqrt{A^2+1}$ in \cref{lem:nn_ntk_ini_k}, we require that $m$ should satisfy $m\geq\mathrm{poly}(n,\lambda_0^{-1},\ln(1/\delta))$ and $m\geq\sqrt{A^2+1}$, where the coefficients of $\mathrm{poly}$ depend on $A$ and $B$. Hence, we get that $m\geq\max(\mathrm{poly}(n,\lambda_0^{-1},\ln(1/\delta)),\sqrt{A^2+1})$. After redefining $\mathrm{poly}(n,\lambda_0^{-1},\ln(1/\delta))$ to be the polynomial that dominates both lower bounds on $m$, we simplify the requirement to $m\geq\mathrm{poly}(n,\lambda_0^{-1},\ln(1/\delta))$. By \cref{cor:up_bound_lam_in}, since $n\geq 3$, the following holds with probability at least $1-\delta$ with respect to random samples: $\lambda_0^{-1}\leq P(n,1/\delta,\ln n,\ln(1/\delta))$, where $P(n,1/\delta,\ln n,\ln(1/\delta))\coloneqq C'n^{8/p}\delta^{-4/p}\left(\ln n+\ln(1/\delta)\right)^3$ is a polynomial whose coefficients depend on $A, M_\mu,$ and $p$. Consequently, it suffices to choose $m\geq\mathrm{poly}(n,P(n,1/\delta,\ln n,\ln(1/\delta)),\ln(1/\delta))$. After redefining $\mathrm{poly}(n,1/\delta,\ln n,\ln(1/\delta))$ to be the polynomial $\mathrm{poly}(n,P(n,1/\delta,\ln n,\ln(1/\delta)),\ln(1/\delta))$, we obtain that $m\geq\mathrm{poly}(n,1/\delta,\ln n,\ln(1/\delta))$ where the coefficients of $\mathrm{poly}$ depend on $A$, $B$, $M_\mu$, and $p$. Since we want both events in \cref{lem:nn_ntk_ini_k,cor:up_bound_lam_in} to hold simultaneously, then, using a union bound, they both hold with probability at least $1-2\delta$. Redefining $\delta$ to be $2\delta$ concludes the proof.
\end{proof}

\begin{lemma}[Uniform NTK approximation of the neural network predictor]\label{lem:nn_ntk_ini}
    Consider $n\geq3$ i.i.d. samples $\{(x_i,y_i)\}_{i=1}^n\sim\rho$ and fix $p\geq 2$. Under \Cref{ass:bound_in_res,ass:bound_la_res,ass:bounded_density}, there exist a constant $C>0$ and a polynomial $\mathrm{poly}$ such that, for any $\delta\in(0,1)$, when the width satisfies $m\geq\mathrm{poly}(n,1/\delta,\ln n,\ln(1/\delta))$, with probability at least $1-\delta$ with respect to random samples and random initialization, we have:
    \begin{equation*}
        \sup_{t\geq0}\sup_{\|x\|\leq A}\left|\ftt(x)-\fttntk(x)\right|\leq C m^{-\frac{1}{24}}\sqrt{\ln m},
    \end{equation*}
    where the constant $C>0$ is independent of $\delta$ and $n$, but depends on $A$, and the coefficients of $\mathrm{poly}$ depend on $A$, $B$, $M_\mu$, and $p$.
\end{lemma}

\begin{proof}
    We start by recalling that $f_0^\mathrm{NN}$ and $f_0^\mathrm{NTK}$ are assumed to be equal to $0$. Using this, for any $x\in\mathcal{X}$, we have:
    \begin{align*}
        \left|\ftt(x)-\fttntk(x)\right|=&\left|\int_0^t\left(\frac{\text{d}}{\text{d}s}f_s^\mathrm{NN}(x)-\frac{\text{d}}{\text{d}s}f_s^\mathrm{NTK}(x)\right)\text{d}s\right|\\
        \leq&\int_0^t\left|\frac{\text{d}}{\text{d}s}f_s^\mathrm{NN}(x)-\frac{\text{d}}{\text{d}s}f_s^\mathrm{NTK}(x)\right|\text{d}s,
    \end{align*}
    where we use the triangle inequality in the second line. We recall that the gradient flows of $f_s^\mathrm{NN}$ and $f_s^\mathrm{NTK}$ are the following:
    \begin{equation*}
    \frac{\text{d}}{\text{d}s}f_s^\mathrm{NN}(x)=-\frac{1}{n}K_s^m(x,\boldsymbol{X})^\top r^\mathrm{NN}(s)\quad\text{and}\quad\frac{\text{d}}{\text{d}s}f_s^\mathrm{NTK}(x)=-\frac{1}{n}K^\mathrm{NTK}(x,\boldsymbol{X})^\top r^\mathrm{NTK}(s),
    \end{equation*}
    where:
    \begin{align*}
        K_s^m(x,\boldsymbol{X})=&\left[K_s^m(x,x_1),\ldots,K_s^m(x,x_n)\right]^\top\in\mathbb{R}^n,\\
        K^\mathrm{NTK}(x,\boldsymbol{X})=&[K^\mathrm{NTK}(x,x_1),\ldots,K^\mathrm{NTK}(x,x_n)]^\top\in\mathbb{R}^n,\\
        r^\mathrm{NN}(s)=&f_s^\mathrm{NN}(\boldsymbol{X})-\boldsymbol{Y}=[f_s^\mathrm{NN}(x_1)-y_1,\ldots,f_s^\mathrm{NN}(x_n)-y_n]^\top\in\mathbb{R}^n,\\
        r^\mathrm{NTK}(s)=&f_s^\mathrm{NTK}(\boldsymbol{X})-\boldsymbol{Y}=[f_s^\mathrm{NTK}(x_1)-y_1,\ldots,f_s^\mathrm{NTK}(x_n)-y_n]^\top\in\mathbb{R}^n.
    \end{align*}
    Consequently, we get:
    \begin{align}\label{eq:ftnn_ftntk_in}
        &\left|\ftt(x)-\fttntk(x)\right|\nonumber\\
        \leq&\int_0^t\left|\frac{\text{d}}{\text{d}s}f_s^\mathrm{NN}(x)-\frac{\text{d}}{\text{d}s}f_s^\mathrm{NTK}(x)\right|\text{d}s\nonumber\\
        =&\int_0^t\left|-\frac{1}{n}K_s^m(x,\boldsymbol{X})^\top r^\mathrm{NN}(s)+\frac{1}{n}K^\mathrm{NTK}(x,\boldsymbol{X})^\top r^\mathrm{NTK}(s)\right|\text{d}s\nonumber\\
        =&\frac{1}{n}\int_0^t\left|K_s^m(x,\boldsymbol{X})^\top r^\mathrm{NN}(s)-\left(K^\mathrm{NTK}(x,\boldsymbol{X})-K^m_s(x,\boldsymbol{X})+K^m_s(x,\boldsymbol{X})\right)^\top r^\mathrm{NTK}(s)\right|\text{d}s\nonumber\\
        =&\frac{1}{n}\int_0^t\left|K_s^m(x,\boldsymbol{X})^\top(r^\mathrm{NN}(s)-r^\mathrm{NTK}(s))+\left(K^m_s(x,\boldsymbol{X})-K^\mathrm{NTK}(x,\boldsymbol{X})\right)^\top r^\mathrm{NTK}(s)\right|\text{d}s\nonumber\\
        \leq&\frac{1}{n}\int_0^t\left(\left\|K_s^m(x,\boldsymbol{X})-K^\mathrm{NTK}(x,\boldsymbol{X})\right\|+\left\|K^\mathrm{NTK}(x,\boldsymbol{X})\right\|\right)\left\|r^\mathrm{NN}(s)-r^\mathrm{NTK}(s)\right\|\text{d}s\nonumber\\
        &+\frac{1}{n}\int_0^t\left\|K^m_s(x,\boldsymbol{X})-K^\mathrm{NTK}(x,\boldsymbol{X})\right\|\left\|r^\mathrm{NTK}(s)\right\|\text{d}s,
    \end{align}
    where we use the triangle inequality in the last inequality. Let:
    \begin{equation*}
        \Delta=\sup_{t\geq0}\sup_{\|x\|\leq A,\|x'\|\leq A}\left|K_t^m(x,x')-K^\mathrm{NTK}(x,x')\right|.
    \end{equation*}
    We notice that:
    \begin{align}
        \left\|K_s^m(x,\boldsymbol{X})-K^\mathrm{NTK}(x,\boldsymbol{X})\right\|=&\sqrt{\sum_{i=1}^n\left|K_s^m(x,x_i)-K^\mathrm{NTK}(x,x_i)\right|^2}\leq\sqrt{n}\Delta,\label{eq:knn_kntk_b}\\
        \left\|K^\mathrm{NTK}(x,\boldsymbol{X})\right\|=&\sqrt{\sum_{i=1}^n\left|K^\mathrm{NTK}(x,x_i)\right|^2}\leq\sqrt{n}\kappa^\mathrm{NTK},\label{eq:kntk_b}
    \end{align}
    where in the second line we use \cref{lem:bounded_ntk_kernel} to bound $\left|K^\mathrm{NTK}(x,x_i)\right|$ by $\kappa^\mathrm{NTK}\coloneqq(p+2)(A^2+1)+1$. Moreover, by \cref{lem:bound_rntk}, we get:
\begin{align}
    \left\|r^\mathrm{NTK}(s)\right\|\leq&\exp\left(-\frac{\lambda_0}{n}s\right)\sqrt{n}B,\label{eq:rntk_b}\\
    \left\|r^\mathrm{NN}(s)-r^\mathrm{NTK}(s)\right\|\leq&\Delta s\exp\left(\gamma s\right)\sqrt{n}B\label{eq:rnn_rntk_b},
\end{align}
where $\lambda_0$ is the smallest eigenvalue of the matrix $\boldsymbol{K}^\mathrm{NTK}$ and $\gamma=\Delta -\frac{\lambda_0}{n}$. Substituting \cref{eq:knn_kntk_b,eq:kntk_b,eq:rntk_b,eq:rnn_rntk_b} into \cref{eq:ftnn_ftntk_in}, we obtain:
\begin{equation*}
    \left|\ftt(x)-\fttntk(x)\right|\leq\left(\Delta+\kappa^\mathrm{NTK}\right)\Delta B\int_0^t s\exp\left(\gamma s\right)\text{d}s+\Delta B\int_0^t\exp\left(-\frac{\lambda_0}{n}s\right)\text{d}s.
\end{equation*}
Applying integration by parts gives:
\begin{align*}
    \int_0^t s\exp\left(\gamma s\right)\text{d}s=&\left[\frac{s}{\gamma}\exp\left(\gamma s\right)\right]_{s=0}^{s=t}-\frac{1}{\gamma}\int_0^t \exp\left(\gamma s\right)\text{d}s=\frac{t}{\gamma}\exp\left(\gamma t\right)-\frac{1}{\gamma^2}\left(\exp\left(\gamma t\right)-1\right)\\
    =&\frac{\exp\left(\gamma t\right)}{\gamma^2}\left(\gamma t-1\right)+\frac{1}{\gamma^2}.
\end{align*}
Also, we have:
\begin{equation*}
    \int_0^t\exp\left(-\frac{\lambda_0}{n}s\right)\text{d}s=\left[-\frac{n}{\lambda_0}\exp\left(-\frac{\lambda_0}{n}s\right)\right]_{s=0}^{s=t}=\frac{n}{\lambda_0}\left[1-\exp\left(-\frac{\lambda_0}{n}t\right)\right]\leq\frac{n}{\lambda_0}.
\end{equation*}
Thus:
\begin{equation}\label{eq:fttftntk_2}
    \left|\ftt(x)-\fttntk(x)\right|\leq\Delta B\left[\left(\Delta+\kappa^\mathrm{NTK}\right)\left(\frac{\exp\left(\gamma t\right)}{\gamma^2}\left(\gamma t-1\right)+\frac{1}{\gamma^2}\right)+\frac{n}{\lambda_0}\right].
\end{equation}
By \cref{lem:nn_ntk}, we know that with probability at least $1-\delta$ with respect to random samples and random initialization, it holds that $\Delta\leq Cm^{-\frac{1}{12}}\sqrt{\ln m}$ when $m$ is large enough such that $m\geq\mathrm{poly}(n,1/\delta,\ln n,\ln(1/\delta))$, where the constant $C>0$ is independent of $\delta$ and $n$, but depends on $A$, and the coefficients of $\mathrm{poly}$ depend on $A$, $B$, $M_\mu$, and $p$. When $m\geq\max\left(\left(\frac{2C\sqrt{12}n}{\lambda_0}\right)^{24},1\right)$, we obtain by \cref{lem:tech_lem_b} that $\Delta\leq\frac{\lambda_0}{2n}$, which implies that $\gamma\leq-\frac{\lambda_0}{2n}<0$. Similarly, when $m\geq\max\left(\left(C\sqrt{12}\right)^{24},1\right)$, it holds by \cref{lem:tech_lem_b} that $\Delta\leq 1$. Hence, when $m\geq\max\left(\mathrm{poly}(n,1/\delta,\ln n,\ln(1/\delta)),\left(\frac{2C\sqrt{12}n}{\lambda_0}\right)^{24},\left(C\sqrt{12}\right)^{24},1\right)$, it holds that $\Delta\leq\min\left(Cm^{-\frac{1}{12}}\sqrt{\ln m},\frac{\lambda_0}{2n},1\right)$. Consequently, with probability at least $1-\delta$ with respect to random samples and random initialization, \cref{eq:fttftntk_2} simplifies to:
\begin{align}
    \left|\ftt(x)-\fttntk(x)\right|\leq& Cm^{-\frac{1}{12}}\sqrt{\ln m} B\left[\left(1+\kappa^\mathrm{NTK}\right)\frac{4n^2}{\lambda_0^2}+\frac{n}{\lambda_0}\right]\nonumber\\
    \leq&Cm^{-\frac{1}{12}}\sqrt{\ln m} B n^2\left[\left(1+\kappa^\mathrm{NTK}\right)\frac{4}{\lambda_0^2}+\frac{1}{\lambda_0}\right].\label{eq:ftt_fttntk_eq_int}
\end{align}
By \cref{cor:up_bound_lam_in}, since $n\geq 3$, the following holds with probability at least $1-\delta$ with respect to random samples: $\lambda_0^{-1}\leq\mathrm{poly}_1(n,1/\delta,\ln n,\ln(1/\delta))$, where $\mathrm{poly}_1(n,1/\delta,\ln n,\ln(1/\delta))\coloneqq C'n^{8/p}\delta^{-4/p}\left(\ln n+\ln(1/\delta)\right)^3$ is a polynomial whose coefficients depend on $A, M_\mu,$ and $p$. Consequently, we obtain:
    \begin{equation}\label{lem:lambda0_1_2}
     \lambda_0^{-1}\leq\mathrm{poly}_1(n,1/\delta,\ln n,\ln(1/\delta))\quad\text{and}\quad\lambda_0^{-2}\leq\left(\mathrm{poly}_1(n,1/\delta,\ln n,\ln(1/\delta))\right)^2.
    \end{equation}
    Using this, we require that $m\geq P_1(n,1/\delta,\ln n,\ln(1/\delta))$ where $P_1$ is a polynomial that dominates:
    \begin{equation*}
        \max\left(\mathrm{poly}(n,1/\delta,\ln n,\ln(1/\delta)),\left(2C\sqrt{12}n\mathrm{poly}_1\left(n,\frac{1}{\delta},\ln n,\ln\frac{1}{\delta}\right)\right)^{24},\left(C\sqrt{12}\right)^{24},1\right).
    \end{equation*}
    Plugging \cref{lem:lambda0_1_2} in \cref{eq:ftt_fttntk_eq_int}, we get:
    \begin{equation*}
        \left|\ftt(x)-\fttntk(x)\right|\leq Cm^{-\frac{1}{12}}\sqrt{\ln m} P_2\left(n,\frac{1}{\delta},\ln n,\ln\frac{1}{\delta}\right),
    \end{equation*}
    where $P_2$ is a polynomial expressed as:
    \begin{multline*}
        P_2\left(n,\frac{1}{\delta},\ln n,\ln\frac{1}{\delta}\right)=B n^2\Bigg[4(1+\kappa^\mathrm{NTK})\left(\mathrm{poly}_1\left(n,\frac{1}{\delta},\ln n,\ln\frac{1}{\delta}\right)\right)^2\\+\mathrm{poly}_1\left(n,\frac{1}{\delta},\ln n,\ln\frac{1}{\delta}\right)\Bigg].
    \end{multline*}
    When $m\geq\max\left(P_1(n,1/\delta,\ln n,\ln(1/\delta)),\left(P_2(n,1/\delta,\ln n,\ln(1/\delta))\right)^{24}\right)$, we obtain:
    \begin{equation*}
        \left|\ftt(x)-\fttntk(x)\right|\leq Cm^{-\frac{1}{24}}\sqrt{\ln m}.
    \end{equation*}
    We conclude the proof by redefining $\mathrm{poly}$ to be the polynomial that dominates:
    \begin{equation*}
        \max\left(P_1(n,1/\delta,\ln n,\ln(1/\delta)),\left(P_2(n,1/\delta,\ln n,\ln(1/\delta))\right)^{24}\right).\qedhere
    \end{equation*}
\end{proof}

\subsection{Generalization guarantees}\label{sec:gen_si}
\subsubsection{High-probability generalization bounds}
\cref{cor:gen_li_al}, taken from \citep{lin2020optimal}, provides a high-probability bound on the generalization error of a broad class of \emph{spectral algorithms} \citep[Algorithm~1]{lin2020optimal} applied to kernel regression in an RKHS. This class includes gradient methods \citep[Example~3.2]{lin2020optimal} and ridge regression \citep[Example~3.3]{lin2020optimal} as special cases, unified through the notion of a \emph{filter function} \citep[Definition~3.1]{lin2020optimal}. The result shows that the estimator produced by any of these algorithms converges to the best approximation of the regression function in the RKHS at a rate that depends on the smoothness of the target function through the \emph{source condition}, and on the complexity of the hypothesis space through the \emph{capacity condition}. This is the key result that we use in \cref{lem:adap_gen_ntk} to control the estimation error of the NTK gradient flow, by identifying the RKHS with the one associated to the \emph{infinite-width} NTK and verifying that the assumptions required by this corollary are satisfied in our setting.

\begin{corollary}[Probabilistic generalization bound of the NTK predictor -- Corollary~4.4 in \citep{lin2020optimal}]\label{cor:gen_li_al}
Let $\mathcal{H}$ be an RKHS associated with a Mercer kernel $K:\mathcal{X}\times\mathcal{X}\to\mathbb{R}$, $\mathcal{T}:\LL\to\LL$ be the integral operator, and $\Sigma:\mathcal{H}\to\mathcal{H}$ be the covariance operator, all defined in \cref{sec:int_space}. Denote by $f_\mathcal{H}$ the projection of the regression function $f^\star$ onto the closure of $\mathcal{H}$ in $\LL$. Given i.i.d. samples $\{(x_i,y_i)\}_{i=1}^n$ drawn from a distribution $\rho$ with marginal $\mu$ on $\mathcal{X}$, we consider the following empirical risk minimization problem:
\begin{equation*}
    f_\lambda=\argmin_{f\in\mathcal{H}}\frac{1}{2n}\sum_{i=1}^n\left(f(x_i)-y_i\right)^2,
\end{equation*}
which we want to solve using a spectral algorithm with filter function $G_\lambda$ indexed with $\lambda>0$. We make the following assumptions:
\begin{assumpar}
\item\label{ass:in_bound}(Bounded kernel) There exists a constant $\kappa\geq 1$ such that $K(x,x')\leq\kappa^2$ for any $x,x'\in\mathcal{X}$.
\item\label{ass:mom_targets}(Moment condition) There exist positive constants $Q,J>0$ such that for all integers $l\geq 2$, we have $\mathbb{E}_{y\sim\rho}[|y|^l\mid x]\leq\frac{1}{2}l!J^{l-2}Q^2$.
\item\label{ass:reg_smo_fH}(Regularity/smoothness of $f_\mathcal{H}$) $f_\mathcal{H}$ satisfies $\mathbb{E}_{x\sim\mu}[|f_\mathcal{H}(x)-f^\star(x)|^2 K(x,\cdot)\otimes K(x,\cdot)]\preceq U^2 \Sigma$ and the following source condition $f_\mathcal{H}=\phi(\mathcal{T})g_0$ with $\|g_0\|_{\LL}\leq S$, where $U,S\geq 0$, $\phi(u)=\kappa^{-2\max(0,\zeta-1)}u^\zeta$, and $\zeta\geq0$.
\item\label{ass:cap_cond}(Capacity condition) For some $\gamma\in]0,1]$ and $c_\gamma>0$, the effective dimension of $\mathcal{H}$ satisfies $d_{\mathrm{eff}}(\lambda)\leq c_\gamma\lambda^{-\gamma}$ for all $\lambda>0$, where $d_{\mathrm{eff}}(\lambda)=\mathrm{Trace}(\Sigma(\Sigma+\lambda\mathbbm{1}_\mathcal{H})^{-1})$.
\end{assumpar}
Let $a\in[0,\min(\frac{1}{2},\zeta)]$ and $\lambda_\mathrm{op}=n^{-\frac{1}{2\zeta+\gamma}}$. There exists a constant $C>0$ such that, for any $\delta\in(0,1)$, with probability at least $1-\delta$ with respect to random samples, we have:
\begin{equation*}
    \left\|\mathcal{T}^{-a}(f_{\lambda_\mathrm{op}}-f_\mathcal{H})\right\|_{\LL}\leq C\left(\ln\frac{6}{\delta}\right)^{2-a}n^{-\frac{\zeta-a}{2\zeta+\gamma}}\quad\text{if }2\zeta+\gamma>1,
\end{equation*}
where the constant $C>0$ is independent of $\delta$, $n$, and $\lambda_\mathrm{op}$, but depends on $\kappa$, $c_\gamma$, $\gamma$, $\zeta$, $U$, $J$, $Q$, $S$, and $a$.
\end{corollary}
\begin{lemma}[Probabilistic generalization bound of the NTK predictor -- Application of \cref{cor:gen_li_al} to our setting]\label{lem:adap_gen_ntk}
    Consider $n$ i.i.d. samples $\{(x_i,y_i)\}_{i=1}^n\sim\rho$. Let \Cref{ass:bound_in_res,ass:bound_la_res,ass:source_cond,ass:bounded_density} hold. If we choose a stopping time $t_{\mathrm{op}}=n^{\frac{p+1}{s(p+1)+p}}$ with $\frac{1}{p+1}<s\leq 2$, then there exists a constant $C>0$ such that, for any $\delta\in(0,1)$, with probability at least $1-\delta$ with respect to random samples, we have:
    \begin{equation*}
        \left\|\fttopntk-f^\star\right\|_{\LL}\leq C\left(\ln\frac{6}{\delta}\right)^2 n^{-\frac{1}{2}\frac{s(p+1)}{s(p+1)+p}},
    \end{equation*}
    where the constant $C>0$ is independent of $\delta$, $n$, and $t_{\mathrm{op}}$, but depends on $s$, $p$, $A$, $B$, $M_\mu$, and $R$.
\end{lemma}

\begin{proof}
In this proof, we apply the result in \cref{cor:gen_li_al} to our setting using our assumptions. For that, we set $a=0$, $s=2\zeta$ to be the smoothness parameter of the interpolation space $[\hntk]^s$, and $\beta\coloneqq\frac{p+1}{p}=\frac{1}{\gamma}$ to be the eigenvalue decay rate of $\kntk$ as shown in \cref{lem:edr_set} under \cref{ass:bound_in_res,ass:bounded_density}, where $a$, $\zeta$, and $\gamma$ are the notations used in \cref{cor:gen_li_al}. We verify that the assumptions in \cref{cor:gen_li_al} hold in our setting for $\kntk$.
\begin{itemize}
    \item \cref{ass:in_bound} is satisfied in our setting under \cref{ass:bound_in_res}, as shown in \cref{lem:bounded_ntk_kernel}, since it suffices to choose $\kappa=\sqrt{(p+2)(A^2+1)+1}\geq1$.
    \item \cref{ass:mom_targets} holds in our setting under \cref{ass:bound_la_res} since the fact that $|y|\leq B$ implies $\mathbb{E}_{y\sim\rho}[|y|^l\mid x]\leq B^{l}=B^{l-2}B^2\leq \frac{1}{2}l!B^{l-2}B^2$ for all $l\geq 2$, so it suffices to choose $J=B$ and $Q=B$.
    \item The first part of \cref{ass:reg_smo_fH} is satisfied in our setting under \cref{ass:bound_in_res,ass:source_cond} for $U=0$, because $f^\star$ coincides with its projection onto the closure of $\hntk$ in $\LL$. In fact, we show in \cref{lem:fstar_closure} that $f^\star$ belongs to the closure of $\hntk$ in $\LL$, which means that its projection onto that closure is itself. The second part of \cref{ass:reg_smo_fH} is satisfied for $S=R$ with $0\leq\zeta\leq 1$ under \cref{ass:source_cond}. The condition on $\zeta$ implies that $0\leq s\leq 2$.
    \item \cref{ass:cap_cond} is a consequence of \cref{lem:edr_set}. In fact, letting $(\lambda_i)_{i\geq 1}$ denote the eigenvalues of $\Sigma$, we have the following upper-bound on the effective dimension of $\mathcal{H}$:
\begin{align*}
    d_\mathrm{eff}(\lambda)=&\mathrm{Trace}(\Sigma(\Sigma+\lambda\mathbbm{1}_\mathcal{H})^{-1})=\sum_{i\geq 1}\frac{\lambda_i}{\lambda_i+\lambda}\leq\sum_{i\geq1}\min\left(1,\frac{k_2i^{-\beta}}{\lambda}\right)\\
    \leq&\int_0^\infty\min\left(1,\frac{k_2 x^{-\beta}}{\lambda}\right)\mathrm{d}x=\frac{\beta}{\beta-1}k_2^\frac{1}{\beta}\lambda^{-\frac{1}{\beta}},
\end{align*}
where we use $\frac{\lambda_i}{\lambda_i+\lambda}\leq\min\left(1,\frac{\lambda_i}{\lambda}\right)$, the upper bound $\lambda_i\leq k_2i^{-\beta}$ of \cref{lem:edr_set}, and the fact that $x\mapsto\min(1,k_2x^{-\beta}/\lambda)$ is non-increasing. The integral equals $x_0+\frac{x_0}{\beta-1}$ with $x_0=(k_2/\lambda)^{1/\beta}$. This is precisely \cref{ass:cap_cond} with $c_\gamma=\frac{\beta}{\beta-1}k_2^{1/\beta}=(p+1)k_2^\frac{p}{p+1}$, since $\beta=\frac{p+1}{p}$. 
\end{itemize}
It remains to identify the filter function and the optimal stopping time. According to \citep[Lemma~D.1]{lai2023generalization}, for a given $\lambda>0$, the filter function of the gradient flow is given by:
\begin{equation*}
    G_\lambda(u)=\frac{1-\exp(-\frac{u}{\lambda})}{u},
\end{equation*}
where $0\leq u\leq\kappa^\mathrm{NTK}$ with $\kappa^\mathrm{NTK}\coloneqq(p+2)(A^2+1)+1$ by \cref{lem:bounded_ntk_kernel}. From \citep[Proof~of~Lemma~D.1]{lai2023generalization}, the continuous time parameter is $t=\frac{1}{\lambda}$. Using the fact that $\lambda_\mathrm{op}^{-1}=n^{\frac{1}{2\zeta+\gamma}}=n^{\frac{\beta}{s\beta+1}}$ allows us to obtain that $t_{\mathrm{op}}=n^{\frac{\beta}{s\beta+1}}=n^{\frac{p+1}{s(p+1)+p}}$, where $t_\mathrm{op}$ is the stopping time of the gradient flow algorithm. It is important to note that the result in \cref{cor:gen_li_al} only holds for $s>1-\frac{1}{\beta}=\frac{1}{p+1}$. Combining this with $0\leq s\leq 2$ provided earlier gives $\frac{1}{p+1}<s\leq 2$.
\end{proof}

\begin{proposition}[Probabilistic generalization bound]\label{prop:gen_bound}
    Consider $n\geq 3$ i.i.d. samples $\{(x_i,y_i)\}_{i=1}^n\sim\rho$ and fix $p\geq 2$. Let \Cref{ass:bound_in_res,ass:bound_la_res,ass:source_cond,ass:bounded_density} hold. If we choose a stopping time $t_{\mathrm{op}}=n^{\frac{p+1}{s(p+1)+p}}$ with $\frac{1}{p+1}<s\leq 2$, then there exist a constant $C>0$ and a polynomial $\mathrm{poly}$ such that, for any $\delta\in(0,1)$, when the width satisfies $m\geq\mathrm{poly}(n,1/\delta,\ln n,\ln(1/\delta))$, with probability at least $1-\delta$ with respect to random samples and random initialization, we have:
    \begin{equation*}
        \left\|\fttop-f^\star\right\|_{\LL}\leq C\left(\ln\frac{6}{\delta}\right)^2 n^{-\frac{1}{2}\frac{s(p+1)}{s(p+1)+p}},
    \end{equation*}
    where the constant $C>0$ is independent of $\delta$, $n$ and $t_{\mathrm{op}}$, but depends on $s$, $p$, $A$, $B$, $M_\mu$, and $R$, and the coefficients of $\mathrm{poly}$ depend on $A$, $B$, $M_\mu$, and $p$.
\end{proposition}
\begin{proof}
    By the triangle inequality and the identity $\|\cdot\|_{\LL}\leq\|\cdot\|_\infty$, we have:
    \begin{align}\label{eq:ftt_op_ntk}
        \left\|\fttop-f^\star\right\|_{\LL}&\leq\left\|\fttop-\fttopntk\right\|_{\LL}+\left\|\fttopntk-f^\star\right\|_{\LL}\nonumber\\
        &\leq\left\|\fttop-\fttopntk\right\|_\infty+\left\|\fttopntk-f^\star\right\|_{\LL}.
    \end{align}
    Applying \cref{lem:adap_gen_ntk}, we obtain under \Cref{ass:bound_in_res,ass:bound_la_res,ass:bounded_density,ass:source_cond} that there exists a constant $C_1>0$ such that, for any $\delta\in(0,1)$, the following holds with probability at least $1-\frac{\delta}{2}$ with respect to random samples:
    \begin{equation}\label{eq:ntk_grad_flow_bound}
        \left\|\fttopntk-f^\star\right\|_{\LL}\leq C_1\left(\ln\frac{6}{\delta}\right)^2n^{-\frac{1}{2}\frac{s(p+1)}{s(p+1)+p}},
    \end{equation}
    where $C_1>0$ is a constant independent of $\delta$, $n$, and $t_{\mathrm{op}}$, and that depends on $s$, $p$, $A$, $B$, $M_\mu$, and $R$. By \Cref{lem:nn_ntk_ini}, under \Cref{ass:bound_in_res,ass:bound_la_res,ass:bounded_density}, we know that there exist a constant $C_2>0$ and a polynomial $P$ such that, for any $\delta\in(0,1)$, when $m\geq P(n,1/\delta,\ln n,\ln(1/\delta))$, with probability at least $1-\frac{\delta}{2}$ with respect to random samples and random initialization, we have:
    \begin{equation*}
        \left\|\fttop-\fttopntk\right\|_\infty\leq\sup_{\|x\|\leq A}\left|\fttop(x)-\fttopntk(x)\right|\leq C_2 m^{-\frac{1}{24}}\sqrt{\ln m},
    \end{equation*}
    where the constant $C_2>0$ is independent of $\delta$ and $n$, but depends on $A$, and the coefficients of $P$ depend on $A$, $B$, $M_\mu$, and $p$. Choosing $m\geq\max\left(P(n,1/\delta,\ln n,\ln(1/\delta)),n^{24}(\ln n)^{12}\right)$ yields $m^{-\frac{1}{24}}\sqrt{\ln m}\leq 6n^{-1}$ by \cref{lem:tech_b} since $n\geq 3$. This implies that:
    \begin{equation}\label{eq:inf_ftop_fttntk}
        \left\|\fttop-\fttopntk\right\|_\infty\leq 6C_2n^{-1}.
    \end{equation}
    Substituting the bounds in \cref{eq:ntk_grad_flow_bound,eq:inf_ftop_fttntk} into \cref{eq:ftt_op_ntk}, we obtain that:
    \begin{equation*}
        \left\|\fttop-f^\star\right\|_{\LL}\leq 6C_2n^{-1}+C_1\left(\ln\frac{6}{\delta}\right)^2 n^{-\frac{1}{2}\frac{s(p+1)}{s(p+1)+p}}.
    \end{equation*}
    Noticing that $n^{-1}\leq n^{-\frac{1}{2}\frac{s(p+1)}{s(p+1)+p}}$, we get:
    \begin{equation*}
        \left\|\fttop-f^\star\right\|_{\LL}\leq\left(6C_2+C_1\left(\ln\frac{6}{\delta}\right)^2\right) n^{-\frac{1}{2}\frac{s(p+1)}{s(p+1)+p}}.
    \end{equation*}
    Given that $\delta\in(0,1)$, it holds that $\ln\frac{6}{\delta}>\ln 6>1$. Hence, we get $\left(\ln\frac{6}{\delta}\right)^2>1$. Therefore, $6C_2\leq6C_2\left(\ln\frac{6}{\delta}\right)^2$. As a result, combining the terms gives:
    \begin{equation*}
        \left\|\fttop-f^\star\right\|_{\LL}\leq\left(6C_2+C_1\right)\left(\ln\frac{6}{\delta}\right)^2 n^{-\frac{1}{2}\frac{s(p+1)}{s(p+1)+p}}.
    \end{equation*}
    Defining the constant $C\coloneqq 6C_2+C_1$ establishes the desired convergence rate. Since we want both bounds in \cref{eq:ntk_grad_flow_bound,eq:inf_ftop_fttntk} to hold simultaneously, then, using a union bound, they both hold with probability at least $1-\delta$. We conclude the proof by choosing $m\geq\mathrm{poly}\left(n,1/\delta,\ln n,\ln(1/\delta)\right)$, where $\mathrm{poly}$ is a polynomial that dominates $\max\left(P(n,1/\delta,\ln n,\ln(1/\delta)),n^{24}(\ln n)^{12}\right)$.
\end{proof}

\subsubsection{Expected generalization bounds}\label{par:events}\label{subsec:exp_gen}
Deriving generalization bounds in expectation is essential to characterize the convergence rate of optimization algorithms, whose guarantees are naturally stated in expectation. However, the standard approach of converting high-probability bounds to expected bounds via $\mathbb{E}[X]=\int_0^\infty\mathrm{Pr}(X>t)\mathrm{d}t$ for $X\geq 0$ does not apply here, as the required network width $m$ depends on $\delta$ (see \cref{prop:gen_bound}) in a way that prevents a direct integration over $\delta$. To overcome this, we introduce truncated neural network predictors, obtained by setting the networks to zero on a carefully defined low-probability event, which allows us to pass from high-probability bounds to expected bounds while maintaining control over the contribution of the low-probability event. To this end, we introduce six events that will be used to control the generalization error of the early-stopped neural network in expectation. 

\textbf{High-probability events governing the generalization error. }Recall that $\fttop$ denotes the neural network trained by gradient flow and stopped at time $t_\mathrm{op}$. To analyze the generalization error of $\fttop$, we decompose it as:
\begin{equation*}
    \left\|\fttop-f^\star\right\|_{\LL}\leq\left\|\fttop-\fttopntk\right\|_{\LL}+\left\|\fttopntk-f^\star\right\|_{\LL},
\end{equation*}
where $\fttopntk$ denotes the corresponding NTK gradient flow stopped at the same time $t_\mathrm{op}$. The first term measures how well the finite-width neural network remains uniformly close to its infinite-width NTK surrogate, while the second term is the estimation error of the NTK gradient flow itself. The following are the two events under which each of these two terms can be made small.
\begin{enumerate}
    \item $\displaystyle{\mathcal{E}_\mathrm{NTK}=\left\{\left\|\fttopntk-f^\star\right\|_{\LL}\leq C_1\left(\ln\frac{6}{\delta}\right)^2 n^{-\frac{1}{2}\frac{s(p+1)}{s(p+1)+p}}\right\}}$, where $\delta\in(0,1)$ and $C_1>0$ is a constant that is independent of $\delta$, $n$, and $t_{\mathrm{op}}$, but depends on $s$, $p$, $A$, $B$, $M_\mu$, and $R$. It is the event that the NTK gradient flow achieves the target estimation rate. By \cref{lem:adap_gen_ntk}, $\mathcal{E}_\mathrm{NTK}$ holds with probability at least $1-\delta$ with respect to random samples.
    \item $\displaystyle{\mathcal{E}_\mathrm{NN}=\left\{\left\|\fttop-\fttopntk\right\|_{\LL}\leq C_2 m^{-\frac{1}{24}}\sqrt{\ln m}\right\}}$, where $C_2>0$ is a constant that is independent of $\delta$, $n$ and $t_\mathrm{op}$ with $\delta\in(0,1)$, but depends on $A$. By \cref{lem:nn_ntk_ini}, for $n\geq3$, $\mathcal{E}_\mathrm{NN}$ holds with probability at least $1-\delta$ with respect to random samples and random initialization, provided that $m\geq\mathrm{poly}(n,1/\delta,\ln n,\ln(1/\delta))$, where the coefficients of $\mathrm{poly}$ depend on $A$, $B$, $M_\mu$, and $p$.
\end{enumerate}
Rather than bounding the probability of $\mathcal{E}_\mathrm{NN}$ directly, we instead exhibit the following events, each depending only on the random initialization or the empirical NTK at initialization, and whose intersection is included in $\mathcal{E}_\mathrm{NN}$, as we discuss in \cref{rem:inc_events}.
\begin{enumerate}[resume]
    \item $\displaystyle{\mathcal{E}_{\theta_0}=\left\{\max_{1\leq i\leq m}\max\left(\left|v_{0,i}\right|,\left|b_{0,i}\right|,\left\|\boldsymbol{u}_{0,i}\right\|_\infty\right)\leq \sqrt{3\ln m}\right\}}$ is the event that no parameter of the initialization $\theta_0$ is not atypically large, ensuring that the parameters remain in a controlled neighborhood of $\theta_0$ during training. By \cref{lem:high_prob_upp_bnd_ini}, this event holds with probability at least $1-\frac{p+2}{\sqrt{2\pi}}m^{-1/2}$ with respect to random initialization.
    \item\label{ev:R} $\displaystyle{\mathcal{R}=\left\{\mathrm{Card}\left(\left\{i\in\{1,\ldots,m\}:\abs{\tilde{\boldsymbol{u}}_{0,i}^\top \tilde{x}}\leq\frac{\sqrt{2\pi}}{4}m^{-\eta}\right\}\right)\leq\lfloor m^\kappa\rfloor,\;\forall x\in\mathcal{X}\right\}}$, where $\tilde{\boldsymbol{u}}_{0,i}=[\boldsymbol{u}_{0,i}^\top,b_{0,i}]^\top$ are the augmented first-layer weights at initialization, $\tilde{x}=[x^\top,1]^\top$ is the augmented input, $0<\eta<\varrho\leq 1$, $\tau>\frac12$, and $\kappa>\max(1-\eta,\tau)$. It is the event that, uniformly over all inputs, at most $\lfloor m^\kappa \rfloor$ neurons have a \emph{small pre-activation} value $\abs{\tilde{\boldsymbol{u}}_{0,i}^\top \tilde{x}}$ at initialization. For the remaining neurons, the pre-activation value is large enough that changing their activation pattern would require the parameters to travel far from their initialization. Consequently, $\mathcal{R}$ guarantees that the activation pattern of the majority of neurons is effectively frozen throughout training, which is precisely the condition needed for the network to operate in the \emph{lazy training NTK regime}. By \citep[Lemma~C.2]{lai2023generalization}, when the width $m$ is sufficiently large such that:
    \begin{equation}\label{eq:cond_m_wid}
        m^{1-\eta}+m^\tau\leq\lfloor m^\kappa\rfloor\qquad\text{and}\qquad\frac{2Ap\sqrt{\ln m}}{\lfloor m^\varrho\rfloor}\leq\frac{\sqrt{2\pi}}{4}m^{-\eta},
    \end{equation}
    the event $\mathcal{R}$ holds with probability at least $1-m^{p\varrho}\exp\left(-2m^{2\tau -1}\right)$ with respect to random initialization. By \cref{lem:tech_lem_b_2}, the conditions on $m$ in \cref{eq:cond_m_wid} are satisfied when $m$ is sufficiently large such that $m\geq\max\left(2^\frac{1}{\varrho},\left(\frac{128A^2p^2}{\pi(\varrho-\eta)}\right)^\frac{1}{\varrho-\eta},3^{\frac{1}{\kappa-\max(1-\eta,\tau)}}\right)$.
    \item\label{ev:C} $\displaystyle{\mathcal{C}=\left\{\sup_{z,z'\in\mathcal{N}_\epsilon}\abs{K_0^m(z,z')-K^\mathrm{NTK}(z,z')}\leq C_3\sqrt{\frac{\ln m}{m}}\right\}}$, where $\mathcal{N}_{\epsilon}$ denotes a uniform grid on the compact $\mathcal{X}$ with mesh size $\epsilon=\frac{2A}{\lfloor m^\varrho \rfloor}$ where $0<\varrho\leq1$, and $C_3>0$ is a constant that depends on $p$, $A$, and $\varrho$. It is the event that the empirical NTK at initialization $K_0^m$ concentrates uniformly around the infinite-width NTK $K^\mathrm{NTK}$ over the grid $\mathcal{N}_\epsilon$ with a discrepancy of order $\sqrt{\frac{\ln m}{m}}$. By \citep[Lemma~C.3]{lai2023generalization}, when $m$ is large enough such that $\frac{3p\varrho}{c_0}\ln m\leq m$ with $c_0>0$ being an absolute positive constant, the event $\mathcal{C}$ holds with probability at least $1-4m^{-p\varrho}$ with respect to random initialization. By \cref{lem:tech_lem_b_2}, it is enough to require $m\geq\max\left(\frac{36p^2\varrho^2}{c_0^2},1\right)$ to ensure that $\frac{3p\varrho}{c_0}\ln m\leq m$.
\end{enumerate}
Controlling the smallest eigenvalue $\lambda_0$ of the infinite-width NTK matrix $\boldsymbol{K}^\mathrm{NTK}$ is essential for the convergence of the gradient flow. To see this, recall from \cref{lem:bound_rntk} that the explicit solution of the NTK gradient flow at the training points $(\boldsymbol{X},\boldsymbol{Y})=\{(x_i,y_i)\}_{i=1}^n$ satisfies at $t\geq 0$:
\begin{equation*}
    \left\|f_t^\mathrm{NTK}(\boldsymbol{X})-\boldsymbol{Y}\right\|\leq\exp\left(-\frac{\lambda_0}{n}t\right)\sqrt{n}B.
\end{equation*}
A positive lower bound $\lambda_0\geq\lambda_\star>0$ therefore guarantees exponential convergence of the training error at rate $\frac{\lambda_\star}{n}$. If $\lambda_0=0$, the training error never decays, and the gradient flow fails to converge entirely. Since the samples $x_1,\ldots,x_n$ are drawn randomly, the smallest eigenvalue $\lambda_0$ of the infinite-width NTK at initialization is itself a random quantity, and must therefore be lower-bounded with high probability, which motivates the following event.

\begin{enumerate}[resume]
    \item $\displaystyle{\mathcal{E}_{\lambda_0}=\left\{\lambda_0 \geq \lambda_\star(n,\delta)\right\}}$, where $\lambda_\star(n,\delta)\coloneqq C_4\bigl(1+g(\epsilon_\star(n,\delta))\bigr)^{-3}\,\epsilon_\star(n,\delta)^{4}>0$, $\epsilon_\star(n,\delta)\coloneqq\min\bigl(\epsilon_A,\ \epsilon_0(n,\delta)\bigr)\in(0,1]$, $\epsilon_A\coloneqq\min\left(1,\frac{2}{\sqrt{A^2+1}}\right)$, $\epsilon_0(n,\delta)\coloneqq\frac{2}{\pi}\left(\frac{2\delta}{M_\mu (A^2+1)^{(p+1)/2} v_p n^2}\right)^{\frac{1}{p}}$, $g(\epsilon)\coloneqq\frac{(p+1)\ln(1/\epsilon)}{\ln(p+1)}$, $\delta\in(0,1)$, $v_{p}=\mathrm{Vol}(B(0,1))$ is the volume of the unit ball in $\mathbb{R}^{p}$, $M_\mu$ is the bound on the probability density function $f_\mu$, and $C_4>0$ is an absolute positive constant. It is the event that the smallest eigenvalue of the infinite-width NTK at initialization is bounded away from zero, a necessary condition for the gradient flow to converge. By \cref{lem:proba_low_bound_min_eig}, $\mathcal{E}_{\lambda_0}$ holds with probability at least $1-\delta$ with respect to random samples.
\end{enumerate}

\begin{remark}[Inclusion of events]\label{rem:inc_events}
    According to \citep[Proof~of~Proposition~3.3]{lai2023generalization}, we have the following inclusion:
\begin{equation*}
    \mathcal{E}_{\theta_0}\cap\mathcal{R}\cap\mathcal{C}\subset\left\{\sup_{t\geq0}\sup_{\|x\|\leq A,\|x'\|\leq A}\left|K_t^m(x,x')-K^\mathrm{NTK}(x,x')\right|\leq C m^{-\frac{1}{12}}\sqrt{\ln m}\right\}.
\end{equation*}
Furthermore, in the proof of \cref{lem:nn_ntk_ini}, we show that:
\begin{equation*}
    \left\{\sup_{t\geq0}\sup_{\|x\|\leq A,\|x'\|\leq A}\left|K_t^m(x,x')-K^\mathrm{NTK}(x,x')\right|\leq C m^{-\frac{1}{12}}\sqrt{\ln m}\right\}\cap\mathcal{E}_{\lambda_0}\subset\mathcal{E}_\mathrm{NN}.
\end{equation*}
Consequently, we obtain $\mathcal{E}\coloneqq\mathcal{E}_{\theta_0}\cap\mathcal{R}\cap\mathcal{C}\cap\mathcal{E}_{\lambda_0}\subset\mathcal{E}_\mathrm{NN}$. This inclusion motivates the following definition, in which we truncate the neural network on $\mathcal{E}$ rather than on $\mathcal{E}_\mathrm{NN}$ directly, since $\mathcal{E}$ is contained in $\mathcal{E}_\mathrm{NN}$ and its probability can be bounded explicitly from its four constituent parts.
\end{remark}

\begin{definition}[Truncated neural network]\label{def:trun_net}
    Let $\mathcal{E}=\mathcal{E}_{\theta_0}\cap\mathcal{R}\cap\mathcal{C}\cap\mathcal{E}_{\lambda_0}$. For a chosen stopping time $t_\mathrm{op}$, the truncated neural network associated with $\fttop$ is defined by $\ftttop\coloneqq \fttop \mathbbm{1}_{\mathcal{E}}$.
\end{definition}

The event $\mathcal{E}$ and the truncated predictor depend on a deterministic confidence parameter $\delta\in(0,1)$, through $\lambda_\star(n,\delta)$. The inner and outer estimators in \cref{def:in_cond_trunc_main} choose $\delta$ to be the corresponding minimizer $\delta^\star$, specified in \cref{eq:deltastarh,eq:deltast} in the proofs of \cref{prop:err_h,prop:exp_gen_bound}, respectively.

By truncating the neural network on $\mathcal{E}$, the truncated predictor $\ftttop$ vanishes on $\mathcal{E}^\complement$, which allows us to pass from the high-probability bound of \Cref{prop:gen_bound} to the expected bound of \Cref{prop:exp_gen_bound} by decomposing the expected error into a contribution on $\mathcal{E}$, controlled by the high-probability bound, and a contribution on $\mathcal{E}^\complement$, which vanishes by definition of the truncation. The explicit probability bounds on each constituent event of $\mathcal{E}$ are then used to show that $\mathrm{Pr}(\mathcal{E}^\complement)$ is small enough to be absorbed into the overall rate. 

\begin{proposition}[Expected generalization bound]\label{prop:exp_gen_bound}
Consider $n$ i.i.d. samples $\{(x_i,y_i)\}_{i=1}^n\sim\rho$ and fix $p\geq 2$. Let \Cref{ass:bound_in_res,ass:bound_la_res,ass:source_cond,ass:bounded_density} hold. If we choose a stopping time $t_{\mathrm{op}}=n^{\frac{p+1}{s(p+1)+p}}$ with $\frac{1}{p+1}<s\leq 2$, then, considering the early-stopped truncated neural network predictor $\ftttop$ for the target function $f^\star$, there exist a constant $C>0$ and a polynomial $\mathrm{poly}$ such that when $n\geq\max\left(3,\left\lceil\left(\frac{3C_1^2 e^3}{2(\kappa^\mathrm{NTK})^2B^2}\right)^\frac{s(p+1)+p}{(s+2)(p+1)}\right\rceil,\left\lceil\left(\frac{1}{3(\kappa^\mathrm{NTK})^2}\right)^\frac{s(p+1)+p}{2(p+1)}\right\rceil,\left\lfloor\left(\frac{C_1^2(\ln 6)^3}{3(\kappa^\mathrm{NTK})^2B^2}\right)^{\frac{s(p+1)+p}{(s+2)(p+1)}}\right\rfloor+1\right)$ and the width satisfies $m\geq\mathrm{poly}(n,\ln n)$, the following holds:
    \begin{equation*}
        \mathbb{E}\left[\left\|\ftttop-f^\star\right\|_{\LL}^2\right]\leq C\left(\ln n\right)^4 n^{-\frac{s(p+1)}{s(p+1)+p}},
    \end{equation*}
    where $\kappa^\mathrm{NTK}\coloneqq(p+2)(A^2+1)+1$, the constant $C>0$ is independent of $n$ and $t_{\mathrm{op}}$, but depends on $s$, $p$, $A$, $B$, $M_\mu$, and $R$, and the coefficients of $\mathrm{poly}$ depend on $s$, $p$, $A$, $B$, $M_\mu$, and $R$. The expectation here is with respect to random samples and random initialization.
\end{proposition}

\begin{proof}
    By the triangle inequality and the identity $(a+b)^2\leq 2(a^2+b^2)$, we have:
    \begin{equation}\label{eq:tri_ineq_ftttop_fstar}
        \mathbb{E}\left[\left\|\ftttop-f^\star\right\|_{\LL}^2\right]\leq2\left(\mathbb{E}\left[\left\|\ftttop-\fttopntk\right\|_{\LL}^2\right]+\mathbb{E}\left[\left\|\fttopntk-f^\star\right\|_{\LL}^2\right]\right).
    \end{equation}
    Next, we upper-bound each expectation on the right-hand side of the inequality, and then combine the resulting bounds.\\
    \textbf{Upper-bound on $\mathbb{E}\left[\left\|\ftttop-\fttopntk\right\|_{\LL}^2\right]$. }Let $\mathcal{E}=\mathcal{E}_{\theta_0}\cap\mathcal{R}\cap\mathcal{C}\cap\mathcal{E}_{\lambda_0}$. By the law of total expectation, we decompose the error over the disjoint events $\mathcal{E}$ and $\mathcal{E}^\complement$ as follows:
    \begin{align}\label{eq:inter_ftttop_fttopntk}
        \mathbb{E}\left[\left\|\ftttop-\fttopntk\right\|_{\LL}^2\right]=&\mathbb{E}\left[\left\|\ftttop-\fttopntk\right\|_{\LL}^2\mathbbm{1}_\mathcal{E}\right]+\mathbb{E}\left[\left\|\ftttop-\fttopntk\right\|_{\LL}^2\mathbbm{1}_{\mathcal{E}^\complement}\right]\nonumber\\
        =&\mathbb{E}\left[\left\|\fttop-\fttopntk\right\|_{\LL}^2\mathbbm{1}_\mathcal{E}\right]+\mathbb{E}\left[\left\|\fttopntk\right\|_{\LL}^2\mathbbm{1}_{\mathcal{E}^\complement}\right]\nonumber\\
        \leq&\mathbb{E}\left[\left\|\fttop-\fttopntk\right\|_{\LL}^2\mathbbm{1}_{\mathcal{E}_\mathrm{NN}}\right]+(\kappa^\mathrm{NTK})^2B^2t_\mathrm{op}^2\mathrm{Pr}\left(\mathcal{E}^\complement\right)\nonumber\\
        \leq&C_2^2 m^{-\frac{1}{12}}\ln m+(\kappa^\mathrm{NTK})^2B^2 n^\frac{2(p+1)}{s(p+1)+p}\mathrm{Pr}\left(\mathcal{E}^\complement\right),
    \end{align}
    where, in the second line, we use the definition of the truncated neural network in \cref{def:trun_net} which gives $\ftttop=0$ on $\mathcal{E}^\complement$; in the third line, we use $\mathcal{E}\subset\mathcal{E}_\mathrm{NN}$ as explained in \cref{rem:inc_events}, and \cref{lem:bound_ftntk} to upper-bound $\left\|\fttopntk\right\|_{\LL}$ by $\kappa^\mathrm{NTK} B t_\mathrm{op}$ with $\kappa^\mathrm{NTK}\coloneqq(p+2)(A^2+1)+1$; and in the fourth line, we upper-bound $\left\|\fttop-\fttopntk\right\|_{\LL}^2$ by $C_2^2 m^{-\frac{1}{12}}\ln m$ on the event $\mathcal{E}_\mathrm{NN}$ and use $t_\mathrm{op}=n^\frac{p+1}{s(p+1)+p}$. Furthermore, by De Morgan's law and a union bound, we have that:
    \begin{align*}
        \mathrm{Pr}\left(\mathcal{E}^\complement\right)=&\mathrm{Pr}\left(\left(\mathcal{E}_{\theta_0}\cap\mathcal{R}\cap\mathcal{C}\cap\mathcal{E}_{\lambda_0}\right)^\complement\right)\\
        =&\mathrm{Pr}\left(\mathcal{E}_{\theta_0}^\complement\cup\mathcal{R}^\complement\cup\mathcal{C}^\complement\cup\mathcal{E}_{\lambda_0}^\complement\right)\\
        \leq&\mathrm{Pr}\left(\mathcal{E}_{\theta_0}^\complement\right)+\mathrm{Pr}\left(\mathcal{R}^\complement\right)+\mathrm{Pr}\left(\mathcal{C}^\complement\right)+\mathrm{Pr}\left(\mathcal{E}_{\lambda_0}^\complement\right)\\
        \leq&\frac{p+2}{\sqrt{2\pi}}m^{-1/2}+m^{p\varrho}\exp\left(-2m^{2\tau -1}\right)+4m^{-p\varrho}+\delta.
    \end{align*}
    We note that $\frac{p+2}{\sqrt{2\pi}}m^{-1/2}\leq\delta$ and $4m^{-p\varrho}\leq\delta$ when we choose $m$ large enough such that $m\geq\frac{(p+2)^2}{2\pi\delta^2}$ and $m\geq\left(\frac{4}{\delta}\right)^\frac{1}{p\varrho}$, respectively. Moreover, by \cref{lem:cond_m}, $m^{p\varrho}\exp\left(-2m^{2\tau -1}\right)\leq\delta$ holds whenever $m$ is large enough such that $m\geq\max(1,M_1(1/\delta))$ where:
    \begin{equation*}
        M_1\left(\frac{1}{\delta}\right)=\left(\left[\frac{p\varrho}{2\tau-1}\left(\ln\left(\frac{p\varrho}{2\tau-1}\right)-1\right)+\ln\left(\frac{1}{\delta}\right)\right]_+\right)^{\frac{1}{2\tau-1}},\qquad\text{with }[t]_+\coloneqq\max(t,0).
    \end{equation*}
    $\mathcal{R}$ and $\mathcal{C}$ hold with high-probability when $m\geq\max\left(2^\frac{1}{\varrho},\left(\frac{128A^2p^2}{\pi(\varrho-\eta)}\right)^\frac{1}{\varrho-\eta},3^{\frac{1}{\kappa-\max(1-\eta,\tau)}}\right)$ and $m\geq\max\left(\frac{36p^2\varrho^2}{c_0^2},1\right)$, respectively, where $c_0$ is an absolute positive constant, as discussed in \cref{ev:R,ev:C}. Combining all these conditions, we conclude that when $m$ is large enough such that $m\geq M_2(1/\delta)$ where:
    \begin{equation*}
        M_2\left(\frac{1}{\delta}\right)=\max\left(\frac{(p+2)^2}{2\pi\delta^2},\left(\frac{4}{\delta}\right)^\frac{1}{p\varrho},M_1\left(\frac{1}{\delta}\right),2^\frac{1}{\varrho},\left(\frac{128A^2p^2}{\pi(\varrho-\eta)}\right)^\frac{1}{\varrho-\eta},3^{\frac{1}{\kappa-\max(1-\eta,\tau)}},\frac{36p^2\varrho^2}{c_0^2}\right),
    \end{equation*}
    we get $\mathrm{Pr}(\mathcal{E}^\complement)\leq 4\delta$. Substituting this into \cref{eq:inter_ftttop_fttopntk}, we obtain:
    \begin{equation}\label{eq1:ftttop_fttopntk}
        \mathbb{E}\left[\left\|\ftttop-\fttopntk\right\|_{\LL}^2\right]\leq C_2^2 m^{-\frac{1}{12}}\ln m+4(\kappa^\mathrm{NTK})^2B^2 n^\frac{2(p+1)}{s(p+1)+p}\delta.
    \end{equation} 
    Since $m\geq\mathrm{poly}(n,1/\delta,\ln n,\ln(1/\delta))$ should be satisfied for $\mathcal{E}_\mathrm{NN}$ to hold with high-probability, we require $m\geq M_3(1/\delta)$ where $M_3(1/\delta)=\max(\mathrm{poly}(n,1/\delta,\ln n,\ln(1/\delta)),M_2(1/\delta))$. Redefining $\mathrm{poly}$ as the polynomial dominating $M_3(1/\delta)$, we obtain $m\geq\mathrm{poly}(n,1/\delta,\ln n,\ln(1/\delta))$, which should be satisfied so that the bound on $\mathbb{E}\left[\left\|\ftttop-\fttopntk\right\|_{\LL}^2\right]$ holds.
    
    \textbf{Upper-bound on $\mathbb{E}\left[\left\|\fttopntk-f^\star\right\|_{\LL}^2\right]$. }By the law of total expectation, we decompose the error over the disjoint events $\mathcal{E}_\mathrm{NTK}$ and $\mathcal{E}_\mathrm{NTK}^\complement$ as follows:
    \begin{align}\label{eq2:fttopntk_fstar}
        \mathbb{E}\left[\left\|\fttopntk-f^\star\right\|_{\LL}^2\right]=&\mathbb{E}\left[\left\|\fttopntk-f^\star\right\|_{\LL}^2\mathbbm{1}_{\mathcal{E}_\mathrm{NTK}}\right]+\mathbb{E}\left[\left\|\fttopntk-f^\star\right\|_{\LL}^2\mathbbm{1}_{\mathcal{E}_\mathrm{NTK}^\complement}\right]\nonumber\\
        \leq&C_1^2\left(\ln\frac{6}{\delta}\right)^4 n^{-\frac{s(p+1)}{s(p+1)+p}}+\mathbb{E}\left[\left(\left\|\fttopntk\right\|_{\LL}+\left\|f^\star\right\|_\infty\right)^2\mathbbm{1}_{\mathcal{E}_\mathrm{NTK}^\complement}\right]\nonumber\\
        \leq&C_1^2\left(\ln\frac{6}{\delta}\right)^4 n^{-\frac{s(p+1)}{s(p+1)+p}}+2\mathbb{E}\left[\left(\left\|\fttopntk\right\|_{\LL}^2+\left\|f^\star\right\|_\infty^2\right)\mathbbm{1}_{\mathcal{E}_\mathrm{NTK}^\complement}\right]\nonumber\\
        \leq&C_1^2\left(\ln\frac{6}{\delta}\right)^4 n^{-\frac{s(p+1)}{s(p+1)+p}}+2\left((\kappa^\mathrm{NTK})^2B^2t_\mathrm{op}^2+B^2\right)\mathrm{Pr}\left(\mathcal{E}_\mathrm{NTK}^\complement\right)\nonumber\\
        \leq&C_1^2\left(\ln\frac{6}{\delta}\right)^4 n^{-\frac{s(p+1)}{s(p+1)+p}}+2\left((\kappa^\mathrm{NTK})^2B^2 n^\frac{2(p+1)}{s(p+1)+p}+B^2\right)\delta,
    \end{align}
    where, in the second line, we upper-bound $\left\|\fttopntk-f^\star\right\|_{\LL}^2$ by $C_1^2\left(\ln\frac{6}{\delta}\right)^4 n^{-\frac{s(p+1)}{s(p+1)+p}}$ on the event $\mathcal{E}_\mathrm{NTK}$ and use the triangle inequality together with the identity $\|f^\star\|_{\LL}\leq\|f^\star\|_\infty$; in the third line, we apply the identity $(a+b)^2\leq 2(a^2+b^2)$; in the fourth line, we use \Cref{lem:bound_ftntk,ass:tar_b} to upper-bound $\left\|\fttopntk\right\|_{\LL}$ by $\kappa^\mathrm{NTK}Bt_\mathrm{op}$ with $\kappa^\mathrm{NTK}\coloneqq(p+2)(A^2+1)+1$ and $\|f^\star\|_\infty$ by $B$; and in the last line, we use $t_\mathrm{op}=n^\frac{p+1}{s(p+1)+p}$.
    \\
    \textbf{Expected generalization bound. }Plugging \cref{eq1:ftttop_fttopntk,eq2:fttopntk_fstar} into \cref{eq:tri_ineq_ftttop_fstar} yields:
    \begin{align*}
        &\mathbb{E}\left[\left\|\ftttop-f^\star\right\|_{\LL}^2\right]\\
        \leq&2\left(C_2^2 m^{-\frac{1}{12}}\ln m+C_1^2\left(\ln\frac{6}{\delta}\right)^4 n^{-\frac{s(p+1)}{s(p+1)+p}}+\left(6(\kappa^\mathrm{NTK})^2B^2 n^\frac{2(p+1)}{s(p+1)+p}+2B^2\right)\delta\right).
    \end{align*}
    Since $n\geq\left\lceil\left(\frac{1}{3(\kappa^\mathrm{NTK})^2}\right)^\frac{s(p+1)+p}{2(p+1)}\right\rceil$, we have $3(\kappa^\mathrm{NTK})^2B^2n^\frac{2(p+1)}{s(p+1)+p}\geq B^2$, therefore:
    \begin{equation}\label{eq:prel_gen}
        \mathbb{E}\left[\left\|\ftttop-f^\star\right\|_{\LL}^2\right]\leq2\left(C_2^2 m^{-\frac{1}{12}}\ln m+\varphi(\delta)\right),
    \end{equation}
    where $\varphi:(0,1)\to(0,+\infty)$ is the function defined for any $\delta\in(0,1)$ by:
    \begin{equation*}
        \varphi(\delta)=C_1^2\left(\ln\frac{6}{\delta}\right)^4 n^{-\frac{s(p+1)}{s(p+1)+p}}+12(\kappa^\mathrm{NTK})^2B^2n^\frac{2(p+1)}{s(p+1)+p}\delta.
    \end{equation*}
    We minimize $\varphi$ over $\delta\in(0,1)$. By \cref{lem:e_delta}, since $n\geq\left\lfloor\left(\frac{C_1^2(\ln 6)^3}{3(\kappa^\mathrm{NTK})^2B^2}\right)^{\frac{s(p+1)+p}{(s+2)(p+1)}}\right\rfloor+1$, the minimizer $\delta^\star\in(0,1)$ and the corresponding minimum value $\varphi(\delta^\star)$ are given by:
    \begin{align}\label{eq:deltast}
        \delta^\star=\frac{9C_1^2}{(\kappa^\mathrm{NTK})^2B^2}n^{-\frac{(s+2)(p+1)}{s(p+1)+p}}W_0^3\left(n^{\frac{1}{3}\frac{(s+2)(p+1)}{s(p+1)+p}}\left(\frac{2(\kappa^\mathrm{NTK})^2B^2}{3C_1^2}\right)^\frac{1}{3}\right),
    \end{align}
    and
    \begin{align*}
        \varphi(\delta^\star)=&27C_1^2W_0^3\left(n^{\frac{1}{3}\frac{(s+2)(p+1)}{s(p+1)+p}}\left(\frac{2(\kappa^\mathrm{NTK})^2B^2}{3C_1^2}\right)^\frac{1}{3}\right)\\
        &\times\left(3W_0\left(n^{\frac{1}{3}\frac{(s+2)(p+1)}{s(p+1)+p}}\left(\frac{2(\kappa^\mathrm{NTK})^2B^2}{3C_1^2}\right)^\frac{1}{3}\right)+4\right)n^{-\frac{s(p+1)}{s(p+1)+p}},
    \end{align*}
    where $W_0$ is the principal branch of the Lambert $W$ function. Note that since $\frac{1}{p+1}<s\leq 2$, one can verify that $\frac{(s+2)(p+1)}{s(p+1)+p}\leq 4$, so that $\frac{1}{3}\frac{(s+2)(p+1)}{s(p+1)+p}\leq\frac{4}{3}$. Additionally, since: 
    \begin{equation*}
        n\geq\max\left(3,\left\lceil\left(\frac{3C_1^2 e^3}{2(\kappa^\mathrm{NTK})^2B^2}\right)^\frac{s(p+1)+p}{(s+2)(p+1)}\right\rceil\right)\geq\max\left(e,\left(\frac{3C_1^2 e^3}{2(\kappa^\mathrm{NTK})^2B^2}\right)^\frac{s(p+1)+p}{(s+2)(p+1)}\right),
    \end{equation*}
    applying \cref{lem:b_w0} yields:
    \begin{align*}
        &W_0^3\left(n^{\frac{1}{3}\frac{(s+2)(p+1)}{s(p+1)+p}}\left(\frac{2(\kappa^\mathrm{NTK})^2B^2}{3C_1^2}\right)^\frac{1}{3}\right)\left(3W_0\left(n^{\frac{1}{3}\frac{(s+2)(p+1)}{s(p+1)+p}}\left(\frac{2(\kappa^\mathrm{NTK})^2B^2}{3C_1^2}\right)^\frac{1}{3}\right)+4\right)\\
        \leq& A_1\left(\left(\frac{2(\kappa^\mathrm{NTK})^2B^2}{3C_1^2}\right)^\frac{1}{3}\right)(\ln n)^4,
    \end{align*}
    where $A_1$ is defined for any $u>0$ by:
    \begin{equation*}
        A_1(u)=\frac{512}{27}+\frac{448}{9}\abs{\ln u}+48(\ln u)^2+20\abs{\ln u}^3+3(\ln u)^4.
    \end{equation*}
    Substituting this bound into the expression of $\varphi(\delta^\star)$, we obtain:
    \begin{equation*}
        \varphi(\delta^\star)\leq C'(\ln n)^4 n^{-\frac{s(p+1)}{s(p+1)+p}},
    \end{equation*}
    where $C'=27C_1^2 A_1\left(\left(\frac{2(\kappa^\mathrm{NTK})^2B^2}{3C_1^2}\right)^\frac{1}{3}\right)$. Plugging this bound in \cref{eq:prel_gen} yields:
    \begin{equation*}
        \mathbb{E}\left[\left\|\ftttop-f^\star\right\|_{\LL}^2\right]\leq2\left(C_2^2 m^{-\frac{1}{12}}\ln m+C'(\ln n)^4 n^{-\frac{s(p+1)}{s(p+1)+p}}\right).
    \end{equation*}
    Choosing $m\geq\max\left(\mathrm{poly}(n,1/\delta,\ln n,\ln(1/\delta)),n^{12}(\ln n)^{12}\right)$ gives $m^{-\frac{1}{12}}\ln m\leq 24n^{-1}$ by \cref{lem:tech_b}. Since $s>0$ and $p>0$, we have $0<\frac{s(p+1)}{s(p+1)+p}<1$, so $n^{-1}\leq n^{-\frac{s(p+1)}{s(p+1)+p}}$. Moreover, $n\geq3>e$ implies $1\leq(\ln n)^4$. Combining these observations leads to:
    \begin{equation*}
        \mathbb{E}\left[\left\|\ftttop-f^\star\right\|_{\LL}^2\right]\leq2\left(24C_2^2+C'\right)(\ln n)^4 n^{-\frac{s(p+1)}{s(p+1)+p}}.
    \end{equation*}
    The desired bound follows by setting $C=2\left(24C_2^2+C'\right)$. 
    
    We now derive an explicit lower-bound on $\delta^\star$ to characterize the precise quantities on which $m$ depends. Recall from \cref{eq:deltast} that:
    \begin{equation*}
        \delta^\star=\frac{9C_1^2}{(\kappa^\mathrm{NTK})^2B^2}n^{-\frac{(s+2)(p+1)}{s(p+1)+p}}W_0^3\left(z\right),\qquad z\coloneqq n^{\frac{1}{3}\frac{(s+2)(p+1)}{s(p+1)+p}}\left(\frac{2(\kappa^\mathrm{NTK})^2B^2}{3C_1^2}\right)^\frac{1}{3}.
    \end{equation*}
    Since $n\geq\left\lceil\left(\frac{3C_1^2 e^3}{2(\kappa^\mathrm{NTK})^2B^2}\right)^\frac{s(p+1)+p}{(s+2)(p+1)}\right\rceil$, we have $z\geq e$. As $W_0$ is increasing on $[0,+\infty)$ with $W_0(e)=1$, this gives $W_0(z)\geq 1$, hence:
    \begin{equation*}
        \delta^\star\geq\frac{9C_1^2}{(\kappa^\mathrm{NTK})^2B^2}n^{-\frac{(s+2)(p+1)}{s(p+1)+p}}.
    \end{equation*}
    Using $\frac{(s+2)(p+1)}{s(p+1)+p}\leq 4$, we obtain:
    \begin{equation*}
        \frac{1}{\delta^\star}\leq\frac{(\kappa^\mathrm{NTK})^2B^2}{9C_1^2}n^{\frac{(s+2)(p+1)}{s(p+1)+p}}\leq\frac{(\kappa^\mathrm{NTK})^2B^2}{9C_1^2}n^4,
    \end{equation*}
    and, taking logarithms and using $\ln n\geq1$ for $n\geq 3$, we get:
    \begin{equation*}
        \ln\frac{1}{\delta^\star}\leq\left|\ln\frac{(\kappa^\mathrm{NTK})^2B^2}{9C_1^2}\right|+4\ln n\leq\left(\left|\ln\frac{(\kappa^\mathrm{NTK})^2B^2}{9C_1^2}\right|+4\right)\ln n.
    \end{equation*}
    Both $1/\delta^\star$ and $\ln(1/\delta^\star)$ are therefore dominated by a polynomial in $(n,\ln n)$ whose coefficients depend only on $s,p,A,B,M_\mu$ and $R$. Redefining $\mathrm{poly}$ accordingly, the condition $m\geq\mathrm{poly}(n,1/\delta^\star,\ln n,\ln(1/\delta^\star))$ is implied by $m\geq\mathrm{poly}(n,\ln n)$, which concludes the proof.
\end{proof}
\section{Differentiability of the Objectives and Properties of the Optimal Solutions}\label{sec:diff_ob_op_sol}
The next two Lemmas follow directly from Lemma D.3 and Lemma D.4 in \citep{petrulionyte2024functional}, respectively. Here, differentiability is understood in the Fr\'echet sense. We write $\partial_h L_{out}(h)\in\LL$ and $\partial_h L_{in}(\omega,h)\in\LL$ for the partial derivatives with respect to $h$, $\partial^2_h L_{in}(\omega,h):\LL\to\LL$ and $\partial^2_{\omega,h}L_{in}(\omega,h):\LL\to\mathbb{R}^d$ for the second-order partial derivatives of $L_{in}$ with respect to $h$ and to $(\omega,h)$.

\begin{lemma}[Differentiability of $L_{out}$]\label{prop:diff_lout}
Under \Cref{ass:bound_la}, $L_{out}$ is differentiable on $\LL$, with partial derivative given, for any $h\in\LL$, by:
    \begin{equation*}
    \partial_h L_{out}(h)\in\LL:x\mapsto h(x)-\mathbb{E}_\mathbb{P}\left[y\mid x\right].
\end{equation*}
\end{lemma}

\begin{lemma}[Differentiability of $\partial_h L_{in}$]\label{prop:diff_lin}
Under \Cref{ass:gome}, the function $(\omega, h)\mapsto\partial_h L_{in}(\omega, h)$ is differentiable on $ U\times\LL$, with partial derivatives given, for any $\omega\in U$ and $h\in\LL$, by:
\begin{equation*}
    \partial_h^2 L_{in}(\omega, h)=\mathrm{Id}_{\LL},\quad\partial_{\omega,h}^2 L_{in}(\omega, h)f=-\mathbb{E}_\mathbb{P}\left[\partial_\omega g_\omega(z) f(x)\right],\;\text{for any }f\in\LL.
\end{equation*}
\end{lemma}

\begin{lemma}[Uniform boundedness of target functions]\label{lem:bound_targets}
    Under \Cref{ass:bound_la,ass:gome}, we have:
    \begin{itemize}
        \item $h^\star_\omega\in L^\infty(\mathcal{X})$ for any $\omega\in U$, and $\sup_{\omega\in U}\|h^\star_\omega\|_\infty\leq D$;
        \item $b^\star\in L^\infty(\mathcal{X})$ and $\|b^\star\|_\infty\leq B$;
        \item $a^\star_\omega\in L^\infty(\mathcal{X})$ for any $\omega\in U$, and $\sup_{\omega\in U}\|a^\star_\omega\|_\infty\leq D+B$.
    \end{itemize}
\end{lemma}

\begin{proof}
    \textbf{Uniform boundedness of $h^\star_\omega$. }Let $\omega\in U$. By \Cref{ass:gome}, $|g_\omega(z)|\leq D$ for any $z\in\mathcal{Z}$, so that, by Jensen's inequality, for $\mathbb{P}_X$-almost every $x\in\mathcal{X}$, $\left|h^\star_\omega(x)\right|=\left|\mathbb{E}_\mathbb{P}\left[g_\omega(z)\mid x\right]\right|\leq\mathbb{E}_\mathbb{P}\left[\left|g_\omega(z)\right|\mid x\right]\leq D$. Taking the essential supremum over $x$ and then the supremum over $\omega\in U$ gives $\sup_{\omega\in U}\|h^\star_\omega\|_\infty\leq D$.

    \textbf{Boundedness of $b^\star$. }Similarly, by \Cref{ass:bound_la}, $|y|\leq B$, so that $|b^\star(x)|=|\mathbb{E}_\mathbb{P}[y\mid x]|\leq\mathbb{E}_\mathbb{P}[|y|\mid x]\leq B$ for $\mathbb{P}_X$-almost every $x\in\mathcal{X}$, hence $\|b^\star\|_\infty\leq B$.

    \textbf{Uniform boundedness of $a^\star_\omega$. }Let $\omega\in U$. By the triangle inequality, $\|a^\star_\omega\|_\infty=\|-h^\star_\omega+b^\star\|_\infty\leq\|h^\star_\omega\|_\infty+\|b^\star\|_\infty$. Taking the supremum over $\omega\in U$ together with the two previous bounds gives $\sup_{\omega\in U}\|a^\star_\omega\|_\infty\leq D+B$.
\end{proof}

\begin{proposition}[Lipschitz continuity of $\omega\mapsto h^\star_\omega$]\label{prop:lip_con_hstar}
    Under \Cref{ass:gome}, the function $\omega\mapsto h^\star_\omega$ is $G$-Lipschitz continuous on $U$.
\end{proposition}

\begin{proof}
    Let $\omega_1,\omega_2\in U$. Since $U$ is convex, the segment $\omega_s\coloneqq \omega_2+s(\omega_1-\omega_2)$, $s\in[0,1]$, lies in $U$, so that applying the Cauchy-Schwarz inequality and \cref{ass:gome} gives for any $z\in\mathcal{Z}$:
    \begin{equation*}
        \left|g_{\omega_1}(z)-g_{\omega_2}(z)\right|\leq\left|\int_0^1\left\langle\partial_\omega g_{\omega_s}(z),\omega_1-\omega_2\right\rangle\mathrm{d}s\right|\leq G\left\|\omega_1-\omega_2\right\| .
    \end{equation*}
    Hence, by Jensen's inequality, for $\mathbb{P}_X$-almost every $x\in\mathcal{X}$, we have:
    \begin{equation*}
        \left|h^\star_{\omega_1}(x)-h^\star_{\omega_2}(x)\right|=\left|\mathbb{E}_\mathbb{P}\left[g_{\omega_1}(z)-g_{\omega_2}(z)\mid x\right]\right|\leq\mathbb{E}_\mathbb{P}\left[\left|g_{\omega_1}(z)-g_{\omega_2}(z)\right|\mid x\right]\leq G\left\|\omega_1-\omega_2\right\| .
    \end{equation*}
    Since $\mathbb{P}_X$ is a probability measure, $\|\cdot\|_{\LL}\leq\|\cdot\|_\infty$, and therefore:
    \begin{equation*}
        \left\|h^\star_{\omega_1}-h^\star_{\omega_2}\right\|_{\LL}\leq G\left\|\omega_1-\omega_2\right\| ,
    \end{equation*}
    which concludes the proof.
\end{proof}
\section{A Sobolev Space Characterization of the Infinite-Width NTK's RKHS}\label{app:sob}
 In this section, our goal is to show that the RKHS $\mathcal{H}^\mathrm{NTK}$ of the infinite-width NTK $K^\mathrm{NTK}$ of the two-layer ReLU network of the form in \cref{eq:shallow}, and its corresponding interpolation spaces, are Sobolev spaces when the input space $\mathcal{X}$ is a bounded Lipschitz domain (\cref{def:lipschitz}).

\subsection{Preliminaries}
For a positive semi-definite kernel $k$ on a set $E$, we write $\mathcal{H}(k)$ for its RKHS. Standard facts about $\mathcal{H}(k)$ are used repeatedly: \emph{(restriction)} for $E_0\subset E$, $\mathcal{H}(k|_{E_0})=\{f|_{E_0}:f\in\mathcal{H}(k)\}$ with $\|g\|_{\mathcal{H}(k|_{E_0})}=\min\{\|f\|_{\mathcal{H}(k)}:f|_{E_0}=g\}$; \emph{(sums)} $\mathcal{H}(k_1+k_2)=\mathcal{H}(k_1)+\mathcal{H}(k_2)$ with $\|f\|^2_{\mathcal{H}(k_1+k_2)}=\min\{\|f_1\|^2_{\mathcal{H}(k_1)}+\|f_2\|^2_{\mathcal{H}(k_2)}:f=f_1+f_2\}$; and \emph{(scaling)} for $c>0$, $\mathcal{H}(ck)=\mathcal{H}(k)$ as sets, with $\|f\|_{\mathcal{H}(ck)}=c^{-1/2}\|f\|_{\mathcal{H}(k)}$ \citep{aronszajn1950theory}. For a subset $F$ of a normed space $\mathcal{H}$, we say $F$ \emph{embeds continuously} into a normed space $\mathcal{H}'$ if $F\subset\mathcal{H}'$ and there is a constant $C>0$ such that $\|f\|_{\mathcal{H}'}\le C\|f\|_{\mathcal{H}}$ for every $f\in F$; we write this $F\hookrightarrow\mathcal{H}'$. For a function $f$ on a topological space $E$, the \emph{support} of $f$, written $\mathrm{supp}\,f$, is the closure of $\{x\in E:f(x)\neq0\}$. $f$ is said to be \emph{compactly supported} in an open set $U\subset E$ if $\mathrm{supp}\,f$ is a compact subset of $U$. Let $\phi:\mathbb{R}^p\rightarrow\mathbb{S}^p\subset\mathbb{R}^{p+1}$ be the map defined for any $x\in\mathbb{R}^p$ by $\phi(x)=\tilde{x}/\|\tilde{x}\|$ with $\tilde{x}=[x^\top,1]^\top$. We write $\mathbb{S}^p_+\coloneqq\{y\in\mathbb{S}^p:y_{p+1}>0\}$ and $\mathbb{S}^p_-\coloneqq\{y\in\mathbb{S}^p:y_{p+1}<0\}$ for the open upper and lower hemispheres of $\mathbb{S}^p$. For any $u\in[-1,1]$, we define $\kappa(u)\coloneqq u\kappa_0(u)+\kappa_1(u)$ where $\kappa_0(u)\coloneqq\frac{1}{\pi}(\pi-\arccos(u))$ and $\kappa_1(u)\coloneqq\frac{1}{\pi}\big(u(\pi-\arccos(u))+\sqrt{1-u^2}\big)$. Using \cref{eq:explicit_ntk}:
\begin{equation*}
    K^\mathrm{NTK}(x,x')=\frac{1}{2}\|\tilde x\|\|\tilde x'\|\kappa\left(\langle\phi(x),\phi(x')\rangle\right)+1,
\end{equation*}
Setting $K_0(y,y')\coloneqq\kappa(\langle y,y'\rangle)$ for $y,y'\in\mathbb{S}^p$, and using the notation $M_g[k](x,x')\coloneqq g(x)k(x,x')g(x')$ for a function $g$ and kernel $k$ and $\phi^*K_0(x,x')\coloneqq K_0(\phi(x),\phi(x'))$ for $x,x'\in\mathcal{X}$:
\begin{equation}\label{eq:kernel_factorization}
    K^\mathrm{NTK}=\frac{1}{2}M_q[\phi^*K_0]+\mathcal{I}_{\mathcal{X}},\quad\text{where }q(x)\coloneqq\|\tilde{x}\|\text{ and }\mathcal{I}_{\mathcal{X}}(x,x')=1\text{ for }x,x'\in\mathcal{X}.
\end{equation}
In this section, $\mathcal{X}$ is a \emph{bounded Lipschitz domain}, in the sense of \cref{def:lipschitz} below. We write $L^2(\mathcal{X},\mathrm{d}x)$ for the space of square-integrable functions on $\mathcal{X}$ with respect to the Lebesgue measure $\mathrm{d}x$, and recall $\LL\coloneqq L^2(\mathcal{X},\mu)$. Let $L^1_{\mathrm{loc}}(\mathcal{X},\mathrm{d}x)$ denote the space of \emph{locally integrable} functions on $\mathcal{X}$, \emph{i.e.}, functions $f:\mathcal{X}\to\mathbb{R}$ such that $\int_K|f(x)|\mathrm{d}x<\infty$ for every compact $K\subset\mathcal{X}$, and let $C_c^\infty(\mathcal{X})$ denote the space of \emph{smooth, compactly supported} functions on $\mathcal{X}$. Write $\mathbb{N}_0\coloneqq\{0,1,2,\dots\}$ for the nonnegative integers.

\begin{definition}[Bounded Lipschitz domain {\citep[Definition~1.2.1.1]{grisvard1985elliptic}}]\label{def:lipschitz}
$\mathcal{X}$ is \emph{bounded Lipschitz} if it is open, bounded, and nonempty, and its boundary $\partial\mathcal{X}$ is locally the graph of a Lipschitz function: for every $x_0\in\partial\mathcal{X}$, there exist $r>0$, an orthonormal coordinate system $(z_1,\dots,z_{p-1},z_p)$ centered at $x_0$, and a Lipschitz function $\gamma:\mathbb{R}^{p-1}\to\mathbb{R}$ such that:
\begin{align*}
\mathcal{X}\cap B(x_0,r)=&\{(z_1,\dots,z_p)\in B(x_0,r) : z_p > \gamma(z_1,\dots,z_{p-1})\},\\
\text{and }\partial\mathcal{X}\cap B(x_0,r)=&\{(z_1,\dots,z_p)\in B(x_0,r) : z_p = \gamma(z_1,\dots,z_{p-1})\},
\end{align*}
where $B(x_0,r)$ is the open ball centered at $x_0$ of radius $r$.
\end{definition}

\begin{definition}[Weak derivative]\label{def:weak}
For $f\in L^1_{\mathrm{loc}}(\mathcal{X},\mathrm{d}x)$ and a multi-index $\alpha=(\alpha_1,\ldots,\alpha_p)\in\mathbb{N}_0^p$ with $|\alpha|=\alpha_1+\ldots+\alpha_p$, $v\in L^1_{\mathrm{loc}}(\mathcal{X},\mathrm{d}x)$ is the \emph{weak $\alpha$-th derivative} of $f$, written $v=D^\alpha f$, if:
\begin{equation*}
\int_{\mathcal{X}} f(x)D^\alpha\phi(x)\mathrm{d}x = (-1)^{|\alpha|}\int_{\mathcal{X}} v(x)\phi(x)\mathrm{d}x \qquad \text{for all } \phi\in C_c^\infty(\mathcal{X}).
\end{equation*}
\end{definition}

The following are the definitions of Sobolev spaces used throughout this section, following the standard treatment in, \emph{e.g.}, \citep[Section~5.1]{agranovich2015sobolev}.
\begin{definition}[Sobolev space, integer order]\label{def:sobint}
For $m\in\mathbb{N}_0$, the Sobolev space of integer order $m$ is:
\begin{equation*}
H^m(\mathcal{X}) \coloneqq \big\{f\in L^2(\mathcal{X},\mathrm{d}x) : D^\alpha f\in L^2(\mathcal{X},\mathrm{d}x) \text{ for every } \alpha \text{ with } |\alpha|\le m\big\},
\end{equation*}
with $\|f\|^2_{H^m(\mathcal{X})} \coloneqq \sum_{|\alpha|\le m}\|D^\alpha f\|^2_{L^2(\mathcal{X},\mathrm{d}x)}$.
\end{definition}

\begin{definition}[Sobolev space, fractional order]\label{def:sobfrac}
For $\tau=m+\vartheta$ with $m\in\mathbb{N}_0$ and $\vartheta\in(0,1)$, define the \emph{Gagliardo (Sobolev--Slobodeckij) seminorm} as:
\begin{equation*}
[f]^2_{\tau,\mathcal{X}} \coloneqq \sum_{|\alpha|=m}\iint_{\mathcal{X}\times\mathcal{X}} \frac{|D^\alpha f(x)-D^\alpha f(y)|^2}{\|x-y\|^{p+2\vartheta}}\mathrm{d}x\mathrm{d}y.
\end{equation*}
The Sobolev space of fractional order $\tau$ is $H^\tau(\mathcal{X}) \coloneqq \big\{f\in H^m(\mathcal{X}) : [f]_{\tau,\mathcal{X}}<\infty\big\}$, with $\|f\|^2_{H^\tau(\mathcal{X})} \coloneqq \|f\|^2_{H^m(\mathcal{X})} + [f]^2_{\tau,\mathcal{X}}$. For $\tau\in\mathbb{N}_0$, \emph{i.e.}, $\vartheta=0$, $H^\tau(\mathcal{X})$ is understood as in \cref{def:sobint}.
\end{definition}

\begin{corollary}[Domination absorbs a summand]\label{cor:absorb}
Let $k_1,k_2$ be positive semi-definite kernels on the same set. If $\mathcal{H}(k_1)\hookrightarrow\mathcal{H}(k_2)$ continuously, then $\mathcal{H}(k_1+k_2)\cong\mathcal{H}(k_2)$.
\end{corollary}
\begin{proof}
By the sum property of RKHSs, $\mathcal{H}(k_1+k_2)=\mathcal{H}(k_1)+\mathcal{H}(k_2)$. Since $\mathcal{H}(k_1)\subset\mathcal{H}(k_2)$, this equals $\mathcal{H}(k_2)$ as a set. Let $C$ be the embedding constant, then $\|f_1\|_{\mathcal{H}(k_2)}\le C\|f_1\|_{\mathcal{H}(k_1)}$. For $f=f_1+f_2\in\mathcal{H}(k_1)+\mathcal{H}(k_2)$, we have:
\begin{equation*}
\|f\|_{\mathcal{H}(k_2)}\le\|f_1\|_{\mathcal{H}(k_2)}+\|f_2\|_{\mathcal{H}(k_2)}\le C\|f_1\|_{\mathcal{H}(k_1)}+\|f_2\|_{\mathcal{H}(k_2)}\le(C+1)\sqrt{\|f_1\|^2_{\mathcal{H}(k_1)}+\|f_2\|^2_{\mathcal{H}(k_2)}}.
\end{equation*}
Taking the infimum over decompositions gives $\|f\|_{\mathcal{H}(k_2)}\le(C+1)\|f\|_{\mathcal{H}(k_1+k_2)}$, and the trivial decomposition $f_1=0$ gives the reverse.
\end{proof}

\begin{lemma}\label{lem:transport}
Let $k$ be a positive semi-definite kernel on a set $E_2$, let $\varphi_0:E_1\to E_2$ be a surjection, and let $g:E_1\to\mathbb{R}$ be nowhere zero. Then the map $f\mapsto g\cdot(f\circ\varphi_0)$ is an isometric isomorphism from $\mathcal{H}(k)$ onto $\mathcal{H}\big(M_g[\varphi_0^*k]\big)$.
\end{lemma}

\begin{proof}
If $\Psi:E_2\to W$ is a feature map for $k$, \emph{i.e.}, $k(y,y')=\langle\Psi(y),\Psi(y')\rangle_W$, then $\Psi'(x)\coloneqq g(x)\Psi(\varphi_0(x))$ is a feature map for $M_g[\varphi_0^*k]$, since $\langle\Psi'(x),\Psi'(x')\rangle_W=g(x)g(x')k(\varphi_0(x),\varphi_0(x'))=M_g[\varphi_0^*k](x,x')$. Write $\Phi(w)\coloneqq\langle w,\Psi(\cdot)\rangle_W$ and $\Phi'(w)\coloneqq\langle w,\Psi'(\cdot)\rangle_W$ for the corresponding representation maps, so that $\mathcal{H}(k)=\{\Phi(w):w\in W\}$, $\mathcal{H}(M_g[\varphi_0^*k])=\{\Phi'(w):w\in W\}$, and, for $f\in\mathcal{H}(k)$, $\|f\|_{\mathcal{H}(k)}=\min\{\|w\|_W:\Phi(w)=f\}$, with the analogous expression for $\mathcal{H}(M_g[\varphi_0^*k])$ via $\Phi'$. Since $\Phi'(w)(x)=\langle w,\Psi'(x)\rangle_W=g(x)\langle w,\Psi(\varphi_0(x))\rangle_W=g(x)\Phi(w)(\varphi_0(x))$ for every $w\in W$, we have $\Phi'=T\circ\Phi$, where $T(f)\coloneqq g\cdot(f\circ\varphi_0)$. In particular, any $w_0$ representing $f$ on the left (\emph{i.e.}, $\Phi(w_0)=f$) automatically represents $T(f)$ on the right (\emph{i.e.}, $\Phi'(w_0)=T(f)$). We claim that, for such $w_0$, the set of \emph{all} $w\in W$ representing $f$ on the left coincides with the set of all $w\in W$ representing $T(f)$ on the right. Since $\Phi$ and $\Phi'$ are linear, the former set is the coset $w_0+N$ and the latter is $w_0+N'$, where $N\coloneqq\ker\Phi$ and $N'\coloneqq\ker\Phi'$; so it suffices to show $N=N'$. Indeed, $w\in N$ means $\langle w,\Psi(y)\rangle_W=0$ for every $y\in E_2$, which, since $\varphi_0$ is surjective, holds if and only if $\langle w,\Psi(\varphi_0(x))\rangle_W=0$ for every $x\in E_1$; and since $g$ is nowhere zero, this holds if and only if $g(x)\langle w,\Psi(\varphi_0(x))\rangle_W=\langle w,\Psi'(x)\rangle_W=0$ for every $x\in E_1$, \emph{i.e.}, $w\in N'$. Hence $N=N'$, and the two cosets $w_0+N$, $w_0+N'$ are identical as subsets of $W$. Consequently, for every $f\in\mathcal{H}(k)$, the set of $w\in W$ representing $f$ on the left, namely $w_0+N$, is exactly the set of $w\in W$ representing $T(f)$ on the right, namely $w_0+N'=w_0+N$. Minimizing $\|w\|_W$ over this common set gives:
\begin{equation*}
\|f\|_{\mathcal{H}(k)}=\min_{w\in w_0+N}\|w\|_W=\min_{w\in w_0+N'}\|w\|_W=\|T(f)\|_{\mathcal{H}(M_g[\varphi_0^*k])}=\|g\cdot(f\circ\varphi_0)\|_{\mathcal{H}(M_g[\varphi_0^*k])}.
\end{equation*}
In addition, $T$ is a bijection from $\mathcal{H}(k)$ onto $\mathcal{H}(M_g[\varphi_0^*k])$. Indeed, if $T(f_1)=T(f_2)$, then $g(x)f_1(\varphi_0(x))=g(x)f_2(\varphi_0(x))$ for all $x\in E_1$. Since $g$ is nowhere zero, $f_1(\varphi_0(x))=f_2(\varphi_0(x))$ and since $\varphi_0$ is surjective, $f_1=f_2$. Thus $T$ is injective. Surjectivity holds because every $h\in\mathcal{H}(M_g[\varphi_0^*k])$ equals $\Phi'(w)$ for some $w$, hence $h=T(\Phi(w))$. Together with the norm identity above, $T$ is an isometric isomorphism.
\end{proof}

\begin{lemma}\label{lem:gnomonic}
The map $\phi$ is a real-analytic diffeomorphism of $\mathbb{R}^p$ onto the open upper hemisphere $\mathbb{S}^p_+\coloneqq\{y\in\mathbb{S}^p:y_{p+1}>0\}$, with inverse $\phi^{-1}(y)=(y_1,\dots,y_p)/y_{p+1}$. Moreover, under the assumption that there exists $A>0$ such that $\|x\|\leq A$ for $x\in\mathcal{X}$, it holds that:
\begin{equation}\label{eq:cap_bound}
    \phi(x)_{p+1}=\big(1+\|x\|^2\big)^{-1/2}\;\geq\;\big(1+A^2\big)^{-1/2}\;>\;0 \qquad\text{for all } x\in\mathcal{X},
\end{equation}
so $\widehat{\mathcal{X}}\coloneqq\phi(\mathcal{X})$ has closure contained in the compact cap $\{y\in\mathbb{S}^p:y_{p+1}\geq(1+A^2)^{-1/2}\}\subset\mathbb{S}^p_+$. Consequently, there exists $\chi\in C^\infty(\mathbb{S}^p)$ with $0\leq\chi\leq1$, $\chi\equiv1$ on $\widehat{\mathcal{X}}$, and $\mathrm{supp}\,\chi\subset\mathbb{S}^p_+$.
\end{lemma}

\begin{proof}
The map $\phi$ is real-analytic on $\mathbb{R}^p$, since its coordinate functions are compositions of polynomials with the real-analytic function $t\mapsto t^{-1/2}$ on $(0,\infty)$. Likewise, $\phi^{-1}$ is real-analytic on $\mathbb{S}^p_+$, since $y_{p+1}>0$ there, and hence $t\mapsto t^{-1}$ is real-analytic on $(0,\infty)$. A direct calculation shows that $\phi^{-1}\circ\phi=\mathrm{Id}_{\mathbb{R}^p}$ and $\phi\circ\phi^{-1}=\mathrm{Id}_{\mathbb{S}^p_+}$, so $\phi$ is a real-analytic diffeomorphism. \cref{eq:cap_bound} follows from $\|x\|\leq A$ since $t\mapsto(1+t)^{-1/2}$ is decreasing. Let $C_A\coloneqq\{y\in\mathbb{S}^p:y_{p+1}\geq(1+A^2)^{-1/2}\}$. By \cref{eq:cap_bound}, $\widehat{\mathcal{X}}\subset C_A$, and since $C_A$ is closed (the preimage of a closed half-line under the continuous map $y\mapsto y_{p+1}$), $\overline{\widehat{\mathcal{X}}}\subset C_A$ as well. In particular, $\overline{\widehat{\mathcal{X}}}$ is compact being a closed subset of the compact $C_A$. Moreover, $C_A\subset\mathbb{S}^p_+$ since $(1+A^2)^{-1/2}>0$. A smooth bump function for the pair $\big(\overline{\widehat{\mathcal{X}}},\mathbb{S}^p_+\big)$, which exists since $\overline{\widehat{\mathcal{X}}}$ is compact and contained in the open set $\mathbb{S}^p_+$, gives $\chi$ \citep[Proposition~2.26]{lee2013smooth}.
\end{proof}

\begin{corollary}\label{cor:reduction}
Define $\widehat{\mathcal{X}}\coloneqq\phi(\mathcal{X})$. It holds that:
\begin{equation}\label{eq:reduction}
    \hntk= q\cdot\Big[\mathcal{H}\big(K_0|_{\widehat{\mathcal{X}}}\big)\circ\phi\Big]+ \mathbb{R}\cdot\mathbbm{1}_{\mathcal{X}},
\end{equation}
where $\mathbbm{1}_\mathcal{X}:\mathcal{X}\to\mathbb{R}$ denotes the constant-one function, i.e., $\mathbbm{1}_\mathcal{X}(x)=1$ for all $x\in\mathcal{X}$.
\end{corollary}
\begin{proof}
Since $\widehat{\mathcal{X}}\coloneqq\phi(\mathcal{X})$, the restriction $\phi|_{\mathcal{X}}$ is a surjection onto $\widehat{\mathcal{X}}$, and $q\geq1>0$ on $\mathcal{X}$. \cref{lem:transport} therefore applies to the kernel $M_q[\phi^*K_0]$ appearing in \cref{eq:kernel_factorization}, giving $\mathcal{H}\big(M_q[\phi^*K_0]\big)=q\cdot[\mathcal{H}(K_0|_{\widehat{\mathcal{X}}})\circ\phi]$ isometrically. The factor $\tfrac12$ rescales the norm only, and the sum property of $\mathcal{H}(\cdot)$ recalled above assembles the two terms of \cref{eq:kernel_factorization} into \cref{eq:reduction}.
\end{proof}

\subsection{The sphere kernel: parity gap and the hemisphere lemma}
Denote by $L^2(\mathbb{S}^p)$ the space of square-integrable functions on $\mathbb{S}^p$ with respect to its surface measure. Let $\{Y_{k,j}\}_{k\ge0,\,1\le j\le a_k}$ denote an orthonormal basis of $L^2(\mathbb{S}^p)$, with $Y_{k,j}$ a spherical harmonic of degree $k$ and $a_k$ the dimension of the degree-$k$ harmonics $\mathcal{Y}_k$. Every $f\in L^2(\mathbb{S}^p)$ decomposes uniquely as $f=\sum_{k\ge0}f_k$, $f_k\coloneqq\sum_{j=1}^{a_k}\langle f,Y_{k,j}\rangle Y_{k,j}\in\mathcal{Y}_k$. For $\tau\ge0$, the Sobolev space $H^\tau(\mathbb{S}^p)$ is defined spectrally by $\|f\|^2_{H^\tau(\mathbb{S}^p)}\coloneqq\sum_{k\ge0}(1+k(k+p-1))^\tau\|f_k\|^2_{L^2(\mathbb{S}^p)}\asymp\sum_{k\ge0}(1+k)^{2\tau}\|f_k\|^2_{L^2(\mathbb{S}^p)}$. Given a dot-product kernel $\kappa(\langle y,y'\rangle)$ on $\mathbb{S}^p$ with Mercer decomposition $\kappa(\langle y,y'\rangle)=\sum_{k\ge0}\mu_k\sum_{j=1}^{a_k}Y_{k,j}(y)Y_{k,j}(y')$, its RKHS is:
\begin{equation*}
\mathcal{H}(\kappa)\coloneqq\Big\{f=\sum_{k\ge0}\sum_{j=1}^{a_k}c_{k,j}Y_{k,j}\ :\ \|f\|^2_{\mathcal{H}(\kappa)}\coloneqq\sum_{k\ge0}\sum_{j=1}^{a_k}\frac{c_{k,j}^2}{\mu_k}<\infty\Big\},
\end{equation*}
with the convention, forced by finiteness, that $c_{k,j}=0$ whenever $\mu_k=0$. Moreover, for an open subset $\mathcal{V}\subset\mathbb{S}^p$, the Sobolev space $H^\tau(\mathcal{V})$ is defined by restriction $H^\tau(\mathcal{V})\coloneqq\big\{U|_{\mathcal{V}}\ :\ U\in H^\tau(\mathbb{S}^p)\big\}$ with norm $\|g\|_{H^\tau(\mathcal{V})}\coloneqq\inf\big\{\|U\|_{H^\tau(\mathbb{S}^p)}\ :\ U\in H^\tau(\mathbb{S}^p),\ U|_{\mathcal{V}}=g\big\}$. For more details, we refer the reader to \citep{atkinson2012spherical,barcelo2020characterization,hewett2017note}.

\begin{corollary}[Parity gap on $\mathbb{S}^p$]\label{cor:paritygap}
Recall $K_0(y,y')\coloneqq\kappa(\langle y,y'\rangle)$ for $y,y'\in\mathbb{S}^p$, with Mercer eigenvalues $(\mu_k)_{k\ge0}$. Then, we have:
\begin{equation*}
    \mu_0,\mu_1>0,\qquad \mu_k=0\ \text{for every odd } k\ge3,\qquad \mu_k\sim C(p+1)\,k^{-(p+1)}\ \text{otherwise as }k\to\infty,    
\end{equation*}
for a constant $C(p+1)>0$. Consequently, writing $\mathcal{E}\coloneqq\overline{\bigoplus_{k\text{ even}}\mathcal{Y}_k}\oplus\mathcal{Y}_1\subset L^2(\mathbb{S}^p)$, it holds that:
\begin{equation*}
\mathcal{H}(K_0)\cong\mathcal{E}\cap H^{\frac{p+1}{2}}(\mathbb{S}^p)\ \subsetneq\ H^{\frac{p+1}{2}}(\mathbb{S}^p).
\end{equation*}
\end{corollary}

\begin{proof}
The first part of this corollary comes from \citep[Proposition~5]{bietti2019inductive}. 

\textbf{Identifying $\mathcal{E}$.} Since $L^2(\mathbb{S}^p)=\bigoplus_{k\ge0}\mathcal{Y}_k$ is an orthogonal decomposition, the closed subspace $\mathcal{E}$ consists exactly of those $f=\sum_{k\geq 0}f_k\in L^2(\mathbb{S}^p)$ with $f_k=0$ for every odd $k\ge3$. We note that the spherical-harmonic expansion already converges in $L^2(\mathbb{S}^p)$, since we have by Parseval's identity that $\|f\|^2_{L^2(\mathbb{S}^p)}=\sum_k\|f_k\|^2_{L^2(\mathbb{S}^p)}<\infty$ for any $f\in L^2(\mathbb{S}^p)$.

\textbf{A two-sided bound on $\mu_k^{-1}$.} Inverting $\mu_k\sim C(p+1)k^{-(p+1)}$ gives $\mu_k^{-1}\sim C(p+1)^{-1}k^{p+1}$ as $k\to\infty$. By the definition of asymptotic equivalence, there exist constants $c_0,C_0>0$ depending only on $p$ and $K\ge1$ such that $c_0 k^{p+1}\leq\mu_k^{-1}\leq C_0 k^{p+1}$, for $k\geq K$. It remains to control the finitely many indices $k<K$. Since $\mu_k>0$ for every $k\in\{1\}\cup2\mathbb N_0$ with $2\mathbb{N}_0=\{0,2,4,6,\ldots\}$, the quantities $\frac{\mu_k^{-1}}{(1+k)^{p+1}}$, with $k\in\{1\}\cup2\mathbb N_0$ and $k<K$, form a finite collection of strictly positive finite numbers. Thus, their minimum is positive and their maximum is finite. Combining these bounds with the preceding estimate for $k\ge K$, and using $k^{p+1}\asymp(1+k)^{p+1}$ for $k\ge1$, yields constants $c,C>0$, depending only on $p$, such that $c(1+k)^{p+1}\leq\mu_k^{-1}\leq C(1+k)^{p+1}$ for $k\in\{1\}\cup2\mathbb N_0$.

\emph{($\subseteq$).} Let $f\in\mathcal{H}(K_0)$, so $f=\sum_{k,j}c_{k,j}Y_{k,j}$ with $\sum_{k,j}c_{k,j}^2/\mu_k<\infty$ and $c_{k,j}=0$ forced wherever $\mu_k=0$, \emph{i.e.}, at every odd $k\ge3$. By the identification above this already gives $f\in\mathcal{E}$. Summing the norm condition over $j$ (Parseval within each degree, $\sum_jc_{k,j}^2=\|f_k\|^2_{L^2(\mathbb{S}^p)}$) and using the lower-bound on $\mu_k^{-1}$, we get:
\begin{align*}
    \|f\|_{H^\frac{p+1}{2}(\mathbb{S}^p)}^2=&\sum_{k\ge0}(1+k)^{p+1}\|f_k\|^2_{L^2(\mathbb{S}^p)}=\sum_{k\in\{1\}\cup2\mathbb{N}_0}(1+k)^{p+1}\|f_k\|^2_{L^2(\mathbb{S}^p)}\\
    \le& \frac1c\sum_{k\in\{1\}\cup2\mathbb{N}_0}\mu_k^{-1}\|f_k\|^2_{L^2(\mathbb{S}^p)}=\frac1c\|f\|^2_{\mathcal{H}(K_0)}<\infty,
\end{align*}
where the first equality holds because $f_k=0$ for $k\notin\{1\}\cup2\mathbb{N}_0$; so $f\in H^{\frac{p+1}2}(\mathbb{S}^p)$. Hence $f\in\mathcal{E}\cap H^{\frac{p+1}2}(\mathbb{S}^p)$.

\emph{($\supseteq$).} Let $f\in\mathcal{E}\cap H^{\frac{p+1}2}(\mathbb{S}^p)$. Then, $f_k=0$ for every odd $k\ge3$, which is exactly the forcing condition defining $\mathcal{H}(K_0)$ above, so $f$ is an admissible candidate for $\mathcal{H}(K_0)$. Since $f\in H^{\frac{p+1}2}(\mathbb{S}^p)$, $\sum_{k\ge0}(1+k)^{p+1}\|f_k\|^2_{L^2(\mathbb{S}^p)}<\infty$. As $f\in\mathcal{E}$, this sum is already supported on $\{1\}\cup2\mathbb{N}_0$, so the upper-bound on $\mu_k^{-1}$ gives:
\begin{equation*}
\|f\|^2_{\mathcal{H}(K_0)}=\sum_{k\in\{1\}\cup2\mathbb{N}_0}\mu_k^{-1}\|f_k\|^2_{L^2(\mathbb{S}^p)}\le C\sum_{k\in\{1\}\cup2\mathbb{N}_0}(1+k)^{p+1}\|f_k\|^2_{L^2(\mathbb{S}^p)}=C\|f\|_{H^\frac{p+1}{2}(\mathbb{S}^p)}^2<\infty,
\end{equation*}
where we use the fact that $f\in H^{\frac{p+1}2}(\mathbb{S}^p)$. Hence, $f\in\mathcal{H}(K_0)$.

The two inclusions give $\mathcal{H}(K_0)=\mathcal{E}\cap H^{\frac{p+1}2}(\mathbb{S}^p)$ as \emph{sets}, and the same constants $c,C$, one used in each direction, give the equivalence of norms. Together, this gives $\mathcal{H}(K_0)\cong\mathcal{E}\cap H^{\frac{p+1}2}(\mathbb{S}^p)$.

\textbf{Properness.} A nonzero $Y\in\mathcal{Y}_3$ is a polynomial, hence lies in every $H^\tau(\mathbb{S}^p)$ for $\tau\geq0$. Its only nonzero component is at degree $3$, which is odd and $\ge3$, so $Y\notin\mathcal{E}$ by the identification above, hence $Y\notin\mathcal{H}(K_0)$. So the containment $\mathcal{H}(K_0)\subsetneq H^{\frac{p+1}2}(\mathbb{S}^p)$ is strict.
\end{proof}

For $y\in\mathbb{S}^p$, write $\mathfrak{a}(y)\coloneqq-y$ for the antipodal map. We first record how composition with $\mathfrak{a}$ acts on the spherical-harmonic expansion of a function.

\begin{lemma}\label{lem:antipodal}
For any $h\in L^2(\mathbb{S}^p)$ with expansion $h=\sum_{k\ge0}h_k$, $h_k\in\mathcal{Y}_k$, the function $h\circ\mathfrak{a}$ has expansion $(h\circ\mathfrak{a})_k=(-1)^kh_k$. Consequently, $\|h\circ\mathfrak{a}\|_{H^\tau(\mathbb{S}^p)}=\|h\|_{H^\tau(\mathbb{S}^p)}$ for every $\tau\ge0$.
\end{lemma}
\begin{proof}
Every $Y\in\mathcal{Y}_k$ is the restriction to $\mathbb{S}^p$ of a polynomial on $\mathbb{R}^{p+1}$ that is homogeneous of degree $k$, \emph{i.e.}, $Y(\lambda x)=\lambda^kY(x)$ for every scalar $\lambda$. Taking $\lambda=-1$ gives $Y(-y)=(-1)^kY(y)$ for every $y\in\mathbb{S}^p$. Applying this to each term of $h=\sum_{k\geq0}h_k$ yields:
\begin{equation*}
    h(\mathfrak{a}(y))=h(-y)=\sum_{k\ge0}h_k(-y)=\sum_{k\ge0}(-1)^kh_k(y),
\end{equation*}
so $h\circ\mathfrak{a}=\sum_{k\geq0}(-1)^kh_k$. Since $(-1)^kh_k\in\mathcal{Y}_k$, this is precisely the spherical-harmonic expansion of $h\circ\mathfrak{a}$, giving $(h\circ\mathfrak{a})_k=(-1)^kh_k$. Since $\|(-1)^kh_k\|_{L^2(\mathbb{S}^p)}=\|h_k\|_{L^2(\mathbb{S}^p)}$ for every $k$, we obtain:
\begin{align*}
\|h\circ\mathfrak{a}\|^2_{H^\tau(\mathbb{S}^p)}=&\sum_{k\ge0}\big(1+k(k+p-1)\big)^\tau\|(h\circ\mathfrak{a})_k\|^2_{L^2(\mathbb{S}^p)}\\
=&\sum_{k\ge0}\big(1+k(k+p-1)\big)^\tau\|h_k\|^2_{L^2(\mathbb{S}^p)}
=\|h\|^2_{H^\tau(\mathbb{S}^p)}.\qedhere
\end{align*}
\end{proof}

\begin{lemma}[Hemisphere lemma]\label{lem:hemisphere}
Let $\mathcal{V}\subset\mathbb{S}^p$ be open, bounded, with Lipschitz boundary, with $\overline{\mathcal{V}}$ compactly contained in $\mathbb{S}^p_+$. Then, it holds that $\mathcal{H}\big(K_0|_{\mathcal{V}}\big)\cong H^{\frac{p+1}{2}}(\mathcal{V})$.
\end{lemma}
\begin{proof}
\emph{($\subseteq$).} Let $g\in\mathcal{H}(K_0|_{\mathcal{V}})$. By the restriction property of RKHSs, there exists $F\in\mathcal{H}(K_0)$ with $F|_{\mathcal{V}}=g$ and $\|F\|_{\mathcal{H}(K_0)}=\|g\|_{\mathcal{H}(K_0|_{\mathcal{V}})}$. By \cref{cor:paritygap}, it holds that $\mathcal{H}(K_0)\subsetneq H^{\frac{p+1}{2}}(\mathbb{S}^p)$ and there is a constant $C_1>0$, depending only on $p$, with $\|F\|_{H^{\frac{p+1}{2}}(\mathbb{S}^p)}\le C_1\,\|F\|_{\mathcal{H}(K_0)}$. Since $F$ is in particular one extension of $g$ to $\mathbb{S}^p$, the definition of the restriction norm on $\mathcal{V}$ as an infimum over all such extensions gives $\|g\|_{H^{\frac{p+1}{2}}(\mathcal{V})}\le\|F\|_{H^{\frac{p+1}{2}}(\mathbb{S}^p)}$. Chaining the three displays, we get:
\begin{equation*}
\|g\|_{H^{\frac{p+1}{2}}(\mathcal{V})}\ \le\ \|F\|_{H^{\frac{p+1}{2}}(\mathbb{S}^p)}\ \le\ C_1\,\|F\|_{\mathcal{H}(K_0)}\ =\ C_1\|g\|_{\mathcal{H}(K_0|_{\mathcal{V}})}<\infty,
\end{equation*}
so $g\in H^{\frac{p+1}{2}}(\mathcal{V})$ with $\|g\|_{H^{\frac{p+1}{2}}(\mathcal{V})}\le C_1\|g\|_{\mathcal{H}(K_0|_{\mathcal{V}})}$.

\emph{($\supseteq$).} Let $g\in H^{\frac{p+1}{2}}(\mathcal{V})$. Our goal is to produce a function on the whole sphere, in $\mathcal{H}(K_0)$, that restricts to $g$ on $\mathcal{V}$.

\textbf{Step 1: extend $g$ to the sphere, then confine the extension to the upper hemisphere.} Since $\mathcal{V}$ is bounded Lipschitz, there is a bounded linear extension operator \citep[Chapter~VI, Theorem~5]{stein1970singular}, giving $\widetilde{G}\in H^{\frac{p+1}{2}}(\mathbb{S}^p)$ with $\widetilde{G}|_{\mathcal{V}}=g$ and $\|\widetilde{G}\|_{H^{\frac{p+1}{2}}(\mathbb{S}^p)}\le C_2\|g\|_{H^{\frac{p+1}{2}}(\mathcal{V})}$ for a constant $C_2>0$ depending only on $p$ and the Lipschitz character of $\mathcal{V}$. Let $\chi\in C^\infty(\mathbb{S}^p)$ be the smooth bump function from \cref{lem:gnomonic}, equal to $1$ on $\mathcal{V}$ and supported in $\mathbb{S}^p_+$, and set $G\coloneqq\chi\widetilde{G}$. By the multiplier property of Sobolev spaces \citep[\S4.2]{Triebel1992}, there is a constant $C_3>0$, depending only on $\chi$, such that:
\begin{equation*}
\|G\|_{H^{\frac{p+1}{2}}(\mathbb{S}^p)}\le C_3\|\widetilde{G}\|_{H^{\frac{p+1}{2}}(\mathbb{S}^p)} \le C_2C_3\|g\|_{H^{\frac{p+1}{2}}(\mathcal{V})}.
\end{equation*}
Moreover, $\mathrm{supp}\,G\subset\mathrm{supp}\,\chi\subset\mathbb{S}^p_+$, and since $\chi\equiv1$ on $\mathcal{V}$, we have $G|_{\mathcal{V}}=\chi|_{\mathcal{V}}\cdot\widetilde{G}|_{\mathcal{V}}=\widetilde{G}|_{\mathcal{V}}=g$.

\textbf{Step 2: symmetrize across the equator.} Define $F\coloneqq G+G\circ\mathfrak{a}$. We verify three properties of $F$.

\emph{(i) $F$ restricts to $g$ on $\mathcal{V}$.} For $y\in\mathcal{V}\subset\mathbb{S}^p_+$, the antipodal point $-y$ lies in the open lower hemisphere $\mathbb{S}^p_-\coloneqq\{y:y_{p+1}<0\}$, which is disjoint from $\mathrm{supp}\,G\subset\mathbb{S}^p_+$. Hence, $G(-y)=0$ and $F(y)=G(y)+G(-y)=G(y)=g(y)$.

\emph{(ii) $F\in H^{\frac{p+1}{2}}(\mathbb{S}^p)$, with $\|F\|_{H^{\frac{p+1}{2}}(\mathbb{S}^p)}\le2\|G\|_{H^{\frac{p+1}{2}}(\mathbb{S}^p)}$.} By \cref{lem:antipodal}, $\|G\circ\mathfrak{a}\|_{H^{\frac{p+1}{2}}(\mathbb{S}^p)}=\|G\|_{H^{\frac{p+1}{2}}(\mathbb{S}^p)}$. The triangle inequality gives the bound on $F=G+G\circ\mathfrak{a}$.

\emph{(iii) $F$ has no spherical-harmonic component of odd degree.} By \cref{lem:antipodal}, $F_k=G_k+(G\circ\mathfrak{a})_k=G_k+(-1)^kG_k=\big(1+(-1)^k\big)G_k$. For odd $k$, $1+(-1)^k=0$, so $F_k=0$.

By (ii) and (iii), $F\in\mathcal{E}\cap H^{\frac{p+1}{2}}(\mathbb{S}^p)$. By \cref{cor:paritygap}, $F\in\mathcal{H}(K_0)$ with $\|F\|_{\mathcal{H}(K_0)}\leq C\|F\|_{H^{\frac{p+1}{2}}(\mathbb{S}^p)}\leq2CC_2C_3\|g\|_{H^{\frac{p+1}{2}}(\mathcal{V})}$, where $C>0$ depends only on $p$. By (i), $F$ is an admissible extension of $g$, so by the restriction property of RKHSs, $g\in\mathcal{H}(K_0|_{\mathcal{V}})$ with $\|g\|_{\mathcal{H}(K_0|_{\mathcal{V}})}\le\|F\|_{\mathcal{H}(K_0)}\leq 2CC_2C_3\|g\|_{H^{\frac{p+1}{2}}(\mathcal{V})}$.
\end{proof}
 
\subsection{\texorpdfstring{The RKHS $\hntk$ is a Sobolev space}{The RKHS HNTK is a Sobolev space}}
\begin{theorem}[The RKHS of $K^{\mathrm{NTK}}$ is a Sobolev space]\label{thm:sobolev_ntk}
Let $p\geq1$ and let $\mathcal{X}\subset\mathbb{R}^p$ be a nonempty bounded Lipschitz domain, with $\|x\|\leq A$ for any $x\in\mathcal{X}$. It holds that $\hntk\cong H^{\frac{p+1}{2}}(\mathcal{X})$ with equivalence constants depending only on $p$, $A$, and the Lipschitz character of $\mathcal{X}$.
\end{theorem}

\begin{proof}
By \cref{cor:reduction}, we have $\hntk=q\cdot\big[\mathcal{H}(K_0|_{\widehat{\mathcal{X}}})\circ\phi\big]+\mathbb{R}\cdot\mathbbm{1}_{\mathcal{X}}$.

\textbf{Step 1: identify the spherical factor.} We check the three hypotheses of \cref{lem:hemisphere} for $\widehat{\mathcal{X}}=\phi(\mathcal{X})$.

\emph{$\widehat{\mathcal{X}}$ is open.} By \cref{lem:gnomonic}, $\phi$ is a diffeomorphism, hence an open map. Since $\mathcal{X}$ is open, so is $\widehat{\mathcal{X}}$.

\emph{$\widehat{\mathcal X}$ is bounded with Lipschitz boundary.} Boundedness is immediate, since $\mathbb{S}^p$ is compact and $\widehat{\mathcal{X}}\subset\mathbb{S}^p$. For the boundary, let $\mathcal A$ be the maximal real-analytic atlas of $\mathbb{S}^p$ (for its standard structure as a submanifold of $\mathbb{R}^{p+1}$), and declare an open $\widehat\Omega\subset\mathbb{S}^p$ to have Lipschitz boundary if there is a chart $(\psi,U)\in\mathcal A$ with $\overline{\widehat\Omega}\subset U$ such that $\psi(\widehat\Omega)$ is a bounded Lipschitz domain in $\mathbb{R}^p$. This does not depend on the chart. Indeed, let $(\psi_1,U_1),(\psi_2,U_2)\in\mathcal A$ both satisfy $\overline{\widehat\Omega}\subset U_1$ and $\overline{\widehat\Omega}\subset U_2$. Since $\mathbb S^p$ is compact, $\overline{\widehat\Omega}$ is compact. Given that $\psi_1:U_1\to\mathbb{R}^p$ and $\psi_2:U_2\to\mathbb{R}^p$ are homeomorphisms, we have $\overline{\psi_1(\widehat\Omega)}=\psi_1(\overline{\widehat\Omega})$ and $\overline{\psi_2(\widehat\Omega)}=\psi_2(\overline{\widehat\Omega})$. Since $\overline{\widehat\Omega}\subset U_1\cap U_2$, we get $\overline{\psi_1(\widehat\Omega)}\subset\psi_1(U_1\cap U_2)$ and $\overline{\psi_2(\widehat\Omega)}\subset\psi_2(U_1\cap U_2)$. The transition map $\tau_0=\psi_2\circ\psi_1^{-1}\colon\psi_1(U_1\cap U_2)\to\psi_2(U_1\cap U_2)$ is a real-analytic $C^1$ diffeomorphism, with inverse $\tau_0^{-1}=\psi_1\circ\psi_2^{-1}$. By the two displayed inclusions, $\tau_0$ is defined on a neighborhood of $\overline{\psi_1(\widehat\Omega)}$ and $\tau_0^{-1}$ is defined on a neighborhood of $\overline{\psi_2(\widehat\Omega)}$, so \cite[Theorem~4.1]{HofmannMitreaTaylor2007} applies to each: it shows $\psi_1(\widehat\Omega)$ is a bounded Lipschitz domain if and only if $\tau_0(\psi_1(\widehat\Omega))=\psi_2(\widehat\Omega)$ is, which proves the chart-independence property. 

We now specialize to $\phi^{-1}$. The map $\phi^{-1}\colon\mathbb{S}^p_+\to\mathbb{R}^p$, $\phi^{-1}(y)=(y_1,\dots,y_p)/y_{p+1}$, is a real-analytic diffeomorphism with real-analytic inverse $\phi(x)=\tilde{x}/\sqrt{1+\|x\|^2}$, so $(\phi^{-1},\mathbb{S}^p_+)\in\mathcal A$. Since $\mathcal X$ is bounded, $\overline{\mathcal X}$ is compact. The fact that $\phi:\mathbb{R}^p\to\mathbb{S}^p_+$ is a homeomorphism gives $\overline{\widehat{\mathcal X}}=\overline{\phi(\mathcal{X})}=\phi(\overline{\mathcal X})$. Boundedness of $\overline{\mathcal X}$ also keeps $\phi(x)_{p+1}=1/\sqrt{1+\|x\|^2}$ bounded away from $0$ on $\overline{\mathcal X}$, so $\overline{\widehat{\mathcal X}}=\phi(\overline{\mathcal X})\subset\mathbb{S}^p_+$. Thus $(\phi^{-1},\mathbb{S}^p_+)$ is a chart with $\overline{\widehat{\mathcal X}}\subset\mathbb{S}^p_+$, and $\phi^{-1}(\widehat{\mathcal X})=\mathcal X$ is a bounded Lipschitz domain by hypothesis. Hence, $\widehat{\mathcal X}$ has Lipschitz boundary.

\emph{$\overline{\widehat{\mathcal{X}}}$ is compactly contained in $\mathbb{S}^p_+$.} By \cref{lem:gnomonic}, $\overline{\widehat{\mathcal{X}}}$ lies in the closed cap $\{y\in\mathbb{S}^p:y_{p+1}\ge(1+A^2)^{-1/2}\}$, which is compact and contained in $\mathbb{S}^p_+$. Hence, $\overline{\widehat{\mathcal{X}}}$ is compactly contained in $\mathbb{S}^p_+$.

All three hypotheses hold, so \cref{lem:hemisphere} applies and gives $\mathcal{H}\big(K_0|_{\widehat{\mathcal{X}}}\big)\ \cong\ H^{\frac{p+1}{2}}(\widehat{\mathcal{X}})$.

\textbf{Step 2: transport back to $\mathcal{X}$.} We prove in this step that $q\cdot\big[\mathcal{H}(K_0|_{\widehat{\mathcal{X}}})\circ\phi\big]\ \cong\ H^{\frac{p+1}{2}}(\mathcal{X})$.

\emph{Undoing $\phi$.} Let $\mathcal A$ be the maximal real-analytic atlas of $\mathbb S^p$, fixed in Step~1. For any two charts $(\psi_1,U_1),(\psi_2,U_2)\in\mathcal{A}$, the transition map $\tau_0\coloneqq\psi_2\circ\psi_1^{-1}$ is a bijection from $\psi_1(U_1\cap U_2)$ onto $\psi_2(U_1\cap U_2)$. Therefore, $\tau_0$ and its inverse $\tau_0^{-1}=\psi_1\circ\psi_2^{-1}$ are real-analytic, since $\psi_1,\psi_2$ are. Thus, $\tau_0$ is a real-analytic diffeomorphism between these two open subsets of $\mathbb{R}^p$, and in particular, being real-analytic, $\tau_0$ is $C^\infty$, matching the hypothesis of \citep[Theorem~80]{behzadan2022sobolev}. For a bounded open $\widehat\Omega\subset\mathbb{S}^p$ with $\overline{\widehat\Omega}\subset U$ for some chart $(\psi,U)\in\mathcal{A}$, define $\|g\|_{H^{\frac{p+1}{2}}(\widehat\Omega),\psi}\coloneqq\|g\circ\psi^{-1}\|_{H^{\frac{p+1}{2}}(\psi(\widehat\Omega))}$. We claim there is a constant $C_1>0$, depending only on $p$ and the pair of charts $(\psi_1,U_1),(\psi_2,U_2)$, such that:
\begin{equation}\label{eq:chartindep}
\frac{1}{C_1}\|g\|_{H^{\frac{p+1}{2}}(\widehat\Omega),\psi_1}\le\|g\|_{H^{\frac{p+1}{2}}(\widehat\Omega),\psi_2}\le C_1\|g\|_{H^{\frac{p+1}{2}}(\widehat\Omega),\psi_1},
\end{equation}
whenever $(\psi_1,U_1),(\psi_2,U_2)\in\mathcal{A}$ both contain $\overline{\widehat\Omega}$. Indeed, since $\widehat\Omega\subset U_1\cap U_2$, and a diffeomorphism restricted to an open subset of its domain remains a diffeomorphism onto the corresponding open subset of its codomain, $\tau_0$ restricts to a real-analytic diffeomorphism $\tau_0\big|_{\psi_1(\widehat\Omega)}:\psi_1(\widehat\Omega)\to\psi_2(\widehat\Omega)$, with real-analytic inverse $\tau_0^{-1}\big|_{\psi_2(\widehat\Omega)}:\psi_2(\widehat\Omega)\to\psi_1(\widehat\Omega)$. We retain the notation $\tau_0,\tau_0^{-1}$ for these restrictions. We have $g\circ\psi_2^{-1}=(g\circ\psi_1^{-1})\circ\tau_0^{-1}$, and \citep[Theorem~80]{behzadan2022sobolev} applied to $\tau_0^{-1}$ between $\psi_2(\widehat\Omega)$ and $\psi_1(\widehat\Omega)$ gives a constant $C_1'>0$ with:
\begin{equation*}
    \|(g\circ\psi_1^{-1})\circ\tau_0^{-1}\|_{H^{\frac{p+1}{2}}(\psi_2(\widehat\Omega))}\le C_1'\|g\circ\psi_1^{-1}\|_{H^{\frac{p+1}{2}}(\psi_1(\widehat\Omega))},
\end{equation*}
\emph{i.e.}, $\|g\|_{H^{\frac{p+1}{2}}(\widehat\Omega),\psi_2}\le C_1'\|g\|_{H^{\frac{p+1}{2}}(\widehat\Omega),\psi_1}$. Conversely, $g\circ\psi_1^{-1}=(g\circ\psi_2^{-1})\circ\tau_0$, and applying \citep[Theorem~80]{behzadan2022sobolev} directly to $\tau_0$ between $\psi_1(\widehat\Omega)$ and $\psi_2(\widehat\Omega)$ gives a constant $C_1''>0$ with:
\begin{equation*}
    \|(g\circ\psi_2^{-1})\circ\tau_0\|_{H^{\frac{p+1}{2}}(\psi_1(\widehat\Omega))}\le C_1''\,\|g\circ\psi_2^{-1}\|_{H^{\frac{p+1}{2}}(\psi_2(\widehat\Omega))},
\end{equation*}
\emph{i.e.}, $\|g\|_{H^{\frac{p+1}{2}}(\widehat\Omega),\psi_1}\le C_1''\|g\|_{H^{\frac{p+1}{2}}(\widehat\Omega),\psi_2}$. Taking $C_1\coloneqq\max(C_1',C_1'')$ gives \cref{eq:chartindep}.

\emph{Specialization to $\phi^{-1}$.} The map $\phi^{-1}\colon\mathbb{S}^p_+\to\mathbb{R}^p$, $\phi^{-1}(y)=(y_1,\dots,y_p)/y_{p+1}$, is a real-analytic diffeomorphism with real-analytic inverse $\phi(x)=\tilde{x}/\sqrt{1+\|x\|^2}$, so $(\phi^{-1},\mathbb{S}^p_+)\in\mathcal A$ by maximality of $\mathcal A$. As shown above, $\overline{\widehat{\mathcal X}}=\phi(\overline{\mathcal X})\subset\mathbb{S}^p_+$, so $(\phi^{-1},\mathbb S^p_+)$ is a chart of $\mathcal A$ containing $\overline{\widehat{\mathcal X}}$ in its domain, exactly as required by \cref{eq:chartindep}. Fix any anchor chart $(\psi_0,U_0)\in\mathcal A$ with $\overline{\widehat{\mathcal X}}\subset U_0$, and set $\|g\|_{H^{\frac{p+1}{2}}(\widehat{\mathcal X})}\coloneqq\|g\|_{H^{\frac{p+1}{2}}(\widehat{\mathcal X}),\psi_0}$, which by Step~1 is independent of this choice of chart. Applying \cref{eq:chartindep} directly with $\widehat\Omega\coloneqq\widehat{\mathcal X}$, $\psi_1\coloneqq\psi_0$, $\psi_2\coloneqq\phi^{-1}$ gives a constant $C_2>0$ with:
\begin{equation}\label{eq:spectral-chart}
\frac{1}{C_2}\|g\|_{H^{\frac{p+1}{2}}(\widehat{\mathcal X})}\le\|g\|_{H^{\frac{p+1}{2}}(\widehat{\mathcal X}),\phi^{-1}}\le C_2\|g\|_{H^{\frac{p+1}{2}}(\widehat{\mathcal X})}.
\end{equation}
Since $\|g\|_{H^{\frac{p+1}{2}}(\widehat{\mathcal X}),\phi^{-1}}=\|g\circ\phi\|_{H^{\frac{p+1}{2}}(\mathcal X)}$, \cref{eq:spectral-chart} gives $\|g\|_{H^{\frac{p+1}{2}}(\widehat{\mathcal X})}\asymp\|g\circ\phi\|_{H^{\frac{p+1}{2}}(\mathcal X)}$, \emph{i.e.}, $H^{\frac{p+1}{2}}(\widehat{\mathcal X}) \cong H^{\frac{p+1}{2}}(\mathcal X)$ via the pullback $g\mapsto g\circ\phi$.

\emph{Undoing $q$.} The function $q(x)=\|\tilde{x}\|$ is real-analytic with $1\le q\le\sqrt{1+A^2}$ on $\mathcal{X}$. Since $\mathcal X$ is a bounded Lipschitz domain, the multiplication theorem for Sobolev spaces on Lipschitz domains \citep[Corollary~3]{behzadan2022sobolev} shows $H^{\frac{p+1}{2}}(\mathcal{X})$ is a Banach algebra under pointwise multiplication. Applying this to $q$ and to $1/q$ (likewise real-analytic and bounded on $\mathcal X$, since $q\ge1$), there is a constant $C_3>0$, depending only on $A$ and $p$, such that multiplication by $q$ is an isomorphism of $H^{\frac{p+1}{2}}(\mathcal{X})$, with:
\begin{equation*}
    \frac{1}{C_3}\|f\|_{H^{\frac{p+1}{2}}(\mathcal{X})}\le\|qf\|_{H^{\frac{p+1}{2}}(\mathcal{X})}\le C_3\|f\|_{H^{\frac{p+1}{2}}(\mathcal{X})}.
\end{equation*}

\emph{Composing.} Combining the two isomorphisms above with the one from Step~1 ($\mathcal{H}(K_0|_{\widehat{\mathcal{X}}})\cong H^{\frac{p+1}{2}}(\widehat{\mathcal{X}})$), we get $q\cdot\big[\mathcal{H}(K_0|_{\widehat{\mathcal{X}}})\circ\phi\big]\ \cong\ H^{\frac{p+1}{2}}(\mathcal{X})$.

\textbf{Step 3: absorb the outer bias.} Recall from \cref{cor:reduction} that $\hntk=q\cdot\big[\mathcal{H}(K_0|_{\widehat{\mathcal{X}}})\circ\phi\big] \;+\; \mathbb{R}\cdot\mathbbm{1}_{\mathcal{X}}$. Write $k_1\coloneqq\mathbbm{1}_{\mathcal{X}}$ for the constant kernel and $k_2$ for the kernel underlying the first summand, so that $\mathcal{H}(k_2)=q\cdot[\mathcal{H}(K_0|_{\widehat{\mathcal{X}}})\circ\phi]$. By Step~2, $\mathcal{H}(k_2)\cong H^{\frac{p+1}{2}}(\mathcal{X})$. Since $\mathbbm{1}_{\mathcal{X}}$ has vanishing weak derivatives of every order, only the $L^2$ term of the Sobolev norm survives, and $\|\mathbbm{1}_{\mathcal{X}}\|^2_{H^{\frac{p+1}{2}}(\mathcal{X})}=\|\mathbbm{1}_{\mathcal{X}}\|^2_{L^2(\mathcal{X})}=\int_{\mathcal{X}}1\,dx=\mathrm{Vol}(\mathcal{X})<\infty$, where $\mathrm{Vol}(\cdot)$ denotes the volume. Finiteness follows from the boundedness of $\mathcal{X}$. Thus $\mathbbm{1}_{\mathcal{X}}\in H^{\frac{p+1}{2}}(\mathcal{X})$. Since $\mathcal{H}(k_1)=\mathbb{R}\cdot\mathbbm{1}_{\mathcal{X}}$ with $\|c\mathbbm{1}_{\mathcal{X}}\|_{\mathcal{H}(k_1)}=|c|$ for $c\in\mathbb{R}$, the homogeneity of the $H^{\frac{p+1}{2}}(\mathcal{X})$ norm and the previous display give $\|c\mathbbm{1}_{\mathcal{X}}\|_{H^{\frac{p+1}{2}}(\mathcal{X})}=|c|\sqrt{\mathrm{Vol}(\mathcal{X})}=\sqrt{\mathrm{Vol}(\mathcal{X})}\|c\mathbbm{1}_{\mathcal{X}}\|_{\mathcal{H}(k_1)}$. Hence, $\mathcal{H}(k_1)$ embeds continuously into $H^{\frac{p+1}{2}}(\mathcal{X})\cong\mathcal{H}(k_2)$, with embedding constant $\sqrt{\mathrm{Vol}(\mathcal{X})}$. Applying \cref{cor:absorb} with this $k_1$ and $k_2$ yields $\hntk=\mathcal{H}(k_1)+\mathcal{H}(k_2)\cong\mathcal{H}(k_2)\cong H^{\frac{p+1}{2}}(\mathcal{X})$, which is the claimed identification.
\end{proof}

\subsection{\texorpdfstring{The interpolation spaces of $\hntk$ are Sobolev spaces}{The interpolation spaces of HNTK are Sobolev spaces}}

For two Hilbert spaces $E_0,E_1$ continuously embedded in a common space, we denote the \emph{real interpolation space} by $(E_0,E_1)_{s,2}$, $s\in(0,1)$, defined via the standard $K$-method \citep[Chapter~3]{BerghLofstrom1976}.

\begin{corollary}[A Sobolev characterization of ${[\hntk]^s}$]\label{cor:interp}
Let $p\geq1$ and $\mathcal{X}\subset\mathbb{R}^p$ be a nonempty bounded Lipschitz domain, with $\|x\|\leq A$ for any $x\in\mathcal{X}$. Assume that $\mu$, the data distribution on $\mathcal{X}$, has a density $f_\mu$ bounded away from $0$ and $\infty$, i.e., there exist $c_\mu,M_\mu>0$ such that $c_\mu\le f_\mu\le M_\mu$. Then, for every $s\in[0,1]$, $[\hntk]^s\cong H^{\frac{s(p+1)}{2}}(\mathcal{X})$, with equivalence constants depending on $p$, $s$, $A$, the Lipschitz character of $\mathcal{X}$, and the density bounds $c_\mu$, $M_\mu$.
\end{corollary}

\begin{proof}
By \cref{thm:sobolev_ntk}, we have $\hntk\cong H^{\frac{p+1}{2}}(\mathcal{X})$. Since $0<c_\mu\le f_\mu\le M_\mu$, the norms $\|\cdot\|_{L^2(\mathcal{X},\mathrm{d}x)}$ and $\|\cdot\|_{\LL}$ are equivalent, so $L^2(\mathcal{X},\mathrm{d}x)\cong\LL$. We use this identification below.

\textbf{Interior case $s\in(0,1)$.} Let $C_c^\infty(\mathcal{X})$ be the space of smooth, compactly supported functions on $\mathcal{X}$. It is standard that $C_c^\infty(\mathcal{X})$ is dense in $L^2(\mathcal{X},\mathrm{d}x)$. By the identification above, $C_c^\infty(\mathcal{X})$ is therefore dense in $\LL$. Since $H^{\frac{p+1}{2}}(\mathcal{X})\supset C_c^\infty(\mathcal{X})$, it follows that $H^{\frac{p+1}{2}}(\mathcal{X})$ is dense in $\LL$.

The embedding $H^{\frac{p+1}{2}}(\mathcal{X})\hookrightarrow\LL$ is compact. Write $\kappa\coloneqq\lfloor\frac{p+1}{2}\rfloor$. If $p$ is odd, $\frac{p+1}{2}=\kappa$ is already an integer, and the Rellich--Kondrachov theorem \citep[Theorem.~6.3, Part~II, Eq.~(6)]{adams2003sobolev} gives directly that $H^{\frac{p+1}{2}}(\mathcal{X})=H^\kappa(\mathcal{X})\hookrightarrow L^2(\mathcal{X},\mathrm{d}x)$ is compact. If $p$ is even, $\kappa=\frac{p}{2}<\frac{p+1}{2}$, and the Rellich--Kondrachov theorem \citep[Theorem.~6.3, Part~I, Eq.~(4)]{adams2003sobolev} shows that $H^\kappa(\mathcal{X})\hookrightarrow L^2(\mathcal{X},\mathrm{d}x)$ is compact. By definition, $H^{\frac{p+1}{2}}(\mathcal{X})\hookrightarrow H^\kappa(\mathcal{X})$ is continuous, and composing the two gives $H^{\frac{p+1}{2}}(\mathcal{X})\hookrightarrow L^2(\mathcal{X},\mathrm{d}x)$ compact. In either case, by the identification $L^2(\mathcal{X},\mathrm{d}x)\cong\LL$, $H^{\frac{p+1}{2}}(\mathcal{X})\hookrightarrow\LL$ is compact.

By \citep[Theorem~4.6]{steinwart2012mercer}, $[\hntk]^s\cong(\LL,\hntk)_{s,2}$. Since $\hntk\cong H^{\frac{p+1}{2}}(\mathcal{X})$, \citep[Lemma~H.5]{chen2024impacts} gives $(\LL,\hntk)_{s,2}\cong\big(\LL,H^{\frac{p+1}{2}}(\mathcal{X})\big)_{s,2}$, so $[\hntk]^s\cong\big(\LL,H^{\frac{p+1}{2}}(\mathcal{X})\big)_{s,2}$. Using the identification $L^2(\mathcal{X},\mathrm{d}x)\cong\LL$ and the fact that $L^2(\mathcal{X},\mathrm{d}x)\cong H^0(\mathcal{X})$, we get $\LL\cong H^0(\mathcal{X})$. \citep[Lemma~H.6]{chen2024impacts} then gives $\big(\LL,H^{\frac{p+1}{2}}(\mathcal{X})\big)_{s,2}\cong\big(H^0(\mathcal{X}),H^{\frac{p+1}{2}}(\mathcal{X})\big)_{s,2}$. Therefore, by \citep[Theorem~6.4.5]{BerghLofstrom1976}, we obtain $[\hntk]^s\cong\big(H^0(\mathcal{X}),H^{\frac{p+1}{2}}(\mathcal{X})\big)_{s,2}=H^{\frac{s(p+1)}{2}}(\mathcal{X})$.

\textbf{Endpoints $s\in\{0,1\}$.} Directly from the definition: $[\hntk]^0=\LL\cong H^0(\mathcal{X})$, and $[\hntk]^1=\hntk\cong H^{\frac{p+1}{2}}(\mathcal{X})$ by \cref{thm:sobolev_ntk}.
\end{proof}
\section{Background on Integral/Covariance Operators and Interpolation Spaces}\label{sec:int_space}
Let $\mathcal{X}\subset\mathbb{R}^p$ be the input space, $\mathcal{Y}\subset\mathbb{R}$ be the target space, and $\rho$ be an unknown probability distribution supported on $\mathcal{X}\times\mathcal{Y}$ with marginal distribution $\mu$ on $\mathcal{X}$. Let $\LL$ denote the space of square-integrable functions with respect to $\mu$ defined on the input space $\mathcal{X}$ and taking values in $\mathcal{Y}$. Let $\mathcal{H}$ be an RKHS defined on $\mathcal{X}$ and associated with a kernel $K:\mathcal{X}\times\mathcal{X}\to\mathbb{R}$. We assume that $K$ is a \emph{Mercer} kernel, \emph{i.e.}, a continuous symmetric positive semi-definite kernel defined on a compact space $\mathcal{X}$. This implies that $K$ is bounded, \emph{i.e.}, there exists $\kappa\geq0$ such that $K(x,x)\leq\kappa^2$ for any $x\in\mathcal{X}$, and that $\mathcal{H}$ is separable, \emph{i.e.}, it contains a countable dense subset \citep[Lemma~4.33]{steinwart2008support}. Under this boundedness condition, it holds that $\mathcal{H}\subset\LL$. Indeed, for any $f\in\mathcal{H}$ and $x\in\mathcal{X}$, we can write by the reproducing property that $f(x)=\langle f,K(x,\cdot)\rangle_\mathcal{H}$, which implies by Cauchy-Schwarz inequality that $|f(x)|\leq\|f\|_\mathcal{H}\sqrt{K(x,x)}\leq\kappa\|f\|_\mathcal{H}$. This gives that $\|f\|_{\LL}^2\leq\kappa^2\|f\|_\mathcal{H}^2<+\infty$, and hence that $f\in\LL$.

\textbf{Integral and covariance operators.} Let $\mathcal{I}:\mathcal{H}\to\LL$ be the natural embedding inclusion operator defined for any $f\in\mathcal{H}$ by $\mathcal{I}(f)=f$. Its adjoint $\mathcal{I}^\star:\LL\to\mathcal{H}$ is defined for any $f\in\LL$ by $\mathcal{I}^\star(f)=\int_\mathcal{X}K(x,\cdot)f(x)\mathrm{d}\mu(x)\in\mathcal{H}$. Both $\mathcal{I}$ and $\mathcal{I}^\star$ are Hilbert-Schmidt operators, with Hilbert-Schmidt norms:
\begin{equation*}
    \left\|\mathcal{I}\right\|_\mathrm{HS}=\left\|\mathcal{I}^\star\right\|_\mathrm{HS}=\sqrt{\int_\mathcal{X}K(x,x)\mathrm{d}\mu(x)}\leq\kappa,
\end{equation*}
and are in particular compact. Let $\Sigma=\mathcal{I}^\star\mathcal{I}:\mathcal{H}\to\mathcal{H}$ be the covariance operator defined by $\Sigma=\mathbb{E}_{x\sim\mu}[K(x,\cdot)\otimes K(x,\cdot)]$, where $(a\otimes b)c=\langle b,c\rangle_\mathcal{H}a$ for $a,b,c\in\mathcal{H}$. Let $\mathcal{T}=\mathcal{I}\mathcal{I}^\star:\LL\to\LL$ be the integral operator defined for any $f\in\LL$ by $\mathcal{T}f=\int_\mathcal{X}K(\cdot,t)f(t)\mathrm{d}\mu(t)$. Since $K$ is a Mercer kernel, it follows that $\mathcal{T}$ is a compact, self-adjoint, and positive operator on $\LL$ \citep[Theorem~4.27]{steinwart2008support}. By the \emph{spectral theorem}, there exists an at most countable index set $N$, a non-increasing summable sequence $(\lambda_i)_{i\in N}\subset(0,+\infty)$, and an orthonormal family $(e_i)_{i\in N}\subset\LL$ forming an orthonormal basis of $\overline{\mathrm{Range}(\mathcal{I})}\subset\LL$, such that $\{\sqrt{\lambda_i}e_i\}_{i\in N}$ is an orthonormal basis of $\mathcal{H}$ \citep[Section~4.5]{steinwart2008support}. The operators $\mathcal{T}$ and $\Sigma$ admit the spectral decompositions:
\begin{equation*}
    \mathcal{T}=\sum_{i\in N}\lambda_i\langle\cdot,e_i\rangle_{\LL}e_i\quad\text{and}\quad\Sigma=\sum_{i\in N}\lambda_i\langle\cdot,\sqrt{\lambda_i}e_i\rangle_\mathcal{H}\sqrt{\lambda_i}e_i,
\end{equation*}
and share the same non-zero eigenvalues $(\lambda_i)_{i\in N}$, since $\mathcal{T}=\mathcal{I}\mathcal{I}^\star$ and $\Sigma=\mathcal{I}^\star\mathcal{I}$ have the same non-zero spectrum. Moreover, by Mercer's theorem \citep[Theorem~4.49]{steinwart2008support}, we have for any $x,x'\in\mathcal{X}$:
\begin{equation*}
    K(x,x')=\sum_{i\in N}\lambda_i e_i(x)e_i(x'),
\end{equation*}
where the convergence is absolute and uniform. Since $N$ is at most countable, there exists a bijection $\phi:\mathbb{N}\to N$; hence, by reindexing via $\phi$, we may write the eigenpairs as $(\lambda_{\phi(i)},e_{\phi(i)})_{i\geq 1}$, and, by a slight abuse of notation, simply denote them by $(\lambda_i,e_i)_{i\geq 1}$.

\textbf{Interpolation space. }For $s\geq 0$, we define the \emph{interpolation space} $[\mathcal{H}]^s$ by \citep[Section~2]{fischer2020sobolev}:
\begin{equation*}
    [\mathcal{H}]^s=\mathrm{Range}(\mathcal{T}^\frac{s}{2})=\left\{\sum_{i\geq 1}a_i\lambda_i^\frac{s}{2}e_i\;\Big |\;\sum_{i\geq 1}a_i^2<+\infty\right\}\subset\LL
\end{equation*}
equipped with the norm $\left\|\sum_{i\geq 1}a_i\lambda_i^\frac{s}{2}e_i\right\|_{[\mathcal{H}]^s}=\left(\sum_{i\geq 1}a_i^2\right)^\frac{1}{2}$. Note that $[\mathcal{H}]^0\subset\LL$ and $[\mathcal{H}]^1=\mathcal{H}$. The functions in $[\mathcal{H}]^s$ with smaller values of $s$ exhibit \emph{lower smoothness} and are therefore more difficult for kernel regression algorithms to estimate.
\section{Auxiliary Technical Lemmas}\label{sec:aux_lem}
In \cref{lem:e_delta,lem:upp_b_w0,lem:b_w0}, let $W_0$ be the principal branch of the Lambert $W$ function, \emph{i.e.}, the real-valued solution $v=W_0(z)\geq0$ of an equation of the form $v\exp(v)=z$ for $z\geq 0$.

\begin{lemma}\label{lem:e_delta}
Let $C_1,C_2>0$, $0<\alpha<1$, $0<\gamma<2$, and $x\geq\left\lfloor\left(\frac{4C_1(\ln 6)^3}{C_2}\right)^{\frac{1}{\alpha+\gamma}}\right\rfloor+1$. Let $\varphi:(0,1)\to(0,+\infty)$ be the function defined for any $\delta\in(0,1)$ by:
\begin{equation*}
    \varphi(\delta)=C_1\left(\ln\frac{6}{\delta}\right)^4x^{-\alpha}+C_2 x^\gamma\delta.
\end{equation*}
Then, $\varphi$ has a unique minimizer $\delta^\star\in(0,1)$, given by:
\begin{equation*}
    \delta^\star=\frac{108C_1}{C_2}x^{-(\alpha+\gamma)}W_0^3\left(x^{\frac{\alpha+\gamma}{3}}\left(\frac{C_2}{18C_1}\right)^{\frac{1}{3}}\right),
\end{equation*}
and the corresponding minimum value is:
\begin{equation*}
    \varphi(\delta^\star)=27C_1W_0^3\left(x^{\frac{\alpha+\gamma}{3}}\left(\frac{C_2}{18C_1}\right)^{\frac{1}{3}}\right)\left(3W_0\left(x^{\frac{\alpha+\gamma}{3}}\left(\frac{C_2}{18C_1}\right)^{\frac{1}{3}}\right)+4\right)x^{-\alpha}.
\end{equation*}
\end{lemma}

\begin{proof}
    \textbf{Existence and uniqueness of $\delta^\star\in(0,1)$. }The derivative of $\varphi$ for any $\delta\in(0,1)$ is:
    \begin{equation*}
        \varphi'(\delta)=-\frac{4C_1\left(\ln\frac{6}{\delta}\right)^3x^{-\alpha}}{\delta}+C_2x^\gamma.
    \end{equation*}
    The first term is strictly increasing in $\delta$ on $(0,1)$, since $\ln(6/\delta)$ and $1/\delta$ are positive and decrease as $\delta$ increases, while the second term is constant. Hence, $\varphi'$ is strictly increasing on $(0,1)$. Moreover, $\varphi'(\delta)\to-\infty$ as $\delta\to0^+$, and $\varphi'(\delta)\to-4C_1(\ln 6)^3x^{-\alpha}+C_2x^\gamma$ as $\delta\to1^-$. Since $x\geq\left\lfloor\left(\frac{4C_1(\ln 6)^3}{C_2}\right)^{\frac{1}{\alpha+\gamma}}\right\rfloor+1>\left(\frac{4C_1(\ln 6)^3}{C_2}\right)^{\frac{1}{\alpha+\gamma}}$, we have $x^{\alpha+\gamma}>\frac{4C_1(\ln 6)^3}{C_2}$, so that the limit $-4C_1(\ln 6)^3x^{-\alpha}+C_2x^\gamma$ is positive, and $\varphi'(\delta)>0$ for every $\delta$ close enough to $1$. Therefore, by the intermediate value theorem applied to the continuous function $\varphi'$, there exists $\delta^\star\in(0,1)$ such that $\varphi'(\delta^\star)=0$. Since $\varphi'$ is strictly increasing, this zero is unique, and since $\varphi'$ changes sign from negative to positive at $\delta^\star$, it is the unique minimizer of $\varphi$ on $(0,1)$.
    
    \textbf{Formula for $\delta^\star$. }The minimizer $\delta^\star$ satisfies $\varphi'(\delta^\star)=0$, that is:
    \begin{equation*}
        \frac{4C_1x^{-\alpha}}{\delta^\star}\left(\ln\frac{6}{\delta^\star}\right)^3=C_2x^\gamma.
    \end{equation*}
    Setting $y=\ln\frac{6}{\delta^\star}$, so that $\delta^\star=6\exp(-y)$, we obtain:
    \begin{equation*}
        y^3\exp(y)=\frac{3C_2}{2C_1}x^{\alpha+\gamma}\Longleftrightarrow y\exp\left(\frac{y}{3}\right)=\left(\frac{3C_2}{2C_1}x^{\alpha+\gamma}\right)^{\frac{1}{3}}.
    \end{equation*}
    Equivalently, setting $z=x^{\frac{\alpha+\gamma}{3}}\left(\frac{C_2}{18C_1}\right)^{\frac{1}{3}}$, $\frac{y}{3}\exp\left(\frac{y}{3}\right)=z$, since $\frac{1}{3}\left(\frac{3C_2}{2C_1}\right)^{\frac{1}{3}}=\left(\frac{C_2}{18C_1}\right)^{\frac{1}{3}}$. Using the identity $W_0(w)\exp(W_0(w))=w$, we get $y=3W_0(z)$. Since $\delta^\star=6\exp(-y)$, the identity $\exp(-W_0(w))=\frac{W_0(w)}{w}$ gives $\delta^\star=6\left[\exp\left(-W_0(z)\right)\right]^3=\frac{6W_0^3(z)}{z^3}=\frac{108C_1}{C_2}x^{-(\alpha+\gamma)}W_0^3(z)$, where we used $z^3=\frac{C_2}{18C_1}x^{\alpha+\gamma}$ in the last equality.
    
    \textbf{Formula for $\varphi(\delta^\star)$. }From the optimality condition, $C_2x^\gamma\delta^\star=4C_1y^3x^{-\alpha}$. As a consequence:
    \begin{align*}
        \varphi(\delta^\star)&=C_1\left(\ln\frac{6}{\delta^\star}\right)^4x^{-\alpha}+C_2x^\gamma\delta^\star=C_1y^4x^{-\alpha}+4C_1y^3x^{-\alpha}\\
        &=C_1y^3(y+4)x^{-\alpha}=27C_1W_0^3(z)\left(3W_0(z)+4\right)x^{-\alpha}.\qedhere
    \end{align*}
\end{proof}

\begin{lemma}[Upper-bound on $W_0$]\label{lem:upp_b_w0}
     For any $x\geq e$, it holds that $W_0(x)\leq\ln x $.
\end{lemma}
\begin{proof}
    Define $F(x)=\ln x-W_0(x)$ for any $x\geq e$. We have that:
    \begin{equation}\label{eq:der_F}
        \frac{\text{d}}{\text{d}x}F(x)=\frac{1}{x}-\frac{\text{d}}{\text{d}x}W_0(x).
    \end{equation}
    To compute the derivative of $W_0(x)$ with respect to $x$, we differentiate the identity $W_0(x)\exp(W_0(x))=x$. Using the product and chain rules yields:
    \begin{equation*}
        \left(\frac{\text{d}}{\text{d}x}W_0(x)\right)\exp(W_0(x))+W_0(x)\exp(W_0(x))\left(\frac{\text{d}}{\text{d}x}W_0(x)\right)=1.
    \end{equation*}
    After rearranging the terms and using the identity $W_0(x)\exp(W_0(x))=x$, we obtain:
    \begin{equation*}
        \frac{\text{d}}{\text{d}x}W_0(x)=\frac{W_0(x)}{x(1+W_0(x))}.
    \end{equation*}
    Note that this derivative is defined for any $x\geq e$ because $x(1+W_0(x))\neq 0$, which is due to the fact that $W_0(x)\geq 0$ for any $x\geq e$. Substituting the obtained derivative into \Cref{eq:der_F} gives:
    \begin{equation*}
        \frac{\text{d}}{\text{d}x}F(x)=\frac{1}{x}-\frac{W_0(x)}{x(1+W_0(x))}=\frac{1}{x(1+W_0(x))}\geq 0.
    \end{equation*}
    Thus, $F$ is increasing on $[e,+\infty)$. Since $F(e)=\ln e-W_0(e)=0$, we get $F(x)\geq 0$ for any $x\geq e$, which implies that $W_0(x)\leq \ln x$. This concludes the proof.
\end{proof}

\begin{lemma}\label{lem:b_w0}
    Let $0<\alpha<\gamma$ and $\kappa>0$. When $x\geq\max(e,(\frac{e}{\kappa})^{\frac{1}{\alpha}})$, the following holds:
    \begin{align*}
        W_0^3\left(x^\alpha\kappa\right)\left(3W_0\left(x^\alpha\kappa\right)+4\right)\leq&\Big(4\gamma^3+3\gamma^4+12(\gamma^2+\gamma^3)\abs{\ln\kappa}+(12\gamma+18\gamma^2)(\ln\kappa)^2\\
        &\quad+(4+12\gamma)\abs{\ln\kappa}^3+3(\ln\kappa)^4\Big)(\ln x)^4.
    \end{align*}
\end{lemma}

\begin{proof}
Fix $0<\alpha<\gamma$ and $\kappa>0$. We have $x^\alpha\kappa\geq e$ since it is assumed that $x\geq\left(\frac{e}{\kappa}\right)^{\frac{1}{\alpha}}$. Therefore, by \cref{lem:upp_b_w0}, we get $W_0\left(x^\alpha\kappa\right)\leq\ln\left(x^\alpha\kappa\right)=\alpha\ln x+\ln\kappa$. Using this, we have:
\begin{align*}
    W_0^3(x^\alpha\kappa)\leq\left(\alpha\ln x+\ln\kappa\right)^3=\alpha^3(\ln x)^3+3\alpha^2(\ln x)^2\ln\kappa+3\alpha\ln x(\ln\kappa)^2+(\ln\kappa)^3,
\end{align*}
and
\begin{align*}
    &W_0^3\left(x^\alpha\kappa\right)\left(3W_0\left(x^\alpha\kappa\right)+4\right)\leq\left(\alpha\ln x+\ln\kappa\right)^3\left(3\alpha\ln x+3\ln\kappa+4\right)\\
    =&\left(\alpha^3(\ln x)^3+3\alpha^2(\ln x)^2\ln\kappa+3\alpha\ln x(\ln\kappa)^2+(\ln\kappa)^3\right)\left(3\alpha\ln x+3\ln\kappa+4\right)\\
    =&3\alpha^4(\ln x)^4+3\alpha^3\ln\kappa(\ln x)^3+4\alpha^3(\ln x)^3+9\alpha^3(\ln x)^3\ln\kappa+9\alpha^2(\ln x)^2(\ln\kappa)^2\\
    &+12\alpha^2(\ln x)^2\ln\kappa+9\alpha^2(\ln x)^2(\ln\kappa)^2+9\alpha\ln x(\ln\kappa)^3+12\alpha\ln x(\ln\kappa)^2\\
    &+3\alpha\ln x(\ln\kappa)^3+3(\ln\kappa)^4+4(\ln\kappa)^3\\
    =&3\alpha^4(\ln x)^4+12\alpha^3\ln\kappa(\ln x)^3+4\alpha^3(\ln x)^3+18\alpha^2(\ln x)^2(\ln\kappa)^2+12\alpha^2(\ln x)^2\ln\kappa\\
    &+12\alpha\ln x(\ln\kappa)^3+12\alpha\ln x(\ln\kappa)^2+3(\ln\kappa)^4+4(\ln\kappa)^3.
\end{align*}
It is also assumed that $x\geq e$, hence $(\ln x)^i\leq(\ln x)^j$ for $0\leq i<j\leq 4$. Also, we know that $\alpha<\gamma$. Consequently, we get $W_0^3(x^\alpha\kappa)\leq\left(\gamma^3+3\gamma^2\abs{\ln\kappa}+3\gamma(\ln\kappa)^2+\abs{\ln\kappa}^3\right)(\ln x)^3$ and
\begin{align*}
    &W_0^3\left(x^\alpha\kappa\right)\left(3W_0\left(x^\alpha\kappa\right)+4\right)\\
    \leq&\Big(3\gamma^4+12\gamma^3\abs{\ln\kappa}+4\gamma^3+18\gamma^2(\ln\kappa)^2+12\gamma^2\abs{\ln\kappa}+12\gamma\abs{\ln\kappa}^3+12\gamma(\ln\kappa)^2+3(\ln\kappa)^4\\
    &\quad+4\abs{\ln\kappa}^3\Big)(\ln x)^4\\
    =&\left(4\gamma^3+3\gamma^4+12(\gamma^2+\gamma^3)\abs{\ln\kappa}+(12\gamma+18\gamma^2)(\ln\kappa)^2+(4+12\gamma)\abs{\ln\kappa}^3+3(\ln\kappa)^4\right)(\ln x)^4.
\end{align*}
This concludes the proof.
\end{proof}

\begin{lemma}[Surface measure of Euclidean spherical caps]\label{lem:sur_euc_geo_balls}
    Let $\hat{y}\in\mathbb{S}^{p}$ and $\epsilon>0$. Define the Euclidean spherical cap of radius $\epsilon$ centered at $\hat{y}$ as the Euclidean ball of radius $\epsilon$ centered at $\hat{y}$ intersected with the sphere: $B_{\mathbb{S}^{p}}(\hat{y},\epsilon)\coloneqq\{\hat{x}\in\mathbb{S}^{p}\mid\|\hat{x}-\hat{y}\|\leq\epsilon\}$. Then, it holds that $\sigma\left(B_{\mathbb{S}^{p}}(\hat{y},\epsilon)\right)\leq v_{p}\left(\frac{\pi}{2}\epsilon\right)^{p}$, where $\sigma$ is the surface measure on $\mathbb{S}^{p}$ and $v_{p}=\mathrm{Vol}(B(0,1))$ is the volume of the unit ball in $\mathbb{R}^{p}$.
\end{lemma}

\begin{proof}
    We denote by $d_g$ the geodesic distance and fix $\hat{x},\hat{y}\in\mathbb{S}^{p}$. On the unit sphere $\mathbb{S}^{p}$, we have $d_g(\hat{x},\hat{y})=\theta$, where $\theta\in[0,\pi]$ is the angle between $\hat{x}$ and $\hat{y}$. Since $\|\hat{x}\|=\|\hat{y}\|=1$, it follows that:
    \begin{equation*}
        \left\|\hat{x}-\hat{y}\right\|^2=\left\|\hat{x}\right\|^2+\left\|\hat{y}\right\|^2-2\left\langle\hat{x},\hat{y}\right\rangle=2-2\cos\theta=4\sin^2\left(\frac{\theta}{2}\right),
    \end{equation*}
    where we use the identity $1-\cos\theta=2\sin^2\left(\frac{\theta}{2}\right)$ in the last equality. Hence, we obtain $\left\|\hat{x}-\hat{y}\right\|=2\sin\left(\frac{\theta}{2}\right)$. Solving for $\theta$ gives $d_g(\hat{x},\hat{y})\coloneqq\theta=2\arcsin\left(\frac{\left\|\hat{x}-\hat{y}\right\|}{2}\right)$. Jordan's inequality states that $\sin\theta\geq\frac{2\theta}{\pi}$ for $\theta\in[0,\frac{\pi}{2}]$, or equivalently $\arcsin(t)\leq\frac{\pi t}{2}$ for $t\in[0,1]$. Since $\frac{\|\hat{x}-\hat{y}\|}{2}\in[0,1]$ for any $\hat{x},\hat{y}\in\mathbb{S}^p$, we obtain $d_g(\hat{x},\hat{y})\leq\frac{\pi}{2}\left\|\hat{x}-\hat{y}\right\|$. Therefore, we get $B_{\mathbb{S}^{p}}(\hat{y},\epsilon)\subset B_g(\hat{y},\frac{\pi}{2}\epsilon)$, where $B_g(\hat{y},\frac{\pi}{2}\epsilon)\coloneqq\{\hat{x}\in\mathbb{S}^{p}\mid d_g(\hat{x},\hat{y})\leq\frac{\pi}{2}\epsilon\}$ is the geodesic ball on the sphere of radius $\frac{\pi}{2}\epsilon$ and centered at $\hat{y}$. The surface measure of a geodesic ball of radius $r$ on $\mathbb{S}^{p}$ satisfies \citep{Li2011HypersphericalCap}:
    \begin{equation*}
        \sigma\left(B_g(\hat{y},r)\right)=w_{p-1}\int_0^r \sin^{p-1}t\,\mathrm{d}t\leq w_{p-1}\int_0^r t^{p-1}\,\mathrm{d}t=\frac{w_{p-1}}{p}r^{p},
    \end{equation*}
    where we use the fact that $\sin t\leq t$ for $t\geq 0$, with $w_{p-1}$ being the surface area of $\mathbb{S}^{p-1}$. Additionally, we have that $w_{p-1}=\frac{2\pi^{\frac{p}{2}}}{\Gamma\left(\frac{p}{2}\right)}$ and $v_{p}=\frac{\pi^{\frac{p}{2}}}{\Gamma\left(\frac{p}{2}+1\right)}$, where $v_{p}$ is the volume of the unit ball in $\mathbb{R}^{p}$ and $\Gamma$ is the gamma function. Using the identity $\Gamma(x+1)=x\Gamma(x)$, we obtain:
    \begin{equation*}
        v_{p}=\frac{\pi^{\frac{p}{2}}}{\left(\frac{p}{2}\right)\Gamma\left(\frac{p}{2}\right)}=\frac{w_{p-1}}{p},
    \end{equation*}
    which yields $\sigma\left(B_g(\hat{y},r)\right)\leq v_{p}\,r^{p}$. Using this inequality and the already established inclusion $B_{\mathbb{S}^{p}}(\hat{y},\epsilon)\subset B_g(\hat{y},\frac{\pi}{2}\epsilon)$, we get:
    \begin{equation*}
        \sigma\left(B_{\mathbb{S}^{p}}(\hat{y},\epsilon)\right)\leq\sigma\left(B_g\left(\hat{y},\frac{\pi}{2}\epsilon\right)\right)\leq v_{p}\left(\frac{\pi}{2}\epsilon\right)^{p}.
    \end{equation*}
    This concludes the proof.
\end{proof}

\begin{lemma}\label{lem:op_norm_exp_mat}
    Let $\boldsymbol{K}\in\mathbb{R}^{n\times n}$ be a symmetric positive definite matrix. For any $t\geq 0$, it holds that:
    \begin{equation*}
        \left\|\exp\left(-\frac{1}{n}t\boldsymbol{K}\right)\right\|_\mathrm{op}=\exp\left(-\frac{1}{n}\lambda_0 t\right),
    \end{equation*}
    where $\lambda_0$ is the smallest eigenvalue of $\boldsymbol{K}$.
\end{lemma}

\begin{proof}
    Since $\boldsymbol{K}$ is a symmetric positive definite matrix, it can be diagonalized as follows $\boldsymbol{K}=Q\Lambda Q^\top$, where $Q$ is orthogonal (\emph{i.e.}, $Q^{-1}=Q^\top$), $\Lambda=\mathbf{diag}(\lambda_1,\ldots,\lambda_n)$, and $\lambda_1,\ldots,\lambda_n> 0$ are the eigenvalues of $\boldsymbol{K}$. Then, we have:
    \begin{align*}
        \exp\left(-\frac{1}{n}t\boldsymbol{K}\right)&=\exp\left(-\frac{1}{n}tQ\Lambda Q^\top\right)=\sum_{k=0}^{+\infty}\frac{(-\frac{1}{n}tQ\Lambda Q^\top )^k}{k!}=Q\left(\sum_{k=0}^{+\infty}\frac{(-\frac{1}{n}t\Lambda)^k}{k!}\right)Q^\top\\
        &=Q\exp\left(-\frac{1}{n}t\Lambda\right)Q^\top=Q\mathbf{diag}\left(\exp\left(-\frac{1}{n}\lambda_1 t\right),\ldots,\exp\left(-\frac{1}{n}\lambda_n t\right)\right)Q^\top.
    \end{align*}
    Consequently, we get:
    \begin{equation*}
        \left\|\exp\left(-\frac{1}{n}t\boldsymbol{K}\right)\right\|_\mathrm{op}=\max_{1\leq i\leq n}\exp\left(-\frac{1}{n}\lambda_i t\right)=\exp\left(-\frac{1}{n}\lambda_0 t\right),
    \end{equation*}
    where $\lambda_0\coloneqq\min(\lambda_1,\ldots,\lambda_n)$ is the smallest eigenvalue of $\boldsymbol{K}$. This concludes the proof.
\end{proof}

\begin{lemma}\label{lem:identity_ln}
    For any $m\geq 1$ and $\alpha,\beta>0$, it holds that:
    \begin{equation*}
        \ln m\leq\frac{m^\alpha}{\alpha},\qquad\beta\ln m-m^\alpha\leq\frac{\beta}{\alpha}\left(\ln\left(\frac{\beta}{\alpha}\right)-1\right).
    \end{equation*}
\end{lemma}
\begin{proof}
    \textbf{First inequality.} Let $g(m)=\frac{m^\alpha}{\alpha}-\ln m$ for $m>0$. We have:
    \begin{equation*}
        g'(m)=m^{\alpha-1}-\frac{1}{m}=\frac{m^\alpha-1}{m}.
    \end{equation*}
    Since $\frac{1}{m}>0$, the sign of $g'$ is determined by $m^\alpha-1$: $g'(m)<0$ for $m<1$ and $g'(m)>0$ for $m>1$, so $m^\star=1$ is a global minimum of $g$ on $(0,+\infty)$. Since $g(1)=\frac{1}{\alpha}\geq 0$, we have $g(m)\geq\frac{1}{\alpha}$ for all $m>0$, which gives:
    \begin{equation*}
        \ln m\leq\frac{m^\alpha-1}{\alpha}\leq\frac{m^\alpha}{\alpha}.
    \end{equation*}
    In particular, this holds for all $m\geq 1$.

    \textbf{Second inequality.} Let $h(m)=\beta\ln m-m^\alpha$ for $m>0$. We have:
    \begin{equation*}
        h'(m)=\frac{\beta}{m}-\alpha m^{\alpha-1}=\frac{\beta-\alpha m^\alpha}{m}.
    \end{equation*}
    Since $\frac{1}{m}>0$, the sign of $h'$ is determined by $\beta-\alpha m^\alpha$: $h'(m)>0$ for $m<m^\star$ and $h'(m)<0$ for $m>m^\star$, where $m^\star=\left(\frac{\beta}{\alpha}\right)^{\frac{1}{\alpha}}$. Hence $m^\star$ is a global maximum of $h$ on $(0,+\infty)$, so for all $m\geq 1$:
    \begin{equation*}
        \beta\ln m-m^\alpha\leq h(m^\star)=\beta\cdot\frac{1}{\alpha}\ln\frac{\beta}{\alpha}-\frac{\beta}{\alpha}=\frac{\beta}{\alpha}\left(\ln\left(\frac{\beta}{\alpha}\right)-1\right).\qedhere
    \end{equation*}
\end{proof}

\begin{lemma}\label{lem:cond_m}
    Let $p\geq1$, $0<\varrho\leq 1$, $\tau>\frac{1}{2}$, and $\delta\in(0,1)$. Then, it holds that $m^{p\varrho}\exp\left(-2m^{2\tau-1}\right)\leq\delta$, for every $m\geq\max(1,M)$, where:
    \begin{equation*}
        M\coloneqq\left(\left[\frac{p\varrho}{2\tau-1}\left(\ln\left(\frac{p\varrho}{2\tau-1}\right)-1\right)+\ln\left(\frac{1}{\delta}\right)\right]_+\right)^{\frac{1}{2\tau-1}}\quad\text{with}\quad [t]_+\coloneqq\max(t,0).
    \end{equation*}
\end{lemma}

\begin{proof}
    Set $\alpha\coloneqq2\tau-1>0$, $\beta\coloneqq p\varrho>0$ and $u\coloneqq\frac{\beta}{\alpha}$. Taking logarithms, the desired inequality is equivalent to:
    \begin{equation}\label{eq:cond_m_log}
        p\varrho\ln m-2m^{2\tau-1}\leq\ln\delta.
    \end{equation}
    By \cref{lem:identity_ln}, we have $\beta\ln m-m^{\alpha}\leq u\left(\ln u-1\right)$ for any $m\geq1$, hence:
    \begin{equation*}
        p\varrho\ln m-2m^{2\tau-1}=\left(\beta\ln m-m^{\alpha}\right)-m^{2\tau-1}\leq u\left(\ln u-1\right)-m^{2\tau-1}.
    \end{equation*}
    It is therefore sufficient that:
    \begin{equation}\label{eq:cond_m_suff}
        m^{2\tau-1}\geq u\left(\ln u-1\right)+\ln\left(\frac{1}{\delta}\right).
    \end{equation}
    If the right-hand side of \cref{eq:cond_m_suff} is nonpositive, then \cref{eq:cond_m_suff} holds for every $m\geq1$, since $m^{2\tau-1}\geq1$. Otherwise, \cref{eq:cond_m_suff} is equivalent to $m\geq M$. In both situations, $m\geq\max(1,M)$ implies \cref{eq:cond_m_suff}, hence \cref{eq:cond_m_log}, which concludes the proof.
\end{proof}

\begin{lemma}\label{lem:tech_b}
    Let $n\geq 3$, $a\geq 1$, $0<b\leq1$, and define the function $f:(1,+\infty)\to\mathbb{R}$ by $f(x)=x^{-\frac{1}{a}}(\ln x)^b$. Then, for all $x\geq n^a(\ln n)^{ab}$, it holds that $f(x)\leq a^b(1+b)^b n^{-1}$.
\end{lemma}
\begin{proof}
    We first study the monotonicity of $f$. For $x>1$, we have by the product rule:
    \begin{equation*}
        f'(x)=-\frac{1}{a}x^{-\frac{1}{a}-1}(\ln x)^b+x^{-\frac{1}{a}} b(\ln x)^{b-1}\frac{1}{x}=x^{-\frac{1}{a}-1}(\ln x)^{b-1}\left(-\frac{1}{a}\ln x+b\right).
    \end{equation*}
    Since $x^{-\frac{1}{a}-1}(\ln x)^{b-1}>0$ for $x>1$, the sign of $f'$ is entirely determined by $-\frac{1}{a}\ln x+b$. Hence, $f'(x)<0$ whenever $x>e^{ab}$. Therefore, $f$ is decreasing on $(e^{ab},+\infty)$. Let $c=n^a(\ln n)^{ab}$ where $n\geq 3$. We have $\ln c=a\ln n+ab\ln\ln n>ab$, so $c>e^{ab}$. Thus, for all $x\geq c$, we have $f(x)\leq f(c)$. We compute:
    \begin{align*}
        f(c)=&\,n^{-1}(\ln n)^{-b}\left(a\ln n+ab\ln\ln n\right)^b=\,n^{-1}(\ln n)^{-b}a^b\left(\ln n+b\ln\ln n\right)^b\\
        \leq&\,n^{-1}(\ln n)^{-b}a^b\left(\ln n+b(\ln n-1)\right)^b\leq\,n^{-1}(\ln n)^{-b}a^b\left((1+b)\ln n\right)^b\\
        =&\,n^{-1}a^b(1+b)^b,
    \end{align*}
    where in the second line we use $\ln\ln n\leq\ln n-1$ for $n>1$. Therefore, $f(x)\leq a^b(1+b)^b n^{-1}$ for all $x\geq n^a(\ln n)^{ab}$, which concludes the proof.
\end{proof}

\begin{lemma}\label{lem:tech_lem_b}
    Let $C_1,C_2>0$. When $m\geq\max\left(\left(\frac{C_1\sqrt{12}}{C_2}\right)^{24},1\right)$, we have $C_1m^{-\frac{1}{12}}\sqrt{\ln m}\leq C_2$.
\end{lemma}

\begin{proof}
    When $m\geq\left(\frac{C_1\sqrt{12}}{C_2}\right)^{24}$, we get $C_1\sqrt{12}m^{-\frac{1}{24}}\leq C_2$. By \cref{lem:identity_ln}, we have $\ln m\leq 12 m^{\frac{1}{12}}$ when $m\geq 1$. Thus, combining both inequalities, when $m\geq\max\left(\left(\frac{C_1\sqrt{12}}{C_2}\right)^{24},1\right)$, we obtain $C_1 m^{-\frac{1}{12}}\sqrt{\ln m}\leq C_1 m^{-\frac{1}{12}}\sqrt{12}m^{\frac{1}{24}}=C_1\sqrt{12}m^{-\frac{1}{24}}\leq C_2$. This concludes the proof.
\end{proof}

\begin{lemma}\label{lem:tech_lem_b_2}
    Let $0<\alpha<\beta\leq 1$, $\delta>\frac{1}{2}$, $\gamma>\max(1-\alpha,\delta)$, $B>0$, $d\geq1$, and $c_0>0$ be an absolute constant. The following statements hold:
    \begin{enumerate}[label=(\roman*)]
        \item\label{stat1} When $m\geq 3^{\frac{1}{\gamma-\max(1-\alpha,\delta)}}$, it holds that $m^{1-\alpha}+m^\delta\leq\lfloor m^\gamma\rfloor$.
        \item\label{stat2} When $m\geq\max\left(2^{\frac{1}{\beta}},\left(\frac{128B^2d^2}{\pi(\beta-\alpha)}\right)^{\frac{1}{\beta-\alpha}}\right)$, it holds that $\frac{2Bd\sqrt{\ln m}}{\lfloor m^\beta\rfloor}\leq\frac{\sqrt{2\pi}}{4}m^{-\alpha}$.
        \item\label{stat3} When $m\geq\max\left(\frac{36d^2\beta^2}{c_0^2},1\right)$, it holds that $\frac{3d\beta}{c_0}\ln m\leq m$.
    \end{enumerate}
\end{lemma}

\begin{proof}
    \textbf{\cref{stat1}. }Since $m\geq 1$, we have $1\leq m^{\max(1-\alpha,\delta)}$, and therefore:
    \begin{equation*}
        m^{1-\alpha}+m^\delta\leq 2m^{\max(1-\alpha,\delta)}=(2m^{\max(1-\alpha,\delta)}+1)-1\leq 3m^{\max(1-\alpha,\delta)}-1.
    \end{equation*}
    Moreover, since $m\geq 3^{\frac{1}{\gamma-\max(1-\alpha,\delta)}}$, we have $m^{\gamma-\max(1-\alpha,\delta)}\geq 3$, which gives $3m^{\max(1-\alpha,\delta)}\leq m^\gamma$. Combining with the standard inequality $\lfloor m^\gamma\rfloor\geq m^\gamma-1$, we conclude:
    \begin{equation*}
        m^{1-\alpha}+m^\delta\leq 3m^{\max(1-\alpha,\delta)}-1\leq m^\gamma-1\leq\lfloor m^\gamma\rfloor.
    \end{equation*}

    \textbf{\cref{stat2}. }Since $m\geq 2^\frac{1}{\beta}$, we have $\frac{m^\beta}{2}\geq 1$. This implies that $\frac{m^\beta}{2}\leq m^\beta-1\leq\lfloor m^\beta\rfloor$. Consequently, we obtain:
    \begin{equation*}
        \frac{2Bd\sqrt{\ln m}}{\lfloor m^\beta\rfloor}\leq\frac{4Bd\sqrt{\ln m}}{m^\beta}\leq\frac{4Bdm^{-\frac{\beta+\alpha}{2}}}{\sqrt{\beta-\alpha}},
    \end{equation*}
    where, in the second inequality, we use the identity $\ln m\leq\frac{m^{\beta-\alpha}}{\beta-\alpha}$ provided in \cref{lem:identity_ln}. So, it suffices to have:
    \begin{equation*}
        \frac{4Bdm^{-\frac{\beta+\alpha}{2}}}{\sqrt{\beta-\alpha}}\leq\frac{\sqrt{2\pi}}{4}m^{-\alpha}\Longleftrightarrow \frac{16Bd}{\sqrt{2\pi(\beta-\alpha)}}\leq m^\frac{\beta-\alpha}{2},
    \end{equation*}
    which is satisfied since $m\geq\left(\frac{128B^2d^2}{\pi(\beta-\alpha)}\right)^\frac{1}{\beta-\alpha}$.

    \textbf{\cref{stat3}. }By \cref{lem:identity_ln}, for any $m\geq1$, we have $\ln m\leq2\sqrt{m}$. Using this, we have:
    \begin{equation*}
        \frac{3d\beta}{c_0}\ln m\leq\frac{6d\beta}{c_0}\sqrt{m}.
    \end{equation*}
    The result then follows since $m\geq\frac{36 d^2\beta^2}{c_0^2}$.
\end{proof}
\section{Further Details on Experiments and Additional Numerical Results}\label{sec:exp_det}

Our code is available at \href{https://github.com/fareselkhoury/OBO}{\texttt{https://github.com/fareselkhoury/OBO}.}

\subsection{Empirical validation of the theory in the NTK regime on synthetic data}\label{sec:validate-ntk}

\textbf{Statistical model.} We generate synthetic data as follows:
\begin{equation*}
x \sim \mathrm{Unif}([0,1]^2), \qquad z_j = \|x-c_j\|^{q_h} + \xi_j \ \ (j=1,\ldots,4), \qquad y = \|x-x_0\|^{q_b} + \xi_y,
\end{equation*}
where $\xi_j,\xi_y \sim \mathcal N(0,\tau^2)$ \emph{i.i.d.} with $\tau=0.01$, and the centers $c_1,\ldots,c_4$ and the point $x_0$ are as defined in the main text. $x$, $\xi_j$, and $\xi_y$ are all redrawn independently at every sample. The feature map $g_\omega$ is given by $g_\omega(z)=\omega^\top z$, where $z\coloneqq[z_1,z_2,z_3,z_4]^\top$. Since $\xi_j,\xi_y$ are zero-mean and independent of $x$, we obtain $h^\star_\omega(x)=\mathbb{E}[g_\omega(z)\mid x]=\psi(x)^\top\omega$ and $b^\star(x)=\mathbb{E}[y\mid x]=\|x-x_0\|^{q_b}$, where $\psi(x)\coloneqq[\|x-c_1\|^{q_h},\ldots,\|x-c_4\|^{q_h}]^\top$.

\textbf{Setup.} We use a width of $m=8{,}000$ for both networks approximating $h^\star_\omega$ and $b^\star$, with input dimension $p=2$ and outer dimension $d=4$. Both networks are trained by full-batch gradient descent at stepsize $1$, for $n_{\mathrm{steps}}=\max(1,\mathrm{round}(t_{\mathrm{op}}/1.0))$ steps, where $t_{\mathrm{op}}$ is the continuous-time stopping rule of \cref{prop:gen_bounds_main}. Because $h^\star_\omega$ and $b^\star$ are available in closed form here, we also compute the \emph{true} hypergradient $\nabla\mathcal F(\omega_k)$ by a Monte Carlo estimate over $20{,}000$ samples at every outer iteration, giving an accurate near-stationarity measure $\|\mathcal G_{1/\mathcal L}(\omega_k)\|$. We estimate $G=\sup_z\|z\|$ by the empirical maximum over $5{,}000$ Monte Carlo samples, and use $\mathcal L=G^2$ exactly, since $g_\omega(z)=\omega^\top z$ is linear in $\omega$ and $\partial_\omega g_\omega(z)=z$ does not depend on $\omega$. The projection set is $\Omega=\{\omega\in\mathbb R^d:\|\omega\|\le2\}$, the outer stepsize is $\eta=0.05$, the outer loop runs for $K=200$ iterations, and the hypergradient batch size is $n_g=200$. The outer parameter is initialized as $\omega_0\sim\mathcal N(0,\tfrac{1}{d}\mathbbm{1}_{d\times d})$, so that $\mathbb{E}[\|\omega_0\|^2]=1$ regardless of $d$, well inside the projection set $\Omega$. For each smoothness level and each sample size, we repeat the estimation of $h^\star_\omega$ and $b^\star$ over $5$ independent seeds and a held-out test set of $4{,}000$ points, reporting the mean and using the standard error of the mean for the confidence intervals. The experiment is implemented using the PyTorch framework \citep{paszke2019pytorch} and runs on a single NVIDIA RTX A5000 GPU. It takes approximately $3$ hours to complete.

\subsection{Insights into feature learning beyond the NTK regime}\label{sec:beyond-ntk}
The NTK analysis is theoretically sound but often \emph{conservative}, as it assumes fixed features and therefore precludes \emph{feature learning}. Following \citep{woodworth2020kernel}, a single network can transition from the NTK to a feature learning regime depending on initialization scale. The bilevel structure, however, introduces an \emph{additional degree of freedom}: the interplay between inner and outer initializations may jointly govern whether the network operates in the NTK or feature learning regime, potentially inducing feature learning even when a single network would remain in the NTK regime. This raises a natural question: \emph{Could the bilevel structure exhibit a feature learning regime where the network adapts its features and exceeds NTK predictions?} Our experiments explore this possibility.

\textbf{Statistical model.} We generate synthetic data as follows:
\begin{equation*}
    x \sim \mathcal{N}\left(0, \frac{1}{p}\mathbbm{1}_{p\times p}\right),\quad z = 2(\mathbbm{1}_p^\top x + \varepsilon)\mathbbm{1}_{r} + \eta,\quad y = g^\star(z) + \varepsilon,
\end{equation*}
where $\varepsilon \sim \mathcal{N}(0, 0.01^2)$, $\eta \sim \mathcal{N}(0,\frac{0.01^2}{4} \mathbbm{1}_{r\times r})$, $g^\star(z) = \phi(z)^\top \omega^\star$, $\phi(z) = \mathrm{SiLU}(Az + b) \in \mathbb{R}^d$ with $A \in \mathbb{R}^{d\times r}$ having \emph{i.i.d.} entries drawn from $\mathcal{N}(0, \frac{1}{r})$, $b \sim \mathcal{N}(0, 10^{-4} \mathbbm{1}_{d\times d})$, and $\omega^\star \sim \mathcal{N}(0, \frac{1}{d} \mathbbm{1}_{d\times d})$, all drawn once at initialization and fixed throughout.

\textbf{Setup. }We consider the empirical inner and outer objectives defined in \cref{eq:lin_lout_main} with $g_\omega:z\mapsto\phi(z)^\top\omega$, where $\phi(z)=\mathrm{SiLU}(Az+b)\in\mathbb{R}^d$ is a fixed random feature map, $\mathrm{SiLU}(x)=x/(1+\exp(-x))$ is applied entrywise, $A\in\mathbb{R}^{d\times r}$ and $b\in\mathbb{R}^d$ are drawn at initialization. We generate synthetic data with $100$ training samples and $20$ validation and test samples at each level, that remain fixed across all runs. Both $h^\star_\omega$ and $b^\star$ are approximated by neural networks $h_\omega^\mathrm{NN}$ and $b^\mathrm{NN}$ of width $500$, of the form in \cref{eq:nn_main} without second-layer bias, normalized by $m$ rather than $\sqrt{m}$ to be consistent with the feature learning limit as $m$ increases \citep{chizat2019lazy}. The first-layer parameters are initialized from a centered Gaussian of variance $\sigma_0^2$ for $h^\mathrm{NN}_\omega$ and $\tau_0^2$ for $b^\mathrm{NN}$, while the second-layer weights are initialized to 0. Since $b^\mathrm{NN}$ is independent of the outer variable, it is trained once at the beginning; $h^\mathrm{NN}_\omega$ is retrained at every outer iteration with a warm-start strategy \citep{vicol2022implicit}. Both networks are trained with full-batch gradient descent until the training error falls below $10^{-15}$. Early stopping is applied based on validation loss, selecting the iterate with the lowest validation error. At the outer level, we use full-batch gradient descent for $5000$ iterations, which is appropriate here because the sample size is small. We use the PyTorch framework \citep{paszke2019pytorch} on an NVIDIA RTX 6000 ADA GPU. The experiments take around 10 hours to complete for each pair $(\sigma_0^2,\tau_0^2)$. We set $d=30$, $p=20$, and $r=10$. The stepsize to train the two neural networks is fixed at $10$, whereas the stepsize for the outer optimization is $0.01$. We vary the initialization variances $\sigma_0^2$ and $\tau_0^2$ in $\{10^{-3},10^{-2},10^{-1},1,10,100\}$ independently, across training sizes and random seeds, to study the transition between the NTK and feature learning regimes.

\begin{figure}[ht]
    \centering
    \includegraphics[width=\linewidth]{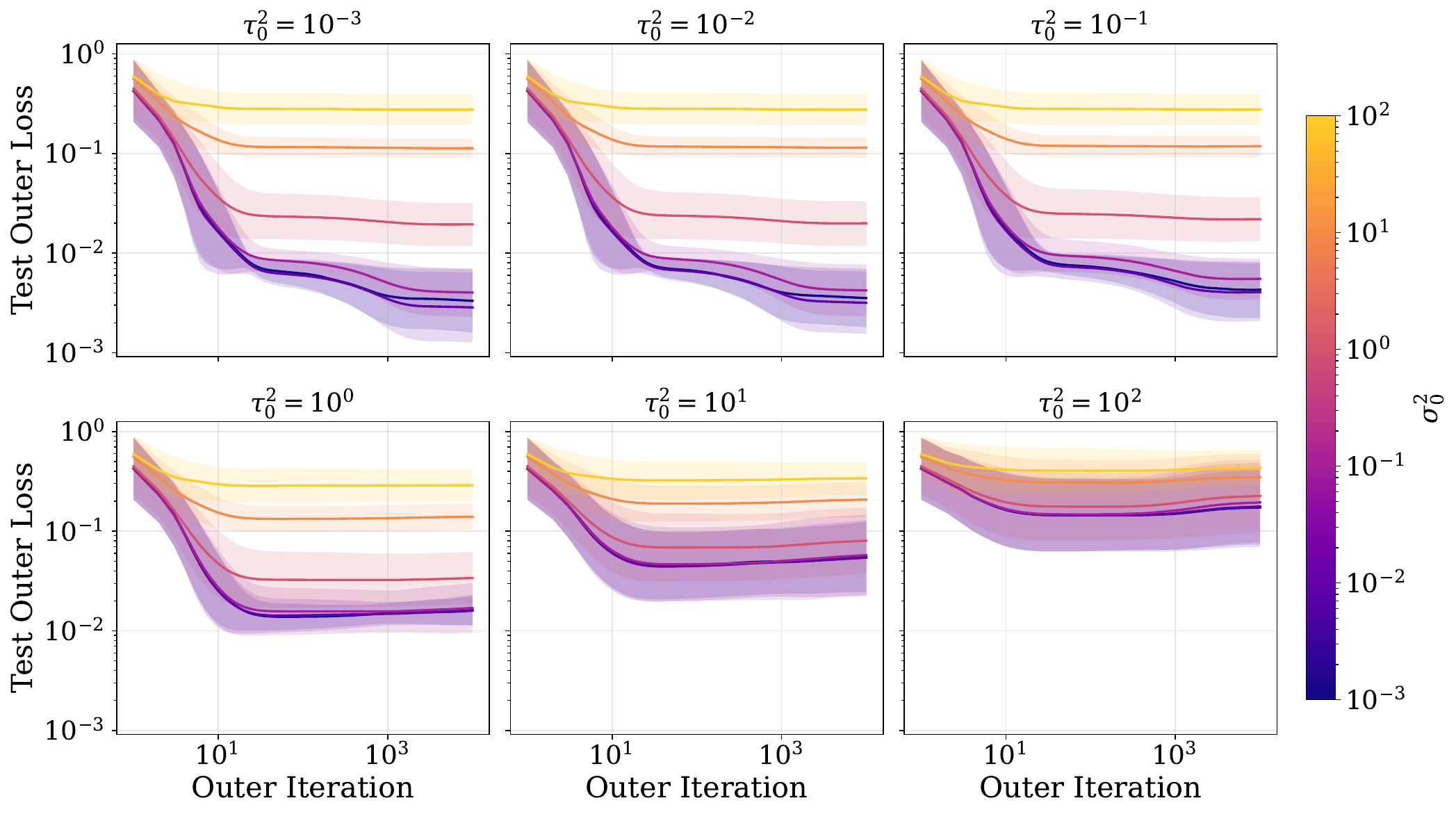}
    \caption{Test outer loss as a function of outer iterations for varying $\sigma_0^2$ and $\tau_0^2$: small initializations (feature learning) converge to lower loss, while large initializations (NTK) plateau at higher values. The plots are averaged over 5 runs and displayed on a log scale. The line represents the mean across all runs, and the shaded region indicates the 95\% confidence interval.}
    \label{fig:test_loss_outer_iter}
\end{figure}

\textbf{Additional numerical results. }\cref{fig:test_loss_outer_iter} shows the test outer loss over outer iterations for all combinations of $(\sigma_0^2, \tau_0^2)$. When both initializations are small, the network operates in the feature learning regime and converges to a low test loss. As either $\sigma_0^2$ or $\tau_0^2$ increases, the network transitions toward the NTK regime, with curves plateauing at higher loss values. This is most visible in the bottom-right panels, where both initializations are large and all curves converge to similarly high loss regardless of $\sigma_0^2$, suggesting that large $\tau_0^2$ alone is sufficient to push the network into the NTK regime.

\begin{figure}[ht]
    \centering
    \includegraphics[width=\linewidth]{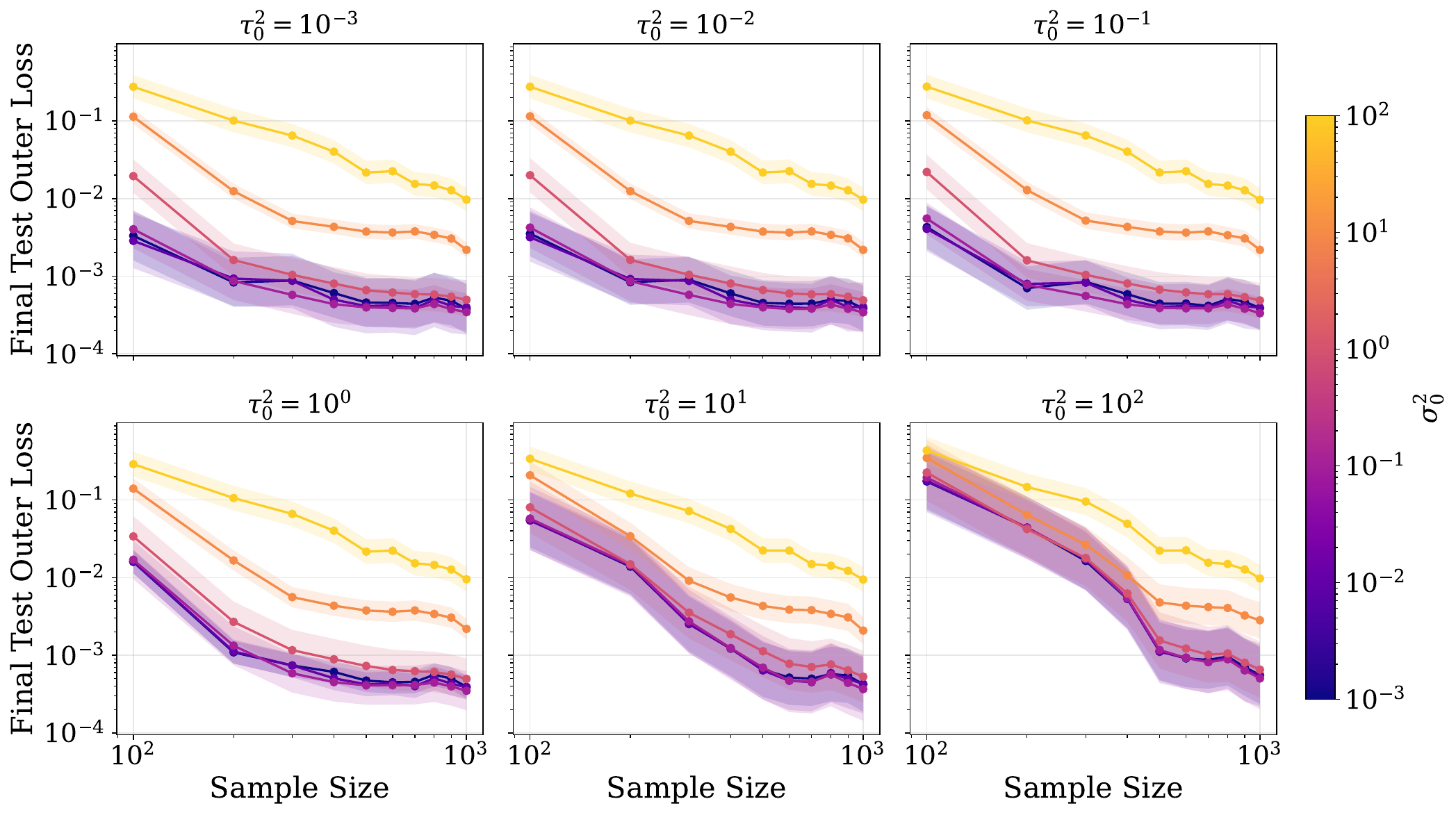}
    \caption{Test outer loss at the last iteration as a function of the sample size for varying $\sigma_0^2$ and $\tau_0^2$: all curves decrease with the sample size, with the ranking between feature learning and NTK regimes preserved throughout. The plots are averaged over 5 runs and displayed on a log scale. The line represents the mean across all runs, and the shaded region indicates the 95\% confidence interval.}
    \label{fig:test_loss_final_sample_size}
\end{figure}

\cref{fig:test_loss_final_sample_size} shows the final test outer loss as a function of the sample size for all combinations of $(\sigma_0^2, \tau_0^2)$. Three observations emerge. First, all curves decrease as the sample size grows, reflecting improved generalization with more data regardless of the regime. Second, the ranking between regimes is preserved at every sample size: when both initializations are small (feature learning), the network consistently achieves lower loss than when either initialization is large (NTK), confirming that the regime effect observed in \cref{fig:heatmaps} is not a finite-sample artifact. Third, when $\tau_0^2$ is large, the curves collapse and become insensitive to $\sigma_0^2$, further confirming that large $\tau_0^2$ alone is sufficient to push the network into the NTK regime, regardless of $\sigma_0^2$.

\begin{figure}[ht]
    \centering
    \includegraphics[width=\linewidth]{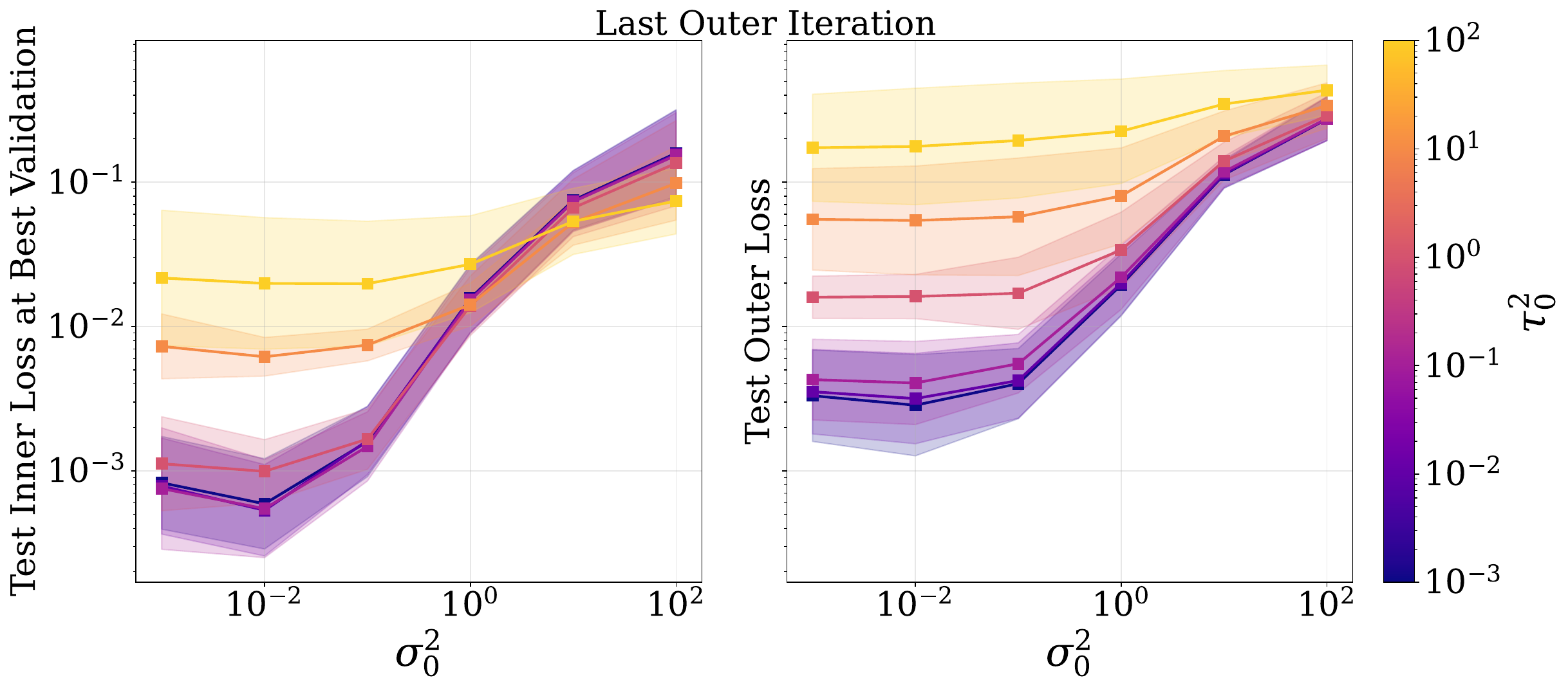}
    \caption{Test inner (left) and outer (right) losses at the last outer iteration as a function of $\sigma_0^2$ for varying $\tau_0^2$: both losses are lowest when both initializations are small (feature learning regime) and degrade as either grows large (NTK regime). The plots are averaged over 5 runs and displayed on a log scale. The line represents the mean across all runs, and the shaded region indicates the 95\% confidence interval.}
    \label{fig:test_inner_outer}
\end{figure}

\cref{fig:test_inner_outer} presents the same results as in \cref{fig:heatmaps} in line plot form, showing the test inner and outer losses at the last outer iteration as a function of $\sigma_0^2$ for varying $\tau_0^2$. The confidence intervals confirm that the ranking between regimes is stable across runs.

\end{document}